\documentclass[11pt]{article}
\usepackage[utf8]{inputenc}

\usepackage[letterpaper,top=2cm,bottom=2cm,left=3cm,right=3cm,marginparwidth=1.75cm]{geometry}

\usepackage{graphicx} 
\usepackage{amsmath}
\usepackage{amssymb}
\usepackage{upgreek}
\usepackage{mathtools}
\usepackage{lineno}
\usepackage{titlesec}
\usepackage{pdflscape}
\usepackage{rotating} 
\usepackage{booktabs}
\usepackage{threeparttable}
\usepackage{caption}
\usepackage{float}
\usepackage[numbers,sort&compress]{natbib}
\usepackage[colorlinks=true, allcolors=blue]{hyperref}
\usepackage{orcidlink}
\usepackage{algorithm}
\usepackage{algorithmicx}
\usepackage{algpseudocode}

\DeclareMathOperator*{\argminA}{arg\,min} 

\DeclareMathOperator{\sign}{sign}

\DeclareMathOperator{\Mean}{mean}
\DeclareMathOperator{\median}{median}
\DeclareMathOperator{\corr}{corr}

\DeclareMathOperator{\mad}{MAD}
\DeclareMathOperator{\Diag}{diag} 
\DeclareMathOperator{\Var}{Var}
\DeclareMathOperator{\Cov}{Cov}

\newcommand{\colvec}[3]{%
  \scalebox{#1}{%
    \renewcommand{\arraystretch}{1.2}%
    $\begin{bmatrix*}[#2]#3\end{bmatrix*}$%
  }%
}%

\newcommand{\ltfrac}[2]{\mbox{\large$\frac{#1}{#2}$}}

\usepackage{stackengine}
\stackMath
\newcommand{\lowwidehat}[1]{%
  \stackon[-1.80ex]{#1}{\smash{\widehat{\phantom{#1}}}}%
}

\usepackage{accents}

\usepackage{tcolorbox}
\usepackage{inconsolata} 

\usepackage{multirow}
\usepackage{array}

\usepackage{listings}
\usepackage{tcolorbox}
\usepackage{xcolor}
\tcbuselibrary{listings,skins}
\usepackage{pdflscape} 
\definecolor{matlabgreen}{rgb}{0,0.5,0}
\definecolor{matlabblue}{rgb}{0,0,1}
\definecolor{matlabpurple}{rgb}{0.58,0,0.82}

\lstdefinestyle{matlab}{
  language=Matlab,
  basicstyle=\ttfamily\small,
  keywordstyle=\color{matlabblue},
  commentstyle=\color{matlabgreen},
  stringstyle=\color{matlabpurple},
  numbers=none,
  showstringspaces=false,
  breaklines=true,
  tabsize=2,
  morekeywords={mean,cov,sqrt,numel,ones,mad,norminv}, 
}

\title{Reclaiming First Principles: A Differentiable Framework for Conceptual Hydrologic Models}

\author{Jasper A. Vrugt, Jonathan M. Frame and Ethan Bollman}
\date{\today}

\begin{document}

\maketitle

\begin{center}
\emph{This paper is dedicated to Professor Soroosh Sorooshian of the University of California, Irvine. His pioneering work on rainfall-discharge modeling, parameter estimation, and machine-learning- and satellite-based estimation of hydrologic fluxes has been a great inspiration to the present work and has helped shape modern hydrologic practice.}
\end{center}

\vspace{1em}

\begin{abstract}
Conceptual hydrologic models remain the cornerstone of rainfall-runoff modeling, yet their calibration is often slow and numerically fragile. Most gradient-based parameter estimation methods rely on finite-difference approximations or automatic differentiation frameworks (e.g., JAX, PyTorch and TensorFlow), which are computationally demanding and introduce truncation errors, solver instabilities, and substantial overhead. These limitations are particularly acute for the ODE systems of conceptual watershed models. Here we introduce a fully analytic and computationally efficient framework for differentiable hydrologic modeling based on exact parameter sensitivities. By augmenting the governing ODE system with sensitivity equations, we jointly evolve the model states and the Jacobian matrix with respect to all parameters. This Jacobian then provides fully analytic gradient vectors for any differentiable loss function. These include classical objective functions such as the sum of absolute and squared residuals, widely used hydrologic performance metrics such as the Nash-Sutcliffe and Kling-Gupta efficiencies, robust loss functions that down-weight extreme events, and hydrograph-based functionals such as flow-duration and recession curves. The analytic sensitivities eliminate the step-size dependence and noise inherent to numerical differentiation, while avoiding the instability of adjoint methods and the overhead of modern machine-learning autodiff toolchains. The resulting gradients are deterministic, physically interpretable, and straightforward to embed in gradient-based optimizers. Overall, this work enables rapid, stable, and transparent gradient-based calibration of conceptual hydrologic models, unlocking the full potential of differentiable modeling without reliance on external, opaque, or CPU-intensive automatic-differentiation libraries.
\end{abstract}

\maketitle

\setcounter{tocdepth}{2} 

\clearpage

\begingroup
\small              
\setlength{\parskip}{0pt}
\tableofcontents
\endgroup
\clearpage

\newpage

\section{Introduction and Scope}
A defining hallmark of enduring scientific progress is not the speed with which new methods appear, but the care with which they are grounded in first principles. Few scholars have embodied this philosophy more consistently than Professor Soroosh Sorooshian of the University of California, Irvine. Throughout his career, he has emphasized rigor over novelty, insight over complexity, and understanding over expedience. His contributions have fundamentally shaped modern hydrology, from establishing statistical foundations for hydrologic model calibration and revealing the intrinsic difficulty of parameter estimation in conceptual models, to transforming global optimization through the SCE-UA algorithm and pioneering the use of neural networks and satellite observations for precipitation estimation. In recognition of this extraordinary legacy, and in celebration of his receipt of the William Bowie Medal of the American Geophysical Union this year, this work is respectfully dedicated to him.


A recent trend in hydrology and in the broader Geosciences is the development of differentiable models, in which gradients of the model output with respect to its parameters are available directly within the simulation framework \citep{shen2023}. A prime example is the work by \citet{feng2022} who implemented the HBV conceptual watershed model in a differentiable-programming environment, embedding neural-network parameterizations within the process-based backbone. The resulting ``$\delta$-models'' are expressed in PyTorch and use automatic differentiation \citep{paszke2017} to enable end-to-end, gradient-based calibration. This capability has opened the door to probabilistic inference, variational methods, hybrid physics-machine-learning models, and real-time data assimilation. Subsequent studies by the same authors have shown that these differentiable models can match or approach the predictive skill of deep-learning models such as LSTMs, while retaining a physical interpretation and providing internal flux and storage outputs \citep{feng2023,feng2024}.

Despite this progress, differentiable modeling in its current form relies almost exclusively on automatic differentiation (AD) frameworks such as PyTorch \citep{paszke2017}, JAX \citep{bradbury2018}, and TensorFlow \citep{abadi2016}. While powerful and convenient, AD software infrastructures have been developed primarily for machine-learning research and may be conceptually distant from the workflow and intuition of hydrologists:
\begin{enumerate}
\item The computational structure of hydrologic models is hidden inside large and opaque computational graphs, making derivative operations difficult to audit or interpret. 
\item The solver implementation is tightly coupled to the AD engine: changes to the ODE solver or state representation often trigger complete re-tracing or re-compilation of the computational graph.     
\item AD frameworks require familiarity with programming patterns (tensor manipulation, computational graphs, device backends) that are not standard in hydrologic modeling.
\end{enumerate}
In addition to this, AD frameworks often impose non-trivial overhead in terms of installation, dependency management, debugging, and data structures. For example, reverse-mode differentiation requires storing intermediate states for backpropagation, which can create significant memory overhead. What is more, hydrologic models often include stiff, nonlinear storage-flux relationships for which naive AD can be orders of magnitude slower or more memory-intensive than necessary. Thus, AD methods provide the gradients but in a way that obscures the mathematics and limits efficiency, flexibility and understanding.

More fundamentally, in physically-based, process-resolved hydrology, analytic Jacobians have long been standard for solving the governing equations themselves. For example, ParFlow one of the most widely used open-source integrated hydrologic models solves a coupled system of three-dimensional variably saturated subsurface flow (Richards’ equation) and two-dimensional overland / surface flow equations, typically via fully implicit time discretization and Newton-Krylov nonlinear solution methods \citep{kuffour2020, maxwell2013,maxwell2015}. The successful use of ParFlow in large-scale and continental-scale hydrologic simulations demonstrates that fully coupled, physically based models with analytic Jacobians of the discretized residuals are not only feasible but also computationally tractable \citep[e.g.,][]{maxwell2015,condon2021}. These works illustrate that detailed knowledge of model sensitivity in terms of the Jacobian of the residual system can be exploited for robust, large-scale hydrologic simulation.

While differentiable hydrology has enabled gradient-based calibration in principle, there remains a striking lack of closed-form analytic derivations of state-Jacobian and parameter-sensitivity matrices for commonly used watershed models—and, in particular, of analytic gradients for widely used hydrologic performance metrics such as the Nash-Sutcliffe efficiency (NSE; \citet{nash1970}) and the Kling-Gupta efficiency (KGE; \citet{gupta2009}). As a result, most hydrologic calibration still relies on derivative-free optimization and/or finite-difference approximations, which sacrifice computational efficiency, obscure the physical and structural properties of the model, and complicate statistical inference under model misspecification \citep{vrugt2025a,vrugt2025b}

The purpose of this paper is to fill this gap by developing a fully differentiable hydrologic modeling framework based on augmented ODE systems that compute state trajectories and parameter sensitivities simultaneously using a single forward integration. The result is a hydrologic simulator that is transparent, fast, stable, and fully analytic, while avoiding the opacity and computational overhead associated with generic automatic differentiation toolboxes such as PyTorch, JAX, or TensorFlow. This analytic-gradient framework enables several key advances

\begin{enumerate}
\item\label{intro:point_1} \textit{Efficient gradient-based model calibration}, including steepest descent, Gauss-Newton, Levenberg-Marquardt \citep{levenberg1944,marquardt1963}, and hybrid schemes that combine derivative-free global search with local gradient-based refinement without finite-difference noise and at substantially reduced computational cost.
\item \textit{Unified treatment of a broad class of loss functions}, ranging from standard, weighted, and generalized least squares objectives, through robust M-estimators, to hydrologic efficiency metrics, scoring rules, and hydrograph-based functionals, all within a single analytic framework.
\item \textit{Rapid estimation of sandwich-adjusted posterior distributions} in a single MCMC experiment using score-based likelihood formulations \citep{Frazier2023a,vrugt2025b}.
\item \textit{Improved insight into parameter identifiability, sensitivity, and equifinality}, as analytic derivatives expose the local geometry of the objective function and parameter space in a way that is inaccessible to black-box optimization methods.
\end{enumerate}
With respect to \ref{intro:point_1}, we purposely use the term calibration for our parameter estimation procedure. Although this procedure is formally equivalent to estimating weights and biases in a machine-learning model via backpropagation, the term calibration reflects the presence of an explicit, physics-informed hypothesis about the data-generating process. This hypothesis is refined through adjustment of its parameters to match observations. By contrast, the term training carries no such prior structural assumption and is therefore more appropriate for data-driven machine-learning models.  

The analytic derivations of the Jacobians and gradient vectors presented in this paper keep the computational structure closely aligned with the governing model equations and their underlying storage-flux relationships. This yields a transparent, efficient, and statistically principled foundation for gradient-based calibration and uncertainty quantification in conceptual hydrologic modeling. The resulting algorithms are straightforward to read, verify, and modify, and can be embedded directly into existing hydrologic codes without the need to recast the entire model within a differentiable-programming environment. In doing so, our approach preserves core strengths of conceptual hydrology, (i) physical interpretability, (ii) compact state representations, and (iii) flexible model structures, while enabling the efficiency and rigor of gradient-based calibration, uncertainty quantification, and robust statistical inference. In short, it delivers many of the benefits of differentiable hydrologic modeling without reliance on heavyweight software frameworks or specialized computer-science tooling.

This manuscript grew out of a summer school on computational methods and machine learning taught by the first author in September 2024 at the University of California, Irvine. Preparing those lectures required revisiting, carefully and explicitly, the chain rule as it is used in modern deep-learning algorithms. Doing so revealed that many ideas now promoted under the banner of differentiable programming are not new to hydrology, but are already implicit in the mathematical structure of conceptual rainfall-runoff models. In that sense, this work emerged not from the wholesale adoption of new tools, but from a deliberate return to first principles.

That perspective traces directly to the foundational contributions of Professor Sorooshian. His early work on rainfall-discharge modeling, pioneering applications of neural networks to watershed response \citep{hsu1995}, rigorous model formulation \citep{sorooshian1983b,guptavk1985}, parameter estimation \citep{gupta1985} and optimization in watershed models \citep{sorooshian1980b,duan1992,sorooshian1993,duan1994} established much of the intellectual groundwork on which this study builds. Long before ``differentiable modeling'' became a term of art, these studies emphasized structure, sensitivity, identifiability, and the careful mathematical treatment of model-data relationships \citep{sorooshian1980,sorooshian1983,gupta1998}. The analytic sensitivity framework developed here is a natural extension of that tradition. It demonstrates the tangible benefits of mathematical and statistical education and training, exposing gradients and sensitivities that are otherwise buried beneath layers of abstraction in modern machine-learning software, where the hydrology itself can become obscured by generic optimization pipelines. 

The remainder of this paper is organized as follows. Section~\ref{sec:core_idea} introduces the central idea of the paper, clarifies the relationships between loss functions, Jacobian matrices, and gradient vectors, and explains their role in derivative-based parameter estimation. This section concludes with a summary of the specific contributions of this work. Section~\ref{sec:theory} presents the theoretical foundation, derives the forward sensitivity equations and augmented ODE system, and develops analytic expressions for Jacobians and gradients associated with commonly used loss functions. Section~\ref{sec:parameterization} discusses model parameterization and parameter transformations suitable for gradient-based optimization. Section~\ref{sec:num_and_automatic_diff} briefly reviews alternative approaches for computing Jacobians and gradients, including numerical and automatic differentiation. Section~\ref{sec:application} presents case studies for several conceptual watershed models, comparing analytic Jacobians and gradients with their numerical and automatic-differentiation counterparts, examining the influence of the numerical solver, and demonstrating the impact of analytic gradients on model calibration. Section~\ref{sec:outlook} discusses the main findings and provides an outlook toward broader applications, and Section~\ref{sec:conclusions} summarizes the principal conclusions of the study. Detailed analytic derivations, algorithmic recipes, and model formulations are provided in the Appendices.

\section{Core Idea and Contribution}\label{sec:core_idea}
We build on both the physically based modeling tradition and recent developments in differentiable hydrology to derive fully analytic gradients for conceptual rainfall-runoff models. Let $y_{t}$ denote the measured streamflow at time $t$, and let $q_{t} = h_{t}(\boldsymbol{\uptheta})$ be the corresponding model-simulated discharge under parameter vector $\boldsymbol{\uptheta} = (\theta_{1},\ldots,\theta_{d})^{\top}$. This notation suppresses the explicit dependence of the model output $q_{t}$ on state variables, exogeneous variables, soil properties and heterogeneity, land-use, and physical constants. 

\subsection{Loss function, Jacobian matrix and gradient vector}
We wish to minimize the distance between the observed and simulated streamflows. We can express this distance using a so-called pointwise loss function $\mathcal{L}_{t}(\cdot)$
\begin{linenomath*}
\begin{align}
\mathcal{L}(\boldsymbol{\uptheta}) = \sum_{t=1}^{n} \mathcal{L}_{t}(y_{t}, q_{t}),
\nonumber
\end{align}
\end{linenomath*}
where $\mathcal{L}(\cdot)$ is referred to as the total loss of the calibration period. Commonly used pointwise loss functions are the absolute error $\mathcal{L}_{t} = |y_{t} - q_{t}|$ and squared error $\mathcal{L}_{t} = \ltfrac{1}{2}(y_{t} - q_{t})^{2}$, and reward-based goodness-of-fit metrics such as the Nash-Sutcliffe efficiency \citep{nash1970} and Kling-Gupta efficiency \citep{gupta2009,kling2012}. 

If the loss function $\mathcal{L}_{t}$ is differentiable then the maximum-likelihood type or M-estimator of the model parameters $\boldsymbol{\uptheta}$ is defined as the solution for which the \emph{total score} $\mathbf{g}_{n}(\boldsymbol{\uptheta})$
\begin{linenomath*}
\begin{equation}
\mathbf{g}_{n}(\boldsymbol{\uptheta}) = \sum_{t=1}^{n}\nabla_{\boldsymbol{\uptheta}} \, \mathcal{L}_{t}(y_{t},q_{t}),
\label{eq:g_n(uptheta)} 
\end{equation}
\end{linenomath*}
vanishes, that is, $\mathbf{g}_{n}(\boldsymbol{\uptheta}) = \mathbf{0}_{d}$ \citep{huber1973}. The term $\mathbf{g}_{y_{t}}(\boldsymbol{\uptheta}) = \nabla_{\boldsymbol{\uptheta}}\, \mathcal{L}_{t}(y_{t},q_{t})$ denotes the \emph{score function} associated with the $t$th loss contribution. We  write this $d \times 1$ vector of partial derivatives as
\begin{linenomath*}
\begin{align}
\mathbf{g}_{y_{t}}(\boldsymbol{\uptheta}) \equiv \nabla_{\boldsymbol{\uptheta}} \, \mathcal{L}_{t}(y_{t},q_{t}) & \equiv \dfrac{\partial \mathcal{L}_{t}(y_{t},q_{t})}{\partial \boldsymbol{\uptheta}} \nonumber \\[1mm]
& = \colvec{1}{c}{ \, \dfrac{\partial \mathcal{L}_{t}(y_{t},q_{t}) }{\partial \theta_{1}}  \;\; \\
\, \vdots \;\; \\
\, \dfrac{\partial \mathcal{L}_{t}(y_{t},q_{t}) }{\partial \theta_{d}} \;\; }. \nonumber
\end{align}
\end{linenomath*}
and designate the bold lowercase letter $\mathbf{g}$ for \emph{gradient vector}. The \emph{total score} $\mathbf{g}_{n}(\boldsymbol{\uptheta})$ for a time series $y_{1}, \ldots, y_{n}$, thus, equals the sum of the individual \emph{score} contributions, $\mathbf{g}_{y_{1}}(\boldsymbol{\uptheta}), \ldots, \mathbf{g}_{y_{n}}(\boldsymbol{\uptheta})$. The entries of this $d \times 1$ vector play a central role in so-called (generalized) estimating equations \citep{godambe1960,liang1986,godambe1991}. An estimating function is unbiased if its expectation vanishes under the data-generating process indexed by $\boldsymbol{\uptheta}_{0}$, i.e., $\mathbb{E}[\mathbf{g}_{n}(\boldsymbol{\uptheta}_{0})] = \mathbf{0}_{d}$ \citep{dawid2005,mameli2015,giummole2018}. This property ensures consistency (under regularity conditions) and guarantees that M-estimators converge to the \emph{true} parameter values $\boldsymbol{\uptheta}_{0}$ as sample size $n$ increases. The autocorrelated and heteroscedastic maximum likelihood estimators of \citet{sorooshian1980} are prototypical M-estimators, since the optimum solution $\lowwidehat{\boldsymbol{\uptheta}}_{n}$ is a zero point of $\mathbf{g}_{n}(\boldsymbol{\uptheta}) = \nabla_{\boldsymbol{\uptheta}} \, \log(L_{n}(\boldsymbol{\uptheta}))$, where $L_{n}(\boldsymbol{\uptheta})$ is the likelihood of $\boldsymbol{\uptheta}$ given data $y_{1}, \ldots, y_{n}$.

Using the chain rule, the $j$th component of the \emph{total score} can be written as
\begin{linenomath*}
\begin{align}
g_{jn}(\boldsymbol{\uptheta}) & = \frac{\partial \mathcal{L}(\boldsymbol{\uptheta})}{\partial \theta_{j}} = \sum_{t=1}^{n} \frac{\partial \mathcal{L}_{t}(y_{t},q_{t})}{\partial q_{t}} \frac{\partial q_{t}}{\partial \theta_{j}}, \qquad (j = 1,\ldots,d)
\label{eq:g_jn(uptheta)_M_est}
\end{align}
\end{linenomath*}
where $\partial \mathcal{L}_{t}/\partial q_{t}$ is the sensitivity of the loss with respect to the simulated discharge. For losses of the form $\mathcal{L}_{t}(y_{t},q_{t}) = \mathcal{L}_{t}(e_{t})$ with residual $e_{t} = y_{t} - q_{t}$, this derivative satisfies $\partial \mathcal{L}_{t}/\partial q_{t} = - \psi(e_{t})$, where $\psi(e) = \mathrm{d}\mathcal{L}(e)/\mathrm{d}e$ is the so-called \textit{influence} (or $\psi$-) function introduced by \citet{hampel1968,hampel1974} and later elaborated in \citet{hampel1986}. This function, measures how a residual influences the parameter estimate(s) and is often of more practical interest than the loss function itself \citep{hampel1971,hampel1974,huber1981}. 

The \emph{total score}, or gradient vector, $\mathbf{g}_{n}(\boldsymbol{\uptheta})$
is obtained by collecting the scalar expressions in Equation~\ref{eq:g_jn(uptheta)_M_est} into vector form, yielding
\begin{linenomath*}
\begin{align}
\mathbf{g}_{n}(\boldsymbol{\uptheta})
& = \nabla_{\boldsymbol{\uptheta}} \mathcal{L}(\boldsymbol{\uptheta}) = \sum_{t=1}^{n}
\frac{\partial \mathcal{L}_{t}(y_{t},q_{t})}{\partial q_{t}}
\biggl(\frac{\partial q_{t}}{\partial \boldsymbol{\uptheta}^{\top}}\biggr)^{\!\top}, \nonumber
\end{align}
\end{linenomath*}
where $\mathbf{j}_{t}^{\top}(\boldsymbol{\uptheta}) = \partial q_{t}/\partial \boldsymbol{\uptheta}^{\top}$
denotes the $t$th row of the $n\times d$ Jacobian matrix $\mathbf{J}_{q}(\boldsymbol{\uptheta})$ of first-order derivatives of the simulated discharge with respect to the individual model parameters
\begin{linenomath*}
\begin{align}
\mathbf{J}_{q}(\boldsymbol{\uptheta}) =  \nabla_{\boldsymbol{\uptheta}} \mathbf{q}_{n}(\boldsymbol{\uptheta}) & = \colvec{1}{c}{ \; \mathbf{j}_{1}^{\top}(\boldsymbol{\uptheta}) \; \\[1mm]
\; \vdots \; \\
\; \mathbf{j}_{n}^{\top}(\boldsymbol{\uptheta}) \; } = \colvec{1}{c}{ \; \dfrac{\partial q_{1}(\boldsymbol{\uptheta})}{\partial \theta_{1}} & \dfrac{\partial q_{1}(\boldsymbol{\uptheta})}{\partial \theta_{2}} & \hdots & \dfrac{\partial q_{1}(\boldsymbol{\uptheta})}{\partial \theta_{d}} \; \\[1mm]
\; \vdots & \vdots &  & \vdots \; \\
\; \dfrac{\partial q_{n}(\boldsymbol{\uptheta})}{\partial \theta_{1}} & \dfrac{\partial q_{n}(\boldsymbol{\uptheta})}{\partial \theta_{2}} & \hdots & \dfrac{\partial q_{n}(\boldsymbol{\uptheta})}{\partial \theta_{d}} \; } \in \mathbb{R}^{n \times d}. \nonumber
\end{align}    
\end{linenomath*}
Introducing the vector $\boldsymbol{\updelta}_{n} = (\partial \mathcal{L}_{1}/\partial q_{1}, \ldots, \partial \mathcal{L}_{n}/\partial q_{n})^{\top}$ the gradient $\mathbf{g}_{n}(\boldsymbol{\uptheta})$ admits the compact matrix-vector form
\begin{linenomath*}
\begin{align}
\mathbf{g}_{n}(\boldsymbol{\uptheta})
& = \mathbf{J}_{q}^{\top}(\boldsymbol{\uptheta})\,\boldsymbol{\updelta}_{n}(\boldsymbol{\uptheta}).
\label{eq:g_n(uptheta)_key}
\end{align}
\end{linenomath*}
Equation~\ref{eq:g_n(uptheta)_key} is the central equation underpinning the methodology developed in this paper. It shows that for \emph{any} pointwise differentiable loss function
$\mathcal{L}_{t}(y_{t},q_{t})$, the total score or gradient vector
$\mathbf{g}_{n}(\boldsymbol{\uptheta}) \in \mathbb{R}^{d\times 1}$ decomposes naturally into two distinct components: (i) the discharge Jacobian $\mathbf{J}_{q}(\boldsymbol{\uptheta})$, which depends on the hydrologic model and its parameter values and (ii) the vector $\boldsymbol{\updelta}_{n}(\boldsymbol{\uptheta})$, which encodes how discrepancies between observations and simulations are propagated back through the model and therefore depends exclusively on the form of the loss function. For this reason, we refer to $\boldsymbol{\updelta}_{n}$ as the \emph{loss-sensitivity vector} or \emph{error-propagation vector}. Importantly, $\boldsymbol{\updelta}_{n}$ is not too difficult to derive by analytic means for a differentiable loss function. For example, for a sum of squared residuals loss $\mathcal{L}_{t}(y_{t},q_{t}) = \tfrac{1}{2}(y_{t} - q_{t})^{2}$, the entries of vector $\boldsymbol{\updelta}_{n}$ follow directly by differentiating $\mathcal{L}_{t}(y_{t},q_{t})$ with respect to $q_{t}$ to yield
\begin{linenomath*}
\begin{align}
\delta_{t} = \frac{\partial \mathcal{L}_{t}}{\partial q_{t}} = \tfrac{1}{2}(2q_{t}-2y_{t})
= -(y_{t}-q_{t}) = - e_{t}, \nonumber
\end{align}
\end{linenomath*}
where $e_{t} = y_{t} - q_{t}$ is the discharge residual at time $t$. Thus, in ordinary least-squares model calibration, the gradient reduces to the familiar expression $\mathbf{g}_{n}(\boldsymbol{\uptheta}) = - \mathbf{J}_{q}^{\top}(\boldsymbol{\uptheta})\,\mathbf{e}_{n}(\boldsymbol{\uptheta})$, with $\mathbf{e}_{n}(\boldsymbol{\uptheta}) = (e_{1}(\boldsymbol{\uptheta}),\ldots,e_{n}(\boldsymbol{\uptheta}))^{\top}$. Thus, given the Jacobian matrix $\mathbf{J}_{q}(\boldsymbol{\uptheta})$ of the simulated streamflows $q_{1}, \ldots, q_{n}$ with respect to the parameters $\theta_{1}, \ldots, \theta_{d}$ the evaluation of the total score becomes straightforward. Specifically, Equation~\ref{eq:g_n(uptheta)_key} allows computation of the gradient $\mathbf{g}_{n}(\boldsymbol{\uptheta})$ for any differentiable loss function. 

\subsection{Derivative-based parameter estimation}
Having access to analytic gradients enables efficient application of derivative-based optimization methods. In particular, gradient descent takes the form 
\begin{linenomath*}
\begin{align}
\boldsymbol{\uptheta}_{(k+1)} & = \boldsymbol{\uptheta}_{(k)} - \eta_{(k)}\,\mathbf{g}_{n}(\boldsymbol{\uptheta}_{(k)}), \nonumber
\end{align}
\end{linenomath*}
where $\eta_{(k)} > 0$ denotes the learning rate at iteration $k$. Substituting the gradient expression of Equation~\ref{eq:g_n(uptheta)_key} and rearranging yields
\begin{linenomath*}
\begin{align}
\Delta \boldsymbol{\uptheta}_{(k)} = \boldsymbol{\uptheta}_{(k+1)} - \boldsymbol{\uptheta}_{(k)} & = - \eta_{(k)}\,
\mathbf{J}_{q}^{\top}(\boldsymbol{\uptheta}_{(k)})
\boldsymbol{\updelta}_{n}(\boldsymbol{\uptheta}_{(k)}).
\label{eq:Delta_uptheta_(k)_GD}
\end{align}
\end{linenomath*}
This expression provides a general recipe for minimizing any pointwise differentiable loss function starting from an arbitrary initial parameter vector $\boldsymbol{\uptheta}_{(0)}$.
Repeated application of Equation~\ref{eq:Delta_uptheta_(k)_GD} iteratively reduces the total calibration loss $\mathcal{L}(\mathbf{y}_{n},\mathbf{q}_{n})$ at relatively low computational cost. The dimensionless scalar $\eta > 0$ or learning rate can be estimated at each iteration $k$ using line search
\begin{linenomath*}
\begin{align}
\eta_{(k)} & = \argminA_{\eta \, \in \, \mathbb{R}_{+}} \mathcal{L}\bigl(\boldsymbol{\uptheta}_{(k)} - \eta \, \mathbf{g}_{n}(\boldsymbol{\uptheta}_{(k)}) \bigr), \nonumber
\end{align}
\end{linenomath*}
thus, by locating the ``best'' point along the gradient direction $\mathbf{g}_{n}(\boldsymbol{\uptheta}_{(k)})$ which minimizes the total loss $\mathcal{L}$. For completeness, Algorithm \ref{algApp:gradient_descent} presents a step by step recipe of gradient descent.  

More robust and rapidly convergent schemes, such as the Levenberg-Marquardt (LM) algorithm \citep{levenberg1944,marquardt1963}, directly exploit the Jacobian $\mathbf{J}_{q}$ to interpolate between gradient descent and Gauss-Newton search through an adaptive damping parameter $\lambda$. At iteration $k$, LM computes the parameter shift vector $\Delta\boldsymbol{\uptheta}_{(k)}$ as follows
\begin{linenomath*}
\begin{align}
\Delta\boldsymbol{\uptheta}_{(k)} & = -\bigl\{\mathbf{J}_{q}^{\top}(\boldsymbol{\uptheta}_{(k)})\mathbf{J}_{q}(\boldsymbol{\uptheta}_{(k)}) + \lambda \, \Diag\bigl(\mathbf{J}_{q}^{\top}(\boldsymbol{\uptheta}_{(k)})\mathbf{J}_{q}(\boldsymbol{\uptheta}_{(k)})\bigr)\bigr\}^{-1}\mathbf{J}^{\top}_{q}(\boldsymbol{\uptheta}_{(k)}) \, \boldsymbol{\updelta}_{n}(\boldsymbol{\uptheta}_{(k)}).
\label{eq:Delta_uptheta_(k)_LM}
\end{align}
\end{linenomath*}
The parameter $\lambda$ is decreased when a trial step reduces the loss function and increased otherwise (see Algorithm \ref{algApp:Levenberg_Marquardt}). As $\lambda \to 0$, LM reduces to the Gauss-Newton method, taking curvature-informed steps that converge rapidly when the residual structure is locally linear. When $\lambda \to \infty$, LM approaches gradient descent, producing small, conservative steps that enhance robustness when far from a minimizer or in regions of strong nonlinearity. 

In textbooks and computer codes, the Jacobian of the model output (e.g., discharge $q_{1},\ldots,q_{n}$) $\mathbf{J}_{q}(\boldsymbol{\uptheta}) = \partial \mathbf{q}_{n}(\boldsymbol{\uptheta}) / \partial \boldsymbol{\uptheta}^{\top}$ is often interchanged with the Jacobian of the residuals, $\mathbf{J}_{e}(\boldsymbol{\uptheta}) = \partial \mathbf{e}_{n}(\boldsymbol{\uptheta}) / \partial \boldsymbol{\uptheta}^{\top}$. Because the residuals differ from the model output only by a minus sign, the two Jacobians satisfy the identity $\mathbf{J}_{e}(\boldsymbol{\uptheta}) = -\mathbf{J}_{q}(\boldsymbol{\uptheta})$. With this substitution, the leading minus sign in Equation~\ref{eq:Delta_uptheta_(k)_LM}
cancels. The switch between $\mathbf{J}_{q}$ and $\mathbf{J}_{e}$ is often made implicitly and without clear notation in textbooks and publications. For this reason, we explicitly retain the subscripts ``$q$’’ and ``$e$’’.

\subsection{Contribution of this paper}
The central contribution of this paper is the development of a fully analytic and computationally efficient framework for gradient-based calibration of conceptual hydrologic models. We derive state- and parameter-sensitivity equations for a class of widely used rainfall-runoff models under standard continuous-time dynamical formulations and exploit these sensitivities to compute the Jacobian matrix
$\mathbf{J}_{\mathbf{q}}(\boldsymbol{\uptheta}) \in \mathbb{R}^{n\times d}$ of simulated discharge with respect to model parameters. A single forward integration of an augmented ODE system yields noise-free, deterministic Jacobians and, consequently, fully analytic gradient vectors
$\mathbf{g}_{n}(\boldsymbol{\uptheta}) \in \mathbb{R}^{d\times1}$. These derivatives require no numerical perturbations, no step-size tuning, and no additional model evaluations per parameter, and they incur negligible additional computational cost even for moderately high-dimensional parameter spaces.

Building on the general relationship
$\mathbf{g}_{n}(\boldsymbol{\uptheta}) = \mathbf{J}_{\mathbf{q}}^{\top}(\boldsymbol{\uptheta})\,\boldsymbol{\updelta}_{n}(\boldsymbol{\uptheta})$, this framework accommodates a broad class of pointwise differentiable loss functions through analytically derived $\boldsymbol{\updelta}_{n}$-vectors. These include absolute and squared residual losses, widely used hydrologic performance metrics such as the Nash--Sutcliffe efficiency \citep{nash1970} and Kling--Gupta efficiency \citep{gupta2009,kling2012}, robust M-estimators that down-weight outliers \citep{vrugt2025a,vrugt2025b}, and hydrograph-based functionals such as flow-duration and recession curves \citep{vrugt2024b}. The resulting gradients enable fast, stable, and fully transparent application of derivative-based optimization methods-including stochastic gradient descent and Gauss-Newton-type algorithms-without reliance on finite differences or opaque automatic differentiation libraries. Beyond computational efficiency, the analytic Jacobians expose the local structure of the parameter space, providing valuable insight into parameter sensitivity, identifiability, equifinality, and uncertainty propagation under model misspecification.

\section{Theory}\label{sec:theory}
According to Equation \ref{eq:g_n(uptheta)_key} the gradient vector $\mathbf{g}_{n}(\boldsymbol{\uptheta}) \in \mathbb{R}^{d \times 1}$ is equal to the matrix-vector product $\mathbf{J}^{\top}_{q}(\boldsymbol{\uptheta})\boldsymbol{\updelta}_{n}(\boldsymbol{\uptheta})$. The Jacobian $\mathbf{J}_{q}(\boldsymbol{\uptheta})$ or $n \times d$ matrix of first-order partial derivatives of simulated discharge $q_{1},\ldots,q_{n}$ with respect to the parameters $\theta_{1},\ldots,\theta_{d}$ does not depend on the loss function. We will first discuss how we can determine this matrix efficiently, without numerical differentiation. Then, in the next section we derive analytic expressions for the $n \times 1$ vector $\boldsymbol{\updelta}_{n}(\boldsymbol{\uptheta})$ for a suite of different loss functions. The matrix-vector product $\mathbf{J}^{\top}_{q}(\boldsymbol{\uptheta})\boldsymbol{\updelta}_{n}(\boldsymbol{\uptheta})$ then produces the gradient vector $\mathbf{g}_{n}(\boldsymbol{\uptheta}) \in \mathbb{R}^{d \times 1}$ for any differentiable loss function. This makes possible rapid and CPU-efficient calibration of hydrologic and machine learning models.

\subsection{Forward sensitivity analysis}\label{subsec:FSA}
An established and mathematically rigorous foundation for the approach pursued here lies in the classical theory of sensitivity analysis for ordinary differential equations (ODEs). In this literature, one does not rely on finite differences or black-box numerical perturbations, but instead derives and integrates sensitivity equations alongside the original state ODEs. These sensitivity equations govern the evolution of the partial derivatives $\partial x_{i}/\partial \theta_{j}$, that is, how state variables respond to infinitesimal changes in parameters, yielding exact and noise-free sensitivity coefficients through time. This \emph{forward sensitivity analysis} is widely used in systems biology, atmospheric chemistry, and chemical kinetics, and enables efficient estimation, identifiability analysis, and uncertainty quantification \citep{cacuci1981,walter1997,perumal2011}.

We now adopt this classical sensitivity-analysis perspective for conceptual hydrologic models and treat a watershed as a general continuous-time dynamical system governed by the ordinary differential equation
\begin{linenomath*}
\begin{align}
\frac{\mathrm{d}\mathbf{x}}{\mathrm{d}t} & = \mathbf{f}(\mathbf{x}, \boldsymbol{\uptheta}, t),
\label{eq:ode}
\end{align}
\end{linenomath*}
where $\mathbf{x}(t) = (x_{1},\ldots,x_{m})^{\top} \in \mathbb{R}^{m \times 1}$ denotes the vector of state variables (e.g., surface water storage, unsaturated-zone moisture, groundwater), and $\boldsymbol{\uptheta} \in \mathbb{R}^{d}$ is the parameter vector. For conceptual rainfall-runoff models such as \texttt{hymod} \citep{boyle2001}, \texttt{hmodel} \citep{schoups2010a}, \texttt{sacsma} \citep{burnash1973}, and \texttt{xinanjiang} \citep{zhou2021}, the state variables $\mathbf{x}(t)$ represent conceptual water storages in interconnected reservoirs or control volumes. These storages are abstractions of surface water, soil moisture and groundwater and encode the dominant hydrologic processes such as infiltration, percolation, interflow, and fast and slow runoff that determine streamflow dynamics. The elements of $\boldsymbol{\uptheta}$ specify soil and reservoir properties such as maximum storage, percolation coefficients, and recession constants, and define the empirical flux equations (transfer functions) that move water between these conceptual stores \citep{sorooshian1983,moore2007}. This representation follows the long-standing tradition of parsimonious conceptual modeling in hydrology \citep{jakeman1993}, where internal storages are treated as physically meaningful even though the boundaries of these control volumes cannot be uniquely delineated in real space \citep{beven2012}. Conceptual models thus provide a parsimonious and interpretable framework for describing the catchment-scale rainfall-runoff transformation while remaining computationally efficient and suitable for gradient-based inference.



Streamflow is a primary output variable of watershed models. This flux is computed during model execution and subsequently returned to the user, typically after aggregation over one or more numerical integration steps. For the continuous sensitivity analysis (CSA) procedure to return the desired Jacobian matrix
$\mathbf{J}_{q}(\boldsymbol{\uptheta}) = \partial \mathbf{q}_{n}/\partial \boldsymbol{\uptheta}^{\top}$, discharge must be treated consistently within the state-space formulation. To this end, we expand the state vector by introducing an additional state variable, $x_{m}$, representing an infinite (non-depleting) reservoir that integrates the catchment outflow. Physically, $x_{m}(t)$ corresponds to the cumulative volume of water that has exited the catchment outlet between the start of the simulation at $t = 0$ and time $t$. This construction allows discharge to be recovered as a flux over the reporting interval rather than being prescribed as an independent model output. After numerical integration, the simulated discharge between two successive output (or measurement) times $t - 1$ and $t$ follows directly from mass conservation
\begin{linenomath*}
\begin{align}
q_{t} & = \frac{x_{m}(t) - x_{m}(t-1)}{t-(t-1)}
= x_{m}(t) - x_{m}(t-1),
\label{eq:q_t_simulation}
\end{align}
\end{linenomath*}
where the second equality assumes unit time steps. In this way, discharge emerges naturally from the evolution of the model’s internal water storages, rather than being introduced as a separate or externally defined variable.

Differentiating the governing ODE system (Equation~\ref{eq:ode}) with respect to the $j$th parameter $\theta_{j}$ yields
\begin{linenomath*}
\begin{align}
\frac{\mathrm{d}}{\mathrm{d}t}
\biggl[\frac{\partial \mathbf{x}}{\partial \theta_{j}}\biggr] & = \frac{\partial}{\partial \theta_{j}}\mathbf{f}(\mathbf{x},\boldsymbol{\uptheta},t).
\label{eq:d_dt_dx_dtheta_j}
\end{align}
\end{linenomath*}
To evaluate the right-hand side, we apply the chain rule
\begin{linenomath*}
\begin{align}
\frac{\partial}{\partial \theta_{j}} \mathbf{f}(\mathbf{x},\boldsymbol{\uptheta},t) & = \overbracket[0.2mm][2mm]{\underbracket[0.2mm][2mm]{\frac{\partial }{\partial \mathbf{x}\vphantom{\theta_{j}}}\mathbf{f}(\mathbf{x},\boldsymbol{\uptheta},t)}_\text{\normalsize $\vphantom{\dfrac{a}{b}} \mathbf{J}_{f}(\mathbf{x})$} \frac{\partial \mathbf{x}}{\partial \theta_{j}}}^\text{\normalsize $\vphantom{\frac{A}{B}}m \times 1$ vector} + \overbracket[0.2mm][2mm]{\underbracket[0.2mm][2mm]{\frac{\partial }{\partial \theta_{j}}\mathbf{f}(\mathbf{x},\boldsymbol{\uptheta},t)}_\text{\normalsize $[\vphantom{\dfrac{a}{b}}\mathbf{J}_{f}(\boldsymbol{\uptheta})]_{\bullet, j}$}}^\text{\normalsize $\vphantom{\frac{A}{B}}m \times 1$ vector},
\label{eq:d_dtheta_j_f(x,theta,t)}
\end{align}
\end{linenomath*}
where 
\begin{linenomath*}
\begin{align}
\mathbf{J}_{f}(\mathbf{x}) = \frac{\partial }{\partial \mathbf{x}^{\top}}\mathbf{f}(\mathbf{x},\boldsymbol{\uptheta},t) \in \mathbb{R}^{m \times m} \qquad \text{and} \qquad \mathbf{J}_{f}(\boldsymbol{\uptheta}) = \frac{\partial }{\partial \boldsymbol{\uptheta}^{\top}}\mathbf{f}(\mathbf{x},\boldsymbol{\uptheta},t) \in \mathbb{R}^{m \times d}, \nonumber
\end{align}
\end{linenomath*}
denote the Jacobian matrices of the system dynamics $\mathrm{d}\mathbf{x}/\mathrm{d}t = \mathbf{f}(\mathbf{x},\boldsymbol{\uptheta},t)$ with respect to states $\mathbf{x} = (x_{1},\ldots,x_{m})^{\top}$ and parameters $\boldsymbol{\uptheta} = (\theta_{1},\ldots,\theta_{d})^{\top}$, respectively, and $[\mathbf{J}_{f}(\boldsymbol{\uptheta})]_{\bullet,j}$ denotes the $j$th column of $\mathbf{J}_{f}(\boldsymbol{\uptheta})$.

Substituting Equation~\ref{eq:d_dtheta_j_f(x,theta,t)} into Equation~\ref{eq:d_dt_dx_dtheta_j} yields the forward sensitivity equations
\begin{linenomath*}
\begin{align}
\frac{\mathrm{d}}{\mathrm{d}t}\biggl[\frac{\partial \mathbf{x}}{\partial \theta_{j}}\biggr]
& = \mathbf{J}_{f}(\mathbf{x})
\frac{\partial \mathbf{x}}{\partial \theta{j}}
+ [\mathbf{J}_{f}(\boldsymbol{\uptheta})]_{\bullet,j},
\label{eq:d_dt_dx_dtheta_j_final}
\end{align}
\end{linenomath*}
which govern the temporal evolution of the sensitivity of the model states with respect to parameter $\theta_{j}$. The sensitivity vector $\partial \mathbf{x} / \partial \theta_{j}$ has the following entries
\begin{linenomath*}
\begin{align}
\frac{\partial \mathbf{x}}{\partial \theta_{j}} & = \colvec{1}{c}{ \; \dfrac{\partial x_{1}}{\partial \theta_{1}} \; \\[1mm] 
\; \vdots \; \\[1mm]
\; \dfrac{\partial x_{m}}{\partial \theta_{1}}} \in \mathbb{R}^{m \times 1}, \nonumber
\end{align}
\end{linenomath*}
and is solved simultaneously with the original state equations in the augmented ODE system of Equation \ref{eq:ode45_aug_system}. A key advantage in the present hydrologic context is that, for conceptual watershed models, the Jacobian matrices $\mathbf{J}_{f}(\mathbf{x})$ and $\mathbf{J}_{f}(\boldsymbol{\uptheta})$ can be derived analytically by direct differentiation of the governing model and flux equations. Appendix~\ref{sec:AppendixB} presents augmented ODE formulations for four widely used conceptual models: \texttt{hymod}, \texttt{hmodel}, \texttt{sacsma}, and \texttt{xinanjiang}. This includes analytic derivation of their respective  $\mathbf{J}_{f}(\mathbf{x})$ and $\mathbf{J}_{f}(\boldsymbol{\uptheta})$ matrices. 

If we combine the vectors $\partial \mathbf{x}/\partial \theta_{j}$ for $j = 1,\ldots,d$ into a single matrix, we obtain the $m \times d$ sensitivity matrix
\begin{linenomath*}
\begin{align}
\mathbf{S}(t) = \frac{\partial \mathbf{x}(t)}{\partial \boldsymbol{\uptheta}^{\top}} & = \colvec{1}{c}{ \; \dfrac{\partial \mathbf{x}}{\partial \theta_{1}} & \dfrac{\partial \mathbf{x}}{\partial \theta_{2}} & \hdots & \dfrac{\partial \mathbf{x}}{\partial \theta_{d}} \;} \nonumber \\
& = \colvec{1}{c}{ \; \dfrac{\partial x_{1}}{\partial \theta_{1}} & \dfrac{\partial x_{1}}{\partial \theta_{2}} & \hdots & \dfrac{\partial x_{1}}{\partial \theta_{d}} \; \\[1mm] 
\; \vdots & \vdots & & \vdots \; \\[1mm]
\; \dfrac{\partial x_{m}}{\partial \theta_{1}} & \dfrac{\partial x_{m}}{\partial \theta_{2}} & \hdots & \dfrac{\partial x_{m}}{\partial \theta_{d}} \;} \in \mathbb{R}^{m \times d}, \nonumber
\end{align}
\end{linenomath*}
then Equation \ref{eq:d_dt_dx_dtheta_j_final} can be written compactly in matrix form as
\begin{linenomath*}
\begin{align}
\frac{\mathrm{d}\mathbf{S}}{\mathrm{d}t} & = \mathbf{J}_{f}(\mathbf{x})\mathbf{S} + \mathbf{J}_{f}(\boldsymbol{\uptheta}). \nonumber
\end{align}
\end{linenomath*}
The sensitivity matrix quantifies how infinitesimal changes in the model parameters affect the evolving system states. Its initial condition is $\mathbf{S}_{0} = \mathbf{0}_{m \times d}$, which is independent of the model states and parameters (as is typical for conceptual hydrologic models).

We can solve for the states and sensitivities simultaneously, by grouping vector $\mathbf{x} \in \mathbb{R}^{d \times 1}$ and matrix $\mathbf{S} \in \mathbb{R}^{m \times d}$ into an augmented ODE system
\begin{linenomath*}
\begin{align}
\frac{\mathrm{d}\mathbf{z}}{\mathrm{d}t} = \frac{\mathrm{d}}{\mathrm{d}t}
\colvec{1}{c}{ \; \mathbf{x} \; \\ 
\; \mathrm{vec}(\mathbf{S}) \; } = \colvec{1}{c}{ \; \mathbf{f}(\mathbf{x},\boldsymbol{\uptheta},t) \; \\ 
\; \mathrm{vec}\bigl(\mathbf{J}_{f}(\mathbf{x})\mathbf{S} + \mathbf{J}_{f}(\boldsymbol{\uptheta})\bigr) \; }, 
\label{eq:ode45_aug_system}
\end{align}
\end{linenomath*}
and integrate for $\mathbf{z} \in \mathbb{R}^{m(d+1) \times 1}$ using a mass-conservative second-order solver with adaptive time step. By solving this expanded system, one can ensure a robust and accurate numerical solution of (i) the storages of the $m$ control volumes, (ii) the routed fluxes into and out of these reservoirs, and (iii) the sensitivities of all state variables to the model parameters. Thus, the sensitivities are computed to the same error tolerance as the state variables $x_{1},\ldots,x_{m}$, and only a single call is required to the ODE solver. This coupled system is known as continuous local sensitivity analysis (CSA) or forward sensitivity analysis (FSA) and is implemented in many large-scale ODE solvers \citep{hindmarsh2005,zhang2014}.

\subsection{The Jacobian matrix, \texorpdfstring{$\mathbf{J}_{q}(\boldsymbol{\uptheta}) \in \mathbb{R}^{n \times d}$}{}}\label{subsec:Jq}
After the numerical solver terminates, the augmented solution vector $\mathbf{z}(t)$ evaluated at the reporting times $t = 1,\ldots,n$ contains all information required to compute both the simulated discharge time series $\mathbf{q}_{n}$ and its Jacobian matrix $\mathbf{J}_{q}(\boldsymbol{\uptheta})$. The same finite-difference rule used in Equation~\ref{eq:q_t_simulation} to recover discharge from the cumulative runoff state applies directly to the sensitivities.

Let $\mathbf{S}(t) = \partial \mathbf{x}(t) / \partial \boldsymbol{\uptheta}^{\top}$ denote the sensitivity matrix returned by the augmented ODE system at time $t$. The $m$th row of $\mathbf{S}(t)$ contains the sensitivities of the discharge reservoir with respect to all parameters. Differentiating the discharge definition yields
\begin{linenomath*}
\begin{align}
\frac{\partial q_{t}}{\partial \theta_{j}} & = s_{m,j}(t) - s_{m,j}(t-1) = \bigl[\mathbf{J}_{q}(\boldsymbol{\uptheta})\bigr]_{t j},
\end{align}
\end{linenomath*}
where $s_{m,j}(t)$ denotes the $(m,j)$th entry of the sensitivity matrix $\mathbf{S}(t)$. Collecting these derivatives for all output times $t = 1,\ldots,n$ yields the exact Jacobian matrix of the discharge time series $\mathbf{J}_{q}(\boldsymbol{\uptheta}) = \partial \mathbf{q}_{n}/\partial  \boldsymbol{\uptheta}^{\top} \in \mathbb{R}^{n \times d}$. Thus, a single forward integration of the augmented ODE system simultaneously produces the simulated discharge and its exact, noise-free Jacobian with respect to all model parameters. No numerical perturbations, finite differences, or adjoint computations are required, and the additional computational cost is modest. The resulting \emph{forward-sensitivity Jacobian} is deterministic, independent of step-size choices or perturbation magnitudes, and free of truncation and round-off errors.

This approach provides a robust and transparent alternative to finite-difference and black-box differentiation techniques \citep{kramer1981,rabitz1989,turanyi1990,borggaard2000,depauw2006,gutenkunst2007,perumal2011,fisher2015}. Because the dimension of the augmented ODE system grows linearly with the number of parameters, forward-mode continuous sensitivity analysis has computational complexity $\mathcal{O}(l,d)$ for a system with $l$ state equations and $d$ parameters. While this scaling is well suited to the moderate parameter dimensions typical of conceptual hydrologic models, problems with very large parameter sets may instead benefit from continuous adjoint sensitivity analysis, which computes gradients with computational complexity $\mathcal{O}(l + d)$.

By grounding our hydrologic calibration framework in this ODE-sensitivity tradition, we combine the mathematical rigor and computational efficiency developed in systems biology and chemical kinetics with the practical demands and physical interpretability of conceptual watershed modeling. 

\subsection{The loss-sensitivity vector, \texorpdfstring{$\boldsymbol{\updelta}_{n}(\boldsymbol{\uptheta})$}{}} \label{subsec:LSV_delta}
Having obtained the Jacobian matrix $\mathbf{J}_{q}(\boldsymbol{\uptheta})$ of the simulated discharge with respect to the model parameters at relatively minimal CPU cost, the remaining step is the computation of the gradient vector $\mathbf{g}_{n}(\boldsymbol{\uptheta}) = \partial \mathcal{L}(\boldsymbol{\uptheta})/\partial \boldsymbol{\uptheta}^{\top} \in \mathbb{R}^{d \times 1}$. Following Equation \ref{eq:g_n(uptheta)_key}, this gradient is the product of the discharge Jacobian and an $n \times 1$ vector $\boldsymbol{\updelta}_{n}(\boldsymbol{\uptheta}) = \partial \mathcal{L}(\boldsymbol{\uptheta})/\partial \mathbf{q}_{n}$ 
which contains the sensitivities of the loss function with respect to the simulated discharge at each time step. Importantly, while the Jacobian $\mathbf{J}_{q}(\boldsymbol{\uptheta})$ is entirely determined by the model dynamics and is therefore independent of the chosen loss function, the vector $\boldsymbol{\updelta}_{n}(\boldsymbol{\uptheta})$ depends explicitly on the form of the loss. Table \ref{table:1} presents closed-form expressions for $\boldsymbol{\updelta}_{n}(\boldsymbol{\uptheta})$ for a broad class of loss functions, thereby completing the analytic construction of the gradient vector $\mathbf{g}_{n}(\boldsymbol{\uptheta})$.
\begin{landscape}
\begin{table}
\centering
\captionsetup[table]{position=bottom}
\begin{threeparttable}
\caption{Loss-sensitivity vectors
$\boldsymbol{\updelta}_{n}(\boldsymbol{\uptheta}) = \partial \mathcal{L}(\boldsymbol{\uptheta})/\partial \mathbf{q}_{n}$ of a suite of different loss functions. Full derivations are presented in Appendix~\ref{sec:AppendixC}.}
\normalsize
\begin{tabular}{p{44mm} p{83mm} p{83mm} p{8mm}}
\toprule
\textbf{Name} & \textbf{Loss function}, $\mathcal{L}(\boldsymbol{\uptheta})$ & \textbf{Loss-sensitivity vector}, $\boldsymbol{\updelta}_{n}(\boldsymbol{\uptheta})$ & \textbf{Eq.} \\
\midrule
Sum Absolute Residuals & $\mathcal{L}_\mathrm{sar} = \sum\limits_{t=1}^{n} \lvert y_{t} - q_{t} \rvert$ &
$\boldsymbol{\updelta}_{n,\mathrm{sar}} = -\sign(\mathbf{y}_{n}-\mathbf{q}_{n})$ & \ref{eqApp:delta_sar} \\[4mm]

Generalized Least Squares & $\mathcal{L}_{\mathrm{gls}}
= \frac{1}{2}(\mathbf{y}_{n} - \mathbf{q}_{n})^{\top}
\boldsymbol{\Sigma}_{\epsilon}^{-1}(\mathbf{y}_{n} - \mathbf{q}_{n})$ & 
$\boldsymbol{\updelta}_{n,\mathrm{gls}} = - \boldsymbol{\Sigma}_{\epsilon}^{-1}(\mathbf{y}_{n} - \mathbf{q}_{n})$ & \ref{eqApp:delta_gls} \\[3mm]

Nash-Sutcliffe Efficiency & $\mathcal{L}_\mathrm{nse} = \dfrac{\mathrm{SS}_\mathrm{r}(\mathbf{y}_{n},\mathbf{q}_{n})}{\mathrm{SS}_\mathrm{t}(\mathbf{y}_{n})}$ &
$\boldsymbol{\updelta}_{n,\mathrm{nse}} = \dfrac{2}{\mathrm{SS}_\mathrm{t}(\mathbf{y}_{n})}(\mathbf{q}_{n} -\mathbf{y}_{n})$ & \ref{eqApp:delta_nse} \\[4mm]

Kling-Gupta Efficiency & $\mathcal{L}_\mathrm{kge} = \sqrt{(r_{qy}-1)^2 + (\nu_{qy} - 1)^2 + (b_{qy} - 1)^2}$ & 
$\boldsymbol{\updelta}_{n,\mathrm{kge}} = \biggl(\dfrac{\partial \mathcal{L}_\mathrm{kge}}{\partial q_{1}}, \ldots, \dfrac{\partial \mathcal{L}_\mathrm{kge}}{\partial q_{n}}\biggr)^{\top}$ & \ref{eqApp:delta_kge} \\[2mm]

Huber robust loss & $\mathcal{L}_\mathrm{huber} = \sum\limits_{t=1}^{n}\mathcal{L}_{c}(\underline{e}_{t}) =
\begin{cases}
\; \frac{1}{2}\underline{e}_{t}^{2}, & \lvert \underline{e}_{t} \rvert \le c,\\[.5mm]
\; c \lvert \underline{e}_{t} \rvert - \frac{1}{2}c^{2}, & \lvert \underline{e}_{t} \rvert > c \end{cases}$ &
$\boldsymbol{\updelta}_{n,\mathrm{huber}} = - \dfrac{1}{S_{y}}\psi_{c}\biggl(\dfrac{\mathbf{y}_{n} -\mathbf{q}_{n}}{S_{y}}\biggr)$ & \ref{eqApp:delta_huber} \\[3mm]

Flow duration curve & $\mathcal{L}_\mathrm{fdc} = \mathbb{E}_{Q,F}\bigl[\vert q - y \vert \bigr] - \ltfrac{1}{2}\bigl(\mathbb{E}_{Q}\bigl[\vert q - q^{\ast} \vert \bigr] + \mathbb{E}_{F}\bigl[\vert y - y^{\ast} \vert \bigr] \bigr)$\tnote{\S} &
$\boldsymbol{\updelta}_{n,\mathrm{fdc}} = \dfrac{1}{n^{2}}
\Biggl[\sum\limits_{j=1}^{n}\sign(q_{i}-y_{j}) -\sum\limits_{j=1}^{n}\sign(q_{i} - q_{j}) \Biggr]_{i=1}^{n}$\tnote{\P} & \ref{eqApp:delta_fdc} \\
\bottomrule
\end{tabular}
\label{table:1}
\begin{tablenotes}
\item[\S] A Monte Carlo approximation is given in Equation \ref{eqApp:d_FDC(Q,F)_monte_carlo}.
\item[\P] $\sign(x)$ is the signum function. This function returns $-1$ if $x < 0$, $0$ if $x = 0$ and $1$ if $x > 0$
\end{tablenotes}
\end{threeparttable}
\end{table}
\end{landscape}

Appendix~\ref{sec:AppendixC} reviews the cost functions considered in this study and presents analytic derivations of their corresponding loss-sensitivity vectors $\boldsymbol{\updelta}_{n}(\boldsymbol{\uptheta})$. These include the sum of absolute residuals ($\ell_{1}$) loss, $\mathcal{L}_\mathrm{sar}$ (Section~\ref{subsecApp:SAR}); generalized least squares ($\ell_{2}$) loss, $\mathcal{L}_\mathrm{gls}$ (Section~\ref{subsecApp:GLS}); the Nash-Sutcliffe loss, $\mathcal{L}_\mathrm{nse}$ (Section~\ref{subsecApp:NSE}); the Kling-Gupta loss, $\mathcal{L}_\mathrm{kge}$ (Section~\ref{subsecApp:KGE}); the Huber robust loss, $\mathcal{L}_\mathrm{huber}$ (Section~\ref{subsecApp:Huber}); and the flow duration curve loss, $\mathcal{L}_\mathrm{fdc}$ (Section~\ref{subsecApp:FDC}). Here, we present condensed derivations of the loss-sensitivity vectors of two of the most widely used performance metrics in hydrology, the Nash-Sutcliffe and Kling-Gupta efficiencies. 

\subsubsection*{Nash-Sutcliffe efficiency}
The \citet{nash1970} efficiency, $\mathrm{NSE}: \ \mathbb{R}^{n} \times \mathbb{R}^{n} \to (-\infty,1]$ is given by
\begin{linenomath*}
\begin{align}
\mathrm{NSE}(\mathbf{y}_{n},\mathbf{q}_{n}) & = 1 - \frac{\sum_{t=1}^{n} (y_{t} - q_{t})^{2}}{\sum_{t=1}^{n} (y_{t} - m_{y})^2} = 1 - \frac{\mathrm{SS}_\mathrm{r}(\mathbf{y}_{n},\mathbf{q}_{n})}{\mathrm{SS}_\mathrm{t}(\mathbf{y}_{n})}, \nonumber
\end{align}
\end{linenomath*}
where $m_{y} = \frac{1}{n}\sum_{t=1}^{n}y_{t}$ denotes the sample mean of the discharge observations and $\mathrm{SS}_\mathrm{r}$ and $\mathrm{SS}_\mathrm{t}$ are the residual and total sum of squares, respectively.

To use NSE in gradient descent, we typically \emph{minimize} the squared-error fraction
\begin{linenomath*}
\begin{align}
\mathcal{L}_\mathrm{nse}(\boldsymbol{\uptheta}) & = 1 - \mathrm{NSE}(\mathbf{y}_{n},\mathbf{q}_{n}) = \frac{\mathrm{SS}_\mathrm{r}(\mathbf{y}_{n},\mathbf{q}_{n})}{\mathrm{SS}_\mathrm{t}(\mathbf{y}_{n})}. \label{eq:L_nse(uptheta)}
\end{align}
\end{linenomath*}
Differentiating w.r.t.\ $q_{1}, \ldots, q_{n}$ gives
\begin{linenomath*}
\begin{align}
\boldsymbol{\updelta}_{n,\mathrm{nse}}(\boldsymbol{\uptheta}) = \frac{\partial \mathcal{L}_\mathrm{nse}(\boldsymbol{\uptheta})}{\partial \mathbf{q}_{n}} & = \frac{2}{\mathrm{SS}_\mathrm{t}(\mathbf{y}_{n})}(\mathbf{q}_{n} - \mathbf{y}_{n}). \nonumber 
\end{align}
\end{linenomath*}
This leaves us with the gradient of $\mathcal{L}_\mathrm{nse}$ with respect to the parameters
\begin{linenomath*}
\begin{align}
\mathbf{g}_{n,\mathrm{nse}}(\boldsymbol{\uptheta}) & = \frac{\partial \mathbf{q}_{n}}{\partial \boldsymbol{\uptheta}^{\top}} \frac{\partial \mathcal{L}_\mathrm{nse}(\boldsymbol{\uptheta})}{\partial \mathbf{q}_{n}} = \mathbf{J}^{\top}_{q}(\boldsymbol{\uptheta})\boldsymbol{\updelta}_{n,\mathrm{nse}}(\boldsymbol{\uptheta}). \nonumber
\end{align}
\end{linenomath*}

\subsubsection*{Kling-Gupta Efficiency}
The Kling--Gupta efficiency or KGE of \citet{gupta2009} is a widely used alternative to the NSE for evaluating hydrologic model performance. The $\mathrm{KGE}: \ \mathbb{R}^{n} \times \mathbb{R}^{n} \to (-\infty,1]$ addresses known NSE limitations related to its disproportionate sensitivity to high flows and combines three quasi-orthogonal measures of model performance
\begin{linenomath*}
\begin{equation}
\mathrm{KGE}(\mathbf{y}_{n}, \mathbf{q}_{n}) = 1 - \sqrt{(r_{qy} - 1)^2 + (\nu_{qy} - 1)^2 + (b_{qy} - 1)^2}, \nonumber
\end{equation}
\end{linenomath*}
where $r_{qy}$ is the \emph{sample} Pearson correlation coefficient $r$ of measured and simulated data and scalars $\nu_{qy} = s_{q} / s_{y}$ and $b_{qy} = m_{q}/m_{y}$ are so-called variability and bias ratios, and $s_{x}$ is the sample standard deviation
\begin{linenomath*}
\begin{align}
s_{x} = \sqrt{\ltfrac{1}{n-1}\sum\nolimits_{t=1}^{n}(x_{i} - m_{x})^{2}}. \nonumber 
\end{align}    
\end{linenomath*}
To use KGE in gradient descent, we must turn this reward-based metric into a cost function
\begin{linenomath*}
\begin{align}
\mathcal{L}_\mathrm{kge}(\boldsymbol{\uptheta}) & = 1 - \mathrm{KGE}(\mathbf{y}_{n}, \mathbf{q}_{n}) = \sqrt{(r_{qy}-1)^2 + (\nu_{qy} - 1)^2 + (b_{qy} - 1)^2}.
\label{eq:L_kge(uptheta)}
\end{align}
\end{linenomath*}
Differentiating $\mathcal{L}_\mathrm{kge}$ w.r.t.\ $q_{t}$ yields
\begin{linenomath*}
\begin{align}
\frac{\partial \mathcal{L}_\mathrm{kge}(\boldsymbol{\uptheta})}{\partial q_{t}} & = 
\frac{1}{\mathcal{L}_\mathrm{kge}(\boldsymbol{\uptheta})}\biggl[(r_{qy}-1)\frac{\partial r_{qy}}{\partial q_{t}} + (\nu_{qy} - 1) \frac{\partial \nu_{qy}}{\partial q_{t}} + (b_{qy} - 1) \frac{\partial b_{qy}}{\partial q_{t}}\biggr], \nonumber
\end{align}
\end{linenomath*}
where $\partial r_{qy}/\partial q_{t}$,
$\partial \nu_{qy}/\partial q_{t}$ and $\partial b_{qy}/\partial q_{t}$ can be derived using standard variance and covariance calculus (see Appendix \ref{sec:AppendixD}). We can collect the individual derivatives in a $n \times 1$ vector $\partial \mathcal{L}_\mathrm{kge}(\boldsymbol{\uptheta}) / \partial \mathbf{q}_{n}$ as follows
\begin{linenomath*}
\begin{align}
\boldsymbol{\updelta}_{n,\mathrm{kge}}(\boldsymbol{\uptheta}) = \frac{\partial \mathcal{L}_\mathrm{kge}(\boldsymbol{\uptheta})}{\partial \mathbf{q}_{n}} & = \left( \frac{\partial \mathcal{L}_\mathrm{kge}(\boldsymbol{\uptheta})}{\partial q_{1}}, \dots, \frac{\partial \mathcal{L}_\mathrm{kge}(\boldsymbol{\uptheta})}{\partial q_{n}} \right)^\top. \nonumber
\end{align}
\end{linenomath*}
This leaves us with the gradient of $\mathcal{L}_\mathrm{kge}$ with respect to the parameters
\begin{linenomath*}
\begin{align}
\mathbf{g}_{n,\mathrm{kge}}(\boldsymbol{\uptheta}) & = \frac{\partial \mathbf{q}_{n}}{\partial \boldsymbol{\uptheta}^{\top}} \frac{\partial \mathcal{L}_\mathrm{kge}(\boldsymbol{\uptheta})}{\partial \mathbf{q}_{n}} = \mathbf{J}^{\top}_{q}(\boldsymbol{\uptheta})\boldsymbol{\updelta}_{n,\mathrm{kge}}(\boldsymbol{\uptheta}). \nonumber
\end{align}
\end{linenomath*}

Then, a final note about the GLS loss function, $\mathcal{L}_\mathrm{gls}(\boldsymbol{\uptheta})$. 
Ordinary least squares (OLS) is recovered as a special case when the discharge measurement errors are assumed independent and homoscedastic, for which $\boldsymbol{\Sigma}_{\epsilon} = \sigma^{2}\mathbf{I}_{n}$. Then, the $n \times n$ weight matrix, $\mathbf{W}_{n} = \boldsymbol{\Sigma}_{\epsilon}^{-1/2}$ is equal to the identity matrix $\mathbf{I}_{n}$. For weighted least squares (WLS), $\boldsymbol{\Sigma}_{\epsilon}$ is diagonal with entries $\sigma^{2}_{t}$, resulting in a diagonal weight matrix with elements $w_{t,t} = 1/\sigma_{t}$, a formulation commonly used in hydrology to accommodate heteroscedastic discharge measurement errors \citep{sorooshian1980}. In the most GLS setting, $\boldsymbol{\Sigma}_{\epsilon}$ is a full covariance matrix, and $\mathbf{W}_{n}$ is therefore a full symmetric matrix that accounts simultaneously for heteroscedasticity and temporal autocorrelation in the discharge errors. This completes our discussion of the score vectors associated with the loss functions considered in this work.

\subsection{Anatomy of the gradient vector \texorpdfstring{$\mathbf{g}_{n}(\boldsymbol{\uptheta})$}{}}\label{subsec:gn_anatomy}
The loss function $\mathcal{L}(\boldsymbol{\uptheta})$ acts as a \emph{governing mechanism} that regulates how discrepancies between observations and simulations are translated into parameter updates, thereby controlling the trade-off between fidelity to the data, numerical stability, and robustness to outliers. The associated gradient vector $\mathbf{g}_{n}(\boldsymbol{\uptheta})$ encodes this regulation quantitatively. It determines both the direction and magnitude of parameter adjustments $\Delta \boldsymbol{\uptheta}$ by combining the sensitivity of the model outputs to the parameters ($= \mathbf{J}_{q}(\boldsymbol{\uptheta}) \in \mathbb{R}^{n \times d}$) with the sensitivity of the loss function to those outputs ($= \boldsymbol{\updelta}_{n} \in \mathbb{R}^{n \times 1}$). In this way, residuals influence the estimator only insofar as they are filtered through the structure of the loss function and the underlying model sensitivities, making the gradient the central object linking model physics, statistical assumptions, and optimization behavior. 

Before we test our analytic gradient expressions for the sum of absolute residuals, generalized least squares, NSE-based, KGE-based, Huber, and FDC-based loss functions, it is important to clarify what can be expected from these gradients in practice. The different loss functions operate on very different numerical scales, and this is directly reflected in the magnitude of their gradients. Squared residual-based losses, such as the generalized least squares objective $\mathcal{L}_\mathrm{gls}(\boldsymbol{\uptheta})$ in Equation~\ref{eqApp:L_GLS(uptheta)}, produce gradients proportional to the size of the residuals. As a result, when model-data misfit is large, the associated gradients are also large, supporting rapid descent in parameter space. In contrast, losses based on flow duration curve discrepancies and efficiency metrics such as NSE and KGE compress residual information through low-order moments and dependence measures, attenuating large residuals and yielding substantially smaller gradient magnitudes in the typical range of $10^{-3}$-$10^{-6}$. 

Small gradients have several undesirable side effects:
\begin{enumerate}
\item Because the parameter update $\Delta \boldsymbol{\uptheta}_{(k)}$ scales with gradient magnitude, weak sensitivities lead to slow movement through parameter space and gradual convergence of gradient-based methods. Adaptive step-size control, such as line search or damping strategies, is therefore required to maintain reasonable convergence rates.

\item Weak sensitivities are also more difficult to characterize accurately when the underlying ODE solution is affected by numerical error. In such cases, gradients computed via finite differences may become dominated by truncation and floating-point noise. This contamination introduces spurious variability in successive gradient estimates, degrades curvature information used by quasi-Newton methods, and can impair search efficiency. These effects are particularly pronounced in conceptual hydrologic models, where threshold behavior and saturating fluxes give rise to extended regions of low sensitivity.
\end{enumerate}
Analytic gradients avoid these problems entirely as they remain accurate in flat regions of the loss surface and faithfully represent genuine insensitivity rather than numerical noise. Consequently, analytic differentiation not only accelerates optimization in terms of CPU time, but should also improve search robustness and reliability across a wide range of loss functions, parameter regimes, and hydrologic conditions, particularly when combined with second-order or quasi-Newton optimization methods.

\section{Model Parameterization}\label{sec:parameterization}
The parameters $\boldsymbol{\uptheta} = (\theta_{1},\ldots,\theta_{d})^{\top}$ of conceptual hydrologic models represent storage capacities, percolation coefficients, depletion and recession constants, areal fractions, and runoff and evaporation coefficients. These quantities have different physical units and span several orders of magnitude, which can make derivative-based optimization in the original bounded parameter space ill-conditioned and numerically unstable. To address this, we map the physical domain onto an unconstrained space. Next, we discuss this reparameterization method and analyze its impact on the Jacobian matrix $\mathbf{J}_{q}(\boldsymbol{\uptheta})$.  

\subsection{Parameter transformations}
\begin{figure}[H]
\centering\includegraphics[width=1\linewidth]{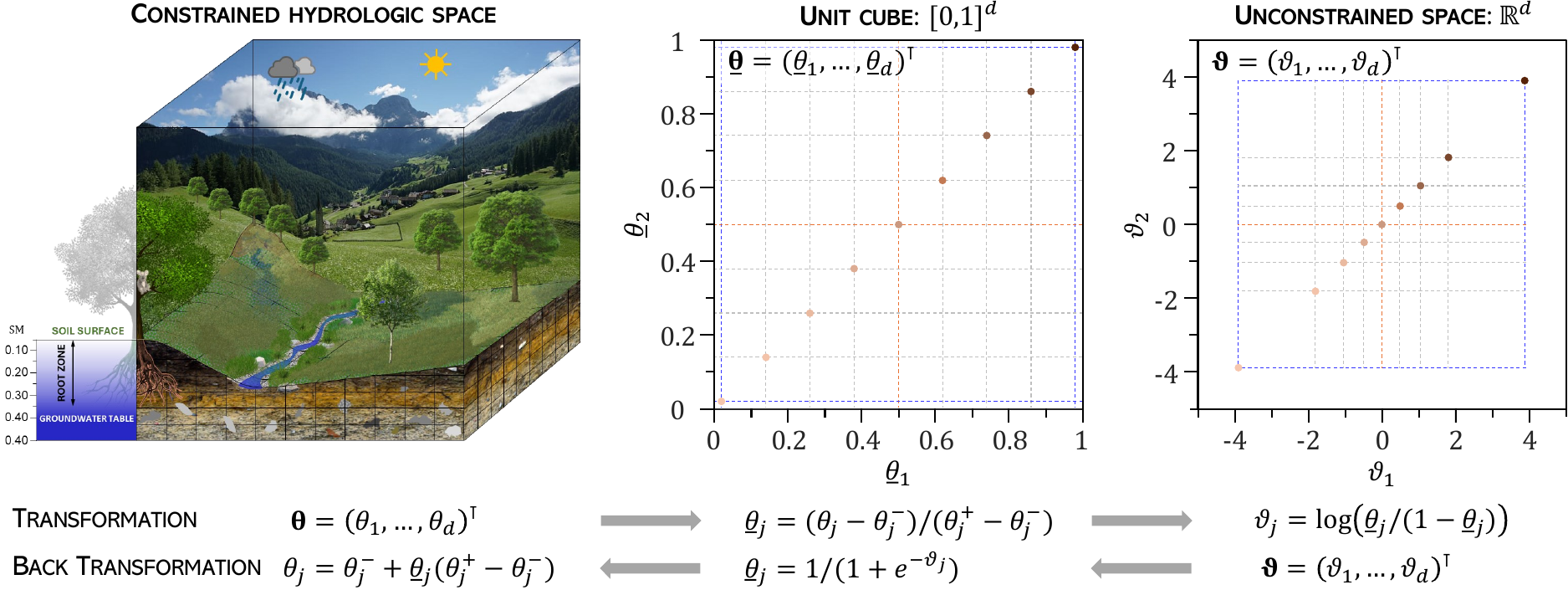}
\caption{Schematic illustration of the parameter transformations used for differentiable hydrologic modeling. The left panel depicts a watershed together with its original, physically interpretable hydrologic parameters $\boldsymbol{\uptheta} = (\theta_{1},\ldots,\theta_{d})^{\top}$. These parameters are first mapped to the unit hypercube, producing $\underline{\boldsymbol{\uptheta}} = (\underline{\theta}_{1},\ldots,\underline{\theta}_{d})^{\top} \in (0,1)^{d}$, shown in the center panel. The right panel displays the corresponding unconstrained parameter vector $\boldsymbol{\upvartheta} = (\vartheta_{1},\ldots,\vartheta_{d})^{\top} \in \mathbb{R}^{d}$, obtained via the logit transformation $\underline{\theta}_{j} = 1/\{(1 + \exp(-\vartheta_{j})\}$. The inverse mapping, $\vartheta_{j} = \log(\underline{\theta}_{j}/\{1 -\underline{\theta}_{j}\})$, returns parameters from the unconstrained space to the unit cube $\mathbb{U}^{d}$. This reparameterization eliminates boundary constraints yet maintains a smooth, invertible mapping to the physical hydrologic parameter space. As a result, gradient-based calibration becomes both more efficient and more robust.}
\label{fig:1}
\end{figure}

The transformation steps are summarized in Figure~\ref{fig:1}.
To stabilize the optimization problem, we first map the admissible parameter domain to the unit hypercube using the affine reparameterization
\begin{linenomath*}
\begin{align}
\theta_{j} & = \theta^{-}_{j} + \underline{\theta}_{j}\,(\theta^{+}_{j} - \theta^{-}_{j}), \qquad j = 1,\ldots,d,
\label{eq:theta_unnormalize}
\end{align}
\end{linenomath*}
where $\theta^{-}_{j}$ and $\theta^{+}_{j}$ are the physically plausible lower and upper bounds (Tables~\ref{tableApp:B1}--\ref{tableApp:B4}), and the normalized parameters $\underline{\boldsymbol{\uptheta}} = (\underline{\theta}_{1},\ldots,\underline{\theta}_{d})^{\top}$ are constrained to the $d$-dimensional unit cube
\begin{linenomath*}
\begin{align}
\underline{\boldsymbol{\uptheta}} \in \mathbb{U}^{d} = [0,1]^{d}. \nonumber
\end{align}
\end{linenomath*}
The Jacobian of this transformation is diagonal and given element-wise by
\begin{linenomath*}
\begin{align}
\frac{\mathrm{d}\theta_{j}}{\mathrm{d}\underline{\theta}_{j}} & = \theta_{j}^{+} - \theta_{j}^{-}.
\label{eq:dtheta_j/dtheta_underj}
\end{align}
\end{linenomath*}

To remove the explicit box constraints for gradient-based parameter estimation, we introduce an unconstrained parameterization via a smooth and bijective mapping $\tau : \mathbb{R}^{d} \to \mathbb{U}^{d}$. Each normalized component $\underline{\theta}_{j} \in [0,1]$ is written as the logistic transformation of an unconstrained value $\vartheta_{j} \in \mathbb{R}$
\begin{linenomath*}
\begin{align}
\underline{\theta}_{j} & = \tau(\vartheta_{j})  = \frac{1}{1 + \exp(-\vartheta_{j})}, \qquad j = 1,\ldots,d. \nonumber 
\end{align}
\end{linenomath*}
and, thus, $\underline{\boldsymbol{\uptheta}} = \tau(\boldsymbol{\upvartheta}) =
\bigl(\tau(\vartheta_{1}),\ldots,\tau(\vartheta_{d})\bigr)^{\top} \in (0,1)^{d}$, where the mapping is applied componentwise. The Jacobian of the logistic transformation is diagonal and given element-wise by
\begin{linenomath*}
\begin{align}
\frac{\mathrm{d}\underline{\theta}_{j}}{\mathrm{d}\vartheta_{j}} & = \underline{\theta}_{j}(1 - \underline{\theta}_{j}).
\label{eq:dundertheta_j/dvartheta_j}
\end{align}
\end{linenomath*}
The overall transformation is therefore a smooth, monotone and bijective mapping
\begin{linenomath*}
\begin{align}
\boldsymbol{\upvartheta} \in \mathbb{R}^{d} \quad \Longrightarrow\quad
\underline{\boldsymbol{\uptheta}} \in (0,1)^{d} \quad \Longrightarrow \quad
\boldsymbol{\uptheta} \in [\theta^{-}_{1},\theta^{+}_{1}] \times \cdots \times[\theta^{-}_{d}, \theta^{+}_{d}]. \nonumber
\end{align}
\end{linenomath*}

The bound constraints no longer need to be enforced by projection or clipping, and any unconstrained gradient-based optimization method (gradient descent, Adam, quasi-Newton, etc.) may be employed on $\boldsymbol{\upvartheta}$. A limitation is that if the optimal solution lies on the boundary $(\theta_{j} = 0 \text{ or } \theta_{j} = 1)$, the logistic map approaches the boundary only asymptotically, and the gradients may become small near the edges of the unit hypercube.

\subsection{Jacobian matrices}
The reparametrization steps yield three expressions for the Jacobian of simulated discharge $q_{1}, \ldots, q_{n}$, namely, $\mathbf{J}_{q}(\boldsymbol{\upvartheta}) = \partial \mathbf{q}_{n}/\partial \boldsymbol{\upvartheta}^{\top}$, $\mathbf{J}_{q}(\underline{\boldsymbol{\uptheta}}) = \partial \mathbf{q}_{n}/\partial \underline{\boldsymbol{\uptheta}}^{\top}$ and $\mathbf{J}_{q}(\boldsymbol{\uptheta}) = \partial \mathbf{q}_{n}/\partial \boldsymbol{\uptheta}^{\top}$, where $\boldsymbol{\upvartheta}$ (unconstrained), $\underline{\boldsymbol{\uptheta}}$ (unit cube) and $\boldsymbol{\uptheta}$ (original domain) were defined previously. Next, we examine the relationships between the three Jacobian matrices and their associated gradients $\mathbf{g}_{n}(\boldsymbol{\upvartheta})$, $\mathbf{g}_{n}(\underline{\boldsymbol{\uptheta}})$ and $\mathbf{g}_{n}(\boldsymbol{\uptheta})$.

The outcome of FSA in Section~\ref{sec:theory} is the Jacobian of the simulated discharge with respect to the hydrologic parameter values $\boldsymbol{\uptheta}$
\begin{linenomath*}
\begin{align}
\mathbf{J}_{q}(\boldsymbol{\uptheta}) &
= \frac{\partial \mathbf{q}_{n}(\boldsymbol{\uptheta})}{\partial \boldsymbol{\uptheta}^{\top}} = \nabla_{\boldsymbol{\uptheta}}\mathbf{q}_{n}(\boldsymbol{\uptheta}).
\nonumber
\end{align}
\end{linenomath*}
If finite-differencing is applied to the unconstrained parameters $\vartheta_{1},\ldots,\vartheta_{d}$, then we yield 
\begin{linenomath*}
\begin{align}
\mathbf{J}_{q}(\boldsymbol{\upvartheta})
& = \frac{\partial \mathbf{q}_{n} (\boldsymbol{\upvartheta})}{\partial \boldsymbol{\upvartheta}^{\top}} = \nabla_{\boldsymbol{\upvartheta}}\mathbf{q}_{n}(\boldsymbol{\upvartheta}).
\nonumber
\end{align}
\end{linenomath*}
These two Jacobian matrices, $\mathbf{J}_{q}(\boldsymbol{\uptheta})$ and $\mathbf{J}_{q}(\boldsymbol{\upvartheta})$, are not equivalent because the discharge sensitivities are taken with respect to different parameterizations. The chain rule establishes the relationship between the two Jacobians 
\begin{linenomath*}
\begin{align}
\frac{\partial \mathbf{q}_{n}}{\partial \vartheta_{j}} & = \frac{\partial \mathbf{q}_{n}}{\partial \theta_{j}}
\frac{\mathrm{d}\theta_{j}}{\mathrm{d}\vartheta_{j}},
\label{eq:dq_n/dvartheta_j}
\end{align}
\end{linenomath*}
or in a more compact form 
\begin{linenomath*}
\begin{align}
\bigl[\mathbf{J}_{q}(\boldsymbol{\upvartheta}) \bigr]_{\bullet,j} & = \frac{\mathrm{d}\theta_{j}}{\mathrm{d}\vartheta_{j}}\bigl[\mathbf{J}_{q}(\boldsymbol{\uptheta}) \bigr]_{\bullet,j}, \nonumber
\end{align}
\end{linenomath*}
where, as before, the subscript ``${\bullet,j}$'' signifies the $j$th column of matrix $\mathbf{J}_{q}$. Variables $\boldsymbol{\uptheta}$ and $\boldsymbol{\upvartheta}$ are related through the intermediate variable $\underline{\theta}_{j} \in [0,1]$ using smooth one-to-one transformations. Indeed, we can write
\begin{linenomath*}
\begin{align}
\frac{\mathrm{d}\theta_{j}}{\mathrm{d}\vartheta_{j}} & = \frac{\mathrm{d}\theta_{j}}{\mathrm{d}\underline{\theta}_{j}}\frac{\mathrm{d}\underline{\theta}_{j}}{\mathrm{d}\vartheta_{j}}. \nonumber
\end{align}
\end{linenomath*}
If we substitute this expression into Equation \ref{eq:dq_n/dvartheta_j} we end up with
\begin{linenomath*}
\begin{align}
\frac{\partial \mathbf{q}_{n}}{\partial \vartheta_{j}} & = \frac{\partial \mathbf{q}_{n}}{\partial \theta_{j}}\frac{\mathrm{d}\theta_{j}}{\mathrm{d}\underline{\theta}_{j}}\frac{\mathrm{d}\underline{\theta}_{j}}{\mathrm{d}\vartheta_{j}}. \nonumber
\end{align}
\end{linenomath*}
where the Jacobians of the intermediate transformations are given in Equations \ref{eq:dtheta_j/dtheta_underj} and \ref{eq:dundertheta_j/dvartheta_j}
\begin{linenomath*}
\begin{align}
\frac{\partial \mathbf{q}_{n}}{\partial \vartheta_{j}} & = \frac{\partial \mathbf{q}_{n}}{\partial \theta_{j}}(\theta^{+}_{j} - \theta^{-1}_{j})\underline{\theta}_{j}(1 - \underline{\theta}_{j}).
\end{align}
\end{linenomath*}
Equivalently, we admit the affine transformation  $\underline{\theta}_{j} = (\theta_{j} - \theta^{-}_{j})/(\theta^{+}_{j} - \theta^{-}_{j})$ and yield
\begin{linenomath*}
\begin{align}
\frac{\partial \mathbf{q}_{n}}{\partial \vartheta_{j}} & = \frac{\partial \mathbf{q}_{n}}{\partial \theta_{j}}(\theta_{j} - \theta^{-}_{j})\Biggl(\frac{\theta^{+}_{j} - \theta_{j}}{\theta^{+}_{j} - \theta^{-}_{j}}\Biggr). \nonumber
\end{align}
\end{linenomath*}
Thus, the Jacobian $\mathbf{J}_{q}(\boldsymbol{\upvartheta})$ in the unconstrained space $\boldsymbol{\upvartheta} = (\vartheta_{1},\ldots,\vartheta_{d})^{\top}$ follows directly from its counterpart $\mathbf{J}_{q}(\boldsymbol{\uptheta})$ in the original bounded parameter space $\boldsymbol{\uptheta} = (\theta_{1},\ldots,\theta_{d})^{\top}$ by the diagonal transformation
\begin{linenomath*}
\begin{align}
\mathbf{J}_{q}(\boldsymbol{\upvartheta})
& = \mathbf{J}_{q}(\boldsymbol{\uptheta})
\Diag\biggl(\frac{\mathrm{d}\theta_{1}}{\mathrm{d}\underline{\theta}_{1}}\frac{\mathrm{d}\underline{\theta}_{1}}{\mathrm{d}\vartheta_{1}},
\ldots, \frac{\mathrm{d}\theta_{d}} {\mathrm{d}\underline{\theta}_{d}}\frac{\mathrm{d}\underline{\theta}_{d}}{\mathrm{d}\vartheta_{d}} \biggr).
\end{align}
\end{linenomath*}
Following Equation \ref{eq:g_n(uptheta)_key}, this also clarifies relationships between $\mathbf{g}_{n}(\boldsymbol{\upvartheta})$, $\mathbf{g}_{n}(\underline{\boldsymbol{\uptheta}})$ and $\mathbf{g}_{n}(\boldsymbol{\uptheta})$. 

\section{Numerical and automatic differentiation}\label{sec:num_and_automatic_diff}
Once the augmented ODE system of Equation \ref{eq:ode45_aug_system} is implemented for a given watershed model, together with analytic expressions for the state and parameter Jacobians $\mathbf{J}_{f}(\mathbf{x})$ and $\mathbf{J}_{f}(\boldsymbol{\uptheta})$, a sufficiently accurate numerical integration yields the exact Jacobian $\mathbf{J}_{q}(\boldsymbol{\uptheta})$ of simulated discharge with respect to the model parameters $\theta_{1},\ldots,\theta_{d}$. The corresponding gradient vectors $\mathbf{g}_{n}(\boldsymbol{\uptheta})$ then follow directly from Equation \ref{eq:g_n(uptheta)_key}.

Strictly speaking, no additional validation is required: the sensitivities are analytic by construction and therefore represent the true derivatives of the model equations. Nevertheless, to build confidence in the proposed framework and to quantify its numerical behavior in practice, we compare the analytic Jacobians and gradient vectors against their counterparts obtained via numerical finite differences and automatic differentiation. These comparisons are carried out for four widely used conceptual hydrologic models, \texttt{hymod}, \texttt{hmodel}, \texttt{sacsma}, and \texttt{xinanjiang}, and across multiple datasets using the six loss functions of Table~\ref{table:1}. 

\subsection{Numerical differentiation}
\begin{figure}[H]
\centering
\includegraphics[width=0.8\linewidth]{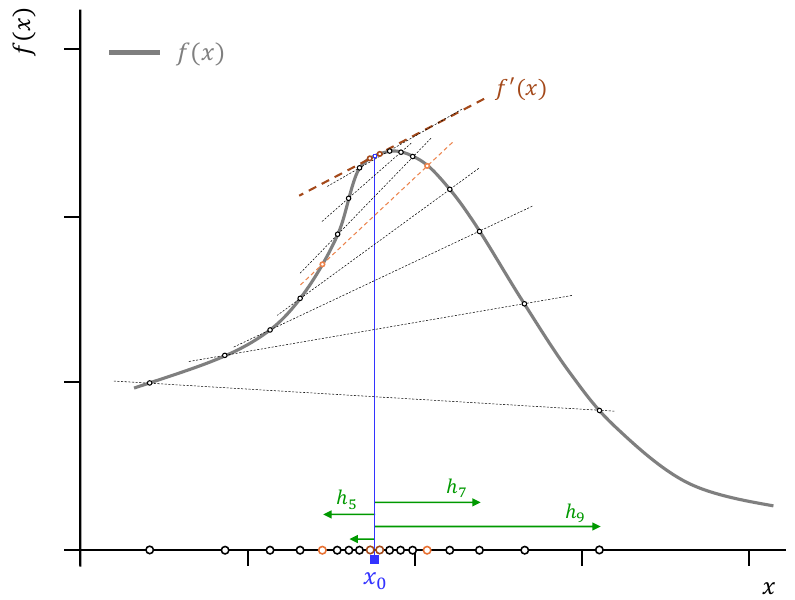}
\caption{Schematic illustration of Richardson extrapolation for numerical differentiation. Starting from an anchor point $x_{0}$ (blue square), the function $f(x)$ (gray curve) is evaluated at a sequence of logarithmically spaced perturbations $x_{0} \pm h_{j}$ (black markers) using a central differencing scheme. Each perturbation pair yields a finite-difference approximation of the derivative, $f'(x_{0}) \approx \frac{1}{2}h^{-1}_{j}\{f(x_{0} + h_{j}) - f(x_{0}-h_{j})\}$ at $x_{0}$, illustrated with the colored secant lines for step sizes $h_{1}$ and $h_{5}$. The derivative estimates for the progressively smaller step sizes $h_{j}$, are recursively combined using multi-term Richardson (Romberg) extrapolation to cancel leading-order truncation errors. This procedure provides a high-order approximation of the derivative in the limit $h \rightarrow 0$, indicated by the brown line.}
\label{fig:2}
\end{figure}

The Jacobian matrix $\mathbf{J}_{q}(\boldsymbol{\upvartheta}) = \nabla_{\boldsymbol{\upvartheta}}\mathbf{q}_{n}(\boldsymbol{\upvartheta}) \in \mathbb{R}^{n\times d}$ of first-order partial derivatives of simulated discharge $q_{1},\ldots,q_{n}$ with respect to the unconstrained parameters $\vartheta_{1},\ldots,\vartheta_{d}$, and the $d \times 1$ gradient vector $\mathbf{g}(\boldsymbol{\upvartheta}) = \nabla_{\boldsymbol{\upvartheta}} \mathcal{L}(\boldsymbol{\upvartheta})$, of the total loss $\mathcal{L} = \sum_{t=1}^{n}\mathcal{L}_{t}(y_{t},q_{t})$ with respect to $\boldsymbol{\upvartheta}$ can be determined by numerical means using repeated evaluation of $q_{1}(\boldsymbol{\upvartheta}),\ldots,q_{n}(\boldsymbol{\upvartheta})$ and $\mathcal{L}(\boldsymbol{\upvartheta})$ in the neighborhood of anchor point $\boldsymbol{\upvartheta}$. We use the \texttt{DERIVESTsuite} toolbox of \citet{derrico2024}, a \textsc{Matlab} collection of fully adaptive finite-difference methods, to obtain numerical reference derivatives of the simulated discharge $q_{1}, \ldots, q_{n}$ and the loss function $\mathcal{L}(\boldsymbol{\upvartheta})$ with respect to the unconstrained parameters $\vartheta_{1},\ldots,\vartheta_{d}$.
\texttt{DERIVESTsuite} combines a high-order central difference scheme with multi-term Romberg extrapolation and an automatic step-size selection based on proportionally cascading perturbations around the anchor point $\boldsymbol{\upvartheta}$ (see  Figure \ref{fig:2}). 
This approach yields accurate numerical approximations of $\mathbf{J}_{q}(\boldsymbol{\upvartheta}) \in \mathbb{R}^{n \times d}$ and $\mathbf{g}(\boldsymbol{\upvartheta}) \in \mathbb{R}^{d \times 1}$, but at the expense of a significant computational cost. A much more efficient procedure is to use fixed-step finite differences, in which each parameter is perturbed by a prescribed absolute step (e.g., $h = 10^{-3}$) or a relative step (e.g., $h = 10^{-2}\,\lvert \vartheta_{j} \rvert$).

\subsection{Automatic differentiation}
A third approach to computing sensitivities, besides analytic and numerical differentiation, is \emph{automatic differentiation}. Automatic differentiation (AD) exploits the fact that every computer program implements a sequence of elementary operations each of which has a known derivative. By applying the chain rule mechanically and exactly throughout the computational graph, AD produces derivatives with machine precision accuracy at a computational cost that is typically within a small constant factor of the model evaluation itself \citep{griewank2008, baydin2018, naumann2011}. Unlike finite differences, AD does not suffer from subtractive cancellation or step-size truncation errors, and unlike symbolic differentiation it does not require closed-form expressions or manual algebra.

Modern software systems provide robust AD implementations, including JAX \citep{jax2018}, PyTorch \citep{paszke2019}, and TensorFlow \citep{abadi2016}. \textsc{Matlab} has recently introduced experimental AD capabilities through the \texttt{differentiation} and \texttt{dlgradient} functions in the Deep Learning Toolbox, enabling reverse-mode differentiation of \textsc{Matlab} functions and models written in \texttt{dlarray} form. These tools provide a convenient reference for evaluating the accuracy and computational efficiency of our analytic sensitivity framework.


\section{Case Studies}\label{sec:application}
We illustrate the performance of analytic differentiation using hydrologic data from four watersheds spanning a range of hydro-climatic conditions and temporal resolutions. Daily precipitation, potential evaporation, and streamflow observations were obtained from (i) the French Broad River at Asheville, North Carolina (drainage area $\approx 2,450$ km\textsuperscript{2}), and (ii) the Leaf River near Collins, Mississippi ($\approx 1,944$ km\textsuperscript{2}). In addition, we use hourly data from the Plynlimon experimental catchments in mid-Wales, UK, which comprise about 20 km\textsuperscript{2} of the headwaters of the (iii) Wye and (iv) Severn rivers. The Wye catchment is primarily grassland, whereas the Severn catchment was dominated by conifer plantations during the 1992-1996 period analyzed here \citep{kirchner2006}.

\subsection{Jacobian matrix}
Figure~\ref{fig:3} compares analytic (solid black lines) and numerical (red squares) entries of the Jacobian matrix $\mathbf{J}_{q}(\boldsymbol{\uptheta})$ for hourly streamflow simulated with the \texttt{xinanjiang} model. Results are shown for 12 of the 14 model parameters using a parameter vector $\boldsymbol{\uptheta}$ sampled randomly from the admissible ranges in Table~\ref{tableApp:B4}. 
\begin{landscape}
\centering
\begin{figure}[H]
\includegraphics[width=1\linewidth]{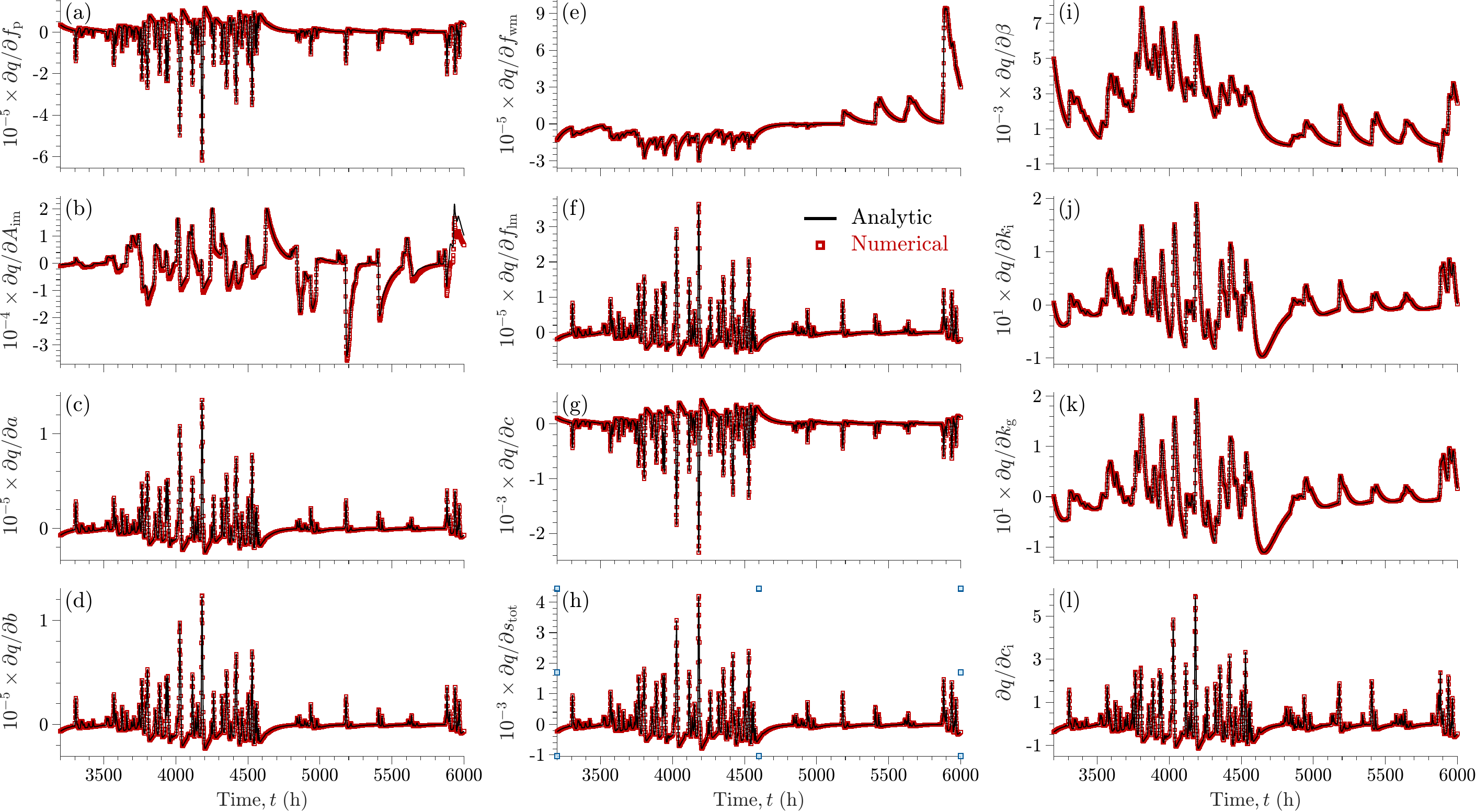}
\caption{Sensitivity of simulated streamflow $q_{1}, \ldots, q_{n}$ to \texttt{xinanjiang} parameters (a) $f_\mathrm{p}$, (b) $A_\mathrm{im}$, (c) $a$, (d) $b$, (e) $f_\mathrm{wm}$, (f) $f_\mathrm{lm}$, (g) $c$, (h) $s_\mathrm{tot}$, (i) $\beta$, (j) $k_\mathrm{i}$, (k) $k_\mathrm{g}$, and (l) $c_\mathrm{i}$ for the Wye river, UK. The entries of the different parameters make up the Jacobian matrix $\mathbf{J}_{q}(\boldsymbol{\uptheta}) = \partial \mathbf{q}_{n}/\partial \boldsymbol{\uptheta}^{\top}$.}
\label{fig:3}
\end{figure}
\end{landscape}
The two Jacobians are virtually indistinguishable, demonstrating excellent agreement between analytic sensitivities obtained via forward sensitivity analysis and finite‐difference approximations. The individual sensitivity trajectories exhibit pronounced temporal variability, reflecting the strong nonlinearity and time-varying dominance of different hydrologic processes. Importantly, non-zero sensitivities are preserved for parameters that act upstream in the model structure and are far removed from the final routing reservoir, confirming that the analytic backpropagation of sensitivities through the augmented ODE system is exact and does not suffer from vanishing or loss of information across interconnected storage elements.

\begin{figure}[H]
\centering\includegraphics[width=1\linewidth]{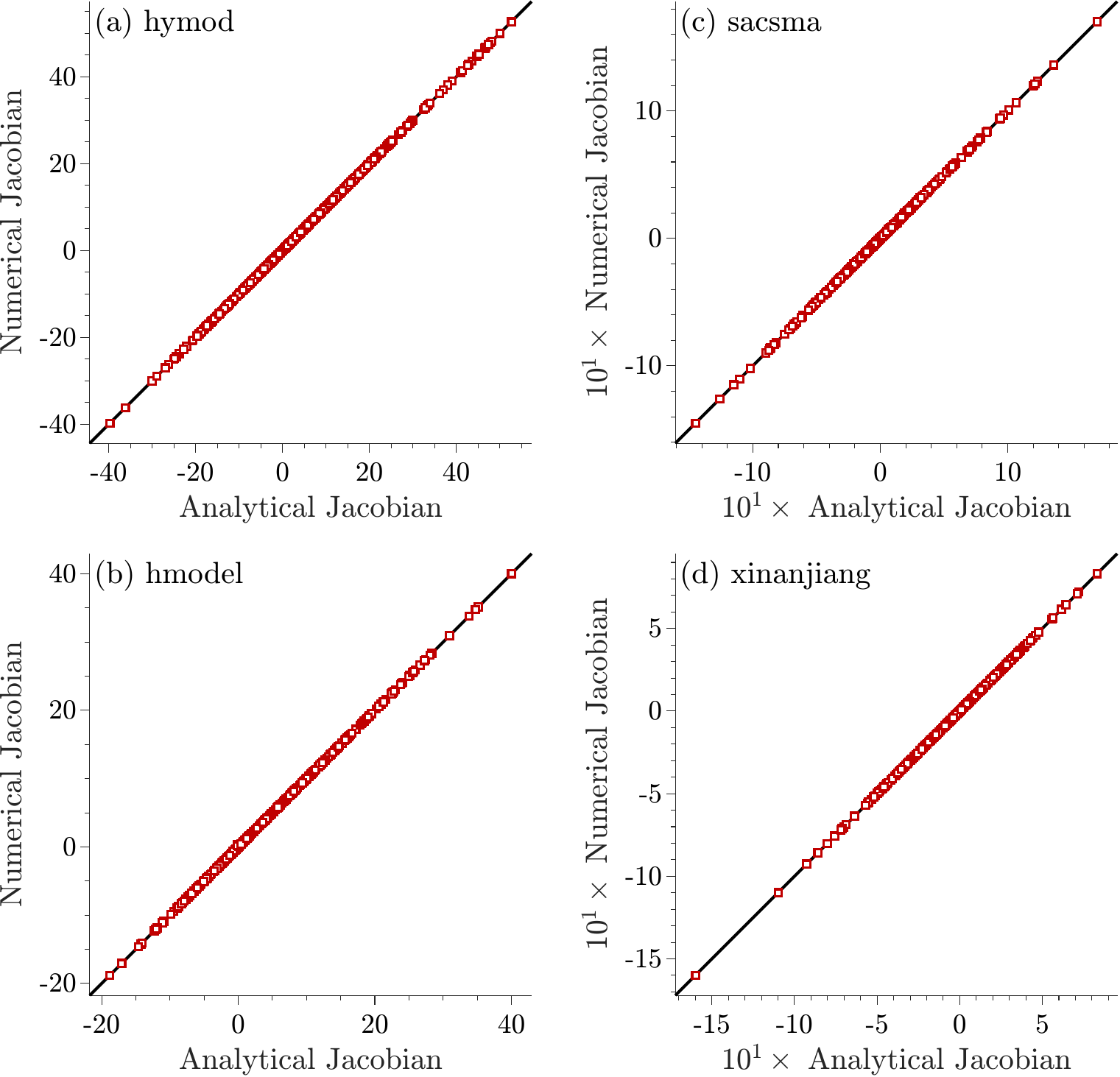}
\caption{Comparison of analytic and numerical Jacobian entries for (a) \texttt{hymod} (
$786,640$ entries), (b) \texttt{hmodel} (
$1,101,296$ entries), (c) \texttt{sacsma} ($2,045,264$ entries), and (d) \texttt{xinanjiang} ($2,202,592$ entries) based on $N = 10$ randomly sampled parameter vectors across the Leaf River, French Broad, Wye, and Severn catchments. In all cases, points fall tightly along the 1:1 line, demonstrating near-perfect agreement between analytic sensitivities and their numerical finite-difference counterparts computed using the \texttt{DERIVESTsuite} toolbox of \citet{derrico2024}.}
\label{fig:4}
\end{figure}

We do not present Jacobian trace plots for the parameters of the other models, as these would merely reproduce the same qualitative findings. Instead, we repeat the above analysis for $N = 10$ random parameter vectors $\underline{\boldsymbol{\uptheta}}_{(1)},\ldots, \underline{\boldsymbol{\uptheta}}_{(N)}$ drawn from the unit hypercube and collect all entries of their corresponding Jacobian matrices into a single vector for the Leaf River, French Broad, Wye, and Severn catchments. This aggregation mixes parameters, catchments, and temporal resolutions (daily and hourly), thereby providing a comprehensive and model-agnostic comparison of analytic and numerical sensitivities. Figure \ref{fig:4} summarizes the results of this experiment and displays scatter plots comparing the numerical and analytic Jacobian entries for   \texttt{hymod}, \texttt{hmodel},  \texttt{sacsma}, and \texttt{xinanjiang}. 

Across all models, catchments, and temporal resolutions, the analytic Jacobian entries align almost perfectly with their numerical counterparts, yielding points that fall tightly on the 1:1 line. This agreement confirms both the correctness and robustness of our analytic sensitivities. In total, this comparison spans $786,640$ Jacobian entries for \texttt{hymod}, $1,101,296$ for \texttt{hmodel}, $2,045,264$ for \texttt{sacsma}, and $2,202,592$ for \texttt{xinanjiang}, demonstrating consistency across models of increasing dimensionality and complexity.


Having established near-perfect agreement between analytic and numerical Jacobians for all considered watershed models and parameter vectors, we now turn to the computational cost of sensitivity-based Jacobian evaluation. Table \ref{table:2} summarizes both the accuracy and efficiency of the analytic Jacobians, reporting (i) the mean absolute difference between analytic and numerical Jacobians and (ii) the corresponding CPU speed-up achieved by analytic differentiation across four conceptual models, multiple catchments, and daily and hourly time resolutions.
\begin{table}[H]
\centering
\captionsetup[table]{position=bottom}
\begin{threeparttable}  
\caption{Mean absolute difference $\Delta \mathbf{J}_{q} = \Mean(\lvert \mathbf{J}^\text{a}_{q}(\underline{\boldsymbol{\uptheta}}) - \mathbf{J}^\text{n}_{q}(\underline{\boldsymbol{\uptheta}}) \rvert)$ between analytic $\mathbf{J}^\text{a}_{q}(\underline{\boldsymbol{\uptheta}})$ and numerical $\mathbf{J}^\text{n}_{q}(\underline{\boldsymbol{\uptheta}})$ Jacobians and CPU speed-up achieved by analytic differentiation for four hydrologic models across daily and hourly datasets. Errors are averaged over all time steps and parameters.}
\begin{tabular}{clcccccc}
\toprule
Model & Dataset & Resolution & $d$ & $m$ & $\Delta \mathbf{J}_{q}$ 
& Speed up \\
\midrule
\multirow{4}{*}{\rotatebox{90}{\parbox{2.0cm}{\centering \texttt{hymod}}}} & Leaf river & daily & \multirow{4}{*}{5} & \multirow{4}{*}{6} & $2.231\cdot10^{-3}$ & $70\times$ \\ 
& French broad & daily &  &  & $3.097\cdot10^{-3}$ & $164\times$ \\ 
& Wye & hourly &  &  & $1.386\cdot10^{-6}$ & $158\times$ \\ 
& Severn & hourly &  &  & $8.936\cdot10^{-7}$ & $105\times$ \\ 
\midrule
\multirow{4}{*}{\rotatebox{90}{\parbox{2.0cm}{\centering \texttt{hmodel}}}} & Leaf river & daily & \multirow{4}{*}{7} & \multirow{4}{*}{5} & $3.066\cdot10^{-3}$ & $261\times$ \\ 
& French broad & daily &  &  & $2.129\cdot10^{-3}$ & $336\times$ \\ 
& Wye & hourly &  & & $1.252\cdot10^{-4}$ & $251\times$ \\ 
& Severn & hourly & & & $1.256\cdot10^{-4}$ & $239\times$ \\ 
\midrule
\multirow{4}{*}{\rotatebox{90}{\parbox{2.0cm}{\centering \texttt{sacsma}}}} & Leaf river & daily & \multirow{4}{*}{13} & \multirow{4}{*}{9}  & $1.060\cdot10^{-2}$ & $402\times$ \\ 
& French broad & daily &  &  & $1.209\cdot10^{-2}$ & $487\times$ \\ 
& Wye & hourly &  & & $1.022\cdot10^{-4}$ & $448\times$ \\ 
& Severn & hourly &  &  & $1.738\cdot10^{-4}$ & $394\times$ \\ 
\midrule
\multirow{4}{*}{\rotatebox{90}{\parbox{2.0cm}{\centering \texttt{xinan} \\[-2mm] \texttt{jiang}}}} & Leaf river & daily & \multirow{4}{*}{14} & \multirow{4}{*}{8} & $3.092\cdot10^{-4}$ & $86\times$ \\ 
& French broad & daily &  &  & $5.193\cdot10^{-4}$ & $387\times$ \\ 
& Wye & hourly &  &  & $9.453\cdot10^{-7}$ & $439\times$ \\ 
& Severn & hourly &  & & $6.411\cdot10^{-7}$ & $490\times$ \\ \bottomrule
\end{tabular}
\label{table:2}
\end{threeparttable}
\end{table}
The results demonstrate excellent numerical agreement between analytic and finite-difference Jacobians, with mean absolute differences typically ranging from $10^{-7}$ to $10^{-3}$ across all models and datasets. These discrepancies are several orders of magnitude smaller than the typical variability of the Jacobian entries themselves and are entirely attributable to truncation and round-off errors inherent to finite-difference approximations. In contrast, the analytic Jacobians are exact by construction and free of step-size sensitivity or numerical noise.

At the same time, analytic differentiation yields substantial and consistent computational gains, reducing CPU time by factors ranging from approximately $70\times$ to nearly $500\times$. These speed-ups are observed systematically for all models and for both daily and hourly datasets. Importantly, the magnitude of the speed-up does not exhibit a clear dependence on model complexity, parameter dimensionality $d$, or number of state variables $m$. While more pronounced trends might emerge in much larger Monte Carlo experiments, such scaling behavior is secondary to the central finding. Analytic differentiation delivers large, reliable, and model-agnostic reductions in computational cost without sacrificing accuracy.

Taken together, these results confirm that forward sensitivity analysis provides an efficient and numerically robust alternative to finite-difference Jacobian estimation. The combination of exact derivatives and dramatic CPU-time savings makes analytic Jacobians particularly well suited for gradient-based calibration, uncertainty quantification, and large-sample hydrologic applications where repeated Jacobian evaluations are unavoidable.

\subsection{Gradient vector}
We can now leverage the analytic discharge Jacobian $\mathbf{J}_{q}(\boldsymbol{\uptheta})$ and compute exact gradient vectors for the $\mathcal{L}_\mathrm{sar}(\boldsymbol{\uptheta})$, $\mathcal{L}_\mathrm{gls}(\boldsymbol{\uptheta})$, $\mathcal{L}_\mathrm{nse}(\boldsymbol{\uptheta})$, $\mathcal{L}_\mathrm{kge}(\boldsymbol{\uptheta})$, $\mathcal{L}_\mathrm{huber}(\boldsymbol{\uptheta})$ and $\mathcal{L}_\mathrm{fdc}(\boldsymbol{\uptheta})$ loss functions of Table~\ref{table:1} using Equations \ref{eqApp:g_nsar(uptheta)}, \ref{eqApp:g_ngls(uptheta)}, \ref{eqApp:g_nnse(uptheta)}, \ref{eqApp:g_nkge(uptheta)}, \ref{eqApp:g_nhuber(uptheta)} and \ref{eqApp:g_nfdc(uptheta)}, respectively. Table \ref{table:3} compares the analytic and numerical gradients of the sum of absolute residuals, sum of squared residuals, NSE-based, KGE-based, Huber and FDC loss functions for the \texttt{sacsma} model using three years of hourly Severn River data. Numerical gradients are computed from finite-differencing using the \texttt{DERIVESTsuite} toolbox. 
\begin{table}[!ht]
\centering
\captionsetup[table]{position=bottom}
\begin{threeparttable}  
\caption{Analytic and numerical gradient vectors of the sum of absolute residuals, squared residuals, NSE-based, KGE-based, Huber and FDC-based loss functions for the \texttt{sacsma} model using 3-years of hourly streamflow data for the Severn river, UK.}
\begin{tabular}{ >{\raggedright\arraybackslash}p{20mm} 
>{\centering\arraybackslash}p{16pt} 
>{\centering\arraybackslash}p{16pt}
>{\centering\arraybackslash}p{6pt}
>{\centering\arraybackslash}p{16pt}
>{\centering\arraybackslash}p{16pt} 
>{\centering\arraybackslash}p{6pt}
>{\centering\arraybackslash}p{16pt}
>{\centering\arraybackslash}p{16pt} 
>{\centering\arraybackslash}p{1pt} }
\toprule
\multirow{2}{*}{Parameter} & \multicolumn{2}{c}{$\mathbf{g}_{n,\mathrm{sar}}(\boldsymbol{\uptheta})$}
&& \multicolumn{2}{c}{$\mathbf{g}_{n,\mathrm{gls}}(\boldsymbol{\uptheta})$}
&& \multicolumn{2}{c}{$\mathbf{g}_{n,\mathrm{nse}}(\boldsymbol{\uptheta})$} \\[1mm]
\cline{2-3}\cline{5-6}\cline{8-9} \\[-4mm]
& \multicolumn{1}{c}{Analytic} & \multicolumn{1}{c}{Numeric} && \multicolumn{1}{c}{Analytic} & \multicolumn{1}{c}{Numeric} && \multicolumn{1}{c}{Analytic} & \multicolumn{1}{c}{Numeric} & \\
\midrule
$u_\mathrm{t,max}$ & \multicolumn{1}{r}{$2.7077$} & \multicolumn{1}{r}{$2.7083$} && \multicolumn{1}{r}{$1.9038$} & \multicolumn{1}{r}{$1.9033$} && \multicolumn{1}{r}{$0.0027$} & \multicolumn{1}{r}{$0.0027$} \\ 
$u_\mathrm{f,max}$ & \multicolumn{1}{r}{$110.39$} & \multicolumn{1}{r}{$108.15$} && \multicolumn{1}{r}{$24.145$} & \multicolumn{1}{r}{$25.182$} && \multicolumn{1}{r}{$0.0341$} & \multicolumn{1}{r}{$0.0355$} \\ 
$l_\mathrm{t,max}$ & \multicolumn{1}{r}{-$1.8764$} & \multicolumn{1}{r}{-$1.8745$} && \multicolumn{1}{r}{-$1.3254$} & \multicolumn{1}{r}{-$1.3251$} && \multicolumn{1}{r}{-$0.0019$} & \multicolumn{1}{r}{-$0.0019$} \\ 
$l_\mathrm{fp,max}$ & \multicolumn{1}{r}{-$3.9700$} & \multicolumn{1}{r}{-$3.9701$} && \multicolumn{1}{r}{-$2.8051$} & \multicolumn{1}{r}{-$2.8044$} && \multicolumn{1}{r}{-$0.0040$} & \multicolumn{1}{r}{-$0.0040$} \\ 
$l_\mathrm{fs,max}$ & \multicolumn{1}{r}{-$358.35$} & \multicolumn{1}{r}{-$358.32$} && \multicolumn{1}{r}{$78.814$} & \multicolumn{1}{r}{$78.813$} && \multicolumn{1}{r}{$0.1112$} & \multicolumn{1}{r}{$0.1112$} \\ 
$\alpha$ & \multicolumn{1}{r}{-$3.5858$} & \multicolumn{1}{r}{-$3.5862$} && \multicolumn{1}{r}{-$2.5468$} & \multicolumn{1}{r}{-$2.5458$} && \multicolumn{1}{r}{-$0.0036$} & \multicolumn{1}{r}{-$0.0036$} \\ 
$\psi$ & \multicolumn{1}{r}{$2.5257$} & \multicolumn{1}{r}{$2.5264$} && \multicolumn{1}{r}{$1.7766$} & \multicolumn{1}{r}{$1.7763$} && \multicolumn{1}{r}{$0.0025$} & \multicolumn{1}{r}{$0.0025$} \\ 
$k_\mathrm{i}$ & \multicolumn{1}{r}{-$0.6613$} & \multicolumn{1}{r}{-$0.6613$} && \multicolumn{1}{r}{-$0.4483$} & \multicolumn{1}{r}{-$0.4483$} && \multicolumn{1}{r}{-$0.0006$} & \multicolumn{1}{r}{-$0.0006$} \\ 
$\kappa$ & \multicolumn{1}{r}{$461.26$} & \multicolumn{1}{r}{$461.58$} && \multicolumn{1}{r}{$77.463$} & \multicolumn{1}{r}{$77.463$} && \multicolumn{1}{r}{$0.1093$} & \multicolumn{1}{r}{$0.1093$} \\ 
$\nu_\mathrm{p}$ & \multicolumn{1}{r}{-$66.316$} & \multicolumn{1}{r}{-$66.401$} && \multicolumn{1}{r}{-$58.956$} & \multicolumn{1}{r}{-$58.956$} && \multicolumn{1}{r}{-$0.0832$} & \multicolumn{1}{r}{-$0.0832$} \\ 
$\nu_\mathrm{s}$ & \multicolumn{1}{r}{-$91.804$} & \multicolumn{1}{r}{-$91.843$} && \multicolumn{1}{r}{-$60.338$} & \multicolumn{1}{r}{-$60.337$} && \multicolumn{1}{r}{-$0.0852$} & \multicolumn{1}{r}{-$0.0852$} \\ 
$a_\mathrm{c,max}$ & \multicolumn{1}{r}{-$203.52$} & \multicolumn{1}{r}{-$203.62$} && \multicolumn{1}{r}{-$130.78$} & \multicolumn{1}{r}{-$130.78$} && \multicolumn{1}{r}{-$0.1846$} & \multicolumn{1}{r}{-$0.1846$} \\ 
$k_\mathrm{f}$ & \multicolumn{1}{r}{-$240.55$} & \multicolumn{1}{r}{-$240.80$} && \multicolumn{1}{r}{-$204.88$} & \multicolumn{1}{r}{-$204.88$} && \multicolumn{1}{r}{-$0.2892$} & \multicolumn{1}{r}{-$0.2892$} \\ 
\midrule
& \multicolumn{2}{c}{$\mathbf{g}_{n,\mathrm{kge}}(\boldsymbol{\uptheta})$}
&& \multicolumn{2}{c}{$\mathbf{g}_{n,\mathrm{huber}}(\boldsymbol{\uptheta})$}
&& \multicolumn{2}{c}{$\mathbf{g}_{n,\mathrm{fdc}}(\boldsymbol{\uptheta})$} \\[1mm]
\cline{2-3}\cline{5-6}\cline{8-9} \\[-4mm]
& \multicolumn{1}{c}{Analytic} & \multicolumn{1}{c}{Numeric} && \multicolumn{1}{c}{Analytic} & \multicolumn{1}{c}{Numeric} && \multicolumn{1}{c}{Analytic} & \multicolumn{1}{c}{Numeric} & \\
\midrule
$u_\mathrm{t,max}$ & \multicolumn{1}{r}{$0.0014$} & \multicolumn{1}{r}{$0.0014$} && \multicolumn{1}{r}{$25.833$} & \multicolumn{1}{r}{$25.832$} && \multicolumn{1}{r}{$0.0000$} & \multicolumn{1}{r}{$0.0000$} \\ 
$u_\mathrm{f,max}$ & \multicolumn{1}{r}{$0.0232$} & \multicolumn{1}{r}{$0.0294$} && \multicolumn{1}{r}{$757.71$} & \multicolumn{1}{r}{$741.65$} && \multicolumn{1}{r}{$0.0046$} & \multicolumn{1}{r}{$0.0047$} \\ 
$l_\mathrm{t,max}$ & \multicolumn{1}{r}{-$0.0010$} & \multicolumn{1}{r}{-$0.0010$} && \multicolumn{1}{r}{-$17.940$} & \multicolumn{1}{r}{-$17.938$} && \multicolumn{1}{r}{-$0.0000$} & \multicolumn{1}{r}{-$0.0000$} \\ 
$l_\mathrm{fp,max}$ & \multicolumn{1}{r}{-$0.0021$} & \multicolumn{1}{r}{-$0.0021$} && \multicolumn{1}{r}{-$37.955$} & \multicolumn{1}{r}{-$37.955$} && \multicolumn{1}{r}{-$0.0000$} & \multicolumn{1}{r}{-$0.0000$} \\ 
$l_\mathrm{fs,max}$ & \multicolumn{1}{r}{$0.5137$} & \multicolumn{1}{r}{$0.5137$} && \multicolumn{1}{r}{-$3562.3$} & \multicolumn{1}{r}{-$3562.2$} && \multicolumn{1}{r}{$0.0009$} & \multicolumn{1}{r}{$0.0009$} \\ 
$\alpha$ & \multicolumn{1}{r}{-$0.0019$} & \multicolumn{1}{r}{-$0.0019$} && \multicolumn{1}{r}{-$34.252$} & \multicolumn{1}{r}{-$34.250$} && \multicolumn{1}{r}{-$0.0000$} & \multicolumn{1}{r}{-$0.0000$} \\ 
$\psi$ & \multicolumn{1}{r}{$0.0013$} & \multicolumn{1}{r}{$0.0013$} && \multicolumn{1}{r}{$24.155$} & \multicolumn{1}{r}{$24.157$} && \multicolumn{1}{r}{$0.0000$} & \multicolumn{1}{r}{$0.0000$} \\ 
$k_\mathrm{i}$ & \multicolumn{1}{r}{-$0.0003$} & \multicolumn{1}{r}{-$0.0003$} && \multicolumn{1}{r}{-$6.2792$} & \multicolumn{1}{r}{-$6.2792$} && \multicolumn{1}{r}{$0.0000$} & \multicolumn{1}{r}{$0.0000$} \\ 
$\kappa$ & \multicolumn{1}{r}{-$0.1081$} & \multicolumn{1}{r}{-$0.1081$} && \multicolumn{1}{r}{$4204.0$} & \multicolumn{1}{r}{$4203.9$} && \multicolumn{1}{r}{-$0.0007$} & \multicolumn{1}{r}{-$0.0007$} \\ 
$\nu_\mathrm{p}$ & \multicolumn{1}{r}{-$0.1044$} & \multicolumn{1}{r}{-$0.1044$} && \multicolumn{1}{r}{-$960.92$} & \multicolumn{1}{r}{-$960.91$} && \multicolumn{1}{r}{$0.0023$} & \multicolumn{1}{r}{$0.0023$} \\ 
$\nu_\mathrm{s}$ & \multicolumn{1}{r}{-$0.0941$} & \multicolumn{1}{r}{-$0.0941$} && \multicolumn{1}{r}{-$1106.8$} & \multicolumn{1}{r}{-$1106.8$} && \multicolumn{1}{r}{$0.0017$} & \multicolumn{1}{r}{$0.0017$} \\ 
$a_\mathrm{c,max}$ & \multicolumn{1}{r}{-$0.0855$} & \multicolumn{1}{r}{-$0.0855$} && \multicolumn{1}{r}{-$1870.7$} & \multicolumn{1}{r}{-$1870.8$} && \multicolumn{1}{r}{-$0.0004$} & \multicolumn{1}{r}{-$0.0004$} \\ 
$k_\mathrm{f}$ & \multicolumn{1}{r}{-$0.1408$} & \multicolumn{1}{r}{-$0.1408$} && \multicolumn{1}{r}{-$2325.8$} & \multicolumn{1}{r}{-$2325.8$} && \multicolumn{1}{r}{-$0.0001$} & \multicolumn{1}{r}{-$0.0001$} \\ 
\bottomrule
\end{tabular}
\label{table:3}
\end{threeparttable}
\end{table}
Across all parameters and loss functions, the analytic and numerical gradients are in close agreement, with only minor discrepancies attributable to finite-difference truncation and floating-point effects. This agreement provides strong confirmation that the forward-sensitivity implementation of the \texttt{sacsma} discharge Jacobian is correct, numerically stable, and fully consistent with independent finite-difference estimates.

Beyond validation, Table~\ref{table:3} provides a direct empirical illustration of the theoretical considerations outlined in Section~\ref{subsec:gn_anatomy}. As anticipated, the magnitude of the gradient vector varies drastically across loss functions as a result of their differing normalization and residual-weighting schemes, reflecting the intrinsic scaling embedded in each objective. The SAR and GLS losses produce gradients of order $10^{0}$ - $10^{2}$, consistent with their linear and quadratic dependence on residual magnitude. In contrast, the NSE-, KGE-, and FDC-based objectives yield gradients that are several orders of magnitude smaller, frequently in the $10^{-3}$ range or below, owing to variance normalization and the use of correlation or distributional measures. 

Robust losses exhibit yet another characteristic behavior. The Huber objective produces large gradients, sometimes exceeding $10^{3}$ for parameters that exert strong control on runoff generation (e.g., $u_\mathrm{f,max}$, $a_\mathrm{c,max}$, and $k_\mathrm{f}$), while simultaneously limiting the influence of extreme residuals. This behavior is fully consistent with its piecewise influence function and highlights how robustness does not necessarily imply small gradients, but rather controlled sensitivity.

These observed patterns are not numerical artifacts but rather implied by the geometry of the loss surfaces discussed earlier. In particular, the extremely small gradients associated with efficiency-based objectives explain their well-known tendency toward slow convergence and stagnation when combined with finite-difference sensitivities. The close agreement between analytic and numerical gradients in Table \ref{table:3} demonstrates that analytic differentiation faithfully captures these weak sensitivities without contamination by numerical noise, thereby preserving genuine flat regions of the objective rather than introducing spurious descent directions. Taken together, these results empirically substantiate the arguments of Section \ref{subsec:gn_anatomy}. Gradient magnitude is an inherent property of the chosen loss function, not a deficiency of the model or its implementation. Analytic gradients therefore offer a dual advantage. They not only reduce CPU-cost, but also enable robust optimization on intrinsically flat response surfaces, making them essential for stable and efficient calibration of hydrologic models across heterogeneous loss landscapes.

\begin{figure}[H]
\centering\includegraphics[width=1\linewidth]{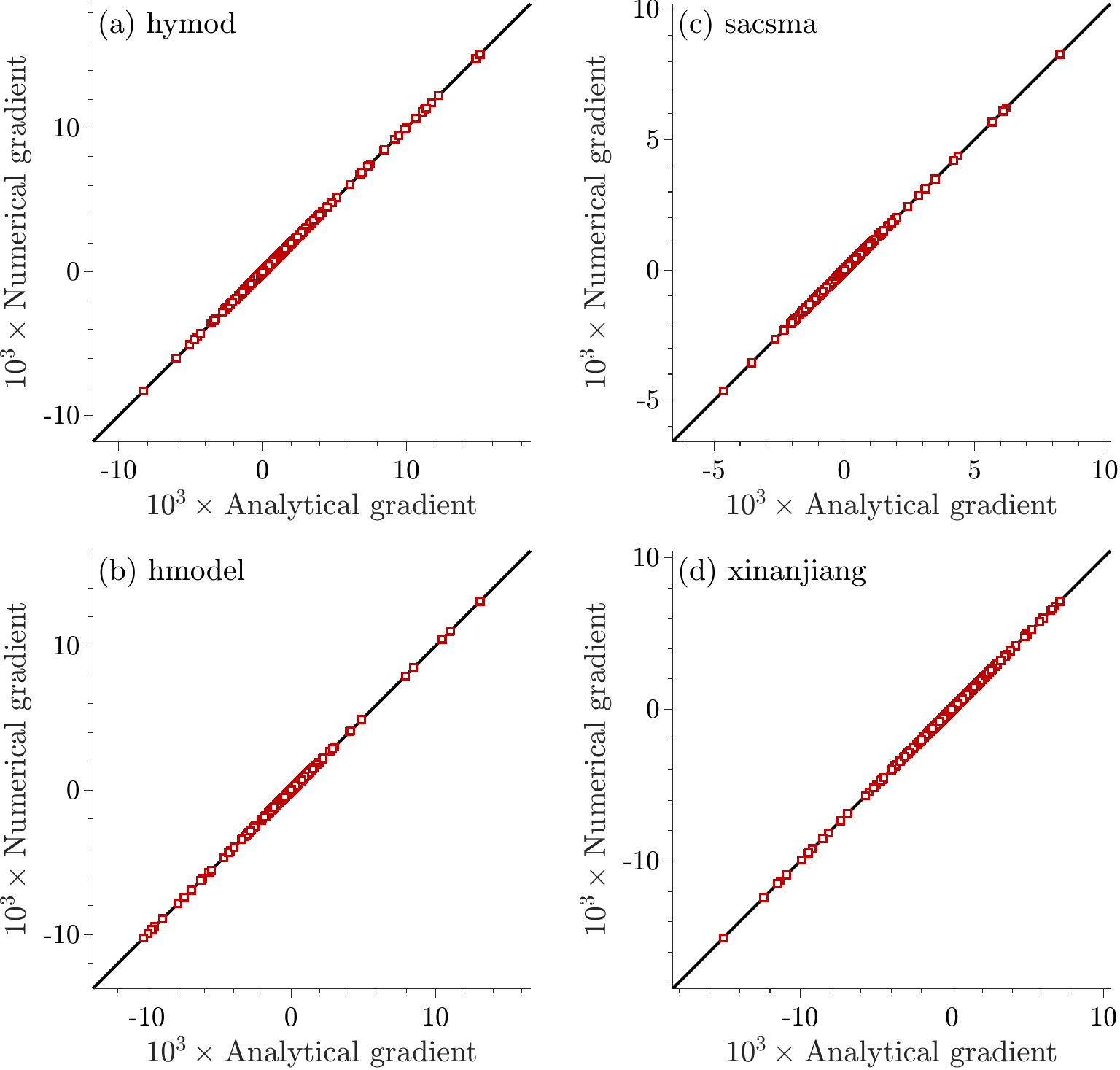}
\caption{Comparison of analytic and numerical gradient vectors for four conceptual hydrologic models using $N = 10$ randomly sampled parameter vectors across the Leaf River, French Broad, Wye, and Severn catchments. Each panel shows a scatter plot of numerical versus analytic gradient values for (a) \texttt{hymod} ($960$ entries), (b) \texttt{hmodel} ($1,344$ entries), (c) \texttt{sacsma} ($2,496$ entries), and (d) \texttt{xinanjiang} ($2,688$ entries).}
\label{fig:5}
\end{figure}

We do not separately present gradient vectors for other models, parameter values and data sets. This would only present similar results. Instead, we extent the Monte Carlo analysis to the gradient vectors of the six loss functions of Table~\ref{table:1}. For each model, we compute analytic and numerical gradients for $N = 10$ different parameter vectors. We then concatenate the gradient entries of of $\mathcal{L}_\mathrm{sar}(\boldsymbol{\uptheta})$, $\mathcal{L}_\mathrm{gls}(\boldsymbol{\uptheta})$, $\mathcal{L}_\mathrm{nse}(\boldsymbol{\uptheta})$, $\mathcal{L}_\mathrm{kge}(\boldsymbol{\uptheta})$, $\mathcal{L}_\mathrm{huber}(\boldsymbol{\uptheta})$, and $\mathcal{L}_\mathrm{fdc}(\boldsymbol{\uptheta})$ into a single vector for the Leaf River, French Broad, Wye, and Severn catchments. This aggregation mixes parameters, catchments, loss functions, and data resolutions (daily and hourly), thereby providing a broad, model-agnostic assessment of gradient accuracy. Figure~\ref{fig:5} summarizes the results of this analysis and displays scatter plots comparing numerical and analytic gradient entries for each watershed model. 
In nearly all cases, the analytic gradients align closely with their numerical finite-difference counterparts, yielding points that cluster tightly around the 1:1 line. By contrast, the magnitude of the gradients varies substantially across models and loss functions, consistent with differences in model structure, complexity, and parameter dimensionality discussed earlier.

Table \ref{table:4} summarizes the accuracy and computational performance of the analytic gradients relative to finite-difference approximations for all four conceptual watershed models, evaluated for different basins and data resolutions. The gradients of the six loss functions are aggregated. 
\begin{table}
\centering
\captionsetup[table]{position=bottom}
\begin{threeparttable}  
\caption{Mean absolute difference between analytic $\mathbf{g}^\mathrm{a}_{n}(\underline{\boldsymbol{\uptheta}})$ and numerical $\mathbf{g}^\mathrm{n}_{n}(\underline{\boldsymbol{\uptheta}})$ gradients and the CPU speed-up achieved by analytic differentiation for the \texttt{hymod}, \texttt{hmodel}, \texttt{sacsma}, and \texttt{xinanjiang} models across daily and hourly datasets. The quantity $\Delta \mathbf{g}_{n} = \Mean(\lvert \mathbf{g}^\mathrm{a}_{n} - \mathbf{g}^\mathrm{n}_{n}\rvert)$ is averaged over all $n$ time steps, $d$ parameters, and the six loss functions of Table~\ref{table:1}.}
\begin{tabular}{clcccccc}
\toprule
Model & Dataset & Resolution & $d$ & $m$ & $\Delta \mathbf{g}$ 
& Speed up \\
\midrule
\multirow{4}{*}{\rotatebox{90}{\parbox{2.0cm}{\centering \texttt{hymod}}}} & Leaf river & daily & \multirow{4}{*}{5} & \multirow{4}{*}{6} & $7.295\cdot10^{-2}$ & $49\times$ \\ 
& French broad & daily &  & & $6.828\cdot10^{-2}$ & $143\times$ \\ 
& Wye & hourly &  &  & $1.639\cdot10^{-2}$ & $178\times$ \\ 
& Severn & hourly &  &  & $1.918\cdot10^{-2}$ & $120\times$ \\ 
\midrule
\multirow{4}{*}{\rotatebox{90}{\parbox{2.0cm}{\centering \texttt{hmodel}}}} & Leaf river & daily & \multirow{4}{*}{7} & \multirow{4}{*}{5}  & $3.819\cdot10^{-2}$ & $260\times$ \\ 
& French broad & daily & &  & $4.076\cdot10^{-2}$ & $340\times$ \\ 
& Wye & hourly &  &  & $4.371\cdot10^{-2}$ & $261\times$ \\ 
& Severn & hourly &  &  & $4.574\cdot10^{-2}$ & $257\times$ \\ 
\midrule
\multirow{4}{*}{\rotatebox{90}{\parbox{2.0cm}{\centering \texttt{sacsma}}}} & Leaf river & daily & \multirow{4}{*}{13} & \multirow{4}{*}{9} & $1.492\cdot10^{-1}$ & $383\times$ \\ 
& French broad & daily &  &  & $1.428\cdot10^{-1}$ & $453\times$ \\ 
& Wye & hourly &  &  & $2.272\cdot10^{-2}$ & $486\times$ \\ 
& Severn & hourly &  &  & $2.069\cdot10^{-2}$ & $396\times$ \\ 
\midrule
\multirow{4}{*}{\rotatebox{90}{\parbox{2.0cm}{\centering \texttt{xinan} \\[-2mm] \texttt{jiang}}}} & Leaf river & daily & \multirow{4}{*}{14} & \multirow{4}{*}{8} & $6.051\cdot10^{-2}$ & $91\times$ \\ 
& French broad & daily &  &  & $6.069\cdot10^{-2}$ & $376\times$ \\ 
& Wye & hourly &  &  & $1.163\cdot10^{-2}$ & $439\times$ \\ 
& Severn & hourly &  &  & $1.117\cdot10^{-2}$ & $515\times$ \\ 
\bottomrule
\end{tabular}
\label{table:4}
\end{threeparttable}
\end{table}
Across all cases, the mean absolute gradient discrepancy is extremely small, typically between $10^{-2}$ and $10^{-1}$, confirming the capabilities of our framework to accurately compute the gradients of differentiable loss functions. 

The efficiency gains are substantial. Even for the simplest daily model, analytic gradients are between $50\times$ and $180\times$ faster than finite differences, and for hourly simulations with long time series the speedup increases to $300$ - $500\times$. These improvements hold across models of differing dimensionality and structural complexity, confirming that the computational burden of numerical differentiation scales unfavorably with problem size, whereas analytic differentiation is not as much affected by these factors. The analytic gradients require only a fraction of the computational cost of numerical finite-difference gradients across a wide range of hydrologic conditions and loss functions.

Finally, we examine in more detail the computational speed-up achieved by analytic evaluation of the gradient vector for individual loss functions. Table~\ref{table:5} reports the ratio of mean computation time for numerical finite-difference gradients to that for analytic gradients for the \texttt{sacsma} model, using discharge data from all four catchments at daily and hourly temporal resolutions.
\begin{table}[H]
\centering
\captionsetup[table]{position=bottom}
\begin{threeparttable}
\caption{Speed-up factors for analytic computation of the gradient vectors $\mathbf{g}_{n}(\boldsymbol{\uptheta})$ of the six loss functions in Table \ref{table:1} for the \texttt{sacsma} model. Results are shown for four watersheds at daily and hourly temporal resolutions.}
\begin{tabular}{cccccccc}
\toprule
\multirow{2}{*}{Dataset}
& \multirow{2}{*}{Res} & 
\multicolumn{6}{c}{Speed up $\times$} \\
\cmidrule(lr){3-8}
& & $\mathcal{L}_\mathrm{sar}$ & $\mathcal{L}_\mathrm{gls}$ & $\mathcal{L}_\mathrm{nse}$ & $\mathcal{L}_\mathrm{kge}$ &  
$\mathcal{L}_\mathrm{huber}$ & $\mathcal{L}_\mathrm{fdc}$ \\  
\midrule
Leaf river & daily & $404$ & $405$ & $403$ & $387$ & $360$ & $338$ \\ 
French broad & daily & $494$ & $490$ & $466$ & $444$ & $417$ & $405$ \\ 
Wye & hourly & $450$ & $474$ & $477$ & $488$ & $493$ & $533$ \\ 
Severn & hourly & $398$ & $399$ & $399$ & $396$ & $372$ & $415$ \\ 
\bottomrule
\end{tabular}
\label{table:5}
\end{threeparttable}
\end{table}

Table~\ref{table:5} shows that analytic differentiation delivers large and consistent computational gains across all six loss functions and all four catchments considered. Speed-up factors range from approximately $340$ to more than $530$, with only modest variation between loss functions for a given dataset. This indicates that the dominant computational cost of numerical differentiation arises from repeated model evaluations rather than from the algebraic complexity of the loss function itself. Consequently, once analytic sensitivities are available, the marginal cost of evaluating alternative loss functions is negligible.

Differences across catchments and temporal resolutions are similarly small. Both daily and hourly datasets exhibit comparable speed-up factors, demonstrating that the efficiency gains of analytic gradients persist for long and high-resolution time series. The slightly larger speed-ups observed for the Wye catchment likely reflect increased numerical stiffness and longer effective integration times, which disproportionately penalize finite-difference schemes. Taken together, these results confirm that analytic gradient computation enables scalable, loss-function-agnostic, and computationally efficient calibration of conceptual hydrologic models.

\subsection{Automatic differentiation}
Having established the accuracy and computational advantages of analytic sensitivities relative to numerical finite-difference approximations, we next compare analytic gradient vectors with those obtained via automatic differentiation (AD). Unlike finite-difference schemes, AD propagates derivatives through the computational graph of the model code and is often promoted as a general-purpose alternative to analytic differentiation. In this comparison, we focus exclusively on gradient vectors rather than Jacobian matrices, as the latter are not a direct by-product of AD and would require repeated scalar evaluations. Moreover, AD must be executed separately for each loss function, whereas the analytic approach requires only a single forward sensitivity calculation; once the discharge Jacobian is available, gradient vectors for any loss function can be assembled at negligible additional cost using the framework developed in this paper.

Table~\ref{table:6} reports the average CPU time (seconds) required to compute the gradient vector $\mathbf{g}_{n}(\boldsymbol{\uptheta})$ using analytic forward sensitivities and automatic differentiation for four conceptual watershed models and six loss functions. Results are averaged over $N = 10$ parameter vectors and four study catchments. 
\begin{table}[H]
\centering
\captionsetup[table]{position=bottom}
\begin{threeparttable}
\caption{CPU time (in seconds) for computing the gradient vector $\mathbf{g}_{n}(\boldsymbol{\uptheta})$ of the four conceptual watershed models using 1-year of daily hydrologic data of the French Broad watershed. Listed values are an average for $N = 10$ parameter vectors and Leaf river, French Broad, Wye and Severn basins.}
\begin{tabular}{llcccccc}
\toprule
& \textbf{Model} & $\mathbf{g}_\mathrm{sar}$ 
& $\mathbf{g}_\mathrm{gls}$ 
& $\mathbf{g}_\mathrm{nse}$ 
& $\mathbf{g}_\mathrm{kge}$ 
& $\mathbf{g}_\mathrm{huber}$ 
& $\mathbf{g}_\mathrm{fdc}$ \\
\midrule \\[-4mm] 
\multirow{4}{*}{\rotatebox{90}{\parbox{2.4cm}{\footnotesize \centering Analytic\\ differentiation}}}
& \texttt{hymod} & $0.038$ & $0.036$ & $0.039$ & $0.043$ & $0.093$ & $0.038$ \\[1mm] 
& \texttt{hmodel} & $0.078$ & $0.076$ & $0.080$ & $0.081$ & $0.128$ & $0.077$ \\[1mm] 
& \texttt{sacsma} & $0.166$ & $0.165$ & $0.169$ & $0.171$ & $0.220$ & $0.167$ \\[1mm] 
& \texttt{xinanjiang} & $0.062$ & $0.060$ & $0.064$ & $0.067$ & $0.115$ & $0.061$ \\[1mm] \midrule 
\multirow{4}{*}{\rotatebox{90}{\parbox{2.4cm}{\footnotesize \centering Automatic\\ differentiation}}} 
& \texttt{hymod} & $320.6$ & $311.8$ & $350.4$ & $291.4$ & $321.0$ & $350.9$ \\[1mm] 
& \texttt{hmodel} & $1406.1$ & $1189.6$ & $1113.7$ & $1141.0$ & $1155.7$ & $1096.0$ \\[1mm] 
& \texttt{sacsma} & $1752.6$ & $1386.2$ & $1348.5$ & $1390.2$ & $1280.2$ & $509.5$ \\[1mm] 
& \texttt{xinanjiang} & $966.6$ & $1008.1$ & $751.0$ & $770.4$ & $756.4$ & $855.4$ \\[1mm] 
\bottomrule  
\end{tabular}
\label{table:6}
\end{threeparttable}
\end{table}
The contrast between the two approaches is striking. Across all models and loss functions, analytic gradients are computed in fractions of a second, whereas AD-based gradients require several minutes up to almost half an hour per evaluation. Speed differences of three to four orders of magnitude are common, particularly for the more complex \texttt{sacsma} and \texttt{xinanjiang} models. These large disparities arise from fundamental differences in how derivatives are propagated. Automatic differentiation effectively re-executes the full model for each loss function while tracking derivative information through every time step and conditional branch. In contrast, the analytic approach decouples model sensitivities from the loss function, so that the dominant computational cost, the forward integration of the sensitivity equations, is incurred only once. As a result, once analytic sensitivities are available, switching between loss functions adds negligible overhead. 

It is not uncommon to use at least 10 years of data for watershed model calibration \citep{yapo1996}. For such record lengths, the computational cost of automatic differentiation increases substantially. Forward-mode AD scales linearly with both the number of parameters and the length of the time series, while reverse-mode AD requires storage of the full computational trajectory, leading to rapidly increasing memory demands. In practice, reverse-mode implementations often rely on checkpointing or batching strategies to mitigate memory usage, at the expense of additional recomputation and wall-clock time. Consequently, the CPU costs reported here for one year of data would increase by at least an order of magnitude for multi-year calibration experiments.  

While computational efficiency is a primary concern, accuracy of the resulting gradients is equally important. Table~\ref{table:7} summarizes the Euclidean ($\ell_2$) norm of the difference between gradient vectors obtained via AD and those computed analytically, averaged over Monte Carlo samples and across catchments. 
\begin{table}[H]
\centering
\captionsetup[table]{position=bottom}
\begin{threeparttable}
\caption{Euclidean norm of the difference between gradient vectors obtained via automatic differentiation and analytic forward-sensitivity differentiation. Listed values represent averages over Monte Carlo parameter samples and watersheds.}
\begin{tabular}{llcccccc}
\toprule
& \textbf{Model} & $\mathbf{g}_\mathrm{sar}$ 
& $\mathbf{g}_\mathrm{gls}$ 
& $\mathbf{g}_\mathrm{nse}$ 
& $\mathbf{g}_\mathrm{kge}$ 
& $\mathbf{g}_\mathrm{huber}$ 
& $\mathbf{g}_\mathrm{fdc}$ \\
\midrule
& \texttt{hymod} & $2.734$ & $3.942$ & $0.008$ & $0.003$ & $0.876$ & $0.001$ \\[1mm]
& \texttt{hmodel} & $0.079$ & $0.059$ & $0.000$ & $0.000$ & $0.020$ & $0.000$ \\[1mm] 
& \texttt{sacsma} & $1.330$ & $7.912$ & $0.015$ & $0.006$ & $4.191$ & $0.002$ \\[1mm] 
& \texttt{xinanjiang} & $20.810$ & $12.393$ & $0.024$ & $0.008$ & $4.473$ & $0.007$ \\[1mm] 
\bottomrule
\end{tabular}
\label{table:7}
\end{threeparttable}
\end{table}
Overall, gradients obtained via AD are of comparable order of magnitude to their analytic counterparts, indicating that AD captures the dominant sensitivity structure in many cases. The closest agreement is observed for smooth and well-scaled loss functions such as NSE and KGE, for which gradient expressions involve relatively simple algebraic operations on model outputs.

Larger discrepancies emerge for more complex models and for loss functions that involve non-smooth operations, distributional comparisons, or conditional weighting, such as GLS, Huber, and FDC. In these cases, differences likely reflect the accumulation of numerical error over long time-stepping loops, sensitivity to branching logic and threshold behavior, and practical limitations of operator-overloading AD implementations in \textsc{Matlab}. These discrepancies do not indicate deficiencies in the analytic sensitivities, which were independently validated against finite-difference derivatives in the previous section. Rather, they highlight the challenges of applying general-purpose AD tools to long hydrologic time series. Reverse-mode AD, while theoretically attractive for scalar-valued loss functions, is rarely practical in \textsc{Matlab} because it requires storage of the full state trajectory for replay, leading to prohibitive memory demands and wall-clock times that often exceed those of forward-mode AD.

Taken together, these results show that while modern automatic differentiation tools can, in principle, deliver gradients at a cost comparable to a small number of forward integrations, our experience with \textsc{Matlab}’s \texttt{dlarray} implementation indicates that, for long hydrologic time series and compiled ODE solvers, analytic forward-sensitivity gradients remain substantially faster and more memory-efficient. By augmenting the ODE system only once and evaluating all derivatives in a single forward pass, the analytic formulation achieves one to several orders of magnitude speedup while preserving machine-precision accuracy. Consequently, analytic differentiation remains the only practical option for scalable gradient-based calibration of conceptual hydrologic models when long records, multiple objectives, or repeated evaluations are required.

\subsection{Impact of Numerical solver}
The augmented ODE systems for the \texttt{hymod}, \texttt{hmodel}, \texttt{sacsma}, and \texttt{xinanjiang} models are integrated using a mass-conservative, second-order Runge-Kutta scheme with adaptive time stepping. The model equations and numerical solver are implemented in \textsc{C++} and coupled to \textsc{Matlab} via the \texttt{mex} interface. In practice, hydrologic models are often implemented with relatively simple time-integration schemes, which may introduce variability in simulated discharge and associated sensitivities. To assess the influence of numerical solver choice, we re-evaluated the augmented ODE system of Equation~\ref{eq:ode45_aug_system} using (i) the built-in \texttt{ode45} solver, and (ii) an explicit Euler scheme with a fixed sub-daily or sub-hourly time step equal to $1/100$ of the observation interval. We also considered a Runge-Kutta implementation in \textsc{Matlab}.

Despite these differences in integration strategy, time-step control, and floating-point behavior, the analytic Jacobians and gradient vectors obtained from the forward-sensitivity formulation were essentially invariant. While different solvers yield slightly different state trajectories, most notably for coarse explicit Euler integration, the resulting sensitivities remained consistent. This indicates that the forward-sensitivity framework is robust to reasonable choices of numerical time integration and does not rely on any specific solver implementation. 

\subsection{Model calibration using numeric and analytic gradients}
\begin{figure}[H]
\centering
\includegraphics[width=1\linewidth]{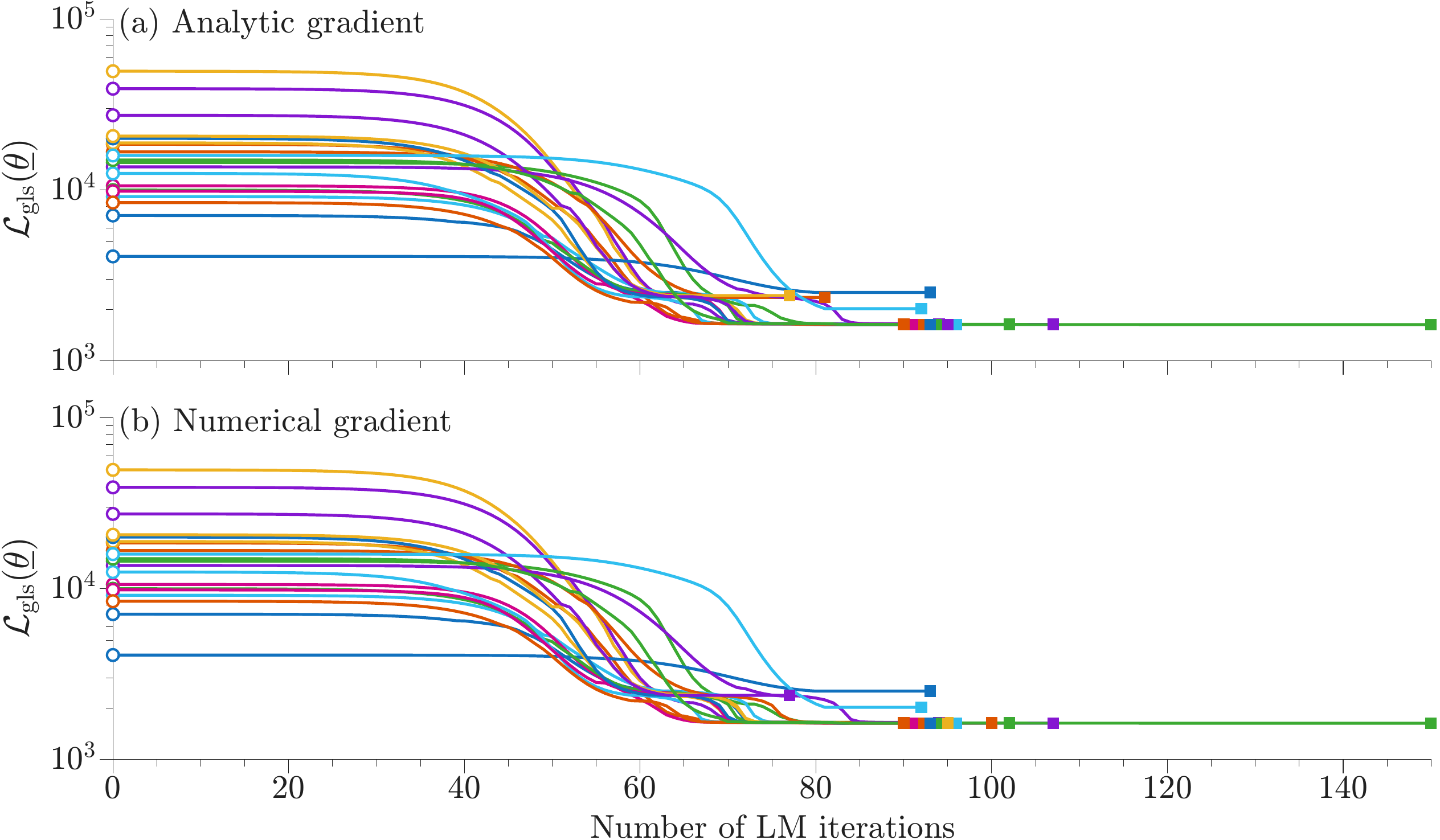}
\caption{Trace plots of the sum of squared residuals (SSR) during gradient-descent calibration of the \texttt{hymod} model using $10$ years of daily discharge data (WY 1992--2001) from the Leaf River. Shown are $N = 20$ optimization trajectories from different starting points using (a) analytic and (b) numerical gradients.}
\label{fig:6}
\end{figure}

We compare numerical and analytic gradient information for calibration of \texttt{hymod}, the most parsimonious of the four watershed models in Appendix~\ref{sec:AppendixB}. We calibrate its $d = 5$ parameters using gradient descent (Algorithm~\ref{algApp:gradient_descent}) and hydrologic data from the Leaf River. The $10$-year calibration period (water years 1982-1991) comprises $n = 3,652$ streamflow observations and is preceded by a $365$-day spin-up to reduce sensitivity to initial state values. Figure \ref{fig:6} shows the evolution of the sum of squared residuals, $\mathcal{L}_\mathrm{gls}(\underline{\boldsymbol{\uptheta}})$ with identity weight matrix $\mathbf{W}_{n} = \mathbf{I}_{n}$. 
We display twenty optimization trajectories, each initialized from a random starting point and shown in a distinct color, using (a) analytic gradients and (b) numerical gradients. Numerical gradients are computed using the \texttt{DERIVESTsuite} toolbox and require, on average, approximately 10 seconds per gradient evaluation. In contrast, analytic gradients are obtained from Equation~\ref{eqApp:g_ngls(uptheta)} and incur only a minor additional computational cost.

Perhaps unexpectedly, the optimization trajectories in the two panels are virtually indistinguishable. At first glance, this appears to contradict the commonly cited shortcomings of numerical gradients. The explanation is twofold. First, the semi-adaptive central differencing scheme combined with Richardson extrapolation provides highly accurate numerical approximations of the \texttt{hymod} gradients. This robustness, however, comes at a substantial computational cost as the optimization trajectories based on numerical gradients require, on average, approximately $150$–$200\times$ more wall-clock time to complete. Second, the Runge-Kutta solver used to integrate the model equations was configured with relatively strict numerical tolerances (\texttt{abstol} = \texttt{reltol} = $10^{-5}$) and a minimum time step of $10^{-5}$ days. These settings yield a smooth, low-noise numerical solution, which is a prerequisite for obtaining near-exact gradients via finite differencing. Together with the low dimensionality of the \texttt{hymod} parameter space, these factors explain why the trajectories in Figs. \ref{fig:6}a and \ref{fig:6}b are effectively identical.  

\begin{figure}
\centering
\includegraphics[width=0.72\linewidth]{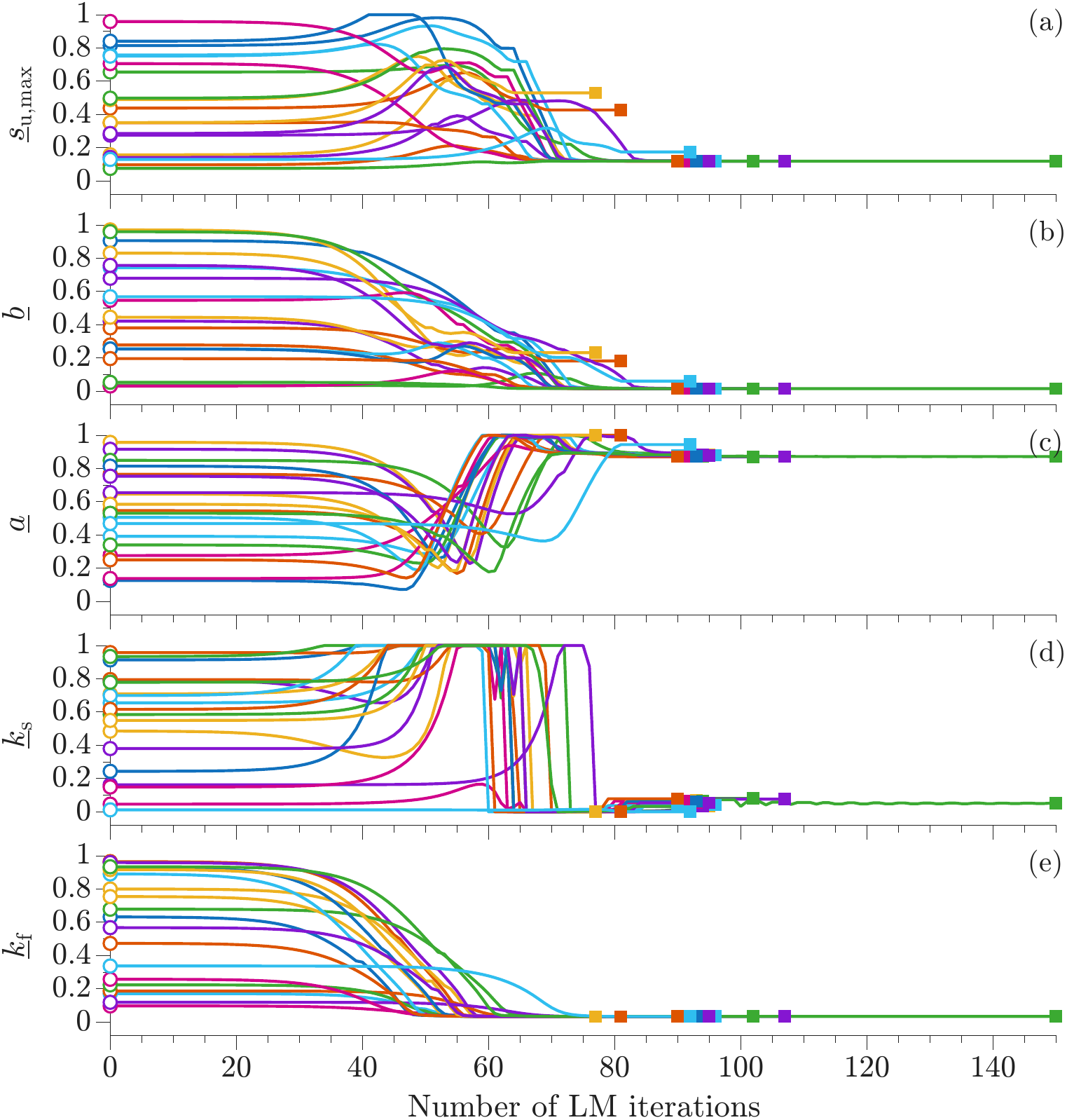}
\caption{Trace plots of the normalized \texttt{hymod} parameter values as a function of LM iteration using analytic gradients: (a) $\underline{s}{\mathrm{u,max}}$, (b) $\underline{b}$, (c) $\underline{a}$, (d) $\underline{k}{\mathrm{s}}$, and (e) $\underline{k}{\mathrm{f}}$. Color coding matches the $\mathcal{L}\mathrm{gls}$ descent trajectories shown in Fig.~\ref{fig:6}. The parameter traces correspond to the successive iterates $\underline{\boldsymbol{\uptheta}}_{(0)},\ldots,\underline{\boldsymbol{\uptheta}}_{(k)}$ generated by Algorithm~\ref{algApp:Levenberg_Marquardt}, where $k$ denotes the iteration index and $\underline{\boldsymbol{\uptheta}} = (\underline{s}_{\mathrm{u,max}}, \, \underline{b}, \, \underline{a}, \, \underline{k}_{\mathrm{s}}, \, \underline{k}_{\mathrm{f}})^{\top}$. The unit cube trust region facilitates comparison of \texttt{hymod} parameters with different physical units and scales.}
\label{fig:7}
\end{figure}

Given \texttt{hymod}’s parsimonious model structure and the use of an accurate second-order numerical time-integration scheme, we expect the response surface of the $\mathcal{L}_\mathrm{gls}$ objective function to be smooth and largely unimodal. Indeed, almost all LM trials converge to approximately the same final loss value. This value, $\mathcal{L}_\mathrm{gls} \approx 1{,}664$, corresponds to a daily root-mean-square error of about $0.94$ mm\,d$^{-1}$ and $\mathrm{NSE} = 0.87$, indicating a well-conditioned calibration problem under the SSR loss function. Consequently, 
both analytic- and numerical-gradient-based optimization are expected to converge to approximately identical parameter estimates for most, if not all, starting points. This expectation is confirmed in Figure \ref{fig:7} which complements the $\mathcal{L}_\mathrm{gls}$ descent trajectories shown in Fig. \ref{fig:6} and visualizes the corresponding pathways in parameter space. 
Consistent with the near-identical convergence behavior observed in $\mathcal{L}_\mathrm{gls}$ loss space, all trajectories collapse toward the same region of parameter space for both gradient formulations. Particularly noteworthy, is the sudden jump of parameter $a$ from its upper to its lower bound at about iteration $k = 60$. Note that is no point in presenting the results for the numerical gradients as the trajectories are nearly indistinguishable. Despite these similarities in convergence behavior, however, substantial differences in computational efficiency are observed. Numerical differentiation incurs a significant per-iteration cost due to repeated model evaluations, such that total wall-clock runtime is dominated by gradient computation. In contrast, analytic gradients reduce this cost by more than two orders of magnitude, enabling rapid convergence and rendering repeated multi-start calibration computationally inexpensive.

These results underscore an important practical distinction. Numerical gradients can approximate the performance of analytic gradients provided that (i) the numerical solution of the governing ODE system is sufficiently accurate and (ii) a robust, semi-adaptive finite-differencing scheme, such as central differencing combined with Richardson extrapolation, is employed. While these conditions can be met for parsimonious models such as \texttt{hymod}, doing so already entails a substantial computational overhead and becomes increasingly difficult as model complexity, parameter dimensionality, or data length increases
Consequently, the computational cost of numerical differentiation can quickly become prohibitive for realistic hydrologic applications. The compromises that must then be made, most notably looser ODE error tolerances and/or less robust gradient estimators, introduce gradient noise, which ultimately leads to divergence between analytic and numerical gradients and degrades optimization performance. In the present example, analytic gradients provide modest but instructive gains, yielding a $100$–$200\times$ speed-up. More importantly, they foreshadow substantially larger benefits for more complex hydrologic models, for which accurate numerical differentiation is no longer computationally viable.

\subsection{Model calibration using analytic gradients}
Having established the correctness of the analytic Jacobians and gradient vectors, we now evaluate their practical utility for hydrologic model calibration. We focus on the \texttt{xinanjiang} model and use one year of hourly discharge observations from the Severn River at Plynlimon, UK, for model calibration. Solving the augmented ODE system in Equation~\ref{eq:ode45_aug_system} for the \texttt{xinanjiang} model, including a 100-day spin-up period, incurs an average computational cost of approximately $0.5$--$1$ second per model evaluation. This corresponds to $465 \times 24 = 11{,}160$ hourly print steps, of which the final $n = 8{,}760$ hours are used to compute the loss function and associated gradient. This data length is equivalent to a 30-year record of daily observations.

\begin{sidewaysfigure}
\centering
\includegraphics[width=1\linewidth]{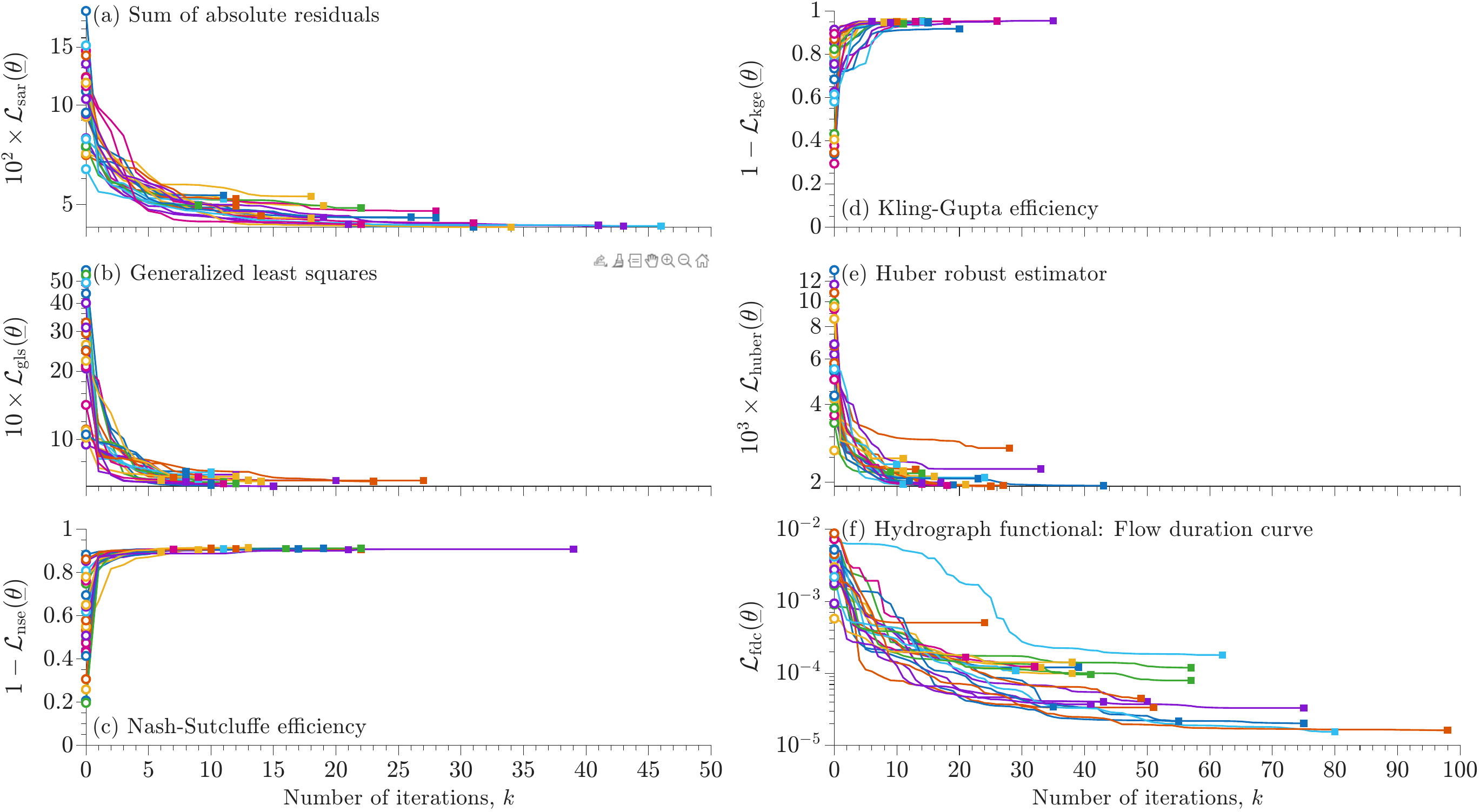}
\caption{Traceplots of six loss functions for the \texttt{xinanjiang} model calibrated to one year of hourly discharge data from the Severn River, Wales. Shown are 20 optimization trials using gradient descent. The CPU time of a single \texttt{xinanjiang} evaluation is on the order of 0.5-0.8 seconds.}
\label{fig:8}
\end{sidewaysfigure}

Figure~\ref{fig:8} portrays the behavior of the six loss functions,
$\mathcal{L}_\mathrm{sar}(\boldsymbol{\uptheta})$, $\mathcal{L}_\mathrm{gls}(\boldsymbol{\uptheta})$,
$\mathcal{L}_\mathrm{nse}(\boldsymbol{\uptheta})$, $\mathcal{L}_\mathrm{kge}(\boldsymbol{\uptheta})$,
$\mathcal{L}_\mathrm{huber}(\boldsymbol{\uptheta})$, and
$\mathcal{L}_\mathrm{fdc}(\boldsymbol{\uptheta})$, during gradient descent of the \texttt{sacsma} model.
Shown are traceplots of the loss value as a function of iteration for $N = 20$ optimization trials per loss function, each using a different starting point, $\underline{\boldsymbol{\uptheta}}_{(0)} \in [0,1]^{d}$ and calibrated using steepest descent (Algorithm~\ref{algApp:gradient_descent}). The figure therefore depicts effects of loss function geometry on optimization behavior.

Several consistent patterns emerge from Figure~\ref{fig:8}. First, for all six objectives, analytic gradients enable rapid and stable descent, with most trials achieving substantial loss reduction within relatively few iterations. This confirms that the derived sensitivities are not only mathematically correct but also numerically effective for large, high-resolution datasets where finite-difference and/or automatic differentiation are prohibitively expensive.
Second, the optimization trajectories do not collapse to a single solution, even under identical loss functions. This phenomenon has long been recognized in conceptual hydrologic modeling and motivated the adoption and development of global optimization methods \citep[e.g.,][]{duan1992,sorooshian1993}. The strongly nonconvex nature of the response surface, together with the sensitivity of gradient-based algorithms to initial conditions, inhibits consistent convergence to a unique optimum. Third, the spread of trajectories varies markedly across loss functions. Reward-based metrics such as NSE, KGE, and FDC exhibit relatively tight clustering of final loss values, indicating that these objectives define broader, flatter basins of attraction with similar optima. In contrast, residual-based objective functions such as SAR, GLS, and the Huber loss show substantially greater variability across trials, consistent with a more rugged loss surface containing multiple local minima. A fourth and closely related observation concerns gradient magnitude. As discussed in Section~\ref{subsec:gn_anatomy} and shown in Table \ref{table:3}, efficiency-based metrics (NSE, KGE, FDC) generate gradients that are orders of magnitude smaller than those of residual-based losses (SAR, GLS, Huber). While analytic gradients faithfully represent these weak sensitivities, they also imply slower parameter updates unless accompanied by careful step-size control or adaptive optimization strategies. Crucially, when gradients are intrinsically small, finite-difference approximations become dominated by numerical noise, leading to unstable or misleading search directions. The analytic gradients used here eliminate this failure mode, accurately revealing genuinely flat regions of the loss surface rather than artifacts of numerical differencing.

It is important to emphasize that the results shown in Figure~\ref{fig:8} are obtained using a single, default implementation of gradient-descent without any tuning of hyperparameters or use of advanced line-search methods. Likewise, we did not explore hybrid second-order schemes such as Gauss-Newton and Levenberg-Marquardt or state-of-the-art machine learning methods. The purpose of this experiment is therefore not to demonstrate optimal optimizer performance, but rather to isolate and assess the intrinsic value of analytic gradients themselves. The fact that substantial and consistent loss reduction is achieved for all six loss functions under such minimal algorithmic assumptions underscores the robustness and practical relevance of the derived sensitivities. Stochastic gradient descent and/or adaptive moment-based optimizers \citep{kingma2015} should further improve convergence speed and robustness, yet do not alter the central conclusion that analytic forward sensitivities provide a practical, scalable, and numerically robust foundation for gradient-based calibration of conceptual hydrologic models across heterogeneous loss functions and long, high-resolution time series. By avoiding both the computational overhead and numerical fragility of finite-difference derivatives and automatic differentiation, the proposed approach enables efficient and fully derivative-based model calibration, making it suitable for operational calibration and large-scale inference.

\begin{figure}
\centering
\includegraphics[width=1\linewidth]{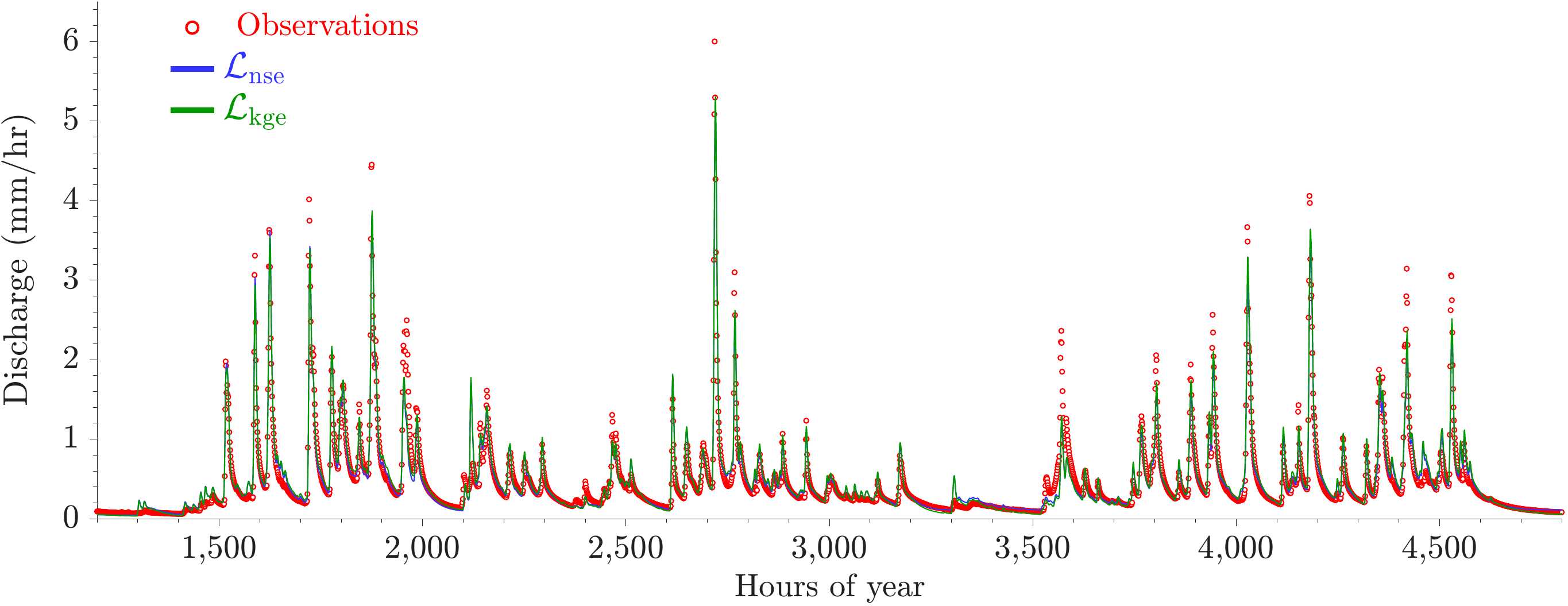}
\caption{Observed and simulated discharge for a representative portion of the one-year hourly calibration period for the Severn River at Plynlimon, UK. The \texttt{xinanjiang} model was calibrated with steepest descent using the NSE (blue) and KGE (green) loss functions. Both calibrations yield similar hydrographs and provide a good overall fit to the observations.}
\label{fig:9}
\end{figure}

Finally, Figure~\ref{fig:9} illustrates the behavior of the calibrated \texttt{xinanjiang} model for a representative subset of the one-year calibration period for the Severn River at Plynlimon, UK. Shown are simulated discharge time series obtained using parameter estimates optimized under the NSE and KGE loss functions, together with the corresponding observed discharge.
Both loss functions yield very similar discharge simulations, despite their different mathematical formulations. For this hourly dataset, the calibrated model reproduces the observed hydrograph well, with overall performance values of $\mathrm{NSE} = 0.914$ and $\mathrm{KGE} = 0.955$. The timing and magnitude of most flow peaks are captured accurately, and the recession behavior following rainfall events is generally well represented. Some mismatches remain, particularly during the large storm event around hour $3,550$, where peak discharge is underestimated. This discrepancy is likely attributable to errors or underestimation in the precipitation forcing rather than deficiencies in the model structure or calibration procedure. Additional deviations are visible during the non-driven portion of the hydrograph immediately after rainfall ceases, when streamflow recedes and catchment storage dominates the dynamics. These differences suggest limitations in the model’s representation of post-event drainage and slow-flow processes.

Overall, Figure~\ref{fig:9} demonstrates that gradient-based calibration using analytic sensitivities yields physically realistic and robust parameter estimates for both NSE- and KGE-based objectives. Despite differences in their loss definitions, both criteria lead to comparable and high-quality simulations for this hourly dataset.

\section{Discussion \& Future Work}\label{sec:outlook}
The large and consistent speed-up factors reported in Tables~\ref{table:5}-\ref{table:7} indicate that hand-coded analytic sensitivities implemented in a compiled core substantially outperform, generic automatic differentiation (AD) frameworks for the class of conceptual hydrologic models considered here.

Table~\ref{table:8} summarizes the relative computational cost, numerical accuracy, and practical limitations of the principal differentiation strategies used in hydrologic modeling. 
\begin{table}[!ht]
\centering
\captionsetup[table]{position=bottom}
\begin{threeparttable}  
\caption{Computational cost, accuracy, and practical limitations of analytic, finite-difference, automatic, and symbolic differentiation. Primal cost (pc) denotes the cost of a single model evaluation. Reported costs for finite-difference and automatic differentiation are lower-bound or idealized estimates. In practice, finite-difference gradients require parameter-specific step-size tuning, and reverse-mode automatic differentiation incurs substantial memory overhead.}
\footnotesize
\begin{tabular}{p{24mm}p{25mm}p{15mm}p{27mm}p{44mm}}
\toprule
\textbf{Method} & \textbf{CPU cost} & \textbf{Slowdown} & \textbf{Accuracy} 
& \textbf{Main limitations} \\
\midrule
\textbf{Analytic} 
& $1.5$--$2\times$pc & \textbf{Fastest} & Exact & Manual derivation; one time \\[1mm]
Finite differences & $(d+1)\times$pc (min.) 
& $10$--$20\times$ & Stepsize dependent
& Expensive for long data; noisy \\[1mm]
AD (forward) & $(d+1)\times(3$--$8)\times$pc & $40$--$120\times$ & Exact & Operator overloading; slow \\[1mm]
AD (reverse) & $5$--$15\times$pc (ideal) & $50$--$200\times$ & Exact & Full storage; poor scalability \\[1mm]
Symbolic & Not feasible & N/A & Exact & No loops or conditionals \\
\bottomrule
\end{tabular}
\label{table:8}
\end{threeparttable}
\end{table}
Analytic forward sensitivities provide the fastest and most reliable means of computing gradient vectors. Once the derivative terms are derived and embedded within the augmented ODE system, the cost of computing all parameter sensitivities scales only modestly, typically by a factor of $1.5$-$2$ relative to a single forward model simulation. The combined state-sensitivity system $(\mathbf{x},\mathbf{S})$ is advanced in a single forward sweep, with all derivatives updated using closed-form expressions and controlled to the same numerical tolerance as the state variables.

Alternative differentiation strategies exhibit markedly different behavior. Numerical finite-difference gradients require at least $(d+1)$ full model evaluations for $d$ parameters when using one-sided schemes and are further affected by truncation and round-off errors. These issues become particularly severe when gradients are intrinsically small, as is common for efficiency-based loss functions such as NSE, KGE, and FDC. In practice, obtaining reliable finite-difference gradients requires careful, parameter-specific step-size selection, since a single global perturbation is rarely adequate. Determining an appropriate step size for each parameter typically involves a cascade of trial evaluations, often on the order of $15-20$ model runs per parameter, substantially increasing CPU cost and rendering finite-difference approaches impractical for high-dimensional or long time-series.

Automatic differentiation is even more expensive in practice. Forward-mode operator-overloading AD differentiates through the entire time-stepping loop and effectively replicates the full computation once per parameter, leading to slowdowns of hundreds to thousands of times for models such as \texttt{sacsma}, \texttt{xinanjiang}, \texttt{hymod}, and \texttt{hmodel} (Table~\ref{table:6}). Reverse-mode AD, while theoretically attractive for scalar loss functions, is rarely practical in \textsc{Matlab} for long time series because it requires storing the full state trajectory for replay, resulting in prohibitive memory demands and wall-clock times.

From a numerical analysis perspective, forward sensitivities must be solved simultaneously with the original ODE system, since each sensitivity depends on the evolving model state. Solving this augmented system requires only a single solver call and ensures that sensitivities are computed with the same numerical accuracy as the state variables. However, because the number of sensitivity equations scales linearly with the number of parameters, forward-mode continuous sensitivity analysis has computational complexity $\mathcal{O}(np)$ for $n$ state variables and $p$ parameters, which can become impractical for high-dimensional parameterizations. In such cases, continuous adjoint sensitivity analysis (CASA) provides a complementary alternative, enabling gradients of scalar objectives to be computed with $\mathcal{O}(n+p)$ complexity. It is worth noting that ODE-based sensitivity analysis is a mature and well-established field, and comparative studies have shown that for small- to medium-sized systems forward-mode sensitivities can outperform discrete adjoint or solver-level automatic differentiation approaches, whereas for larger systems adjoint methods may offer superior scaling \citep{ma2018}.

\begin{figure}[!ht]
\centering\includegraphics[width=0.95\linewidth]{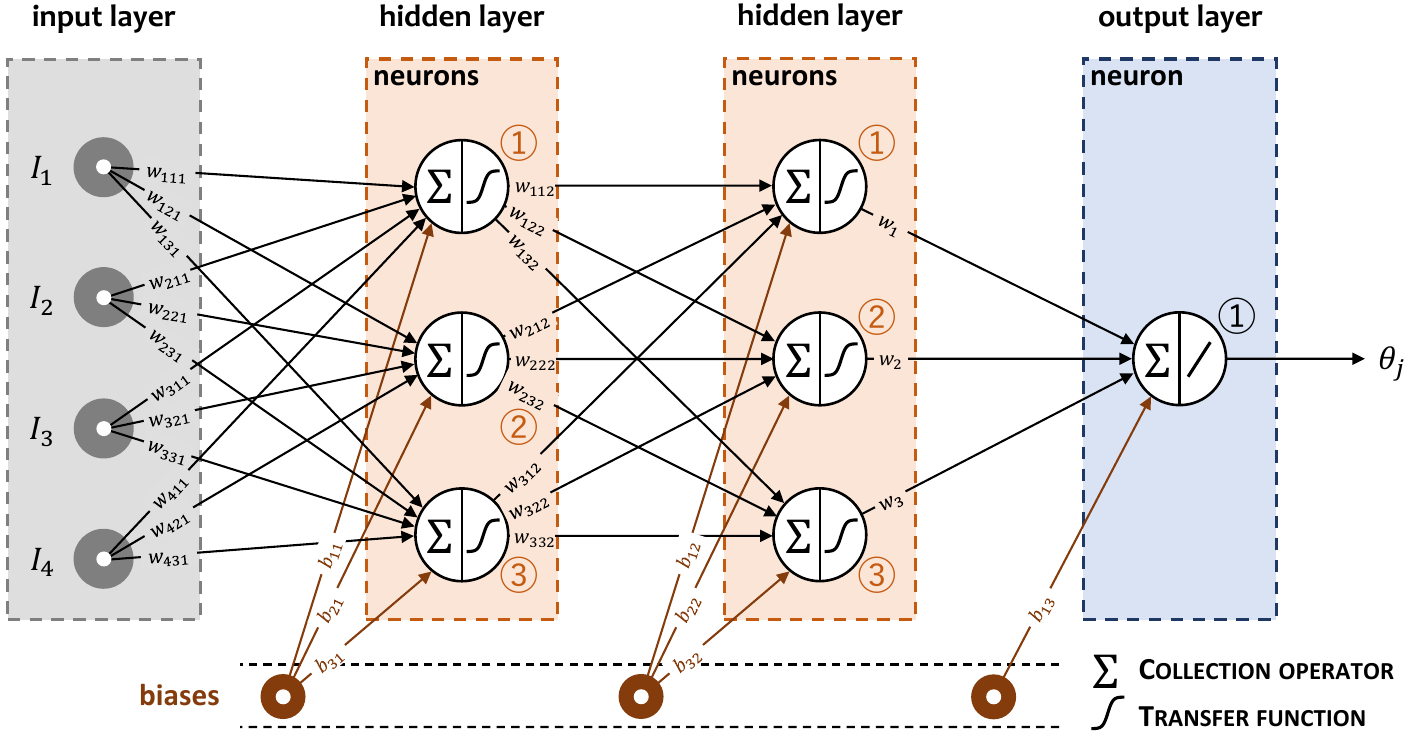}
\caption{Architecture of a feedforward neural network with 2 hidden layers. The catchment attributes $I_{1}, \ldots, I_{4}$ are turned into estimates of the $j$th model parameter $\theta_{j}$. This necessitates the estimation of $24$ weights and $7$ biases.}
\label{fig:10}
\end{figure}

The analytic sensitivity framework developed here opens the door to large-scale, data-driven hydrologic learning problems that are otherwise computationally infeasible. One promising direction is continental-scale parameter estimation, in which model parameters for all watersheds in CONUS are inferred simultaneously from catchment attributes using a feedforward neural network (see Figure \ref{fig:10}). 
In this setting, each model parameter $\theta_j$ is predicted from physiographic attributes $I_1,\ldots,I_{K}$ (e.g., elevation, soil properties, slope), and gradients of discharge with respect to network weights and biases are obtained via the chain rule
\begin{linenomath*}
\begin{align}
\frac{\mathrm{d}q_{t}}{\mathrm{d}w_{i}} & = \frac{\mathrm{d}q_{t}}{\mathrm{d}\theta_{j}}\frac{\mathrm{d}\theta_{j}}{\mathrm{d}w_{i}}, \\
\frac{\mathrm{d}q_{t}}{\mathrm{d}b_{k}} & = \frac{\mathrm{d}q_{t}}{\mathrm{d}\theta_{j}}\frac{\mathrm{d}\theta_{j}}{\mathrm{d}b_{k}},
\end{align}
\end{linenomath*}
enabling efficient gradient-based training of the network parameters. Because the analytic Jacobians $\mathrm{d}q/\mathrm{d}\theta_j$ are already available, this learning problem can be solved with modest computational resources and without relying on large machine-learning software frameworks.

Neural networks such as LSTMs remain fully differentiable, and their gradients are obtained by repeated application of the chain rule through time (backpropagation through time). The distinction with conceptual hydrologic models is therefore not mathematical, but practical. The dimensionality, opacity, and memory requirements of the resulting Jacobians render explicit sensitivity analysis infeasible.

\section{Conclusions}\label{sec:conclusions}
This paper has presented a unified, exact, and computationally efficient framework for evaluating parameter sensitivities in conceptual watershed models. By deriving closed-form expressions for the state Jacobian $\mathbf{J}_{f}(\mathbf{x})$ and parameter Jacobian $\mathbf{J}_{f}(\boldsymbol{\uptheta})$ for four widely used hydrologic models (\texttt{hymod}, \texttt{hmodel}, \texttt{sacsma}, and \texttt{xinanjiang}), and by embedding these derivatives into an augmented ODE system, we obtain continuous-time, noise-free sensitivity trajectories that are valid at any temporal resolution. The resulting analytic Jacobian of simulated discharge, $\mathbf{J}_{q}(\boldsymbol{\uptheta})$, provides exact gradient information for arbitrary differentiable loss functions.

A second contribution of this work is the clarification of the relationship between Jacobians defined in the original bounded parameter space and their counterparts in an unconstrained space. By introducing a smooth, monotone reparameterization that maps $\boldsymbol{\upvartheta} \in \mathbb{R}^{d}$ to the physical parameter bounds, we obtain a transparent chain rule linking $\nabla_{\boldsymbol{\upvartheta}}\mathbf{q}_{n}$ and $\nabla_{\boldsymbol{\uptheta}}\mathbf{q}_{n}$. This transformation enables direct comparison between analytic sensitivities and numerical differentiation, which would otherwise operate in incompatible parameter spaces.

Across all models, data sets, and temporal resolutions, we find excellent agreement between analytic Jacobians and their numerical counterparts once the transformations are applied. However, the computational cost differs dramatically. Analytic Jacobians are orders of magnitude faster and avoid the numerical instability, truncation error, and step-size tuning inherent to finite differences. Hourly data particularly accentuate this performance gap. Numerical Jacobian calculations become prohibitively expensive, whereas the augmented ODE formulation scales linearly and remains robust.

We further demonstrate how the analytic Jacobian can be coupled with a wide range of objective  functions, including $\ell_{1}$ and $\ell_{2}$ loss functions, M-estimators, hydrograph functionals (flow duration curve), and model efficiency metrics such as the NSE and KGE (NSE and KGE), yielding exact gradient vectors that facilitate rapid gradient-based hydrologic model calibration. The framework is loss-agnostic, model-agnostic, and readily extensible to other conceptual and physically based hydrologic models.

Finally, although automatic differentiation provides an attractive and general-purpose derivative mechanism, our results show that carefully derived analytic sensitivities remain the gold standard for conceptual hydrologic models. They are exact, efficient, and fully interpretable in terms of the underlying model physics. Future work will extend the proposed framework to multi-basin and spatially distributed hydrologic models. Through backpropagation, the availability of analytic gradients enables direct estimation of the weights and biases of recurrent neural networks, such as LSTMs, for predicting hydrologic parameter values. Another promising direction is the integration of these sensitivities into score-based likelihoods so as to robustly quantify parameter and predictive uncertainty under model misspecification. 

Overall, this study demonstrates that exact analytic sensitivities offer a powerful and scalable foundation for gradient-based hydrologic model calibration, enabling faster computation, greater numerical stability, and deeper insight into model behavior.

\section*{Acknowledgments}
The comments of the three anonymous reviewers are gratefully acknowledged and have led to a substantially improved manuscript. During the preparation of this work, the authors used GPT-5 (developed by OpenAI) to assist with mathematical derivations and language editing. All AI-generated content was carefully reviewed and edited by the authors, who take full responsibility for the final version of the manuscript.

\section*{Competing Interests}
The authors declare no competing interests.


\section*{Software and Data}
The conceptual watershed models are implemented in \textsc{MATLAB}, \textsc{C++}, and \textsc{Python}. The software will be made publicly available at \url{https://github.com/jaspervrugt/diffhydrology} upon formal acceptance of this paper.

\appendix

\clearpage
\newpage

\appendix

\renewcommand\thesection{\Alph{section}}
\numberwithin{equation}{section}
\numberwithin{figure}{section}
\numberwithin{table}{section}
\numberwithin{algorithm}{section}

\titleformat{\section}{\normalfont\Large\bfseries}{Appendix \thesection:}{0.6em}{}

\section[\appendixname~\thesection]{Algorithmic recipes}\label{sec:AppendixA}
\renewcommand{\theequation}{\thesection.\arabic{equation}}
\setcounter{equation}{0}
\setcounter{algorithm}{0}
In this Appendix we present algorithmic recipes of gradient descent (Algorithm \ref{algApp:gradient_descent}) and the Levenberg-Marquardt method (Algorithm \ref{algApp:Levenberg_Marquardt}). 

We first present the gradient descent method.
\begin{algorithm}[H]
\caption{Gradient descent}
\begin{algorithmic}
\State \textbf{Input:} $\mathcal{L}:\mathbb{R}^{d} \to \mathbb{R}^{1}$ a total loss function such that $\mathcal{L} = \sum_{t=1}^{n} \mathcal{L}_{t}(y_{t},q_{t})$
\State \phantom{\textbf{Input:}} $\boldsymbol{\uptheta}_{(0)}$ an arbitrary initial solution
\State \textbf{Output:} $\boldsymbol{\uptheta}_{(\ast)}$ a local minimum of the total loss function $\mathcal{L}(\boldsymbol{\uptheta})$ \vspace{1mm}
\Statex \textbf{begin}
\Statex \hspace{1em} $k \leftarrow 0$
\Statex \hspace{1em} \textbf{while} not converged \textbf{and} $(k < k_{\max})$ \textbf{do}
\Statex \hspace{2em} $\boldsymbol{\uptheta}_{(k+1)} = \boldsymbol{\uptheta}_{(k)} - \eta_{(k)} \mathbf{g}_{n}(\boldsymbol{\uptheta}_{(k)})$
\Statex \hspace{2em} $\eta_{(k)} = \arg\min_{\eta>0} \mathcal{L}\!\left(\boldsymbol{\uptheta}_{(k)} - \eta\,\mathbf{g}_{n}(\boldsymbol{\uptheta}_{(k)})\right)$
\Statex \hspace{2em} $k \gets k+1$
\Statex \hspace{1em} \textbf{end while}
\Statex \textbf{end} \vspace{1mm}
\State \textbf{Return:} $\boldsymbol{\uptheta}_{(k)}$.
\end{algorithmic}
\label{algApp:gradient_descent}
\end{algorithm}

A more robust search algorithm is Levenberg-Marquardt which can switch adaptively between gradient descent and Gauss-Newton depending on the nature of the response surface.  
\begin{algorithm}[H]
\caption{Levenberg-Marquardt algorithm}
\begin{algorithmic}
\State \textbf{Input:} $\mathcal{L}:\mathbb{R}^{d} \to \mathbb{R}^{1}$ a total loss function such that $\mathcal{L} = \sum_{t=1}^{n} \mathcal{L}_{t}(y_{t},q_{t})$
\State \phantom{\textbf{Input:}} $\nu$ a damping multiplier
\State \phantom{\textbf{Input:}} $\boldsymbol{\uptheta}_{(0)}$ an arbitrary initial solution
\State \textbf{Output:} $\boldsymbol{\uptheta}_{(\ast)}$ a local minimum of the total loss function $\mathcal{L}(\boldsymbol{\uptheta})$ \vspace{1mm}
\Statex \textbf{begin}
\Statex \hspace{1em} $k \leftarrow 0$
\Statex \hspace{1em} $\lambda_{(0)} \leftarrow 10^{-3}\max\bigl\{\Diag\bigl(\mathbf{J}^{\top}_{q}(\boldsymbol{\uptheta}_{(0)})\mathbf{J}_{q}(\boldsymbol{\uptheta}_{(0)})\bigr)\bigr\}$
\Statex \hspace{1em} \textbf{while} not converged \textbf{and} $(k < k_{\max})$ \textbf{do}
\Statex \hspace{2em} $\boldsymbol{\uptheta}_{(k+1)} = \boldsymbol{\uptheta}_{(k)} - \eta_{(k)} \mathbf{g}_{n}(\boldsymbol{\uptheta}_{(k)})$
\Statex \hspace{2em} Compute $\Delta \boldsymbol{\uptheta}_{(k)} = - \bigl\{\mathbf{J}^{\top}_{q}(\boldsymbol{\uptheta}_{(k)})\mathbf{J}_{q}(\boldsymbol{\uptheta}_{(k)}) + \lambda_{(k)}\Diag \bigl(\mathbf{J}^{\top}_{q}(\boldsymbol{\uptheta}_{(k)})\mathbf{J}_{q}(\boldsymbol{\uptheta}_{(k)})\bigr)\bigr\}^{-1} \mathbf{J}^{\top}_{q}(\boldsymbol{\uptheta}_{(k)}) \, \mathbf{e}_{n}(\boldsymbol{\uptheta}_{(k)})$
\Statex \hspace{2em} \textbf{if} $F(\boldsymbol{\uptheta}_{(k)} + \Delta \boldsymbol{\uptheta}_{(k)}) < F(\boldsymbol{\uptheta}_{(k)})$ \textbf{then}
\Statex \hspace{3em} $\boldsymbol{\uptheta}_{(k+1)} =  \boldsymbol{\uptheta}_{(k)} + \Delta \boldsymbol{\uptheta}_{(k)}$ 
\Statex \hspace{3em} $\lambda_{(k+1)} = \lambda_{(k)}/\nu$
\Statex \hspace{2em} \textbf{else}
\Statex \hspace{3em} $\boldsymbol{\uptheta}_{(k+1)} =  \boldsymbol{\uptheta}_{(k)}$ 
\Statex \hspace{3em} $\lambda_{(k+1)} = \nu \lambda_{(k)}$
\Statex \hspace{2em} \textbf{endif} 
\Statex \hspace{2em} $k \gets k+1$
\Statex \hspace{1em} \textbf{end while} 
\Statex \textbf{end} \vspace{1mm}
\State \textbf{Return:} $\boldsymbol{\uptheta}_{(k)}$.
\end{algorithmic}
\label{algApp:Levenberg_Marquardt}
\end{algorithm}
This concludes the algorithmic recipes.

\newpage

\section[\appendixname~\thesection]{Jacobian matrices of system dynamics}\label{sec:AppendixB}
In this Appendix we review the \texttt{hymod}, \texttt{hmodel}, \texttt{sacsma} and \texttt{xinanjiang} conceptual watershed models and present analytic expressions of their Jacobian matrices of the system dynamics with respect to their states and parameters, respectively. The models are coded in \textsc{Matlab} and \textsc{C++} and use a mass-conservative second-order integration method with adaptive time step. This guarantees a robust and accurate numerical solution of the simulated fluxes, state variables and sensitivity matrices. Next, we discuss each of the models separately. 

\subsection{HYdrologic MODel}
\begin{figure}[H]
\centering\includegraphics[width=0.7\linewidth]{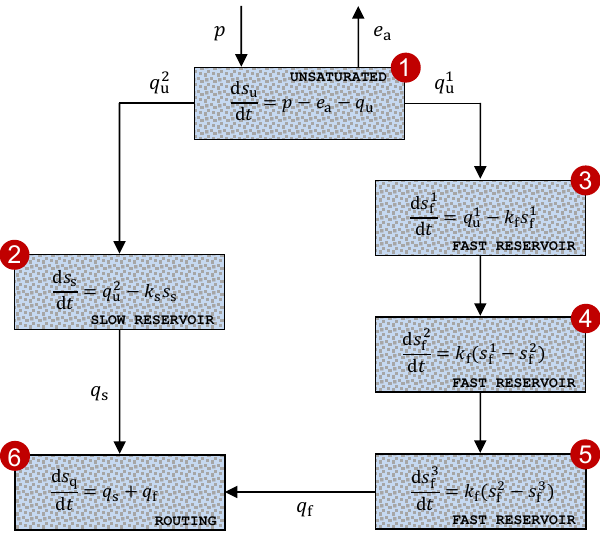}
\caption{Schematic illustration of the HYdrologic MODel of \cite{boyle2001}. Grey boxes, labeled in red, correspond to fictitious control volumes of the watershed which govern the rainfall-runoff transformation. The state variables, $s_\mathrm{u}$, $s_\mathrm{s}$, $s^{1}_\mathrm{f}$, $s^{2}_\mathrm{f}$, $s^{3}_\mathrm{f}$ and $s_{q}$, correspond to the water storage in each compartment. Arrows portray the fluxes into and out of the compartments, including precipitation, $p$, evaporation, $e_\mathrm{a}$, precipitation converted into flow, $q_\mathrm{u}$, fast flow, $q_\mathrm{f}$, and baseflow, $q_\mathrm{s}$. The fluxes are computed as follows, $q_\mathrm{u} = p( 1 - (1-\overline{s}_\mathrm{u})^{b} )$, $e_\mathrm{a} = e_\mathrm{p}\overline{s}_\mathrm{u}(1 + c)/(\overline{s}_\mathrm{u} + c)$, $q^{1}_\mathrm{u} = a q_\mathrm{u}$, $q^{2}_\mathrm{u} = (1-a) q_\mathrm{u}$, $q_\mathrm{f} = k_\mathrm{f}s^{3}_\mathrm{f}$ and $q_\mathrm{s} = k_\mathrm{s}s_\mathrm{s}$, where $e_\mathrm{p}$ signifies the potential evapotranspiration, $c = 10^{-2}$, $\overline{s}_\mathrm{u} = s_\mathrm{u}/s_\mathrm{u,max}$ and $s_\mathrm{u,max}$, $a$, $b$, $k_\mathrm{s}$ and $k_\mathrm{f}$ are unknown parameters.} 
\label{figApp:B1}
\end{figure}

The HYdrologic MODel (\texttt{hymod}) originates from the PhD thesis of \cite{boyle2001} and describes the rainfall-discharge relationship using five fictitious control volumes. These reservoirs simulate processes such as evaporation, percolation, river inflow and baseflow (see Figure \ref{figApp:B1}).
Thus, \texttt{hymod} has $m = 6$ state variables $\mathbf{x} = (s_\mathrm{u}, s_\mathrm{s}, s^{1}_\mathrm{f}, s^{2}_\mathrm{f}, s^{3}_\mathrm{f}, s_\mathrm{q})^{\top}$, where the sixth control volume $s_\mathrm{q}$ is an infinite reservoir which accumulates the discharge. The \texttt{hymod} state equations follow from conservation of mass in each control volume
\begin{linenomath*}
\begin{align}
\frac{\mathrm{d}\mathbf{x}}{\mathrm{d}t} =
\colvec{1}{c}{ \; p - e_\mathrm{a} - q_\mathrm{u}  \; \\[1mm]
\; (1 - a)\,q_\mathrm{u} - k_\mathrm{s} s_\mathrm{s} \; \\[1mm]
\; a \, q_\mathrm{u} - k_\mathrm{f} s^{1}_\mathrm{f} \; \\[1mm]
\; k_\mathrm{f}( s^{1}_\mathrm{f} - s^{2}_\mathrm{f}) \; \\[1mm]
\; k_\mathrm{f}(s^{2}_\mathrm{f} - s^{3}_\mathrm{f}) \; \\[1mm]
\; k_\mathrm{f} s^{3}_\mathrm{f} + k_\mathrm{s} s_\mathrm{s} \;} \in \mathbb{R}^{m \times 1}.
\end{align}
\end{linenomath*}
Table \ref{tableApp:B1} presents the $d = 5$ \texttt{hymod} parameters $\boldsymbol{\uptheta} = (s_\mathrm{u,max}, b, a, k_\mathrm{s}, k_\mathrm{f})^\top$ with their corresponding symbols, units, and lower and upper bounds.
\begin{table}[H]
\centering
\captionsetup[table]{position=bottom}
\begin{threeparttable}
\caption{Summary of \texttt{hymod} parameters and their symbols, units, and lower and upper bounds.}
\label{tableApp:B1}
\begin{tabular}{l c c c c }
\toprule
\multicolumn{1}{c}{ Parameter } & Symbol & Units & Min. & Max. \\
\midrule
Maximum storage unsaturated zone            & $s_\mathrm{u,max}$      & mm    & 50     & 1000  \\
Spatial variability of soil moisture capacity & $b$      & $-$    & $10^{-1}$  & 10 \\
Flow partitioning coefficient   & $a$      
& $-$  & 0     & 1 \\
Recession constant, slow reservoir   & $k_\mathrm{s}$        & d$^{-1}$     & $10^{-4}$     & 1 \\
Recession constant, fast reservoir   & $k_\mathrm{f}$        & d$^{-1}$     & $10^{-1}$     & 5 \\
\bottomrule
\end{tabular}
\end{threeparttable}
\end{table}

To construct the augmented ODE system used for analytic sensitivities, we require the $m \times m$ Jacobian matrix $\mathbf{J}_{f}(\mathbf{x})$ of the system dynamics with respect to the states
\begin{linenomath*}
\begin{align}
\mathbf{J}_{f}(\mathbf{x}) & = \frac{\partial \mathbf{f}(\mathbf{x},\boldsymbol{\uptheta},t)}{\partial \mathbf{x}^{\top}} \in \mathbb{R}^{6 \times 6}, \nonumber
\end{align}
\end{linenomath*}
and the $m \times d$ Jacobian matrix $\mathbf{J}_{f}(\boldsymbol{\uptheta})$ of the system state with respect to the parameters
\begin{linenomath*}
\begin{align}
\mathbf{J}_{f}(\boldsymbol{\uptheta}) & = \frac{\partial \mathbf{f}(\mathbf{x},\boldsymbol{\uptheta},t)}{\partial \boldsymbol{\uptheta}^{\top}} \in \mathbb{R}^{6 \times 5}. \nonumber
\end{align}
\end{linenomath*}
We already introduced the dimensionless storage, $\overline{s}_\mathrm{u} = s_\mathrm{u}/s_\mathrm{u,max}$, and the fluxes
\begin{linenomath*}
\begin{align}
q_\mathrm{u} & = p[1 - (1-\overline{s}_\mathrm{u})^{b}] \qquad \text{and} \qquad e_\mathrm{a} = e_\mathrm{p}\,\overline{s}_\mathrm{u}\,\frac{1+c}{\overline{s}_\mathrm{u} + c}. \nonumber
\intertext{using $c = 10^{-2}$. The derivatives for the two Jacobian matrices follow directly}
\frac{\partial \overline{s}_\mathrm{u}}{\partial s_\mathrm{u}} & = \frac{1}{s_{\mathrm{u,max}}} \nonumber \\
\frac{\partial q_\mathrm{u}}{\partial \overline{s}_\mathrm{u}} & = p\,b\,(1-\overline{s}_\mathrm{u})^{b-1} \nonumber \\
\frac{\partial q_\mathrm{u}}{\partial s_\mathrm{u}} & = \frac{p\,b\,(1 -\overline{s}_\mathrm{u})^{b-1}}{s_{\mathrm{u,max}}} \nonumber \\
\frac{\partial e_\mathrm{a}}{\partial \overline{s}_\mathrm{u}} & = e_\mathrm{p}(1+c)\,\frac{c}{(\overline{s}_\mathrm{u}+c)^{2}} \nonumber \\
\frac{\partial e_\mathrm{a}}{\partial s_\mathrm{u}} & = \frac{e_\mathrm{p}(1+c)c}{(\overline{s}_\mathrm{u}+c)^{2}s_{\mathrm{u,max}}}.\nonumber
\end{align}
\end{linenomath*}

\subsubsection{Jacobian of system dynamics with respect to states}
Using the \texttt{hymod} state equations, the $6 \times 6$ Jacobian of the system dynamics $\mathbf{f}(\mathbf{x},\boldsymbol{\uptheta},t)$ with respect to the state variables is
\begin{equation}
\mathbf{J}_{f}(\mathbf{x}) = \nabla_{\mathbf{x}}\mathbf{f}(\mathbf{x},\boldsymbol{\uptheta},t) = 
\colvec{1}{c}{ \; - \dfrac{\partial q_\mathrm{u}}{\partial s_\mathrm{u}} - \dfrac{\partial e_\mathrm{a}}{\partial s_\mathrm{u}} & 0 & 0 & 0 & 0 & 0 \; \\[3mm]
\; (1-a)\dfrac{\partial q_\mathrm{u}}{\partial s_\mathrm{u}} & - k_\mathrm{s} & 0 & 0 & 0 & 0 \; \\[3mm]
\; a \dfrac{\partial q_\mathrm{u}}{\partial s_\mathrm{u}} & 0 & - k_\mathrm{f} & 0 & 0 & 0 \; \\[3mm]
\; 0 & 0 & k_\mathrm{f} & - k_\mathrm{f} & 0 & 0 \; \\[3mm]
\; 0 & 0 & 0 & k_\mathrm{f} & - k_\mathrm{f} & 0 \; \\[3mm]
0 & k_\mathrm{s} & 0 & 0 & k_\mathrm{f} & 0 \;} \in \mathbb{R}^{m \times m}.
\label{eq:Jx_full}
\end{equation}
where
\begin{linenomath*}
\begin{align}
\frac{\partial q_\mathrm{u}}{\partial s_\mathrm{u}} & = \frac{p\,b(1-\overline{s}_\mathrm{u})^{b-1}}{s_{\mathrm{u,max}}} \qquad \text{and} \qquad \frac{\partial e_\mathrm{a}}{\partial s_\mathrm{u}} = \frac{e_\mathrm{p}(1+c)c}{(\overline{s}_\mathrm{u}+c)^{2}s_{\mathrm{u,max}}}. \nonumber
\end{align}
\end{linenomath*}
All partial derivatives with respect to $s_\mathrm{q}$ in the last column are zero because the routed discharge does not feed back into the other reservoirs.

\subsubsection{Jacobian of system dynamics with respect to parameters}
The parameter Jacobian $\mathbf{J}_{f}(\boldsymbol{\uptheta})$ is sparse and follows directly from the flux equations. Let us first look at the derivatives w.r.t. parameter $s_\mathrm{u,max}$
\begin{linenomath*}
\begin{align}
& \frac{\partial \, \overline{s}_\mathrm{u}}{\partial s_\mathrm{u,max}} = - \frac{s_\mathrm{u}}{s_\mathrm{u,max}^{2}}, \nonumber \\
& \frac{\partial q_\mathrm{u}}{\partial s_\mathrm{u,max}} = p \, b(1-\overline{s}_\mathrm{u})^{b-1}
\!\left(-\frac{s_\mathrm{u}} {s_\mathrm{u,max}^{2}}\right), \nonumber \\
& \frac{\partial e_\mathrm{a}}{\partial s_\mathrm{u,max}} = e_\mathrm{p}(1+c)\frac{c}{(\overline{s}_\mathrm{u} + c)^{2}}
\!\biggl(-\frac{s_\mathrm{u}}{s_\mathrm{u,max}^{2}}\biggr). \nonumber
\intertext{Next, we look at the derivatives w.r.t. parameter $b$}
& \frac{\partial q_\mathrm{u}}{\partial b} = - p (1 - \overline{s}_\mathrm{u})^{b}\log(1 -\overline{s}_\mathrm{u}) \qquad \frac{\partial e_\mathrm{a}}{\partial b} = 0 \nonumber
\intertext{and w.r.t. parameter $a$}
& \frac{\partial f_{2}}{\partial a} = - q_\mathrm{u} \qquad \frac{\partial f_{3}}{\partial a} = q_\mathrm{u} \nonumber
\intertext{and w.r.t. parameter $k_\mathrm{s}$}
& \frac{\partial f_{2}}{\partial k_\mathrm{s}} = -s_\mathrm{s} \qquad \frac{\partial f_{6}}{\partial k_\mathrm{s}} = s_\mathrm{s}. \nonumber
\intertext{Finally, we back out the derivatives w.r.t. parameter $k_\mathrm{f}$}
& \frac{\partial f_{3}}{\partial k_\mathrm{f}} = - s^{1}_\mathrm{f} \qquad \frac{\partial f_{4}}{\partial k_\mathrm{f}} = s^{1}_\mathrm{f} - s^{2}_\mathrm{f} \qquad \frac{\partial f_{5}}{\partial k_\mathrm{f}} = s^{2}_\mathrm{f} - s^{3}_\mathrm{f} \qquad \frac{\partial f_{6}}{\partial k_\mathrm{f}} = s^{3}_\mathrm{f}. \nonumber
\end{align}
\end{linenomath*}
We can now collect all these expressions and define the Jacobian of the model states with respect to parameters
\begin{linenomath*}
\begin{align} 
\mathbf{J}_{f}(\boldsymbol{\uptheta}) =
\colvec{1}{c}{ \; - \dfrac{\partial q_\mathrm{u}}{\partial s_\mathrm{u,max}} - \dfrac{\partial e_\mathrm{a}}{\partial s_\mathrm{u,max}} & - \dfrac{\partial q_\mathrm{u}}{\partial b} & 0 & 0 & 0 \; \\[3mm]
\; (1-a)\dfrac{\partial q_\mathrm{u}}{\partial s_\mathrm{u,max}} & (1-a)\dfrac{\partial q_\mathrm{u}}{\partial b} & - \, q_\mathrm{u} &
- \, s_\mathrm{s} & 0 \; \\[3mm]
\; a \dfrac{\partial q_\mathrm{u}}{\partial s_\mathrm{u,max}} & a \dfrac{\partial q_\mathrm{u}}{\partial b} & q_\mathrm{u} & 0 &
- \, s^{1}_\mathrm{f} \; \\[3mm]
\; 0 & 0 & 0 & 0 & s^{1}_\mathrm{f} - s^{2}_\mathrm{f} \; \\[3mm]
\; 0 & 0 & 0 & 0 & s^{2}_\mathrm{f} - s^{3}_\mathrm{f} \; \\[3mm]
\; 0 & 0 & 0 & s_\mathrm{s} & s^{3}_\mathrm{f} \; }
\in \mathbb{R}^{m \times d}.
\label{eq:Jtheta_final}
\end{align}
\end{linenomath*}

These matrices complete the analytic specification of the augmented ODE system used to compute the sensitivity matrix $\mathbf{S} = \partial\mathbf{x}/\partial\boldsymbol{\uptheta}^{\top}$ and, by differencing the accumulated discharge state, the exact Jacobian $\mathbf{J}_{q}(\boldsymbol{\uptheta})$ of the simulated hydrograph.

\subsection{Hydrologic model}
\begin{figure}[H]
\centering\includegraphics[width=0.7\linewidth]{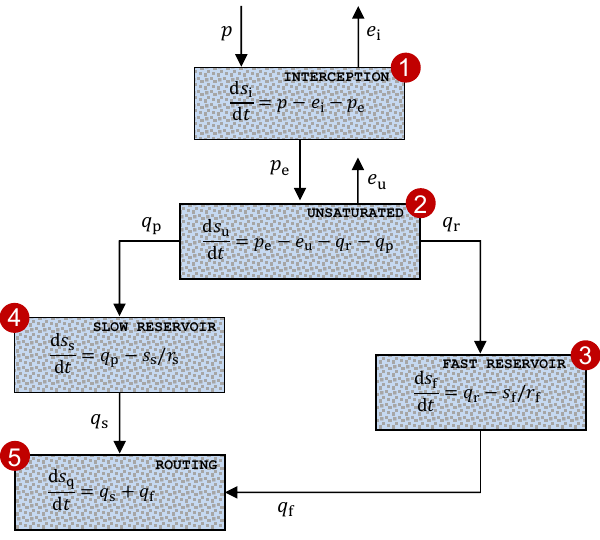}
\caption{Schematic illustration of the \texttt{hmodel} after \cite{schoups2010a}. Grey boxes, labeled in red, correspond to fictitious control volumes of the watershed which govern the rainfall-runoff transformation. The state variables, $s_\mathrm{i}$, $s_\mathrm{u}$, $s_\mathrm{f}$, $s_\mathrm{s}$ and $s_\mathrm{q}$, correspond to the water storage in each compartment. Arrows portray the fluxes into and out of the compartments, including precipitation, $p$, interception evaporation, $e_\mathrm{i}$, excess precipitation, $p_\mathrm{e}$, evaporation, $e_\mathrm{u}$, surface runoff, $q_\mathrm{r}$, percolation, $q_\mathrm{p}$, fast flow, $q_\mathrm{f}$ and baseflow, $q_\mathrm{s}$. The fluxes are computed as follows, $e_\mathrm{i} = e_\mathrm{p} \, \phi(\overline{s}_\mathrm{i},\alpha_\mathrm{i})$, $p_\mathrm{e} = p\,\phi(\overline{s}_\mathrm{i},\alpha_\mathrm{p})$, $e_\mathrm{u} = (e_\mathrm{p} - e_\mathrm{i})\,\phi(\overline{s}_\mathrm{u},\alpha_\mathrm{e})$, $q_\mathrm{r} = p_\mathrm{e} \, \phi(\overline{s}_\mathrm{u},\alpha_\mathrm{f})$, $q_\mathrm{p} = q_\mathrm{max}\,\phi(\overline{s}_\mathrm{u},\alpha_\mathrm{f})$, $q_\mathrm{f} = s_\mathrm{f}/r_\mathrm{f}$ and $q_\mathrm{s} = s_\mathrm{s}/r_\mathrm{s}$, where $e_\mathrm{p}$ signifies the potential evapotranspiration, the functions, $\phi(x,y) = \{1 - \varphi(- x y)\}/\{1 - \varphi(-x)\}$ and $\varphi(x) = \exp(\min\{x,300\})$, protect against overflow, $\alpha_\mathrm{i} = 50$, $\alpha_\mathrm{p} = -50$, $\overline{s}_\mathrm{i} = s_\mathrm{i} / i_\mathrm{max}$, $\overline{s}_\mathrm{u} = s_\mathrm{u}/s_\mathrm{max}$ and $i_\mathrm{max}$, $s_\mathrm{max}$, $q_\mathrm{max}$, $\alpha_\mathrm{e}$, $\alpha_\mathrm{f}$, $r_\mathrm{f}$ and $r_\mathrm{s}$ are unknown parameters.}
\label{figApp:B2}
\end{figure}

The Hydrologic model (\texttt{hmodel}) is a parsimonious conceptual watershed model originally developed by \cite{schoups2010a}. This model transforms rainfall into runoff at the watershed outlet using an interception, unsaturated zone, fast and slow flow reservoir, respectively, which simulate interception, throughfall, evaporation, surface runoff, percolation, fast streamflow and baseflow (see Figure \ref{figApp:B2}).
Thus, \texttt{hmodel} has $m = 5$ state variables $\mathbf{x} = (s_\mathrm{i}, s_\mathrm{u}, s_\mathrm{f}, s_\mathrm{s}, s_\mathrm{q})^{\top}$, where the fifth control volume $s_\mathrm{q}$ is an infinite reservoir which accumulates the discharge. The \texttt{hmodel} state equations follow from conservation of mass in each control volume
\begin{linenomath*}
\begin{align}
\frac{\mathrm{d}\mathbf{x}}{\mathrm{d}t} =
\colvec{1}{c}{\; p - e_\mathrm{i} - p_\mathrm{e} \; \\[1mm]
\; p_\mathrm{e} - e_\mathrm{u} - q_\mathrm{r} - q_\mathrm{p} \; \\[1mm]
\; q_\mathrm{r} - q_\mathrm{f} \; \\[1mm]
\; q_\mathrm{p} - q_\mathrm{s} \; \\[1mm]
\; q_\mathrm{s} + q_\mathrm{f} \; }
\end{align}
\end{linenomath*}
Table \ref{tableApp:B2} lists the $d = 7$ \texttt{hmodel} parameters $\boldsymbol{\uptheta} = (i_\mathrm{max}, s_\mathrm{max}, q_\mathrm{max}, \alpha_\mathrm{e}, \alpha_\mathrm{f}, r_\mathrm{f}, r_\mathrm{s})^\top$ and their corresponding symbols, units and upper and lower bounds. 
\begin{table}[H]
\centering
\captionsetup[table]{position=bottom}
\begin{threeparttable}
\caption{Description of \texttt{hmodel} parameters, including symbols, units, lower and upper bounds.}
\label{tableApp:B2}
\begin{tabular}{l c c c c }
\toprule
\multicolumn{1}{c}{ Parameter } & Symbol & Units & Min. & Max. \\
\midrule
Maximum interception            & $i_\mathrm{max}$      & mm    & 0.1     & 10  \\
Soil water storage capacity     & $s_\mathrm{max}$      & mm    & 10    & 1000 \\
Maximum percolation rate        & $q_\mathrm{max}$      & mm\,d$^{-1}$  & $10^{-1}$ & 100 \\
Evaporation parameter           & $\alpha_\mathrm{e}$   & $-$   & 0     & 100 \\
Runoff parameter                & $\alpha_\mathrm{f}$   & $-$   &  -10   & 10 \\
Time constant, fast reservoir   & $r_\mathrm{f}$        & d     & $10^{-1}$     & 10 \\
Time constant, slow reservoir   & $r_\mathrm{s}$        & d     & 1     & 150 \\
\bottomrule
\end{tabular}
\end{threeparttable}
\end{table}

To construct the augmented ODE system used for analytic sensitivities, we require the $m \times m$ Jacobian matrix $\mathbf{J}_{f}(\mathbf{x})$ of the system dynamics with respect to the states
\begin{linenomath*}
\begin{align}
\mathbf{J}_{f}(\mathbf{x}) & = \frac{\partial \mathbf{f}(\mathbf{x},\boldsymbol{\uptheta},t)}{\partial \mathbf{x}^{\top}} \in \mathbb{R}^{5 \times 5}, \nonumber
\end{align}
\end{linenomath*}
and the $m \times d$ Jacobian matrix $\mathbf{J}_{f}(\boldsymbol{\uptheta})$ of the system state with respect to the parameters
\begin{linenomath*}
\begin{align}
\mathbf{J}_{f}(\boldsymbol{\uptheta}) & = \frac{\partial \mathbf{f}(\mathbf{x},\boldsymbol{\uptheta},t)}{\partial \boldsymbol{\uptheta}^{\top}} \in \mathbb{R}^{5 \times 7}. \nonumber
\end{align}
\end{linenomath*}
We already introduced the dimensionless storages, $
\overline{s}_\mathrm{i} = s_\mathrm{i}/i_\mathrm{max}$ and $\overline{s}_\mathrm{u} = s_\mathrm{u}/s_\mathrm{max}$ and the nonlinear transformation functions
\begin{linenomath*}
\begin{align}
& \phi(x,y) = \frac{1 - \varphi(-xy)}{1 - \varphi(-x)} \qquad \text{and} \qquad \varphi(x) = \exp(\min\{x,300\}). \nonumber
\end{align}
\end{linenomath*}

\subsubsection{Jacobian of system dynamics with respect to states}
We first compute the basic derivatives needed for the Jacobian. We start with the derivatives of the nonlinear $\phi$-function
\begin{linenomath*}
\begin{align}
& \frac{\partial \phi}{\partial x} =
\frac{-\varphi^{\prime}(-xy)(-y)(1-\varphi(-x)) - (1-\varphi(-xy))(+\varphi^{\prime}(-x))}{(1-\varphi(-x))^{2}} \nonumber \\[2mm]
& \frac{\partial \phi}{\partial y} =
\frac{-\varphi^{\prime}(-xy)(-x)}{1-\varphi(-x)}, \nonumber
\intertext{where $\varphi^{\prime}(x) = \exp(\min\{x,300\})$ when $x < 300$ and $0$ otherwise. The chain-rule gives}
& \frac{\partial \phi(\overline{s}_\mathrm{i},y)}{\partial s_\mathrm{i}} = \frac{1}{i_\mathrm{max}}\frac{\partial \phi}{\partial x} \qquad \frac{\partial \phi(\overline{s}_\mathrm{u},y)}{\partial s_\mathrm{u}} = \frac{1}{s_\mathrm{max}}\frac{\partial \phi}{\partial x}. \nonumber 
\intertext{We use these identities to develop analytic expressions for the derivatives of fluxes. For example}
& \frac{\partial e_\mathrm{i}}{\partial s_\mathrm{i}}
= e_\mathrm{p}\frac{\partial \phi(\overline{s}_\mathrm{i},\alpha_\mathrm{i})}{\partial s_\mathrm{i}} \nonumber \\[1mm]
& \frac{\partial p_\mathrm{e}}{\partial s_\mathrm{i}}
= p\frac{\partial \phi(\overline{s}_\mathrm{i},\alpha_\mathrm{p})}{\partial s_\mathrm{i}} \nonumber \\[1mm]
& \frac{\partial e_\mathrm{u}}{\partial s_\mathrm{u}}
= (e_\mathrm{p} - e_\mathrm{i})\frac{\partial \phi(\overline{s}_\mathrm{u},\alpha_\mathrm{e})}{\partial s_\mathrm{u}} \nonumber \\[1mm]
& \frac{\partial q_\mathrm{r}}{\partial s_\mathrm{u}} = p_\mathrm{e}\frac{\partial \phi(\overline{s}_\mathrm{u},\alpha_\mathrm{f})}{\partial s_\mathrm{u}} \nonumber \\[1mm]
& \frac{\partial q_\mathrm{p}}{\partial s_\mathrm{u}} = q_\mathrm{max}\frac{\partial \phi(\overline{s}_\mathrm{u},\alpha_\mathrm{f})}{\partial s_\mathrm{u}} \nonumber \\[1mm]
& \frac{\partial q_\mathrm{f}}{\partial s_\mathrm{f}} = \frac{1}{r_\mathrm{f}} \nonumber \\[1mm]
& \frac{\partial q_\mathrm{s}}{\partial s_\mathrm{s}} = \frac{1}{r_\mathrm{s}}. \nonumber
\end{align}
\end{linenomath*}
All other cross-derivatives are zero.

We can now enter these derivatives in the Jacobian state matrix 
\begin{equation}
\mathbf{J}_{f}(\mathbf{x}) = \nabla_{\mathbf{x}}\mathbf{f}(\mathbf{x},\boldsymbol{\uptheta},t) = \colvec{1}{c}{ \; - \dfrac{\partial e_\mathrm{i}}{\partial s_\mathrm{i}} - \dfrac{\partial p_\mathrm{e}}{\partial s_\mathrm{i}} & 0 & 0 & 0 &  0 \; \\[3mm]
\; \dfrac{\partial p_\mathrm{e}}{\partial s_\mathrm{i}} & - \dfrac{\partial e_\mathrm{u}}{\partial s_\mathrm{u}} - \dfrac{\partial q_\mathrm{r}}{\partial s_\mathrm{u}} - \dfrac{\partial q_\mathrm{p}}{\partial s_\mathrm{u}} & 0 & 0 & 0 \; \\[3mm]
\; 0 & \dfrac{\partial q_\mathrm{r}}{\partial s_\mathrm{u}} & - \dfrac{\partial q_\mathrm{f}}{\partial s_\mathrm{f}} & 0 & 0 \; \\[3mm]
\; 0 & \dfrac{\partial q_\mathrm{p}}{\partial s_\mathrm{u}} & 0 & -\dfrac{\partial q_\mathrm{s}}{\partial s_\mathrm{s}} & 0 \; \\[3mm]
\; 0 & 0 & \dfrac{\partial q_\mathrm{f}}{\partial s_\mathrm{f}} & \dfrac{\partial q_\mathrm{s}}{\partial s_\mathrm{s}} & 0 \;}.
\end{equation}

\subsubsection{Jacobian of system dynamics with respect to parameters}
We list the parameter derivatives needed for the augmented ODE. We start with the derivatives w.r.t. parameter $i_\mathrm{max}$
\begin{linenomath*}
\begin{align}
\frac{\partial \overline{s}_\mathrm{i}}{\partial i_\mathrm{max}} & = - \frac{s_\mathrm{i}}{i^{2}_\mathrm{max}} & \frac{\partial e_\mathrm{i}}{\partial i_\mathrm{max}} & = e_\mathrm{p}\frac{\partial \phi}{\partial x}\biggl( - \frac{s_\mathrm{i}}{i^{2}_\mathrm{max}}\biggr) \nonumber \\
\frac{\partial p_\mathrm{e}}{\partial i_\mathrm{max}} & = p\frac{\partial \phi}{\partial x}\biggl( - \frac{s_\mathrm{i}}{i^{2}_\mathrm{max}}\biggr). \nonumber 
\intertext{Then, we proceed with the derivatives w.r.t. parameter $s_\mathrm{max}$}
\frac{\partial \overline{s}_\mathrm{u}}{\partial s_\mathrm{max}} & = - \frac{s_\mathrm{u}}{s^{2}_\mathrm{max}} & \frac{\partial e_\mathrm{u}}{\partial s_\mathrm{max}} & = (e_\mathrm{p} - e_\mathrm{i})\frac{\partial \phi}{\partial x}\biggl(-\frac{s_\mathrm{u}}{s^{2}_\mathrm{max}}\biggr) \nonumber \\
\frac{\partial q_\mathrm{r}}{\partial s_\mathrm{max}} & = p_\mathrm{e}\frac{\partial \phi}{\partial x}\biggl(-\frac{s_\mathrm{u}}{s^{2}_\mathrm{max}}\biggr) \nonumber \\
\frac{\partial q_\mathrm{p}}{\partial s_\mathrm{max}} & = q_\mathrm{max}\frac{\partial \phi}{\partial x}\biggl( -\frac{s_\mathrm{u}}{s^{2}_\mathrm{max}}\biggr). \nonumber 
\intertext{The derivatives w.r.t. parameter $q_\mathrm{max}$}
\frac{\partial q_\mathrm{p}}{\partial q_\mathrm{max}} & = \phi(\overline{s}_\mathrm{u}, \alpha_\mathrm{f}). \nonumber
\intertext{Finally, for the derivatives w.r.t. $\alpha_\mathrm{e}$, and $\alpha_\mathrm{f}$ we must look at} 
\frac{\partial \phi}{\partial y} & = \frac{-\varphi^{\prime}(-xy)(-x)}{1-\varphi(-x)}. \nonumber
\intertext{Thus,}
\frac{\partial e_\mathrm{u}}{\partial \alpha_\mathrm{e}} & = (e_\mathrm{p}-e_\mathrm{i}) \frac{\partial \phi(\overline{s}_\mathrm{u},\alpha_\mathrm{e})}{\partial y} \nonumber \\[2mm]
\frac{\partial q_\mathrm{r}}{\partial \alpha_\mathrm{f}} & = p_\mathrm{e} \frac{\partial \phi(\overline{s}_\mathrm{u},\alpha_\mathrm{f})}{\partial y} \nonumber \\[2mm]
\frac{\partial q_\mathrm{p}}{\partial \alpha_\mathrm{f}} & = q_\mathrm{max} \frac{\partial \phi(\overline{s}_\mathrm{u},\alpha_\mathrm{f})}{\partial y}. \nonumber 
\intertext{Finally, we yield the derivative w.r.t. parameters $r_\mathrm{f}$ and $r_\mathrm{s}$}
\frac{\partial q_\mathrm{f}}{\partial r_\mathrm{f}} & = - \frac{s_\mathrm{f}}{r_\mathrm{f}^{2}} & \frac{\partial q_\mathrm{s}}{\partial r_\mathrm{s}} & = - \frac{s_\mathrm{s}}{r_\mathrm{s}^{2}}. \nonumber 
\end{align}
\end{linenomath*}

The parameter Jacobian is now equal to
\begin{equation}
\mathbf{J}_{f}(\boldsymbol{\uptheta}) =
\colvec{1}{c}{ \; \dfrac{\partial f_\mathrm{1}}{\partial i_\mathrm{max}} & 0 & 0 & 0 & 0 & 0 & 0 \;
\\[4mm]
\dfrac{\partial f_\mathrm{2}}{\partial i_\mathrm{max}} & \dfrac{\partial f_\mathrm{2}}{\partial s_\mathrm{max}} & \dfrac{\partial f_\mathrm{2}}{\partial q_\mathrm{max}} & \dfrac{\partial f_\mathrm{2}}{\partial \alpha_\mathrm{e}} & \dfrac{\partial f_\mathrm{2}}{\partial \alpha_\mathrm{f}} & 0 & 0 \;
\\[4mm]
\; 0 & \dfrac{\partial f_\mathrm{3}}{\partial s_\mathrm{max}} & 0 & 0 & \dfrac{\partial f_\mathrm{3}}{\partial \alpha_\mathrm{f}} & 
\dfrac{\partial f_{3}}{\partial r_\mathrm{f}} & 0 \;
\\[4mm]
\; 0 & \dfrac{\partial f_\mathrm{4}}{\partial s_\mathrm{max}} & \dfrac{\partial f_\mathrm{4}}{\partial q_\mathrm{max}} & 0 & \dfrac{\partial f_\mathrm{4}}{\partial \alpha_\mathrm{f}} & 0 & \dfrac{\partial f_{4}}{\partial r_\mathrm{s}} \;
\\[4mm]
\; 0 & 0 & 0 & 0 & 0 & \dfrac{\partial f_{5}}{\partial r_\mathrm{f}} & \dfrac{\partial f_{5}}{\partial r_\mathrm{s}} \; },
\end{equation}
where the nonzero entries are
\begin{linenomath*}
\begin{align}
\frac{\partial f_{1}}{\partial i_\mathrm{max}}
& = - \frac{\partial e_\mathrm{i}}{\partial i_\mathrm{max}} - \frac{\partial p_\mathrm{e}}{\partial i_\mathrm{max}} \nonumber \\[1mm]
\frac{\partial f_{2}}{\partial i_\mathrm{max}}
& = \frac{\partial p_\mathrm{e}}{\partial i_\mathrm{max}} &  \frac{\partial f_{2}}{\partial s_\mathrm{max}} & = -\frac{\partial e_\mathrm{u}}{\partial s_\mathrm{max}} - \frac{\partial q_\mathrm{r}}{\partial s_\mathrm{max}} - \frac{\partial q_\mathrm{p}}{\partial s_\mathrm{max}} \nonumber \\[1mm]
\frac{\partial f_{2}}{\partial q_\mathrm{max}} & = -\frac{\partial q_\mathrm{p}}{\partial q_\mathrm{max}} = -\phi(\overline{s}_\mathrm{u},\alpha_\mathrm{f}) & \frac{\partial f_{2}}{\partial \alpha_\mathrm{e}} & = - \frac{\partial e_\mathrm{u}}{\partial \alpha_\mathrm{e}} = - (e_\mathrm{p} - e_\mathrm{i}) \frac{\partial \phi(\overline{s}_\mathrm{u},\alpha_\mathrm{e})}{\partial y} \nonumber \\[1mm]
\frac{\partial f_{2}}{\partial \alpha_\mathrm{f}} & = - \frac{\partial q_\mathrm{r}}{\partial \alpha_\mathrm{f}} - \frac{\partial q_\mathrm{p}}{\partial \alpha_\mathrm{f}} \nonumber \\[1mm]
\frac{\partial f_{3}}{\partial s_\mathrm{max}} & = \frac{\partial q_\mathrm{r}}{\partial s_\mathrm{max}} & \frac{\partial f_{3}}{\partial \alpha_\mathrm{f}} & = \frac{\partial q_\mathrm{r}}{\partial \alpha_\mathrm{f}} \nonumber \\[1mm] \displaybreak
\frac{\partial f_{3}}{\partial r_\mathrm{f}} & = -\frac{\partial q_\mathrm{f}}{\partial r_\mathrm{f}} = - \Bigl(-\frac{s_\mathrm{f}}{r_\mathrm{f}^{2}}\Bigr) = \frac{s_\mathrm{f}}{r_\mathrm{f}^{2}} \nonumber \\[1mm]
\frac{\partial f_{4}}{\partial s_\mathrm{max}} & = \frac{\partial q_\mathrm{p}}{\partial s_\mathrm{max}} & \frac{\partial f_{4}}{\partial q_\mathrm{max}} & = \frac{\partial q_\mathrm{p}}{\partial q_\mathrm{max}} = \phi(\overline{s}_\mathrm{u},\alpha_\mathrm{f}) \nonumber \\[1mm] 
\frac{\partial f_{4}}{\partial \alpha_\mathrm{f}} & = \frac{\partial q_\mathrm{p}}{\partial \alpha_\mathrm{f}} \nonumber 
& \frac{\partial f_{4}}{\partial r_\mathrm{s}} & = -\frac{\partial q_\mathrm{s}}{\partial r_\mathrm{s}} = - \Bigl(-\frac{s_\mathrm{s}}{r_\mathrm{s}^{2}}\Bigr) = \frac{s_\mathrm{s}}{r_\mathrm{s}^{2}} \nonumber \\[1mm]
\frac{\partial f_{5}}{\partial r_\mathrm{f}} & = \frac{\partial q_\mathrm{f}}{\partial r_\mathrm{f}} = - \frac{s_\mathrm{f}}{r_\mathrm{f}^{2}} & \frac{\partial f_{5}}{\partial r_\mathrm{s}} & = \frac{\partial q_\mathrm{s}}{\partial r_\mathrm{s}} = - \frac{s_\mathrm{s}}{r_\mathrm{s}^{2}}. \nonumber
\end{align}
\end{linenomath*}
These expressions complete the analytic specification of the Jacobian matrices required for the augmented ODE system used to compute the sensitivity matrix
$\mathbf{S} = \partial\mathbf{x}/\partial\boldsymbol{\uptheta}^{\top}$ and, by differencing the accumulated discharge state, the exact Jacobian of the simulated hydrograph.

\subsection{Sacramento Soil Moisture Accounting model}
\begin{figure}[H]
\centering\includegraphics[width=1.0\linewidth]{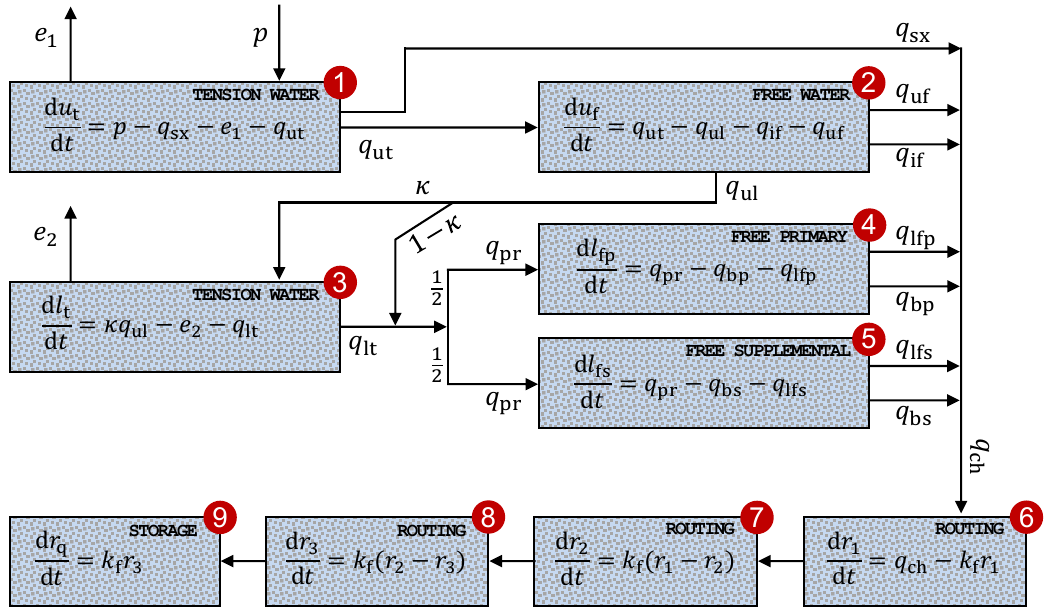}
\caption{Schematic illustration of the \texttt{sacsma} model after \cite{burnash1973} and \cite{clark2008}. Blue boxes labeled in red correspond to fictitious control volumes which govern the rainfall-runoff transformation. The model has eight state variables, including the free water storages of the upper soil layer, $u_\mathrm{f}$, and primary, $l_\mathrm{fp}$, and secondary, $l_\mathrm{fs}$, base flow reservoirs, the tension water storages of the upper, $u_\mathrm{t}$, and lower, $l_\mathrm{t}$, soil layers, the water levels, $r_{1}$, $r_{2}$ and $r_{3}$, of the three routing reservoirs and the storage $r_\mathrm{q}$ of the discharge reservoir. Arrows portray fluxes in and out of the compartments, including precipitation, $p_{t}$, evaporation from the upper soil layer, $e_\mathrm{1}$, overflow from tension storage in upper soil layer, $q_\mathrm{ut}$, surface runoff, $q_\mathrm{sx}$, overflow from free storage in upper soil layer, $q_\mathrm{uf}$, interflow, $q_\mathrm{if}$, percolation from upper to lower layer, $q_\mathrm{ul}$, evaporation from lower soil layer, $e_{2}$, overflow from tension storage in lower soil layer, $q_\mathrm{lt}$, flow into primary and supplemental storage, $q_\mathrm{pr}$, overflow from primary $q_\mathrm{lfp}$ and secondary $q_\mathrm{lfs}$ base flow storage in the lower soil layer and base flow from primary $q_\mathrm{bp}$ and secondary $q_\mathrm{bs}$ reservoirs. These fluxes are computed as follows,
$e_\mathrm{1} = e_\mathrm{p}(u_\mathrm{t}/u_\mathrm{t,max})$, $q_\mathrm{sx} = a_\mathrm{c,max}(u_\mathrm{t}/u_\mathrm{t,max})p_{t}$, $q_\mathrm{ut} = (p_{t} - q_\mathrm{sx})\phi(u_\mathrm{t},u_\mathrm{t,max})$, $q_\mathrm{ul} = q_{0}d_\mathrm{lz}(u_\mathrm{f}/u_\mathrm{f,max})$, $q_\mathrm{uf} = q_\mathrm{ut}\phi(u_\mathrm{f},u_\mathrm{f,max})$, $q_\mathrm{if} = k_\mathrm{i}(u_\mathrm{f}/u_\mathrm{f,max})$, $e_\mathrm{2} = (e_\mathrm{p} - e_{1})(l_\mathrm{t}/l_\mathrm{t,max})$, $q_\mathrm{lt} = \kappa q_\mathrm{ul}\phi(l_\mathrm{t},l_\mathrm{t,max})$, $q_\mathrm{pr} = \tfrac{1}{2}(1-\kappa)q_\mathrm{ul} + \tfrac{1}{2}q_\mathrm{lt}$, $q_\mathrm{lfp} = q_\mathrm{pr} \phi(l_\mathrm{fp},l_\mathrm{fp,max})$, $q_\mathrm{lfs} = q_\mathrm{pr} \phi(l_\mathrm{fs},l_\mathrm{fs,max})$, $q_\mathrm{bp} = \nu_\mathrm{p} l_\mathrm{fp}$, $q_\mathrm{bs} = \nu_\mathrm{s} l_\mathrm{fs}$, where $e_{\text{p}}$ is the potential evapotranspiration, $q_{0} = \nu_\mathrm{p}l_\mathrm{fp,max} + \nu_\mathrm{s}l_\mathrm{fs,max}$, $d_\mathrm{lz} = 1 + \alpha[(l_\mathrm{t} + l_\mathrm{fp} + l_\mathrm{fs})/(l_\mathrm{t,max} + l_\mathrm{fp,max} + l_\mathrm{fs,max})]^{\psi}$, the smoothing function $\phi(x_{1},x_{2}) = \{1 + \exp[(x_{2} - \varsigma \, \varrho \, x_{2} - x_{1})/(\varrho \, x_{2})]\}^{-1}$ with $\varsigma = 5$ and $\varrho = 10^{-2}$, and $u_\mathrm{f,max}$, $u_\mathrm{t,max}$, $l_\mathrm{fp,max}$, $l_\mathrm{fs,max}$, $l_\mathrm{t,max}$, $\alpha$, $\psi$, $k_\mathrm{i}$, $\kappa$, $\nu_\mathrm{p}$, $\nu_\mathrm{s}$ and $a_\mathrm{c,max}$ are unknown parameters. Channel inflow $q_\mathrm{ch} = q_\mathrm{sx} + q_\mathrm{uf} + q_\mathrm{if} + q_\mathrm{lfp} + q_\mathrm{bp} + q_\mathrm{lfs} + q_\mathrm{bs}$ is routed through three linear reservoirs with common recession constant $k_\mathrm{f}$ and yields the streamflow $q_{t} = k_\mathrm{f}r_{3}$ at the watershed outlet.}
\label{figApp:B3}
\end{figure}

The Sacramento Soil Moisture Accounting (\texttt{sacsma}) model is used by the National Weather Service River Forecast System for flood forecasting throughout the United States. The model converts areal average precipitation into streamflow \citep{burnash1973}. Our implementation follows \cite{clark2008} and is presented in Figure \ref{figApp:B3}.
Thus, the \texttt{sacsma} model has $m = 9$ state variables $\mathbf{x} = (u_\mathrm{t}, u_\mathrm{f}, l_\mathrm{t}, l_\mathrm{fp}, l_\mathrm{fs}, r_\mathrm{1}, r_\mathrm{2}, r_\mathrm{3},  r_\mathrm{q})^{\top}$, where the ninth control volume $r_\mathrm{q}$ is an infinite reservoir which accumulates the discharge. From the schematic and the caption the \texttt{sacsma} flux equations are
\begin{linenomath*}
\begin{align}
\frac{\mathrm{d}\mathbf{x}}{\mathrm{d}t}
= \mathbf{f}(\mathbf{x},\boldsymbol{\uptheta},t) = \colvec{1}{c}{ \; p_t - q_\mathrm{sx} - q_\mathrm{ut} - e_\mathrm{1} \; \\[1mm]
\; q_\mathrm{ut} - q_\mathrm{ul} - q_\mathrm{uf} - q_\mathrm{if} \; \\[1mm]
\; \kappa q_\mathrm{ul} - q_\mathrm{lt} - e_\mathrm{2} \; \\[1mm]
\; q_\mathrm{pr} - q_\mathrm{lfp} - q_\mathrm{bp} \; \\[1mm]
\; q_\mathrm{pr} - q_\mathrm{lfs} - q_\mathrm{bs} \; \\[1mm]
\; q_\mathrm{ch} - k_\mathrm{f} r_1 \; \\[1mm]
\; k_\mathrm{f} r_1 - k_\mathrm{f} r_2 \; \\[1mm]
\; k_\mathrm{f} r_2 - k_\mathrm{f} r_3 \; }  \in \mathbb{R}^{m \times 1}.
\end{align}
\end{linenomath*}
A mass-conservative second-order integration method with adaptive time stepping solves the state variables, $u_\mathrm{t}$, $u_\mathrm{f}$, $l_\mathrm{t}$, $l_\mathrm{fp}$, $l_\mathrm{fs}$, $r_{1}$, $r_{2}$, $r_{3}$ and $r_\mathrm{q}$, and fluxes $q_\mathrm{xx}$ of the control volumes using daily time series of areal average rainfall $(p_{1},\ldots,p_{n})^{\top}$ and potential evapotranspiration $(e_{\mathrm{p}1},\ldots,e_{\mathrm{p}n})^{\top}$ and values of the $d = 13$ model parameters $\boldsymbol{\uptheta}$. A one-year spin-up period eliminates the impact of state variable initialization.

Table \ref{tableApp:B3} lists the $d = 13$ \texttt{sacsma} model parameters and their corresponding symbols, units and upper and lower bounds. 
\begin{table}[H]
\centering
\captionsetup[table]{position=bottom}
\begin{threeparttable}
\caption{\texttt{sacsma} model parameters and their symbols, units, lower and upper bounds.}
\label{tableApp:B3}
\begin{tabular}{l l c c c }
\toprule
\multicolumn{1}{l}{ Symbol } & Description & Units & Min. & Max. \\
\midrule
$u_\mathrm{t,max}$ & Upper zone tension water maximum storage & mm & 50 & 500 \\
$u_\mathrm{f,max}$ & Upper zone free water maximum storage & mm & 10 & 500 \\
$l_\mathrm{t,max}$ & Lower zone tension water maximum storage & mm & 10 & 500 \\
$l_\mathrm{fp,max}$ & Lower zone free water primary maximum storage & mm & 10 & 1000 \\
$l_\mathrm{fs,max}$ & Lower zone free water supplemental maximum storage & mm & 10 & 1000 \\
$\alpha$ & Percolation multiplier for the lower layer & - & 1 & 250 \\
$\psi$ & Percolation exponent for the lower layer & - & 1 & 5 \\
$k_\mathrm{i}$ & Upper zone free water lateral depletion rate (interflow) & mm\,d$^{-1}$ & $10^{-2}$ & 100 \\
$\kappa$ & Fraction of percolation to tension storage in lower layer & - & 0.05 & 0.95 \\
$\nu_\mathrm{p}$ & Base flow depletion rate for primary reservoir & d$^{-1}$ & $10^{-3}$ & 0.25 \\
$\nu_\mathrm{s}$ & Base flow depletion rate for secondary reservoir & d$^{-1}$ & $10^{-3}$ & 0.25 \\
$a_\mathrm{c,max}$ & Maximum fraction of saturated area & - & 0.05 & 0.95 \\
$k_\mathrm{f}$ & Recession constant of routing reservoirs & d$^{-1}$ & $10^{-1}$ & 5 \\
\bottomrule
\end{tabular}
\end{threeparttable}
\end{table}

To construct the augmented ODE system used to compute analytic sensitivities, we require the $m \times m$ Jacobian matrix $\mathbf{J}_{f}(\mathbf{x})$ of the system dynamics with respect to the states
\begin{linenomath*}
\begin{align}
\mathbf{J}_{f}(\mathbf{x}) & = \frac{\partial \mathbf{f}(\mathbf{x},\boldsymbol{\uptheta},t)}{\partial \mathbf{x}^{\top}}
\in \mathbb{R}^{9 \times 9}, \nonumber
\end{align}
\end{linenomath*}
and the $m \times d$ Jacobian matrix $\mathbf{J}_{f}(\boldsymbol{\uptheta})$ of the system dynamics with respect to the parameters
\begin{linenomath*}
\begin{align}
\mathbf{J}_{f}(\boldsymbol{\uptheta}) & = \frac{\partial \mathbf{f}(\mathbf{x},\boldsymbol{\uptheta},t)}{\partial \boldsymbol{\uptheta}^{\top}} \in \mathbb{R}^{9 \times 13}. \nonumber
\end{align}
\end{linenomath*}
where $\boldsymbol{\uptheta} = \bigl(u_\mathrm{t,max}, u_\mathrm{f,max}, l_\mathrm{t,max}, l_\mathrm{fp,max}, l_\mathrm{fs,max}, \alpha, \psi, k_\mathrm{i}, \kappa, \nu_\mathrm{p}, \nu_\mathrm{s}, a_\mathrm{c,max}, k_\mathrm{f} \bigr)^{\top}$.

We introduce the following dimensionless storages
\begin{linenomath*}
\begin{align}
\overline{u}_\mathrm{t} & = \frac{u_\mathrm{t}}{u_\mathrm{t,max}}, &
\overline{u}_\mathrm{f} & = \frac{u_\mathrm{f}}{u_\mathrm{f,max}}, &
\overline{l}_\mathrm{t} & = \frac{l_\mathrm{t}}{l_\mathrm{t,max}}, &
\overline{l}_\mathrm{fp} & = \frac{l_\mathrm{fp}}{l_\mathrm{fp,max}}, & \overline{l}_\mathrm{fs} & = \frac{l_\mathrm{fs}}{l_\mathrm{fs,max}}, \nonumber
\end{align}
\end{linenomath*}
and the percolation multiplier
\begin{linenomath*}
\begin{align}
d_\mathrm{lz} & = 1 + \alpha \biggl(\frac{l_\mathrm{tot}}{l_\mathrm{max}}\biggr)^{\psi}, \nonumber
\end{align}
\end{linenomath*}
where $l_\mathrm{tot} = l_\mathrm{t} + l_\mathrm{fp} + l_\mathrm{fs}$ and $l_\mathrm{max} = l_\mathrm{t,max} + l_\mathrm{fp,max} + l_\mathrm{fs,max}$, and the smoothing functions
\begin{linenomath*}
\begin{align}
g_\mathrm{t} & = \phi(u_\mathrm{t},u_\mathrm{t,max}), &
g_\mathrm{f} & = \phi(u_\mathrm{f},u_\mathrm{f,max}), &
g_{l_\mathrm{t}} & = \phi(l_\mathrm{t},l_\mathrm{t,max}), \nonumber \\
g_{l_\mathrm{fp}} & = \phi(l_\mathrm{fp},l_\mathrm{fp,max}), &
g_{l_\mathrm{fs}} & = \phi(l_\mathrm{fs},l_\mathrm{fs,max}), \nonumber
\end{align}
\end{linenomath*}
with
\begin{linenomath*}
\begin{align}
\phi(x_{1},x_{2}) & = \biggl\{1 + \exp\biggl[\frac{x_{2} - \varsigma \, \varrho \, x_{2} - x_{1}}{\varrho \, x_{2}}\biggr]\biggr\}^{-1}, \nonumber
\end{align}
\end{linenomath*}
and $\varsigma = 5$ and $\varrho = 10^{-2}$.

\subsubsection*{Auxiliary derivatives}
The logistic smoothing $\phi(x_{1},x_{2})$ depends on
\begin{linenomath*}
\begin{align}
z(x_{1},x_{2})
& = \frac{x_{2} - \varsigma \, \varrho \, x_{2} - x_{1}}{\varrho \, x_{2}}
= \frac{1-\varsigma \, \varrho}{\varrho} - \frac{x_1}{\varrho \, x_2}, \nonumber
\end{align}
\end{linenomath*}
so that
\begin{linenomath*}
\begin{align}
\frac{\partial \phi}{\partial z} & = - \phi(1-\phi), &
\frac{\partial z}{\partial x_{1}} & = - \frac{1}{\varrho \, x_{2}}, &
\frac{\partial z}{\partial x_{2}} & = \frac{x_{1}}{\varrho \, x^{2}_{2}}, \nonumber
\end{align}
\end{linenomath*}
and therefore
\begin{linenomath*}
\begin{align}
\frac{\partial \phi}{\partial x_{1}} & = \frac{\phi(1-\phi)}{\varrho \, x_{2}}, &
\frac{\partial \phi}{\partial x_{2}} & = -\frac{x_1}{\varrho \, x_{2}^{2}}\,\phi(1-\phi). \nonumber
\end{align}
\end{linenomath*}

Specializing to the \texttt{sacsma} smoothing functions,
\begin{linenomath*}
\begin{align}
\phi^{\prime}_\mathrm{t} & = \frac{\partial g_\mathrm{t}}{\partial u_\mathrm{t}} = \frac{g_\mathrm{t}(1-g_\mathrm{t})}{\varrho \, u_\mathrm{t,max}}, & \frac{\partial g_\mathrm{t}}{\partial u_\mathrm{t,max}} & = - \frac{u_\mathrm{t}}{\varrho \, u_\mathrm{t,max}^{2}} g_\mathrm{t}(1-g_\mathrm{t}), \nonumber \\[1mm]
\phi^{\prime}_\mathrm{f} & = \frac{\partial g_\mathrm{f}}{\partial u_\mathrm{f}} = \frac{g_\mathrm{f}(1-g_\mathrm{f})}{\varrho \, u_\mathrm{f,max}}, & \frac{\partial g_\mathrm{f}}{\partial u_\mathrm{f,max}} & = - \frac{u_\mathrm{f}}{\varrho \, u_\mathrm{f,max}^2} g_\mathrm{f}(1 - g_\mathrm{f}), \nonumber \\[1mm]
\phi^{\prime}_{l\mathrm{t}} & = \frac{\partial g_{l_\mathrm{t}}}{\partial l_\mathrm{t}} = \frac{g_{l_\mathrm{t}}(1-g_{l_\mathrm{t}})}{\varrho \, l_\mathrm{t,max}}, & \frac{\partial g_{l_\mathrm{t}}}{\partial l_\mathrm{t,max}} & = - \frac{l_\mathrm{t}}{\varrho \, l_\mathrm{t,max}^2} g_{l\mathrm{t}}(1-g_{l\mathrm{t}}), \nonumber \\[1mm]
\phi^{\prime}_{l_\mathrm{fp}} & = \frac{\partial g_{l_\mathrm{fp}}}{\partial l_\mathrm{fp}} = \frac{g_{l_\mathrm{fp}}(1-g_{l_\mathrm{fp}})}{\varrho \, l_\mathrm{fp,max}}, & \frac{\partial g_{l_\mathrm{fp}}}{\partial l_\mathrm{fp,max}} & = - \frac{l_\mathrm{fp}}{\varrho \, l_\mathrm{fp,max}^2} g_{l_\mathrm{fp}}(1-g_{l_\mathrm{fp}}), \nonumber \\[1mm]
\phi^{\prime}_{l_\mathrm{fs}} & = \frac{\partial g_{l_\mathrm{fs}}}{\partial l_\mathrm{fs}} = \frac{g_{l_\mathrm{fs}}(1-g_{l\mathrm{fs}})}{\varrho \, l_\mathrm{fs,max}}, & \frac{\partial g_{l_\mathrm{fs}}}{\partial l_\mathrm{fs,max}} & = - \frac{l_\mathrm{fs}}{\varrho \, l_\mathrm{fs,max}^2} g_{l_\mathrm{fs}}(1-g_{l_\mathrm{fs}}). \nonumber
\end{align}
\end{linenomath*}

For the percolation multiplier $d_\mathrm{lz}$ we obtain
\begin{linenomath*}
\begin{align}
D^{\prime}
& = \alpha \, \psi \biggl(\frac{l_\mathrm{tot}}{l_\mathrm{max}}\biggr)^{\psi-1} \frac{1}{l_\mathrm{max}}, \nonumber \\[1mm]
\frac{\partial d_\mathrm{lz}}{\partial l_\mathrm{t}} & = \frac{\partial d_\mathrm{lz}}{\partial l_\mathrm{fp}} = \frac{\partial d_\mathrm{lz}}{\partial l_\mathrm{fs}} = D^{\prime}, \nonumber \\[1mm]
\frac{\partial d_\mathrm{lz}}{\partial \alpha} & = \biggl(\frac{l_\mathrm{tot}}{l_\mathrm{max}}\biggr)^{\psi}, \nonumber \\[1mm]
\frac{\partial d_\mathrm{lz}}{\partial \psi} & = \alpha \biggl(\frac{l_\mathrm{tot}}{l_\mathrm{max}}\biggr)^{\psi}
\log\biggl(\frac{l_\mathrm{tot}}{l_\mathrm{max}}\biggr), \nonumber \\[1mm]
\frac{\partial d_\mathrm{lz}}{\partial l_\mathrm{t,max}} & = \frac{\partial d_\mathrm{lz}}{\partial l_\mathrm{fp,max}} = \frac{\partial d_\mathrm{lz}}{\partial l_\mathrm{fs,max}} = - \alpha \, \psi \biggl(\frac{l_\mathrm{tot}}{l_\mathrm{max}}\biggr)^{\psi - 1}\frac{l_\mathrm{tot}}{l^{2}_\mathrm{max}}. \nonumber
\end{align}
\end{linenomath*}

\subsubsection{Jacobian of the system dynamics with respect to states}
With the state ordering
\begin{linenomath*}
\begin{align}
\mathbf{x} = (u_\mathrm{t}, u_\mathrm{f}, l_\mathrm{t}, l_\mathrm{fp}, l_\mathrm{fs}, r_{1}, r_{2}, r_{3}, r_\mathrm{q})^{\top}, \nonumber
\end{align}
\end{linenomath*}
the Jacobian of the system dynamics with respect to the state variables is
\begin{linenomath*}
\begin{align}
\mathbf{J}_{f}(\mathbf{x}) & = \frac{\partial \mathbf{f}(\mathbf{x},\boldsymbol{\uptheta},t)}{\partial \mathbf{x}^{\top}}
= \colvec{1}{c}{ \; \dfrac{\partial f_{1}}{\partial u_\mathrm{t}} & 0 & 0 & 0 & 0 & 0 & 0 & 0 & 0 \; \\[4mm]
\; \dfrac{\partial f_{2}}{\partial u_\mathrm{t}} & \dfrac{\partial f_{2}}{\partial u_\mathrm{f}} & \dfrac{\partial f_{2}}{\partial l_\mathrm{t}} & \dfrac{\partial f_{2}}{\partial l_\mathrm{fp}} & \dfrac{\partial f_{2}}{\partial l_\mathrm{fs}} & 0 & 0 & 0 & 0 \; \\[4mm]
\; \dfrac{\partial f_{3}}{\partial u_\mathrm{t}} & \dfrac{\partial f_{3}}{\partial u_\mathrm{f}} & \dfrac{\partial f_{3}}{\partial l_\mathrm{t}} & \dfrac{\partial f_{3}}{\partial l_\mathrm{fp}} & \dfrac{\partial f_{3}}{\partial l_\mathrm{fs}} & 0 & 0 & 0 & 0 \; \\[4mm]
\; 0 & \dfrac{\partial f_{4}}{\partial u_\mathrm{f}} & \dfrac{\partial f_{4}}{\partial l_\mathrm{t}} & \dfrac{\partial f_{4}}{\partial l_\mathrm{fp}} & \dfrac{\partial f_{4}}{\partial l_\mathrm{fs}} & 0 & 0 & 0 & 0 \; \\[4mm]
\; 0 & \dfrac{\partial f_{5}}{\partial u_\mathrm{f}} & \dfrac{\partial f_{5}}{\partial l_\mathrm{t}} & \dfrac{\partial f_{5}}{\partial l_\mathrm{fp}} & \dfrac{\partial f_{5}}{\partial l_\mathrm{fs}} & 0 & 0 & 0 & 0 \; \\[4mm]
\; \dfrac{\partial f_{6}}{\partial u_\mathrm{t}} & \dfrac{\partial f_{6}}{\partial u_\mathrm{f}} & \dfrac{\partial f_{6}}{\partial l_\mathrm{t}} & \dfrac{\partial f_{6}}{\partial l_\mathrm{fp}} & \dfrac{\partial f_{6}}{\partial l_\mathrm{fs}} & \dfrac{\partial f_{6}}{\partial r_{1}} & 0 & 0 & 0 \; \\[4mm]
\; 0 & 0 & 0 & 0 & 0 & \dfrac{\partial f_{7}}{\partial r_{1}} & \dfrac{\partial f_{7}}{\partial r_{2}} & 0 & 0 \; \\[4mm]
\; 0 & 0 & 0 & 0 & 0 & 0 & \dfrac{\partial f_{8}}{\partial r_{2}} & \dfrac{\partial f_{8}}{\partial r_{3}} & 0 \; \\[4mm]
\; 0 & 0 & 0 & 0 & 0 & 0 & 0 & \dfrac{\partial f_{9}}{\partial r_{3}} & 0 \; } \in \mathbb{R}^{9 \times 9},
\label{eq:JxSAC}
\end{align}
\end{linenomath*}
where the nonzero entries are
\begin{linenomath*}
\begin{align}
& \frac{\partial f_{1}}{\partial u_\mathrm{t}} = - \frac{\partial q_\mathrm{sx}}{\partial u_\mathrm{t}} - \frac{\partial q_\mathrm{ut}}{\partial u_\mathrm{t}} - \frac{\partial e_{1}}{\partial u_\mathrm{t}}, \nonumber \\[1mm]
& \frac{\partial f_{2}}{\partial u_\mathrm{t}} = \frac{\partial q_\mathrm{ut}}{\partial u_\mathrm{t}} - \frac{\partial q_\mathrm{uf}}{\partial u_\mathrm{t}} = (1 - g_\mathrm{f})\,\frac{\partial q_\mathrm{ut}}{\partial u_\mathrm{t}}, \qquad \frac{\partial f_{2}}{\partial u_\mathrm{f}} = -\frac{\partial q_\mathrm{ul}}{\partial u_\mathrm{f}} - \frac{\partial q_\mathrm{uf}}{\partial u_\mathrm{f}} - \frac{\partial q_\mathrm{if}}{\partial u_\mathrm{f}}, \qquad
\frac{\partial f_{2}}{\partial l_\mathrm{t}} = - \frac{\partial q_\mathrm{ul}}{\partial l_\mathrm{t}}, \nonumber \\[1mm]
& \frac{\partial f_{2}}{\partial l_\mathrm{fp}} = - \frac{\partial q_\mathrm{ul}}{\partial l_\mathrm{fp}}, \qquad \frac{\partial f_{2}}{\partial l_\mathrm{fs}} = - \frac{\partial q_\mathrm{ul}}{\partial l_\mathrm{fs}}, \nonumber \\[1mm]
& \frac{\partial f_{3}}{\partial u_\mathrm{t}} = - \frac{\partial e_\mathrm{2}}{\partial u_\mathrm{t}}, \qquad \frac{\partial f_{3}}{\partial u_\mathrm{f}} = \kappa \frac{\partial q_\mathrm{ul}}{\partial u_\mathrm{f}} - \frac{\partial q_\mathrm{lt}}{\partial u_\mathrm{f}}, \qquad \frac{\partial f_{3}}{\partial l_\mathrm{t}} = \kappa \frac{\partial q_\mathrm{ul}}{\partial l_\mathrm{t}} - \frac{\partial q_\mathrm{lt}}{\partial l_\mathrm{t}} - \frac{\partial e_\mathrm{2}}{\partial l_\mathrm{t}}, \nonumber \\[1mm]
& \frac{\partial f_{3}}{\partial l_\mathrm{fp}} = \kappa \frac{\partial q_\mathrm{ul}}{\partial l_\mathrm{fp}} - \frac{\partial q_\mathrm{lt}}{\partial l_\mathrm{fp}}, \qquad
\frac{\partial f_{3}}{\partial l_\mathrm{fs}} = \kappa \frac{\partial q_\mathrm{ul}}{\partial l_\mathrm{fs}} - \frac{\partial q_\mathrm{lt}}{\partial l_\mathrm{fs}}, \nonumber \\[1mm]
& \frac{\partial f_{4}}{\partial u_\mathrm{f}} = \frac{\partial q_\mathrm{pr}}{\partial u_\mathrm{f}} - \frac{\partial q_\mathrm{lfp}}{\partial u_\mathrm{f}}, \qquad \frac{\partial f_{4}}{\partial l_\mathrm{t}} = \frac{\partial q_\mathrm{pr}}{\partial l_\mathrm{t}} - \frac{\partial q_\mathrm{lfp}}{\partial l_\mathrm{t}}, \nonumber \\[1mm]
& \frac{\partial f_{4}}{\partial l_\mathrm{fp}} = \frac{\partial q_\mathrm{pr}}{\partial l_\mathrm{fp}} - \frac{\partial q_\mathrm{lfp}}{\partial l_\mathrm{fp}} - \frac{\partial q_\mathrm{bp}}{\partial l_\mathrm{fp}}, \qquad \frac{\partial f_{4}}{\partial l_\mathrm{fs}} = \frac{\partial q_\mathrm{pr}}{\partial l_\mathrm{fs}} - \frac{\partial q_\mathrm{lfp}}{\partial l_\mathrm{fs}}, \nonumber \\[1mm]
& \frac{\partial f_{5}}{\partial u_\mathrm{f}} = \frac{\partial q_\mathrm{pr}}{\partial u_\mathrm{f}} - \frac{\partial q_\mathrm{lfs}}{\partial u_\mathrm{f}}, \qquad \frac{\partial f_{5}}{\partial l_\mathrm{t}} = \frac{\partial q_\mathrm{pr}}{\partial l_\mathrm{t}} - \frac{\partial q_\mathrm{lfs}}{\partial l_\mathrm{t}}, \nonumber \\[1mm]
& \frac{\partial f_{5}}{\partial l_\mathrm{fp}} = \frac{\partial q_\mathrm{pr}}{\partial l_\mathrm{fp}} - \frac{\partial q_\mathrm{lfs}}{\partial l_\mathrm{fp}}, \qquad \frac{\partial f_{5}}{\partial l_\mathrm{fs}} = \frac{\partial q_\mathrm{pr}}{\partial l_\mathrm{fs}} - \frac{\partial q_\mathrm{lfs}}{\partial l_\mathrm{fs}} - \frac{\partial q_\mathrm{bs}}{\partial l_\mathrm{fs}}, \nonumber \\[1mm]
& \frac{\partial f_{6}}{\partial u_\mathrm{t}} = \frac{\partial q_\mathrm{ch}}{\partial u_\mathrm{t}}, \qquad \frac{\partial f_{6}}{\partial u_\mathrm{f}} = \frac{\partial q_\mathrm{ch}}{\partial u_\mathrm{f}}, \qquad \frac{\partial f_{6}}{\partial l_\mathrm{t}}
= \frac{\partial q_\mathrm{ch}}{\partial l_\mathrm{t}}, \nonumber \\[1mm] \displaybreak[1]
& \frac{\partial f_{6}}{\partial l_\mathrm{fp}} = \frac{\partial q_\mathrm{ch}}{\partial l_\mathrm{fp}}, \qquad \frac{\partial f_{6}}{\partial l_\mathrm{fs}} = \frac{\partial q_\mathrm{ch}}{\partial l_\mathrm{fs}}, \qquad \frac{\partial f_{6}}{\partial r_{1}} = - k_\mathrm{f}, \nonumber \\[1mm]
& \frac{\partial f_{7}}{\partial r_{1}} = k_\mathrm{f}, \qquad
\frac{\partial f_{7}}{\partial r_{2}} = - k_\mathrm{f}, \nonumber \\[1mm]
& \frac{\partial f_{8}}{\partial r_{2}} = k_\mathrm{f}, \qquad
\frac{\partial f_{8}}{\partial r_{3}} = - k_\mathrm{f}. \nonumber
\end{align}
\end{linenomath*}
The channel inflow derivatives are
\begin{linenomath*}
\begin{align}
\frac{\partial q_\mathrm{ch}}{\partial u_\mathrm{t}}
& = \frac{\partial q_\mathrm{sx}}{\partial u_\mathrm{t}}
  + \frac{\partial q_\mathrm{uf}}{\partial u_\mathrm{t}}, \nonumber \\[1mm]
\frac{\partial q_\mathrm{ch}}{\partial u_\mathrm{f}}
& = \frac{\partial q_\mathrm{uf}}{\partial u_\mathrm{f}}
  + \frac{\partial q_\mathrm{if}}{\partial u_\mathrm{f}}, \nonumber \\[1mm]
\frac{\partial q_\mathrm{ch}}{\partial l_\mathrm{t}}
& = \frac{\partial q_\mathrm{lfp}}{\partial l_\mathrm{t}}
  + \frac{\partial q_\mathrm{lfs}}{\partial l_\mathrm{t}}, \nonumber \\[1mm]
\frac{\partial q_\mathrm{ch}}{\partial l_\mathrm{fp}}
& = \frac{\partial q_\mathrm{lfp}}{\partial l_\mathrm{fp}}
  + \frac{\partial q_\mathrm{lfs}}{\partial l_\mathrm{fp}}
  + \frac{\partial q_\mathrm{bp}}{\partial l_\mathrm{fp}}, \nonumber \\[1mm]
\frac{\partial q_\mathrm{ch}}{\partial l_\mathrm{fs}}
& = \frac{\partial q_\mathrm{lfp}}{\partial l_\mathrm{fs}}
  + \frac{\partial q_\mathrm{lfs}}{\partial l_\mathrm{fs}}
  + \frac{\partial q_\mathrm{bs}}{\partial l_\mathrm{fs}}. \nonumber
\end{align}
\end{linenomath*}
All partial derivatives of the fluxes $q_\mathrm{sx}$, $q_\mathrm{ut}$, $e_{1}$, $q_\mathrm{uf}$, $q_\mathrm{if}$, $q_\mathrm{ul}$, $e_{2}$, $q_\mathrm{lt}$, $q_\mathrm{pr}$, $q_\mathrm{lfp}$, $q_\mathrm{bp}$, $q_\mathrm{lfs}$ and $q_\mathrm{bs}$ with respect to the states follow by combining the auxiliary derivatives above with the constitutive relationships for each flux.

\subsubsection{Jacobian of system dynamic with respect to parameters} 
The $m \times d$ Jacobian $\mathbf{J}_{f}(\boldsymbol{\uptheta})$ of the system dynamics $\mathbf{f}(\mathbf{x},\boldsymbol{\uptheta},t)$ with respect to the parameters $\boldsymbol{\uptheta} = (u_\mathrm{t,max},u_\mathrm{f,max},l_\mathrm{t,max},l_\mathrm{fp,max},l_\mathrm{fs,max},\alpha,\psi,k_\mathrm{i},\kappa,\nu_\mathrm{p},\nu_\mathrm{s},a_\mathrm{c,max},k_\mathrm{f})^{\top}$ is 
\begin{linenomath*}
\begin{align}
\mathbf{J}_{f}(\boldsymbol{\uptheta}) & =
\frac{\partial \mathbf{f}(\mathbf{x},\boldsymbol{\uptheta},t)}{\partial \boldsymbol{\uptheta}^{\top}} = \colvec{1}{c}{ \; \mathbf{j}_{1}^\top \; \\[1mm]
 \; \mathbf{j}_{2}^\top  \; \\[1mm]
 \; \mathbf{j}_{3}^\top  \; \\[1mm]
 \; \mathbf{j}_{4}^\top  \; \\[1mm]
 \; \mathbf{j}_{5}^\top  \; \\[1mm]
 \; \mathbf{j}_{6}^\top  \; \\[1mm]
 \; \mathbf{j}_{7}^\top  \; \\[1mm]
 \; \mathbf{j}_{8}^\top  \; \\[1mm] 
 \; \mathbf{j}_{9}^\top  \; } \in \mathbb{R}^{9 \times 13}
\end{align}
\end{linenomath*}
has the following rows in order of $f_{1}, \ldots, f_{9}$:

\paragraph*{Row 1: $f_1 = \mathrm{d}x_{1}/\mathrm{d}t = p_{t} - q_\mathrm{sx} - q_\mathrm{ut} - e_\mathrm{1}$}
\begin{linenomath*}
\begin{align}
\mathbf{j}^{\top}_{1} & = \biggl( - \frac{\partial q_\mathrm{sx}}{\partial u_\mathrm{t,max}} - \frac{\partial q_\mathrm{ut}}{\partial u_\mathrm{t,max}} - \frac{\partial e_\mathrm{1}}{\partial u_\mathrm{t,max}}, \; 0, \;0,\;0,\;0,\;0,\;0,\;0,\;0,\;0,\;0, \; - \frac{\partial q_\mathrm{sx}}{\partial a_\mathrm{c,max}} - \frac{\partial q_\mathrm{ut}}{\partial a_\mathrm{c,max}} , \; 0 \biggr). \nonumber
\end{align}
\end{linenomath*}

\paragraph*{Row 2: $f_{2} = \mathrm{d}x_{2}/\mathrm{d}t = q_\mathrm{ut} - q_\mathrm{ul} - q_\mathrm{uf} - q_\mathrm{if}$}
\begin{linenomath*}
\begin{align}
\mathbf{j}^{\top}_{2} & = \biggl( \frac{\partial q_\mathrm{ut}}{\partial u_\mathrm{t,max}} - \frac{\partial q_\mathrm{uf}}{\partial u_\mathrm{t,max}}, \; - \frac{\partial q_\mathrm{ul}}{\partial u_\mathrm{f,max}} - \frac{\partial q_\mathrm{uf}}{\partial u_\mathrm{f,max}} - \frac{\partial q_\mathrm{if}}{\partial u_\mathrm{f,max}}, \; - \frac{\partial q_\mathrm{ul}}{\partial l_\mathrm{t,max}}, \; - \frac{\partial q_\mathrm{ul}}{\partial l_\mathrm{fp,max}}, \; - \frac{\partial q_\mathrm{ul}}{\partial l_\mathrm{fs,max}}, \nonumber \\[1mm] & \hspace{7mm} - \frac{\partial q_\mathrm{ul}}{\partial \alpha}, \; - \frac{\partial q_\mathrm{ul}}{\partial \psi}, \; - \frac{\partial q_\mathrm{if}}{\partial k_\mathrm{i}}, \; 0, \; - \frac{\partial q_\mathrm{ul}}{\partial \nu_\mathrm{p}}, \; - \frac{\partial q_\mathrm{ul}}{\partial \nu_\mathrm{s}}, \; \frac{\partial q_\mathrm{ut}}{\partial a_\mathrm{c,max}} - \frac{\partial q_\mathrm{uf}}{\partial a_\mathrm{c,max}}, \; 0 \biggr). \nonumber  
\end{align}
\end{linenomath*}

\paragraph*{Row 3: $f_3 = \mathrm{d}x_{3}/\mathrm{d}t = \kappa q_\mathrm{ul} - q_\mathrm{lt} - e_\mathrm{2}$}
\begin{linenomath*}
\begin{align}
\mathbf{j}^{\top}_{3} & = \biggl( - \frac{\partial e_\mathrm{2}}{\partial u_\mathrm{t,max}}, \; \kappa \frac{\partial q_\mathrm{ul}}{\partial u_\mathrm{f,max}} - \frac{\partial q_\mathrm{lt}}{\partial u_\mathrm{f,max}}, \; \kappa \frac{\partial q_\mathrm{ul}}{\partial l_\mathrm{t,max}} - \frac{\partial q_\mathrm{lt}}{\partial l_\mathrm{t,max}} - \frac{\partial e_\mathrm{2}}{\partial l_\mathrm{t,max}}, \; \kappa \frac{\partial q_\mathrm{ul}}{\partial l_\mathrm{fp,max}} - \frac{\partial q_\mathrm{lt}}{\partial l_\mathrm{fp,max}}, \nonumber \\[1mm] & \hspace{7mm} \kappa \frac{\partial q_\mathrm{ul}}{\partial l_\mathrm{fs,max}} - \frac{\partial q_\mathrm{lt}}{\partial l_\mathrm{fs,max}}, \; \kappa \frac{\partial q_\mathrm{ul}}{\partial \alpha} - \frac{\partial q_\mathrm{lt}}{\partial \alpha}, \; \kappa \frac{\partial q_\mathrm{ul}}{\partial \psi} - \frac{\partial q_\mathrm{lt}}{\partial \psi}, \; 0, \; \frac{\partial f_{3}}{\partial \kappa} = q_\mathrm{ul} - \frac{\partial q_\mathrm{lt}}{\partial \kappa}, \; \kappa \frac{\partial q_\mathrm{ul}}{\partial \nu_\mathrm{p}} - \frac{\partial q_\mathrm{lt}}{\partial \nu_\mathrm{p}}, \nonumber \\[1mm] & \hspace{7mm} \kappa \frac{\partial q_\mathrm{ul}}{\partial \nu_\mathrm{s}} - \frac{\partial q_\mathrm{lt}}{\partial \nu_\mathrm{s}}, \; 0, \; 0 \biggr). \nonumber
\end{align}
\end{linenomath*}

\paragraph*{Row 4: $f_4 = \mathrm{d}x_{4}/\mathrm{d}t  = q_\mathrm{pr} - q_\mathrm{lfp} - q_\mathrm{bp}$}
\begin{linenomath*}
\begin{align}
\mathbf{j}^{\top}_{4} & = \biggl( 0, \; \frac{\partial q_\mathrm{pr}}{\partial u_\mathrm{f,max}} - \frac{\partial q_\mathrm{lfp}}{\partial u_\mathrm{f,max}}, \; \frac{\partial q_\mathrm{pr}}{\partial l_\mathrm{t,max}} - \frac{\partial q_\mathrm{lfp}}{\partial l_\mathrm{t,max}}, \; \frac{\partial q_\mathrm{pr}}{\partial l_\mathrm{fp,max}} - \frac{\partial q_\mathrm{lfp}}{\partial l_\mathrm{fp,max}}, \; \frac{\partial q_\mathrm{pr}}{\partial l_\mathrm{fs,max}} - \frac{\partial q_\mathrm{lfp}}{\partial l_\mathrm{fs,max}}, \nonumber \\[1mm] & \hspace{7mm} \frac{\partial q_\mathrm{pr}}{\partial \alpha} - \frac{\partial q_\mathrm{lfp}}{\partial \alpha}, \; \frac{\partial q_\mathrm{pr}}{\partial \psi} - \frac{\partial q_\mathrm{lfp}}{\partial \psi}, \; 0, \; \frac{\partial q_\mathrm{pr}}{\partial \kappa} - \frac{\partial q_\mathrm{lfp}}{\partial \kappa}, \; \frac{\partial q_\mathrm{pr}}{\partial \nu_\mathrm{p}} - \frac{\partial q_\mathrm{lfp}}{\partial \nu_\mathrm{p}} - \frac{\partial q_\mathrm{bp}}{\partial \nu_\mathrm{p}}, \; \frac{\partial q_\mathrm{pr}}{\partial \nu_\mathrm{s}} - \frac{\partial q_\mathrm{lfp}}{\partial \nu_\mathrm{s}}, \; 0, \; 0 \biggr). \nonumber
\end{align}
\end{linenomath*}

\paragraph*{Row 5: $f_5 = \mathrm{d}x_{5}/\mathrm{d}t = q_\mathrm{pr} - q_\mathrm{lfs} - q_\mathrm{bs}$}
\begin{linenomath*}
\begin{align}
\mathbf{j}^{\top}_{5} & = \biggl(0, \; \frac{\partial q_\mathrm{pr}}{\partial u_\mathrm{f,max}} - \frac{\partial q_\mathrm{lfs}}{\partial u_\mathrm{f,max}}, \; \frac{\partial q_\mathrm{pr}}{\partial l_\mathrm{t,max}} - \frac{\partial q_\mathrm{lfs}}{\partial l_\mathrm{t,max}}, \; \frac{\partial q_\mathrm{pr}}{\partial l_\mathrm{fp,max}} - \frac{\partial q_\mathrm{lfs}}{\partial l_\mathrm{fp,max}}, \; \frac{\partial q_\mathrm{pr}}{\partial l_\mathrm{fs,max}} - \frac{\partial q_\mathrm{lfs}}{\partial l_\mathrm{fs,max}}, \nonumber \\[1mm] & \hspace{7mm} \frac{\partial q_\mathrm{pr}}{\partial \alpha} - \frac{\partial q_\mathrm{lfs}}{\partial \alpha}, \; \frac{\partial q_\mathrm{pr}}{\partial \psi} - \frac{\partial q_\mathrm{lfs}}{\partial \psi}, \; 0, \; \frac{\partial q_\mathrm{pr}}{\partial \kappa} - \frac{\partial q_\mathrm{lfs}}{\partial \kappa}, \; \frac{\partial q_\mathrm{pr}}{\partial \nu_\mathrm{p}} - \frac{\partial q_\mathrm{lfs}}{\partial \nu_\mathrm{p}}, \; \frac{\partial q_\mathrm{pr}}{\partial \nu_\mathrm{s}} - \frac{\partial q_\mathrm{lfs}}{\partial \nu_\mathrm{s}} - \frac{\partial q_\mathrm{bs}}{\partial \nu_\mathrm{s}}, \; 0, \; 0 \biggr). \nonumber
\end{align}
\end{linenomath*}

\paragraph*{Row 6: $f_6 = \mathrm{d}x_{6}/\mathrm{d}t = q_\mathrm{ch} - k_\mathrm{f} r_1$}
\begin{linenomath*}
\begin{align}
\mathbf{j}^{\top}_{6} & = \biggl( \frac{\partial q_\mathrm{ch}}{\partial u_\mathrm{t,max}}, \; \frac{\partial q_\mathrm{ch}}{\partial u_\mathrm{f,max}}, \; \frac{\partial q_\mathrm{ch}}{\partial l_\mathrm{t,max}}, \; \frac{\partial q_\mathrm{ch}}{\partial l_\mathrm{fp,max}}, \; \frac{\partial q_\mathrm{ch}}{\partial l_\mathrm{fs,max}}, \; \frac{\partial q_\mathrm{ch}}{\partial \alpha}, \; \frac{\partial q_\mathrm{ch}}{\partial \psi},
\; \frac{\partial q_\mathrm{ch}}{\partial k_\mathrm{i}}, \; \frac{\partial q_\mathrm{ch}}{\partial \kappa}, \; \frac{\partial q_\mathrm{ch}}{\partial \nu_\mathrm{p}}, \; \frac{\partial q_\mathrm{ch}}{\partial \nu_\mathrm{s}}, \nonumber \\[1mm]
& \hspace{7mm} \frac{\partial q_\mathrm{ch}}{\partial a_\mathrm{c,max}}, \; - r_{1} \biggr), \nonumber 
\end{align}
\end{linenomath*}
where each $\partial q_\mathrm{ch}/\partial \theta$ is the sum of the corresponding derivatives of
$q_\mathrm{sx},q_\mathrm{uf},q_\mathrm{if},q_\mathrm{lfp},q_\mathrm{bp},q_\mathrm{lfs}$ and $q_\mathrm{bs}$.

\paragraph*{Row 7: $f_7 = \mathrm{d}x_{7}/\mathrm{d}t = k_\mathrm{f} r_1 - k_\mathrm{f} r_2$}
\begin{linenomath*}
\begin{align}
\mathbf{j}^{\top}_{7} & = \bigl(0, \; 0, \; 0,\; 0,\; 0,\; 0,\; 0,\; 0,\; 0,\; 0,\; 0,\; 0, \; r_{1} - r_{2} \bigr). \nonumber
\end{align}
\end{linenomath*}

\paragraph*{Row 8: $f_8 = \mathrm{d}x_{8}/\mathrm{d}t = k_\mathrm{f} r_2 - k_\mathrm{f} r_3$}
\begin{linenomath*}
\begin{align}
\mathbf{j}^{\top}_{8} & = \bigl(0, \; 0, \; 0,\; 0,\; 0,\; 0,\; 0,\; 0,\; 0,\; 0,\; 0,\; 0, \; r_{2}- r_{3} \bigr). \nonumber 
\end{align}
\end{linenomath*}

\paragraph*{Row 9: $f_9 = \mathrm{d}x_{9}/\mathrm{d}t = k_\mathrm{f} r_3$}
\begin{linenomath*}
\begin{align}
\mathbf{j}^{\top}_{9} & = \bigl(0, \; 0, \; 0,\; 0,\; 0,\; 0,\; 0,\; 0,\; 0,\; 0,\; 0,\; 0, \; r_{3} \bigr). \nonumber
\end{align}
\end{linenomath*}
This completes the analytic specification of the augmented ODE system for the \texttt{sacsma} model.

\subsection{Xinanjiang conceptual watershed model}
The \texttt{xinanjiang} conceptual watershed model is the result of decades of work by Dr. Renjun Zhao and his colleagues at the Hydrological Bureau of the Ministry of Water Resources in China. The model’s initial formulation, based on a saturation-excess runoff mechanism and a top-down runoff generation approach, was developed in 1963 \citep{zhao1963}. In 1980, it was formally named the \texttt{xinanjiang} model \citep{zhao1980}, reflecting its intended application to the humid \texttt{xinanjiang} river basin in China \citep{zhao1992}. In a second development phase (1980-2002), several structural improvements were made, including a three-layer evapotranspiration module, the introduction of interflow as a runoff component, and the replacement of the original hydrograph method with a linear reservoir and/or lag-routing techniques. 

\begin{figure}[H]
\centering\includegraphics[width=0.99\linewidth]{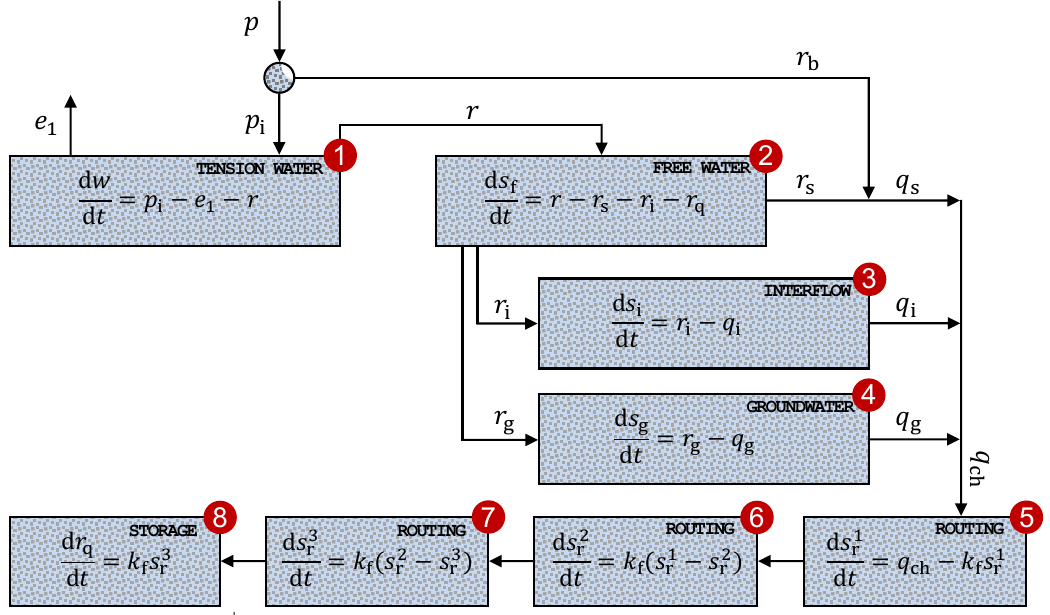}
\caption{Schematic illustration of the \texttt{xinanjiang} conceptual watershed model. Blue boxes labeled in red are fictitious control volumes that govern the rainfall-runoff transformation. The model includes $m = 8$ state variables, including the tension water storage $w$ and free water storage $s_\mathrm{f}$ of the upper soil layer, interflow $s_\mathrm{i}$ and groundwater $s_\mathrm{g}$ reservoirs, water levels $s^{1}_\mathrm{r}$, $s^{2}_\mathrm{r}$, and $s^{3}_\mathrm{r}$ of the routing reservoirs, and the storage $s_\mathrm{q}$ of the discharge reservoir. Fluxes (arrows) describe water movement into and out of compartments: precipitation ($p_{t}$), runoff from impervious areas ($r_\mathrm{b})$, infiltration ($p_\mathrm{i}$), surface runoff from the contributing free area ($r_\mathrm{s}$), evaporation ($e_\mathrm{1}$), runoff ($r$), interflow ($r_\mathrm{i}$), baseflow ($r_\mathrm{g}$), delayed interflow ($q_\mathrm{i}$), delayed baseflow ($q_\mathrm{g}$), and surface runoff ($q_\mathrm{s}$). These fluxes are computed as follows, $r_\mathrm{b} = A_\mathrm{im} p_{t}$, $p_\mathrm{i} = (1-A_\mathrm{im}) p_{t}$, $r_\mathrm{s} = r\{1-(1-s_\mathrm{f}/s_\mathrm{max})^\beta\}$, $e_\mathrm{1} = e_\mathrm{pan}$ if $w > \text{lm}$, $e_\mathrm{1} = (w/\text{lm})e_\mathrm{pan}$ if $c\cdot \text{lm} \leq w \leq \text{lm}$ otherwise $e_\mathrm{1} = c\cdot e_\mathrm{pan}$, $r = p_\mathrm{i} \{(0.5-a)^{(1-b)}(w/w_\mathrm{max})^{b}\}$ if $(w/w_\mathrm{max}) \leq 0.5 - a$ and $r = p_\mathrm{i}\{1-(0.5+a)^{(1-b)}(1 - w/w_\mathrm{max})^{b}\}$ otherwise, $r_\mathrm{i} = k_\mathrm{i}s_\mathrm{f}\{1-(1-s_\mathrm{f}/s_\mathrm{max})^\beta\}$, $r_\mathrm{g} = k_\mathrm{g}s_\mathrm{g}\{1-(1-s_\mathrm{f}/s_\mathrm{max})^\beta\}$, $q_\mathrm{i} = c_\mathrm{i}s_\mathrm{i}$, $q_\mathrm{g} = c_\mathrm{g}s_\mathrm{g}$, and $q_\mathrm{s} = r_\mathrm{b} + r_\mathrm{s}$, where $e_\mathrm{pan} = f_\mathrm{p} e_\mathrm{p}$ is pan evaporation, $e_\mathrm{p}$ denotes the potential evapotranspiration, $w_\mathrm{max} = f_\mathrm{wm}s_\mathrm{tot}$ is the maximum tension water depth, $s_\mathrm{max} = (1-f_\mathrm{wm})s_\mathrm{tot}$ is the maximum free water depth, $\text{lm} = f_\mathrm{lm}w_\mathrm{max}$ is the tension water threshold for evaporation change and $f_\mathrm{p}$, $A_\mathrm{im}$, $a$, $b$, $s_\mathrm{tot}$, $f_\mathrm{wm}$, $f_\mathrm{lm}$, $c$, $\beta$, $k_\mathrm{i}$, $k_\mathrm{g}$, $c_\mathrm{i}$, $c_\mathrm{g}$ and $k_\mathrm{f}$ are free parameters. Total channel inflow $q_\mathrm{ch} = q_\mathrm{s} + q_\mathrm{i} + q_\mathrm{g}$ is routed through three linear reservoirs (with identical recession constant $k_\mathrm{f}$) and produces streamflow at watershed outlet, $q_{t} = k_\mathrm{f}s_\mathrm{r}^{3}$.}
\label{figApp:B4}
\end{figure}

The \texttt{xinanjiang} model transforms areal average precipitation into streamflow by modeling control volumes, state variables, and fluxes as outlined in Figure \ref{figApp:B4}. 
The \texttt{xinanjiang} model is driven by daily time series of areal-average rainfall, $(p_1, \ldots, p_n)^\top$, and potential evapotranspiration, $(e_{\text{p}1}, \ldots, e_{\text{p}n})^\top$. Our implementation follows the formulations of \citet{zhao1992} and \citet{jayawardena2000}, as summarized in ODE form by \citet{knoben2018}, but includes two key additions: (i) an adjustment coefficient, $f_\mathrm{p}$, to convert meteorological estimates of potential evapotranspiration, $e_\mathrm{p}$ (mm/d), into local estimates of actual evaporation; and (ii) a cascade of three linear reservoirs to route channel inflow and convert it into river discharge, $q$ (mm/d). 

Thus, the \texttt{xinanjiang} model has $m = 8$ state variables $\mathbf{x} = (w, s_\mathrm{f}, s_\mathrm{i}, s_\mathrm{g}, s^{1}_\mathrm{r}, s^{2}_\mathrm{r}, s^{3}_\mathrm{r}, s_\mathrm{q})^{\top}$, where the eighth control volume $s_\mathrm{q}$ is an infinite reservoir which accumulates the discharge. The \texttt{xinanjiang} state equations follow from conservation of mass in each control volume
\begin{linenomath*}
\begin{align}
\frac{\mathrm{d}\mathbf{x}}{\mathrm{d}t} = \mathbf{f}(\mathbf{x},\boldsymbol{\uptheta},t) = \colvec{1}{c}{ \;
p_\mathrm{i} - e_\mathrm{1} - r \; \\[1mm]
\; r - r_\mathrm{i} - r_\mathrm{g} \; \\[1mm]
\; r_\mathrm{i} - q_\mathrm{i} \; \\[1mm]
\; r_\mathrm{g} - q_\mathrm{g} \; \\[1mm]
\; q_\mathrm{ch} - k_\mathrm{f} s^{1}_\mathrm{r} \; \\[1mm]
\; k_\mathrm{f} s^{1}_\mathrm{r} - k_\mathrm{f} s^{2}_\mathrm{r} \; \\[1mm]
\; k_\mathrm{f} s^{2}_\mathrm{r} - k_\mathrm{f} s^{3}_\mathrm{r} \; \\[1mm]
\; k_\mathrm{f} s^{3}_\mathrm{r} \;} \in \mathbb{R}^{m \times 1},
\label{eq:Xinanjiang_odes}
\end{align}
\end{linenomath*}
and streamflow at the outlet is $q_{t} = k_\mathrm{f} s^{3}_\mathrm{r}$.

A mass-conservative second-order integration method with adaptive time stepping solves the state variables using daily time series of areal average rainfall $(p_{1},\ldots,p_{n})^{\top}$ and potential evapotranspiration $(e_{\mathrm{p}1},\ldots,e_{\mathrm{p}n})^{\top}$ and values of the model parameters listed in Table~\ref{tableApp:B4}. A spin-up period eliminates the impact of state variable initialization.

The fourteen parameters of the \texttt{xinanjiang} model are listed in Table \ref{tableApp:B4}. 
\begin{table}[H]
\centering
\captionsetup[table]{position=bottom}
\begin{threeparttable}
\caption{Description of \texttt{xinanjiang} model parameters, including symbols, units, lower and upper bounds.}
\label{tableApp:B4}
\begin{tabular}{c l c c c }
\toprule
\multicolumn{1}{l}{ Symbol } & Description & Units & Min. & Max. \\
\midrule
$f_\mathrm{p}$ & Ratio of potential evapotranspiration to pan evaporation & - & $0.5$ & $1.5$ \\
$A_\mathrm{im}$ & Impervious area & - & $10^{-4}$ & $10^{-1}$ \\
$a$ & Tension water distribution inflection parameter & - & $-0.5$ & $0.5$ \\
$b$ & Tension water distribution shape parameter & - & $10^{-1}$ & $2$ \\
$f_\mathrm{wm}$ & Fraction of $s_\mathrm{tot}$ that is $w_\mathrm{max}$ & - & $10^{-3}$ & $1$ \\
$f_\mathrm{lm}$ & First evaporation threshold (fraction of $w_\mathrm{max}$) & - & $10^{-3}$ & $1$ \\
$c$ & Second evaporation threshold (fraction of tension water) & - & $10^{-3}$ & $1$ \\
$s_\mathrm{tot}$ & Total soil moisture storage & mm & $1$ & $10^{3}$ \\
$\beta$ & Free water distribution shape parameter & - & $10^{-3}$ & $2$ \\
$k_\mathrm{i}$ & Free water interflow parameter & d$^{-1}$ & $10^{-3}$ & $3$ \\
$k_\mathrm{g}$ & Free water groundwater parameter & d$^{-1}$ & $10^{-3}$ & $1$ \\
$c_\mathrm{i}$ & Interflow time coefficient & d$^{-1}$ & $10^{-3}$ & $1$ \\
$c_\mathrm{g}$ & Baseflow time coefficient & d$^{-1}$ & $10^{-3}$ & $1$ \\
$k_\mathrm{f}$ & Recession constant of routing reservoirs & d$^{-1}$ & $10^{-1}$ & $5$ \\
\bottomrule
\end{tabular}
\end{threeparttable}
\end{table}

To construct the augmented ODE system used for analytic sensitivities, we require the $m \times m$ Jacobian matrix $\mathbf{J}_{f}(\mathbf{x})$ of the system dynamics with respect to the states
\begin{linenomath*}
\begin{align}
\mathbf{J}_{f}(\mathbf{x}) & = \frac{\partial \mathbf{f}(\mathbf{x},\boldsymbol{\uptheta},t)}{\partial \mathbf{x}^{\top}} \in \mathbb{R}^{8 \times 8}, \nonumber
\end{align}
\end{linenomath*}
and the $m \times d$ Jacobian matrix $\mathbf{J}_{f}(\boldsymbol{\uptheta})$ of the system dynamics with respect to the parameters $\boldsymbol{\uptheta} = ( f_\mathrm{p}, A_\mathrm{im}, a, b, f_\mathrm{wm}, f_\mathrm{lm}, c, s_\mathrm{tot}, \beta, k_\mathrm{i}, k_\mathrm{g}, c_\mathrm{i}, c_\mathrm{g}, k_\mathrm{f} )^{\top}$
\begin{linenomath*}
\begin{align}
\mathbf{J}_{f}(\boldsymbol{\uptheta}) & =
\frac{\partial \mathbf{f}(\mathbf{x},\boldsymbol{\uptheta},t)}{\partial \boldsymbol{\uptheta}^{\top}} \in \mathbb{R}^{8 \times 14}. \nonumber
\end{align}
\end{linenomath*}

\subsubsection*{Auxiliary derivatives}
The dependence of the storage capacities on the parameters is
\begin{linenomath*}
\begin{align}
w_\mathrm{max} & = f_\mathrm{wm} s_\mathrm{tot} &
s_\mathrm{max} & = (1-f_\mathrm{wm}) s_\mathrm{tot} &
\mathrm{lm} & = f_\mathrm{lm} w_\mathrm{max}, \nonumber
\end{align}
\end{linenomath*}
and therefore
\begin{linenomath*}
\begin{align}
\frac{\partial w_\mathrm{max}}{\partial f_\mathrm{wm}} & = s_\mathrm{tot} &
\frac{\partial w_\mathrm{max}}{\partial s_\mathrm{tot}} & = f_\mathrm{wm} \nonumber \\[1mm]
\frac{\partial s_\mathrm{max}}{\partial f_\mathrm{wm}} & = - s_\mathrm{tot} &
\frac{\partial s_\mathrm{max}}{\partial s_\mathrm{tot}} & = 1 - f_\mathrm{wm} \nonumber \\[1mm]
\frac{\partial \mathrm{lm}}{\partial f_\mathrm{lm}} & = w_\mathrm{max} &
\frac{\partial \mathrm{lm}}{\partial f_\mathrm{wm}} & = f_\mathrm{lm} \frac{\partial w_\mathrm{max}}{\partial f_\mathrm{wm}} = f_\mathrm{lm} s_\mathrm{tot} &
\frac{\partial \mathrm{lm}}{\partial s_\mathrm{tot}} & = f_\mathrm{lm} \frac{\partial w_\mathrm{max}}{\partial s_\mathrm{tot}} = f_\mathrm{lm} f_\mathrm{wm}. \nonumber
\end{align}
\end{linenomath*}

The dimensionless storages satisfy
\begin{linenomath*}
\begin{align}
\overline{w} & = \frac{w}{w_\mathrm{max}} &
\overline{s}_\mathrm{f} & = \frac{s_\mathrm{f}}{s_\mathrm{max}}, \nonumber
\end{align}
\end{linenomath*}
with
\begin{linenomath*}
\begin{align}
\frac{\partial \overline{w}}{\partial w} & = \frac{1}{w_\mathrm{max}} & \frac{\partial \overline{w}}{\partial w_\mathrm{max}} & = - \frac{w}{w_\mathrm{max}^{2}} \nonumber \\[1mm]
\frac{\partial \overline{s}_\mathrm{f}}{\partial s_\mathrm{f}} & = \frac{1}{s_\mathrm{max}} & \frac{\partial \overline{s}_\mathrm{f}}{\partial s_\mathrm{max}} & = - \frac{s_\mathrm{f}}{s_\mathrm{max}^{2}}. \nonumber
\end{align}
\end{linenomath*}

The free-water contributing area factor $A_\mathrm{f} = 1 - (1-\overline{s}_\mathrm{f})^{\beta}$ yields
\begin{linenomath*}
\begin{align}
\frac{\partial A_\mathrm{f}}{\partial \overline{s}_\mathrm{f}}
& = \beta (1-\overline{s}_\mathrm{f})^{\beta-1}, &
\frac{\partial A_\mathrm{f}}{\partial \beta}
& = - (1-\overline{s}_\mathrm{f})^{\beta} \log(1-\overline{s}_\mathrm{f}), \nonumber \\[1mm]
\frac{\partial A_\mathrm{f}}{\partial s_\mathrm{f}}
& = \frac{\partial A_\mathrm{f}}{\partial \overline{s}_\mathrm{f}} \frac{\partial \overline{s}_\mathrm{f}}{\partial s_\mathrm{f}}
= \frac{\beta (1-\overline{s}_\mathrm{f})^{\beta-1}}{s_\mathrm{max}}, \nonumber
\end{align}
\end{linenomath*}
and analogous expressions for derivatives with respect to $f_\mathrm{wm}$ and $s_\mathrm{tot}$ via $s_\mathrm{max}$.

The runoff $r$ depends on $p_\mathrm{i}$, $a$, $b$, $f_\mathrm{wm}$ and $s_\mathrm{tot}$ through $\overline{w}$ and $w_\mathrm{max}$. Its partial derivatives follow directly from the piecewise definition and are not written out explicitly here. Likewise, the partial derivatives of $e_\mathrm{1}$ with respect to $w$, $\mathrm{lm}$ and the parameters $f_\mathrm{p}$, $f_\mathrm{lm}$, $f_\mathrm{wm}$, $c$ and $s_\mathrm{tot}$ follow from its piecewise definition.

\subsubsection{Jacobian of the system dynamics with respect to states}
The Jacobian of the system dynamics with respect to the state variables $\mathbf{x} = (w, s_\mathrm{f}, s_\mathrm{i}, s_\mathrm{g}, s^{1}_\mathrm{r}, s^{2}_\mathrm{r}, s^{3}_\mathrm{r}, s_\mathrm{q})^{\top}$ is equal to 
\begin{linenomath*}
\begin{align}
\mathbf{J}_{f}(\mathbf{x}) & = \frac{\partial \mathbf{f}(\mathbf{x},\boldsymbol{\uptheta},t)}{\partial \mathbf{x}^{\top}} =
\colvec{1}{c}{ \;
\dfrac{\partial f_{1}}{\partial w} & \dfrac{\partial f_{1}}{\partial s_\mathrm{f}} & 0 & 0 & 0 & 0 & 0 & 0 \; \\[4mm]
\; \dfrac{\partial f_{2}}{\partial w} & \dfrac{\partial f_{2}}{\partial s_\mathrm{f}} & \dfrac{\partial f_{2}}{\partial s_\mathrm{i}} & \dfrac{\partial f_{2}}{\partial s_\mathrm{g}} & 0 & 0 & 0 & 0 \; \\[4mm]
\; 0 & \dfrac{\partial f_{3}}{\partial s_\mathrm{f}} & \dfrac{\partial f_{3}}{\partial s_\mathrm{i}} & 0 & 0 & 0 & 0 & 0 \; \\[4mm]
\; 0 & \dfrac{\partial f_{4}}{\partial s_\mathrm{f}} & 0 & \dfrac{\partial f_{4}}{\partial s_\mathrm{g}} & 0 & 0 & 0 & 0 \; \\[4mm]
\; \dfrac{\partial f_{5}}{\partial w} & \dfrac{\partial f_{5}}{\partial s_\mathrm{f}} & \dfrac{\partial f_{5}}{\partial s_\mathrm{i}} & \dfrac{\partial f_{5}}{\partial s_\mathrm{g}} & \dfrac{\partial f_{5}}{\partial s^{1}_\mathrm{r}} & 0 & 0 & 0 \; \\[4mm]
\; 0 & 0 & 0 & 0 & \dfrac{\partial f_{6}}{\partial s^{1}_\mathrm{r}} & \dfrac{\partial f_{6}}{\partial s^{2}_\mathrm{r}} & 0 & 0 \; \\[4mm]
\; 0 & 0 & 0 & 0 & 0 & \dfrac{\partial f_{7}}{\partial s^{2}_\mathrm{r}} & \dfrac{\partial f_{7}}{\partial s^{3}_\mathrm{r}} & 0 \; \\[4mm]
\; 0 & 0 & 0 & 0 & 0 & 0 & \dfrac{\partial f_{8}}{\partial s^{3}_\mathrm{r}} & 0 \; } \in \mathbb{R}^{8 \times 8},
\label{eq:Jx_Xinanjiang}
\end{align}
\end{linenomath*}
where the nonzero entries follow from the flux definitions
\begin{linenomath*}
\begin{align}
& f_{1} = \mathrm{d}x_{1}/\mathrm{d}t = p_\mathrm{i} - e_\mathrm{1} - r: \nonumber \\[1mm]
& \frac{\partial f_{1}}{\partial w} = - \frac{\partial e_\mathrm{1}}{\partial w} - \frac{\partial r}{\partial w} \qquad
  \frac{\partial f_{1}}{\partial s_\mathrm{f}} = - \frac{\partial r}{\partial s_\mathrm{f}} \nonumber \\[2mm]
& f_{2} = \mathrm{d}x_{2}/\mathrm{d}t = r - r_\mathrm{i} - r_\mathrm{g}: \nonumber \\[1mm]
& \frac{\partial f_{2}}{\partial w} = \frac{\partial r}{\partial w} \qquad \frac{\partial f_{2}}{\partial s_\mathrm{f}} = \frac{\partial r}{\partial s_\mathrm{f}} - \frac{\partial r_\mathrm{i}}{\partial s_\mathrm{f}} - \frac{\partial r_\mathrm{g}}{\partial s_\mathrm{f}} \qquad \frac{\partial f_{2}}{\partial s_\mathrm{i}} = 0 \qquad \frac{\partial f_{2}}{\partial s_\mathrm{g}} = - \frac{\partial r_\mathrm{g}}{\partial s_\mathrm{g}} \nonumber \\[2mm]
& f_{3} = \mathrm{d}x_{3}/\mathrm{d}t = r_\mathrm{i} - q_\mathrm{i}: \nonumber \\[1mm]
& \frac{\partial f_{3}}{\partial s_\mathrm{f}} = \frac{\partial r_\mathrm{i}}{\partial s_\mathrm{f}} \qquad \frac{\partial f_{3}}{\partial s_\mathrm{i}} = - \frac{\partial q_\mathrm{i}}{\partial s_\mathrm{i}} = - c_\mathrm{i} \nonumber \\[2mm]
& f_{4} = \mathrm{d}x_{4}/\mathrm{d}t = r_\mathrm{g} - q_\mathrm{g}: \nonumber \\[1mm]
& \frac{\partial f_{4}}{\partial s_\mathrm{f}} = \frac{\partial r_\mathrm{g}}{\partial s_\mathrm{f}} \qquad \frac{\partial f_{4}}{\partial s_\mathrm{g}} = \frac{\partial r_\mathrm{g}}{\partial s_\mathrm{g}} - \frac{\partial q_\mathrm{g}}{\partial s_\mathrm{g}} = \frac{\partial r_\mathrm{g}}{\partial s_\mathrm{g}} - c_\mathrm{g} \nonumber \\[2mm]
& f_{5} = \mathrm{d}x_{5}/\mathrm{d}t = q_\mathrm{ch} - k_\mathrm{f} s^{1}_\mathrm{r}: \nonumber \\[1mm]
& \frac{\partial f_{5}}{\partial w} = \frac{\partial q_\mathrm{ch}}{\partial w} \qquad \frac{\partial f_{5}}{\partial s_\mathrm{f}} = \frac{\partial q_\mathrm{ch}}{\partial s_\mathrm{f}} \qquad \frac{\partial f_{5}}{\partial s_\mathrm{i}} = \frac{\partial q_\mathrm{ch}}{\partial s_\mathrm{i}} \qquad \frac{\partial f_{5}}{\partial s_\mathrm{g}} = \frac{\partial q_\mathrm{ch}}{\partial s_\mathrm{g}} \nonumber \\[1mm]  \displaybreak 
& \frac{\partial f_{5}}{\partial s^{1}_\mathrm{r}} = - k_\mathrm{f} \nonumber \\[2mm]
& f_{6} = \mathrm{d}x_{6}/\mathrm{d}t = k_\mathrm{f} s^{1}_\mathrm{r} - k_\mathrm{f} s^{2}_\mathrm{r}: \nonumber \\[1mm]
& \frac{\partial f_{6}}{\partial s^{1}_\mathrm{r}} = k_\mathrm{f} \qquad \frac{\partial f_{6}}{\partial s^{2}_\mathrm{r}} = - k_\mathrm{f} \nonumber \\[2mm]
& f_{7} = \mathrm{d}x_{7}/\mathrm{d}t = k_\mathrm{f} s^{2}_\mathrm{r} - k_\mathrm{f} s^{3}_\mathrm{r}: \nonumber \\[1mm]
& \frac{\partial f_{7}}{\partial s^{2}_\mathrm{r}} = k_\mathrm{f} \qquad \frac{\partial f_{7}}{\partial s^{3}_\mathrm{r}} = - k_\mathrm{f} \nonumber \\[2mm]
& f_{8} = \mathrm{d}x_{8}/\mathrm{d}t = k_\mathrm{f} s^{3}_\mathrm{r}: \nonumber \\[1mm]
& \frac{\partial f_{8}}{\partial s^{3}_\mathrm{r}} = k_\mathrm{f}. \nonumber
\end{align}
\end{linenomath*} \\[-1mm]
The derivatives of $q_\mathrm{ch}$ with respect to the storages are
\begin{linenomath*}
\begin{align}
\frac{\partial q_\mathrm{ch}}{\partial w} & = \frac{\partial q_\mathrm{s}}{\partial w} = \frac{\partial r}{\partial w} A_\mathrm{f} \nonumber \\[1mm]
\frac{\partial q_\mathrm{ch}}{\partial s_\mathrm{f}} & = \frac{\partial q_\mathrm{s}}{\partial s_\mathrm{f}} + \frac{\partial q_\mathrm{i}}{\partial s_\mathrm{f}} + \frac{\partial q_\mathrm{g}}{\partial s_\mathrm{f}} = \frac{\partial r}{\partial s_\mathrm{f}} A_\mathrm{f} + r \frac{\partial A_\mathrm{f}}{\partial s_\mathrm{f}}
+ k_\mathrm{i} s_\mathrm{f} \frac{\partial A_\mathrm{f}}{\partial s_\mathrm{f}} + k_\mathrm{g} s_\mathrm{g} \frac{\partial A_\mathrm{f}}{\partial s_\mathrm{f}} \nonumber \\[1mm]
\frac{\partial q_\mathrm{ch}}{\partial s_\mathrm{i}} & = \frac{\partial q_\mathrm{i}}{\partial s_\mathrm{i}} = c_\mathrm{i} \nonumber \\[1mm]
\frac{\partial q_\mathrm{ch}}{\partial s_\mathrm{g}} & = \frac{\partial q_\mathrm{g}}{\partial s_\mathrm{g}} = c_\mathrm{g}. \nonumber
\end{align}
\end{linenomath*}
All partial derivatives of $r$, $e_\mathrm{1}$, $r_\mathrm{i}$ and $r_\mathrm{g}$ with respect to the states are obtained by combining the above auxiliary derivatives with the constitutive relationships.

\subsubsection{Jacobian of the system dynamics with respect to parameters}
The $m \times d$ Jacobian $\mathbf{J}_{f}(\boldsymbol{\uptheta})$ of the system dynamics $\mathbf{f}(\mathbf{x},\boldsymbol{\uptheta},t)$ with respect to the parameters
\begin{linenomath*}
\begin{align}
\boldsymbol{\uptheta} =
(f_\mathrm{p}, A_\mathrm{im}, a, b, f_\mathrm{wm}, f_\mathrm{lm}, c, s_\mathrm{tot}, \beta, k_\mathrm{i}, k_\mathrm{g}, c_\mathrm{i}, c_\mathrm{g}, k_\mathrm{f})^{\top} \nonumber
\end{align}
\end{linenomath*}
is written row-wise as
\begin{linenomath*}
\begin{align}
\mathbf{J}_{f}(\boldsymbol{\uptheta}) & =
\frac{\partial \mathbf{f}(\mathbf{x},\boldsymbol{\uptheta},t)}{\partial \boldsymbol{\uptheta}^{\top}} =
\colvec{1}{c}{ \; \mathbf{j}_{1}^\top \; \\[1mm]
 \; \mathbf{j}_{2}^\top  \; \\[1mm]
 \; \mathbf{j}_{3}^\top  \; \\[1mm]
 \; \mathbf{j}_{4}^\top  \; \\[1mm]
 \; \mathbf{j}_{5}^\top  \; \\[1mm]
 \; \mathbf{j}_{6}^\top  \; \\[1mm]
 \; \mathbf{j}_{7}^\top  \; \\[1mm]
 \; \mathbf{j}_{8}^\top  \; } \in \mathbb{R}^{8 \times 14},
\end{align}
\end{linenomath*}
with rows ordered according to $f_{1},\ldots,f_{8}$.

\paragraph*{Row 1: $f_1 = \mathrm{d}x_{1}/\mathrm{d}t = p_\mathrm{i} - e_\mathrm{1} - r$}
\begin{linenomath*}
\begin{align}
\mathbf{j}^{\top}_{1} & =
\biggl( - \frac{\partial e_\mathrm{1}}{\partial f_\mathrm{p}} - \frac{\partial r}{\partial f_\mathrm{p}}, \; \frac{\partial p_\mathrm{i}}{\partial A_\mathrm{im}} - \frac{\partial r}{\partial A_\mathrm{im}}, \; - \frac{\partial r}{\partial a}, \; - \frac{\partial r}{\partial b}, \; - \frac{\partial e_\mathrm{1}}{\partial f_\mathrm{wm}} - \frac{\partial r}{\partial f_\mathrm{wm}}, \; - \frac{\partial e_\mathrm{1}}{\partial f_\mathrm{lm}}, \; - \frac{\partial e_\mathrm{1}}{\partial c}, \;
- \frac{\partial e_\mathrm{1}}{\partial s_\mathrm{tot}} - \frac{\partial r}{\partial s_\mathrm{tot}}, \nonumber \\[1mm]
& \hspace{7mm} - \frac{\partial r}{\partial \beta}, \; 0, \; 0, \; 0, \; 0, \; 0 \biggr), \nonumber
\end{align}
\end{linenomath*}
where $\partial p_\mathrm{i}/\partial A_\mathrm{im} = -p_{t}$.

\paragraph*{Row 2: $f_2 = \mathrm{d}x_{2}/\mathrm{d}t = r - r_\mathrm{i} - r_\mathrm{g}$}
\begin{linenomath*}
\begin{align}
\mathbf{j}^{\top}_{2} & = \biggl( \frac{\partial r}{\partial f_\mathrm{p}}, \; \frac{\partial r}{\partial A_\mathrm{im}}, \;
\frac{\partial r}{\partial a}, \; \frac{\partial r}{\partial b}, \;
\frac{\partial r}{\partial f_\mathrm{wm}} - \frac{\partial r_\mathrm{i}}{\partial f_\mathrm{wm}} - \frac{\partial r_\mathrm{g}}{\partial f_\mathrm{wm}}, \; \frac{\partial r}{\partial f_\mathrm{lm}}, \; \frac{\partial r}{\partial c}, \;
\frac{\partial r}{\partial s_\mathrm{tot}} - \frac{\partial r_\mathrm{i}}{\partial s_\mathrm{tot}} - \frac{\partial r_\mathrm{g}}{\partial s_\mathrm{tot}}, \nonumber \\[1mm]
& \hspace{7mm} \frac{\partial r}{\partial \beta} - \frac{\partial r_\mathrm{i}}{\partial \beta} - \frac{\partial r_\mathrm{g}}{\partial \beta}, \; - \frac{\partial r_\mathrm{i}}{\partial k_\mathrm{i}}, \; - \frac{\partial r_\mathrm{g}}{\partial k_\mathrm{g}}, \; 0, \; 0, \; 0 \biggr). \nonumber
\end{align}
\end{linenomath*}

\paragraph*{Row 3: $f_3 = \mathrm{d}x_{3}/\mathrm{d}t = r_\mathrm{i} - q_\mathrm{i}$}
\begin{linenomath*}
\begin{align}
\mathbf{j}^{\top}_{3} & = \biggl( 0, \; 0, \; 0, \; 0, \; \frac{\partial r_\mathrm{i}}{\partial f_\mathrm{wm}}, \; 0, \; 0, \;
\frac{\partial r_\mathrm{i}}{\partial s_\mathrm{tot}}, \; \frac{\partial r_\mathrm{i}}{\partial \beta}, \; \frac{\partial r_\mathrm{i}}{\partial k_\mathrm{i}}, \; 0, \; - \frac{\partial q_\mathrm{i}}{\partial c_\mathrm{i}}, \; 0, \; 0 \biggr). \nonumber
\end{align}
\end{linenomath*}

\paragraph*{Row 4: $f_4 = \mathrm{d}x_{4}/\mathrm{d}t = r_\mathrm{g} - q_\mathrm{g}$}
\begin{linenomath*}
\begin{align}
\mathbf{j}^{\top}_{4} & = \biggl( 0, \; 0, \; 0, \; 0, \; \frac{\partial r_\mathrm{g}}{\partial f_\mathrm{wm}}, \; 0, \; 0, \;
\frac{\partial r_\mathrm{g}}{\partial s_\mathrm{tot}}, \; \frac{\partial r_\mathrm{g}}{\partial \beta}, \; 0, \; \frac{\partial r_\mathrm{g}}{\partial k_\mathrm{g}}, \; 0, \;
- \frac{\partial q_\mathrm{g}}{\partial c_\mathrm{g}}, \; 0
\biggr). \nonumber
\end{align}
\end{linenomath*}

\paragraph*{Row 5: $f_5 = \mathrm{d}x_{5}/\mathrm{d}t = q_\mathrm{ch} - k_\mathrm{f} s^{1}_\mathrm{r}$}
\begin{linenomath*}
\begin{align}
\mathbf{j}^{\top}_{5} & = \biggl( \frac{\partial q_\mathrm{ch}}{\partial f_\mathrm{p}}, \; \frac{\partial q_\mathrm{ch}}{\partial A_\mathrm{im}}, \; \frac{\partial q_\mathrm{ch}}{\partial a}, \;
\frac{\partial q_\mathrm{ch}}{\partial b}, \; \frac{\partial q_\mathrm{ch}}{\partial f_\mathrm{wm}}, \; \frac{\partial q_\mathrm{ch}}{\partial f_\mathrm{lm}}, \; \frac{\partial q_\mathrm{ch}}{\partial c}, \; \frac{\partial q_\mathrm{ch}}{\partial s_\mathrm{tot}}, \; \frac{\partial q_\mathrm{ch}}{\partial \beta}, \; \frac{\partial q_\mathrm{ch}}{\partial k_\mathrm{i}}, \;
\frac{\partial q_\mathrm{ch}}{\partial k_\mathrm{g}}, \; \frac{\partial q_\mathrm{ch}}{\partial c_\mathrm{i}}, \; \frac{\partial q_\mathrm{ch}}{\partial c_\mathrm{g}}, \; - s^{1}_\mathrm{r} \biggr), \nonumber
\end{align}
\end{linenomath*}
where each $\partial q_\mathrm{ch}/\partial \theta$ is the sum of the corresponding derivatives of $q_\mathrm{s}$, $q_\mathrm{i}$ and $q_\mathrm{g}$.

\paragraph*{Row 6: $f_6 = \mathrm{d}x_{6}/\mathrm{d}t = k_\mathrm{f} s^{1}_\mathrm{r} - k_\mathrm{f} s^{2}_\mathrm{r}$}
\begin{linenomath*}
\begin{align}
\mathbf{j}^{\top}_{6} & = \bigl( 0, \; 0, \; 0, \; 0, \; 0, \; 0, \; 0, \; 0, \; 0, \; 0, \; 0, \; 0, \; 0, \; s^{1}_\mathrm{r} - s^{2}_\mathrm{r} \bigr). \nonumber
\end{align}
\end{linenomath*}

\paragraph*{Row 7: $f_7 = \mathrm{d}x_{7}/\mathrm{d}t = k_\mathrm{f} s^{2}_\mathrm{r} - k_\mathrm{f} s^{3}_\mathrm{r}$}
\begin{linenomath*}
\begin{align}
\mathbf{j}^{\top}_{7} & = \bigl( 0, \; 0, \; 0, \; 0, \; 0, \; 0, \; 0, \; 0, \; 0, \; 0, \; 0, \; 0, \; 0, \; s^{2}_\mathrm{r} - s^{3}_\mathrm{r} \bigr). \nonumber
\end{align}
\end{linenomath*}

\paragraph*{Row 8: $f_8 = \mathrm{d}x_{8}/\mathrm{d}t = k_\mathrm{f} s^{3}_\mathrm{r}$}
\begin{linenomath*}
\begin{align}
\mathbf{j}^{\top}_{8} & = \bigl( 0, \; 0, \; 0, \; 0, \; 0, \; 0, \; 0, \; 0, \; 0, \; 0, \; 0, \; 0, \; 0, \; s^{3}_\mathrm{r} \bigr). \nonumber
\end{align}
\end{linenomath*}\\[-2mm]
This completes the analytic specification of the augmented ODE system for the \texttt{xinanjiang} model. 

\subsubsection{Chain-rule corrections for \texorpdfstring{$f_\mathrm{wm}$}{fwm}, \texorpdfstring{$f_\mathrm{lm}$}{flm} and \texorpdfstring{$s_\mathrm{tot}$}{stot}}
The internal storage capacities
\begin{linenomath*}
\begin{align}
w_{\max} = f_\mathrm{wm}\, s_\mathrm{tot}, \qquad s_{\max} = (1 - f_\mathrm{wm})\, s_\mathrm{tot}, \qquad \mathrm{lm} = f_\mathrm{lm}\, w_{\max}, \nonumber
\end{align}
\end{linenomath*}
depend on the parameters $f_\mathrm{wm}$, $f_\mathrm{lm}$ and $s_\mathrm{tot}$. Consequently, any scalar flux $g$ appearing in the vector field $\mathbf{f} = (f_{1},\ldots,f_{8})^{\top}$ (e.g., $g = r, r_\mathrm{i}, r_\mathrm{g}, e_{1}, q_\mathrm{ch}$) must satisfy the chain rule
\begin{linenomath*}
\begin{align}
\frac{\partial g}{\partial f_\mathrm{wm}} & = \frac{\partial g}{\partial w_{\max}} \frac{\partial w_{\max}}{\partial f_\mathrm{wm}}
+ \frac{\partial g}{\partial \mathrm{lm}} \frac{\partial \mathrm{lm}}{\partial f_\mathrm{wm}} + \frac{\partial g}{\partial s_{\max}}
\frac{\partial s_{\max}}{\partial f_\mathrm{wm}}, 
\label{eq:chain_fwm} 
\\[2mm]
\frac{\partial g}{\partial f_\mathrm{lm}} & = \frac{\partial g}{\partial \mathrm{lm}} \, \frac{\partial \mathrm{lm}}{\partial f_\mathrm{lm}}, 
\label{eq:chain_flm}
\\[2mm]
\frac{\partial g}{\partial s_\mathrm{tot}} & = \frac{\partial g}{\partial w_{\max}} \frac{\partial w_{\max}}{\partial s_\mathrm{tot}}
+ \frac{\partial g}{\partial \mathrm{lm}}\frac{\partial \mathrm{lm}}{\partial s_\mathrm{tot}} + \frac{\partial g}{\partial s_{\max}} \frac{\partial s_{\max}}{\partial s_\mathrm{tot}}.
\label{eq:chain_stot}
\end{align}
\end{linenomath*}
The required partial derivatives of $w_{\max}$, $s_{\max}$ and $\mathrm{lm}$ with respect to $f_\mathrm{wm}$, $f_\mathrm{lm}$ and $s_\mathrm{tot}$ are given in the auxiliary-derivatives section and are substituted into \ref{eq:chain_fwm}--\ref{eq:chain_stot} as needed. All parameter derivatives $\partial f_{k}/\partial \theta_{j}$ reported in the parameter Jacobian $\mathbf{J}_{f}(\boldsymbol{\uptheta})$ already include these chain-rule contributions.

This concludes the description of all hydrologic models. 

\newpage

\section[\appendixname~\thesection]{Sensitivity vectors $\boldsymbol{\updelta}_{n}$ of common loss functions}\label{sec:AppendixC}
\renewcommand{\theequation}{\thesection.\arabic{equation}}
\setcounter{equation}{0}
\setcounter{algorithm}{0}

In this Appendix we derive analytic expressions for the loss sensitivity vectors $\boldsymbol{\updelta}_{n}$ of the SAR, GLS, NSE, KGE, Huber and FDC-based cost functions. Then, we also present algorithmic recipes and concise \textsc{Matlab}-codes for their computation.

\subsection{Sum of absolute residuals}\label{subsecApp:SAR}
The sum of absolute residuals or $L_{1}$ loss function is equal to 
\begin{linenomath*}
\begin{align}
\mathcal{L}_\mathrm{sar}(\boldsymbol{\uptheta}) & = \sum_{t=1}^{n} \lvert y_{t} - q_{t}(\boldsymbol{\uptheta}) \rvert
\label{eqApp:L_sar(uptheta)}
\end{align}
\end{linenomath*}
Differentiating this loss w.r.t.\ $q_{t}$ gives
\begin{linenomath*}
\begin{align}
\frac{\partial \mathcal{L}_\mathrm{sar}(\boldsymbol{\uptheta})}{\partial q_{t}} &
= - \sign\bigl(e_{t}(\boldsymbol{\uptheta})\bigr), \qquad t = 1,\dots,n, \nonumber
\end{align}
\end{linenomath*}
where $\sign(x)$ is the signum function. This function returns $-1$ if $x < 0$, $0$ if $x = 0$ and $1$ if $x > 0$. In vector form we yield
\begin{linenomath*}
\begin{align}
\boldsymbol{\updelta}_{n,\mathrm{sar}}(\boldsymbol{\uptheta}) = \frac{\partial \mathcal{L}_\mathrm{sar}(\boldsymbol{\uptheta})}{\partial \mathbf{q}_{n}} & = - \sign\bigl(\mathbf{e}_{n}(\boldsymbol{\uptheta})\bigr).
\label{eqApp:delta_sar}
\end{align}
\end{linenomath*}

Algorithm~\ref{alg:delta_sar} provides a step-by-step recipe for computing the SAR loss sensitivity vector $\boldsymbol{\updelta}_{n,\mathrm{sar}}(\boldsymbol{\uptheta})$
\begin{algorithm}[H]
\caption{Sum of absolute residuals loss-sensitivity vector $\boldsymbol{\updelta}_{n}(\boldsymbol{\uptheta})$}
\begin{algorithmic}
\State \textbf{Input:} $n$-vectors of observed $\mathbf{y}_{n} = (y_{1},\ldots,y_{n})^{\top}$ and simulated $\mathbf{q}_{n} = (q_{1},\ldots,q_{n})^{\top}$ values.
\State \textbf{Output:} SAR loss sensitivity vector $\boldsymbol{\updelta}_{n}(\boldsymbol{\uptheta}) \in \mathbb{R}^{n}$. \vspace{1mm}
\Statex \hspace{1em} Compute $n \times 1$ vector of residuals $\mathbf{e}_{n} = \mathbf{y}_{n} - \mathbf{q}_{n}$.
\Statex \hspace{1em} Compute entries of loss sensitivity vector, $\delta_{t}(\boldsymbol{\uptheta}) = - \sign(e_{t})$ for all $t = 1,\ldots,n$. \vspace{1mm}
\State \textbf{Return:} $\boldsymbol{\updelta}_{n} = \bigl(\delta_{1},\ldots,\delta_{n})^{\top}$.
\end{algorithmic}
\label{alg:delta_sar}
\end{algorithm}
\noindent and the inset presents \textsc{Matlab}-style pseudocode.
\begin{tcolorbox}[colback=black!5!white,colframe=black!40!white]
\begin{lstlisting}[style=matlab]
function delta = delta_sar(y_n,q_n)
% DELTA_SAR Loss sensitivity vector sum of absolute residuals
  e_n = y_n - q_n;         % nx1 vector of residuals
  delta = - sign(e_n);     % SAR loss sensitivity
end
\end{lstlisting}
\end{tcolorbox}

\noindent This leaves us with the gradient of $\mathcal{L}_\mathrm{sar}$ with respect to the parameters
\begin{linenomath*}
\begin{align}
\mathbf{g}_{n,\mathrm{sar}}(\boldsymbol{\uptheta}) & = \frac{\partial \mathbf{q}_{n}}{\partial \boldsymbol{\uptheta}^{\top}} \frac{\partial \mathcal{L}_\mathrm{sar}(\boldsymbol{\uptheta})}{\partial \mathbf{q}_{n}} = \mathbf{J}^{\top}_{q}(\boldsymbol{\uptheta})\boldsymbol{\updelta}_{n,\mathrm{sar}}(\boldsymbol{\uptheta}).
\label{eqApp:g_nsar(uptheta)}
\end{align}
\end{linenomath*}

\subsection{Generalized least squares}\label{subsecApp:GLS}
For a generalized least squares (GLS) or $\ell_{2}$ loss function
\begin{linenomath*}
\begin{align}
\mathcal{L}_{\mathrm{gls}}(\boldsymbol{\uptheta})
& = \tfrac{1}{2}\bigl(\mathbf{y}_{n} - \mathbf{q}_{n}(\boldsymbol{\uptheta})\bigr)^{\top}
\boldsymbol{\Sigma}_{\epsilon}^{-1}\bigl(\mathbf{y}_{n} - \mathbf{q}_{n}(\boldsymbol{\uptheta})\bigr) \nonumber \\
& = \tfrac{1}{2}\,\mathbf{e}_{n}^{\top}(\boldsymbol{\uptheta})\boldsymbol{\Sigma}_{\epsilon}^{-1}\,\mathbf{e}_{n}(\boldsymbol{\uptheta}) \nonumber \\
& = \tfrac{1}{2}\,\mathbf{e}_{n}^{\top}(\boldsymbol{\uptheta})\mathbf{W}_{\!n}^{\top}\mathbf{W}_{\!n}\,\mathbf{e}_{n}(\boldsymbol{\uptheta}) \nonumber \\
& = \tfrac{1}{2}\,\bigl(\mathbf{W}_{\!n}\mathbf{e}_{n}(\boldsymbol{\uptheta})\bigr)^{\top}\bigl(\mathbf{W}_{\!n}\mathbf{e}_{n}(\boldsymbol{\uptheta})\bigr) \nonumber \\
& = \tfrac{1}{2}\,\mathbf{e}^{\ast}_{n}\boldsymbol{\uptheta})^{\top}\mathbf{e}^{\ast}_{n}(\boldsymbol{\uptheta}),
\label{eqApp:L_GLS(uptheta)}
\end{align}
\end{linenomath*}
where $\boldsymbol{\Sigma}_{\epsilon}$ is the $n \times n$ covariance matrix of the discharge measurement errors and
\begin{linenomath*}
\begin{align}
\mathbf{e}_{n}(\boldsymbol{\uptheta}) & = \mathbf{y}_{n} - \mathbf{q}_{n} (\boldsymbol{\uptheta}) \in \mathbb{R}^{n \times 1}
\quad \text{and} \quad \mathbf{e}^{\ast}_{n}(\boldsymbol{\uptheta}) = \mathbf{W}_{\!n} \, \mathbf{e}_{n}(\boldsymbol{\uptheta})
\in \mathbb{R}^{n \times 1}, \nonumber
\end{align}
\end{linenomath*}
denote the ordinary and ``whitened'' residual vectors, respectively, and
\begin{linenomath*}
\begin{align}
\mathbf{W}_{\!n} = \boldsymbol{\Sigma}_{e}^{-1/2} & = \colvec{1}{c}{ \; w_{1,1} & w_{1,2} & \hdots & w_{1,n} \; \\[1mm]
\; w_{2,1} & w_{2,2} & \hdots & w_{1,n} \; \\[1mm]
\; \vdots & \vdots & \ddots & \vdots \; \\
\; w_{n,1} & w_{n,2} & \hdots & w_{n,n} \; } \in \mathbb{R}^{n \times n}, \nonumber
\end{align}    
\end{linenomath*}
is a symmetric square root weight matrix. The whitened residuals $\mathbf{e}^{\ast}_{n}$ are also referred to as partial residuals in the context of distribution-adaptive likelihood functions \citep{schoups2010b,vrugt2022b}.

The derivative of the GLS loss with respect to the simulated discharge $\mathbf{q}_{t}$ follows directly from the chain rule. Since $\mathbf{e}_{t} = \mathbf{y}_{t} - \mathbf{q}_{t}$, we obtain
\begin{linenomath*}
\begin{align}
\frac{\partial \mathcal{L}_\mathrm{gls}(\boldsymbol{\uptheta})}{\partial q_{t}} & = - \boldsymbol{\Sigma}^{-1}_{\epsilon}\bigl(y_{t} - q_{t}(\boldsymbol{\uptheta})\bigr), \qquad t = 1,\dots,n. \nonumber
\end{align}
\end{linenomath*}
In vector form we can write
\begin{linenomath*}
\begin{align}
\boldsymbol{\updelta}_{n,\mathrm{gls}}(\boldsymbol{\uptheta}) = \frac{\partial \mathcal{L}_\mathrm{gls}(\boldsymbol{\uptheta})}{\partial \mathbf{q}_{n}} & = - \boldsymbol{\Sigma}^{-1}_{\epsilon}\bigl(\mathbf{y}_{n} - \mathbf{q}_{n}(\boldsymbol{\uptheta})\bigr).
\label{eqApp:delta_gls}
\end{align}
\end{linenomath*}

Algorithm~\ref{algApp:delta_GLS} provides a step-by-step recipe for computing $\boldsymbol{\updelta}_{n,\mathrm{gls}}(\boldsymbol{\uptheta})$
\begin{algorithm}[H]
\caption{Generalized least squares loss-sensitivity vector $\boldsymbol{\updelta}_{n}(\boldsymbol{\uptheta})$}
\begin{algorithmic}
\State \textbf{Input:} $n$-vectors of observed $\mathbf{y}_{n} = (y_{1},\ldots,y_{n})^{\top}$ and simulated $\mathbf{q}_{n} = (q_{1},\ldots,q_{n})^{\top}$ values.
\State \phantom{\textbf{Input:}} The $n \times n$ measurement error covariance matrix, $\boldsymbol{\Sigma}_{\epsilon}$ of the discharge observations.
\State \textbf{Output:} GLS loss sensitivity vector $\boldsymbol{\updelta}_{n}(\boldsymbol{\uptheta}) \in \mathbb{R}^{n}$. \vspace{1mm}
\Statex \hspace{1em} Compute $n \times 1$ vector of residuals $\mathbf{e}_{n} = \mathbf{y}_{n} - \mathbf{q}_{n}$.
\Statex \hspace{1em} Compute loss sensitivity vector, $\boldsymbol{\updelta}_{n}(\boldsymbol{\uptheta}) = - \boldsymbol{\Sigma}_{\epsilon}\mathbf{e}_{n}$. \vspace{1mm}
\State \textbf{Return:} $\boldsymbol{\updelta}_{n} = \bigl(\delta_{1},\ldots,\delta_{n})^{\top}$.
\end{algorithmic}
\label{algApp:delta_GLS}
\end{algorithm}
\noindent and a \textsc{Matlab}-style pseudocode is presented in the inset below.
\begin{tcolorbox}[colback=black!5!white,colframe=black!40!white]
\begin{lstlisting}[style=matlab]
function delta = delta_gls(y_n,q_n,Sigma_e)
% DELTA_GLS Loss sensitivity vector generalized least squares
  e_n = y_n - q_n;         % nx1 vector of residuals
  delta = - Sigma_e\e_n;   % GLS loss sensitivity
end
\end{lstlisting}
\end{tcolorbox}

For ordinary least squares (OLS), the weight matrix equals the identity, $\mathbf{W}_{n} = \mathbf{I}_{n}$, with zeros everywhere except for ones on the main diagonal. For weighted least squares (WLS), $\mathbf{W}_{n}$ is diagonal with entries $w_{t,t} = 1/\sigma_{t}$ equal to the reciprocal of the discharge measurement error standard deviations \citep{sorooshian1980}. In the most general case of GLS, $\mathbf{W}_{n}$ is a full symmetric matrix that simultaneously accounts for heteroscedasticity and temporal autocorrelation in the discharge measurement errors.

This leaves us with the gradient of $\mathcal{L}_\mathrm{gls}$ with respect to the parameters
\begin{linenomath*}
\begin{align}
\mathbf{g}_{n,\mathrm{gls}}(\boldsymbol{\uptheta}) = \frac{\partial \mathbf{q}_{n}}{\partial \boldsymbol{\uptheta}^{\top}} \frac{\partial \mathcal{L}_\mathrm{gls}(\boldsymbol{\uptheta})}{\partial \mathbf{q}_{n}} & = \mathbf{J}^{\top}_{q}(\boldsymbol{\uptheta})\boldsymbol{\updelta}_{n,\mathrm{gls}}(\boldsymbol{\uptheta}).
\label{eqApp:g_ngls(uptheta)}
\end{align}
\end{linenomath*}

\subsection{Nash-Sutcliffe Efficiency}\label{subsecApp:NSE}
The \citet{nash1970} efficiency, $\mathrm{NSE}: \ \mathbb{R}^{n} \times \mathbb{R}^{n} \to (-\infty,1]$, is defined as
\begin{linenomath*}
\begin{align}
\mathrm{NSE}(\mathbf{y}_{n},\mathbf{q}_{n}) & = 1 - \frac{\sum_{t=1}^{n} (y_{t} - q_{t})^{2}}{\sum_{t=1}^{n} (y_{t} - m_{y})^2} = 1 - \frac{\mathrm{SS}_\mathrm{r}(\mathbf{y}_{n},\mathbf{q}_{n})}{\mathrm{SS}_\mathrm{t}(\mathbf{y}_{n})}, \nonumber
\end{align}
\end{linenomath*}
where $m_{y} = \frac{1}{n}\sum_{t=1}^{n}y_{t}$ denotes the sample mean of the discharge observations and $\mathrm{SS}_\mathrm{r}$ and $\mathrm{SS}_\mathrm{t}$ are the residual and total sum of squares, respectively.

To use NSE in gradient descent, we typically \emph{minimize} the squared-error fraction
\begin{linenomath*}
\begin{align}
\mathcal{L}_\mathrm{nse}(\boldsymbol{\uptheta}) & = 1 - \mathrm{NSE}(\mathbf{y}_{n},\mathbf{q}_{n}) = \frac{\mathrm{SS}_\mathrm{r}(\mathbf{y}_{n},\mathbf{q}_{n})}{\mathrm{SS}_\mathrm{t}(\mathbf{y}_{n})}.
\label{eqApp:L_nse(uptheta)}
\end{align}
\end{linenomath*}
Differentiating w.r.t.\ $q_{t}$ gives
\begin{linenomath*}
\begin{align}
\frac{\partial \mathcal{L}_\mathrm{nse}(\boldsymbol{\uptheta})}{\partial q_{t}} &
= \frac{2}{\mathrm{SS}_\mathrm{t}(\mathbf{y}_{n})}(q_{t} - y_{t}) = - \frac{2}{\mathrm{SS}_\mathrm{t}(\mathbf{y}_{n})} e_{t}(\boldsymbol{\uptheta}), \qquad t = 1,\dots,n. \nonumber
\end{align}
\end{linenomath*}
Collecting in vector form
\begin{linenomath*}
\begin{align}
\boldsymbol{\updelta}_{n,\mathrm{nse}}(\boldsymbol{\uptheta}) = \frac{\partial \mathcal{L}_\mathrm{nse}(\boldsymbol{\uptheta})}{\partial \mathbf{q}_{n}} & = \frac{2}{\mathrm{SS}_\mathrm{t}(\mathbf{y}_{n})}(\mathbf{q}_{n} - \mathbf{y}_{n}),
\label{eqApp:delta_nse}
\end{align}
\end{linenomath*}
and refer to $\boldsymbol{\updelta}_{n,\mathrm{nse}}(\boldsymbol{\uptheta})$ as the NSE score.

Algorithm~\ref{algApp:delta_nse} provides a step-by-step recipe for computing the NSE loss sensitivity vector $\boldsymbol{\updelta}_{n,\mathrm{nse}}(\boldsymbol{\uptheta})$
\begin{algorithm}[H]
\caption{Nash-Sutcliffe loss-sensitivity vector $\boldsymbol{\updelta}_{n}(\boldsymbol{\uptheta})$}
\begin{algorithmic}
\State \textbf{Input:} $n$-vectors of observed $\mathbf{y}_{n} = (y_{1},\ldots,y_{n})^{\top}$ and simulated $\mathbf{q}_{n} = (q_{1},\ldots,q_{n})^{\top}$ values.
\State \textbf{Output:} NSE loss sensitivity vector $\boldsymbol{\updelta}_{n}(\boldsymbol{\uptheta}) \in \mathbb{R}^{n}$. \vspace{1mm}
\Statex \hspace{1em} Compute $n \times 1$ vector of residuals $\mathbf{e}_{n} = \mathbf{y}_{n} - \mathbf{q}_{n}$.
\Statex \hspace{1em} Compute sample mean of  measured data, $m_{y} = \frac{1}{n}\sum_{t=1}^{n} y_{t}$.
\Statex \hspace{1em} Compute total sum of squares, $\mathrm{SS}_\mathrm{t} = \sum_{t=1}^{n} (y_{t} - m_{y})^{2}$.
\Statex \hspace{1em} Compute entries of loss sensitivity vector, $\delta_{t}(\boldsymbol{\uptheta}) = - 2e_{t}/\mathrm{SS}_\mathrm{t}$, for all $t = 1,\ldots,n$.  \vspace{1mm}
\State \textbf{Return:} $\boldsymbol{\updelta}_{n} = \bigl(\delta_{1},\ldots,\delta_{n})^{\top}$.
\end{algorithmic}
\label{algApp:delta_nse}
\end{algorithm}
\noindent and the inset below presents \textsc{Matlab}-style pseudocode.
\begin{tcolorbox}[colback=black!5!white,colframe=black!40!white]
\begin{lstlisting}[style=matlab]
function delta = delta_nse(y_n,q_n)
% DELTA_NSE Loss sensitivity vector Nash-Sutclife efficiency
  e_n = y_n - q_n;         % nx1 vector of residuals
  m_y = mean(y_n);         % sample mean of y_n
  SSt = sum((y_n-m_y).^2); % total sum of squares of y_n
  delta = -2/SSt * e_n     % NSE loss sensitivity 
end
\end{lstlisting}
\end{tcolorbox}

\noindent This leaves us with the gradient of $\mathcal{L}_\mathrm{nse}$ with respect to the parameters
\begin{linenomath*}
\begin{align}
\mathbf{g}_{n,\mathrm{nse}}(\boldsymbol{\uptheta}) & = \frac{\partial \mathbf{q}_{n}}{\partial \boldsymbol{\uptheta}^{\top}} \frac{\partial \mathcal{L}_\mathrm{nse}(\boldsymbol{\uptheta})}{\partial \mathbf{q}_{n}} = \mathbf{J}^{\top}_{q}(\boldsymbol{\uptheta})\boldsymbol{\updelta}_{n,\mathrm{nse}}(\boldsymbol{\uptheta}).
\label{eqApp:g_nnse(uptheta)}
\end{align}
\end{linenomath*}

\subsection{Kling-Gupta Efficiency}\label{subsecApp:KGE}
The Kling--Gupta efficiency or KGE of \citet{gupta2009} is a widely used alternative to the NSE for evaluating hydrologic model performance. The $\mathrm{KGE}: \ \mathbb{R}^{n} \times \mathbb{R}^{n} \to (-\infty,1]$ addresses known NSE limitations related to its disproportionate sensitivity to high flows and combines three quasi-orthogonal measures of model performance
\begin{linenomath*}
\begin{equation}
\mathrm{KGE}(\mathbf{y}_{n}, \mathbf{q}_{n}) = 1 - \sqrt{(r_{qy} - 1)^2 + (\nu_{qy} - 1)^2 + (b_{qy} - 1)^2}, \nonumber
\end{equation}
\end{linenomath*}
where the \emph{sample} Pearson correlation coefficient $r$ of measured and simulated data
\begin{linenomath*}
\begin{align}
r_{qy} & = \frac{\frac{1}{n-1}\sum\nolimits^{n}_{t=1}(q_{t} - m_{q})(y_{t} - m_{y})}{\sqrt{\frac{1}{n-1}\sum\nolimits^{n}_{t=1}(q_{t} - m_{q})^{2}} \, \sqrt{\frac{1}{n-1}\sum\nolimits^{n}_{t=1}(y_{t} - m_{y})^{2}}}, \nonumber
\end{align}    
\end{linenomath*}
is an estimate of the population correlation
\begin{linenomath*}
\begin{align}
\corr[Q,Y] & = \frac{\Cov[Q,Y]}{\Var[Q]^{1/2} \Var[Y]^{1/2}}, \nonumber 
\end{align}    
\end{linenomath*}
of measured $y \in Y$ and simulated $q \in Q$ data and scalars $\nu_{qy} = s_{q} / s_{y}$ and $b_{qy} = m_{q}/m_{y}$ are so-called variability and bias ratios, and $s_{x}$ is the sample standard deviation
\begin{linenomath*}
\begin{align}
s_{x} = \sqrt{\ltfrac{1}{n-1}\sum\nolimits_{t=1}^{n}(x_{i} - m_{x})^{2}}. \nonumber 
\end{align}    
\end{linenomath*}
To improve the separation of bias and variability effects, \citet{kling2012} replaced $\nu_{qy}$ by the ratio of the coefficients of variation,
$\upsilon = \mathrm{CV}_{q}/\mathrm{CV}_{y}$,
where $\mathrm{CV}_{q} = s_{q}/m_{q}$ and $\mathrm{CV}_{y} = s_{y}/m_{y}$ denote the coefficients of variation of the simulated and observed discharge, respectively.

To use KGE in gradient descent, we must turn this reward-based metric into a cost function
\begin{linenomath*}
\begin{align}
\mathcal{L}_\mathrm{kge}(\boldsymbol{\uptheta}) & = 1 - \mathrm{KGE}(\mathbf{y}_{n}, \mathbf{q}_{n}) = \sqrt{(r_{qy}-1)^2 + (\nu_{qy} - 1)^2 + (b_{qy} - 1)^2}.
\label{eqApp:L_kge(uptheta)}
\end{align}
\end{linenomath*}
Before we differentiate $\mathcal{L}_\mathrm{kge}$ we first considers its squared form
\begin{linenomath*}
\begin{align}    
\mathcal{L}^{2}_\mathrm{kge}(\boldsymbol{\uptheta}) & = (r_{qy}-1)^2 + (\nu_{qy} - 1)^2 + (b_{qy} - 1)^2. \nonumber
\end{align}
\end{linenomath*}
which avoids carrying the square root through intermediate steps. Indeed, using the identity 
\begin{linenomath*}
\begin{align}
\frac{\mathrm{d}}{\mathrm{d}x}\sqrt{u(x)} & = \frac{1}{2\sqrt{u(x)}}\frac{\mathrm{d}u(x)}{\mathrm{d}x}, \nonumber
\end{align}
\end{linenomath*}
we obtain
\begin{linenomath*}
\begin{align}
\frac{\partial \mathcal{L}_\mathrm{kge}(\boldsymbol{\uptheta})}{\partial q_t}
& = \frac{1}{2\,\mathcal{L}_\mathrm{kge}(\boldsymbol{\uptheta})}
\frac{\partial \mathcal{L}_\mathrm{kge}^2(\boldsymbol{\uptheta})}{\partial q_t}. \nonumber
\end{align}
\end{linenomath*}
Thus, differentiating $\mathcal{L}_\mathrm{kge}$ w.r.t.\ $q_{t}$ yields
\begin{linenomath*}
\begin{align}
\frac{\partial \mathcal{L}_\mathrm{kge}(\boldsymbol{\uptheta})}{\partial q_{t}} & = 
\frac{1}{\mathcal{L}_\mathrm{kge}(\boldsymbol{\uptheta})}\biggl[(r_{qy}-1)\frac{\partial r_{qy}}{\partial q_{t}} + (\nu_{qy} - 1) \frac{\partial \nu_{qy}}{\partial q_{t}} + (b_{qy} - 1) \frac{\partial b_{qy}}{\partial q_{t}}\biggr], \nonumber
\end{align}
\end{linenomath*}
where $\partial r_{qy}/\partial q_{t}$,
$\partial \nu_{qy}/\partial q_{t}$ and $\partial b_{qy}/\partial q_{t}$ can be derived using standard variance and covariance calculus (see Appendix \ref{sec:AppendixD}) to yield
\begin{linenomath*}
\begin{align}
\frac{\partial r_{qy}}{\partial q_{t}} & = \frac{y_{t} - m_{y}}{(n-1)\,s_{q}\,s_{y}} - r_{qy}\frac{q_{t} - m_{q}}{(n-1)\,s_{q}^2}, \label{eqApp:dr_dq} \\
\frac{\partial \nu_{qy}}{\partial q_{t}} & =
\frac{1}{s_{y}} \frac{\partial s_{q}}{\partial q_{t}} = \frac{q_{t} - m_{q}}{(n-1)\,s_{q}\,s_{y}},
\label{eqApp:dnu_dq} \\
\frac{\partial b_{qy}}{\partial q_{t}} & = \frac{1}{m_{y}} \frac{\partial m_{q}}{\partial q_{t}} = \frac{1}{n\,m_{y}}.
\label{eqApp:db_dq}
\end{align}
\end{linenomath*}
We can collect the individual derivatives in a $n \times 1$ vector $\partial \mathcal{L}_\mathrm{kge}(\boldsymbol{\uptheta}) / \partial \mathbf{q}_{n}$ as follows
\begin{linenomath*}
\begin{align}
\boldsymbol{\updelta}_{n,\mathrm{kge}}(\boldsymbol{\uptheta}) = \frac{\partial \mathcal{L}_\mathrm{kge}(\boldsymbol{\uptheta})}{\partial \mathbf{q}_{n}} & = \left( \frac{\partial \mathcal{L}_\mathrm{kge}(\boldsymbol{\uptheta})}{\partial q_{1}}, \dots, \frac{\partial \mathcal{L}_\mathrm{kge}(\boldsymbol{\uptheta})}{\partial q_{n}} \right)^\top.
\label{eqApp:delta_kge}
\end{align}
\end{linenomath*}

Algorithm~\ref{algApp:delta_kge} provides a step-by-step recipe for computing the KGE loss sensitivity vector $\boldsymbol{\updelta}_{n,\mathrm{kge}}(\boldsymbol{\uptheta})$
\begin{algorithm}[H]
\caption{Kling-Gupta loss-sensitivity vector $\boldsymbol{\updelta}_{n}(\boldsymbol{\uptheta})$}
\begin{algorithmic}
\State \textbf{Input:} $n$-vectors of observed $\mathbf{y}_{n} = (y_{1},\ldots,y_{n})^{\top}$ and simulated $\mathbf{q}_{n} = (q_{1},\ldots,q_{n})^{\top}$ values.
\State \textbf{Output:} KGE loss sensitivity vector $\boldsymbol{\updelta}_{n}(\boldsymbol{\uptheta}) \in \mathbb{R}^{n}$. \vspace{1mm}
\Statex \hspace{1em} Compute sample means, $m_{y}$ and $m_{q}$, of measured and simulated data.
\Statex \hspace{1em} Compute sample variances, $s^{2}_{y}$ and $s^{2}_{q}$, of measured and simulated data.
\Statex \hspace{1em} Compute sample correlation coefficient $r_{qy}$ of measured and simulated data.
\Statex \hspace{1em} Compute bias $b_{qy} = m_{y}/m_{q}$ and variability $\nu_{qy} = s_{y}/s_{q}$ ratios.
\Statex \hspace{1em} Compute KGE loss $\mathcal{L}_\mathrm{kge} = \sqrt{(r_{qy}-1)^{2} + (\nu_{qy}-1)^{2} + (b_{qy}-1)^{2}}$.
\Statex \hspace{1em} Compute $\partial r_{qy}/\partial q_{t}$, $\partial \nu_{qy}/\partial q_{t}$ and $\partial b_{qy}/\partial q_{t}$ using Equations 
\ref{eqApp:dr_dq} - \ref{eqApp:db_dq}.
\Statex \hspace{1em} Compute $\delta_{t}(\boldsymbol{\uptheta}) = \mathcal{L}^{-1}_\mathrm{kge}\{(r_{qy}-1)\partial r_{qy}/\partial q_{t} + (\nu_{qy}-1)\partial \nu_{qy}/\partial q_{t} + (b_{qy}-1)\partial b_{qy}/\partial q_{t}\}$ 
\Statex \hspace{2em} for $t = 1,\ldots,n$. \vspace{1mm}
\State \textbf{Return:} $\boldsymbol{\updelta}_{n} = \bigl(\delta_{1},\ldots,\delta_{n})^{\top}$.
\end{algorithmic}
\label{algApp:delta_kge}
\end{algorithm}
\noindent and a \textsc{Matlab}-style pseudocode is presented below.
\begin{tcolorbox}[colback=black!5!white,colframe=black!40!white]
\begin{lstlisting}[style=matlab]
function delta = delta_kge(y_n,q_n)
% DELTA_KGE Loss sensitivity vector Kling-Gupta efficiency
  n = numel(q_n);        % # elements 
  m_y = mean(y_n);       % sample mean of y_n
  m_q = mean(q_n);       % sample mean of q_n
  C = cov(y_n,q_n);      % sample covariance matrix y_n and q_n
  s_y = sqrt(C(1));      % sample standard deviation of y_n
  s_q = sqrt(C(4));      % sample standard deviation of q_n
  r_qy = C(2)/(s_y*s_q); % sample correlation coefficient y_n, q_n
  v_qy = s_q/s_y;        % variability ratio
  b_qy = m_q/m_y;        % bias ratio
  L_kge = sqrt((r_qy-1)^2+(v_qy-1)^2+(b_qy-1)^2); % KGE loss
  dvdqt = (q_n-m_q)/((n-1)*s_q^2);                % dv_qy/dq_t
  dbdqt = ones(n,1)/(n*m_y);                      % db_qy/dq_t
  drdqt = (y_n-m_y)/((n-1)*s_q*s_y) ...           % dr_qy/dq_t
      - r*dvdqt;
  delta = ((r_qy-1)*drdqt+(v_qy-1)*dvdqt          % KGE loss sentvty
      + (z-1)*dzdqt)/L_kge;               
end
\end{lstlisting}
\end{tcolorbox}

\noindent This leaves us with the gradient of $\mathcal{L}_\mathrm{kge}$ with respect to the parameters
\begin{linenomath*}
\begin{align}
\mathbf{g}_{n,\mathrm{kge}}(\boldsymbol{\uptheta}) & = \frac{\partial \mathbf{q}_{n}}{\partial \boldsymbol{\uptheta}^{\top}} \frac{\partial \mathcal{L}_\mathrm{kge}(\boldsymbol{\uptheta})}{\partial \mathbf{q}_{n}} = \mathbf{J}^{\top}_{q}(\boldsymbol{\uptheta})\boldsymbol{\updelta}_{n,\mathrm{kge}}(\boldsymbol{\uptheta}).
\label{eqApp:g_nkge(uptheta)}
\end{align}
\end{linenomath*}

\subsection{M-estimator}\label{subsecApp:Huber}
The generalized least squares loss $\mathcal{L}_{t} = \frac{1}{2}e_{t}^{2}$ is unique in its mathematical convenience, but it is also maximally fragile. The quadratic penalty grows without bound, causing spurious data (outliers, extremes) to dominate the estimation of the mean. 

Robust estimation replaces this loss with alternative loss functions that control, cap, or reshape the influence of aberrant data points. These alternatives are known as M-estimators and can be viewed as ``generalized regression criteria'' with tunable robustness. The motivating ideas and principles of M-estimation originate from the pioneering work of John Wilder Tukey (1915-2000) one of the most influential mathematical and theoretical statisticians of the 20\textsuperscript{th} century. \citet{tukey1960a} pointed out the excessive sensitivity of classical statistical methods of regression analysis to small departures from idealized hypotheses. Tukey's ideas were developed further by \citet{huber1964,huber1967} and in the PhD thesis of \citet{hampel1968}, which ultimately led to the emergence of the new field of \textit{robust statistics} or M-estimation. \citet{huber1981} defines a robust statistic as one that is (p. 1) ``\ldots resistant to errors in the results, produced by deviations from assumptions''. M-estimation, where M stands for maximum-likelihood-type is a generalization of least squares estimation to situations in which the data has outliers, extreme observations and/or does not follow a normal distribution making such estimators more robust to outliers and misspecification \citep{huber1964,huber1981}. 

The \citet{huber1964} loss, for example
\begin{linenomath*}
\begin{equation}
\mathcal{L}_{c}(\underline{e}) =
\begin{cases}
\; \frac{1}{2}\underline{e}^{2}, & \lvert \underline{e} \rvert \le c,\\[1mm]
\; c \lvert \underline{e} \rvert - \frac{1}{2}c^{2}, & \lvert \underline{e} \rvert > c,
\end{cases}
\label{eqApp:Lc_huber(uptheta)}
\end{equation}
\end{linenomath*}
yields a more robust location functional $T_\mathrm{Huber}(Y)$, where $Y$ is the random variable (discharge) of interest. This functional smoothly interpolates between the mean (as $c \to \infty$) and the median (as $c \to 0$) of $Y$, and illustrates how the choice of loss function determines the parameters being estimated. The loss function $\mathcal{L}_{c}(\cdot)$ is applied to standardized residuals
\begin{linenomath*}
\begin{align}
\underline{e}_{i} = \frac{y_{i} - q_{i}}{S_{y}} = \frac{y_{i}}{S_{y}} - \frac{q_{i}}{S_{y}} = \underline{y}_{i} - \underline{q}_{i}, \qquad i = 1,\ldots,n, \nonumber
\end{align}
\end{linenomath*}
where $S_{y}$ is a robust estimate of scale ensuring that the loss is unit-invariant. A common data-derived choice for $S_{y}$ uses the median absolute deviation (MAD) from the sample median $\widetilde{y}$. The MAD is defined as
\begin{linenomath*}
\begin{align}
\sigma_{n} = \mad(y_{1},\ldots,y_{n}) = \median_{i = 1,\ldots,n} ( \lvert y_i - \widetilde{y} \rvert ). \nonumber
\end{align}
\end{linenomath*}
To obtain a scale estimator consistent under Gaussian errors, one rescales the MAD using the normal-consistency constant
\begin{linenomath*}
\begin{align}
\xi = \frac{1}{\Phi^{-1}(0.75)} \approx 1.4826, \nonumber
\end{align}
\end{linenomath*}
where $\Phi^{-1}(\tau)$ is the quantile function of the standard normal distribution evaluated at $\tau \in (0,1)$ and $S_{y} = \xi \sigma_{n}$ equals the standard deviation when $Y \sim \mathcal{N}(\mu,\sigma^{2})$.\footnote{If $X \sim \mathcal{N}(\mu,\sigma^{2})$, then $\mad(\lvert X - \mu \rvert) = \sigma\,\Phi^{-1}(0.75) \approx 0.67449\,\sigma$.} If an instrument noise level $\sigma_{0}$ is known, one may instead set $\sigma_{n} = \sigma_{0}$ and $\xi = 1$.

The total Huber loss is now equal to
\begin{linenomath*}
\begin{equation}
\mathcal{L}_\mathrm{huber}(\boldsymbol{\uptheta}) = \sum_{t=1}^{n} \mathcal{L}_{c}(\underline{e}_{t})
\label{eqApp:L_huber(uptheta)}
\end{equation}
\end{linenomath*}
The derivative of the loss $\mathcal{L}_{c}(\underline{e})$ with respect to $\underline{e}$ is the so-called influence function
\begin{linenomath*}
\begin{equation}
\psi_{c}(\underline{e}) = \frac{\mathrm{d}\mathcal{L}_{c}(\underline{e})}{\mathrm{d}\underline{e}} = 
\begin{cases}
\; \underline{e} & \lvert \underline{e} \rvert \le c,\\[1mm]
\; c \sign(\underline{e}) & \lvert \underline{e} \rvert > c.
\end{cases}
\label{eqApp:psi_huber}
\end{equation}
\end{linenomath*}
A robust estimator should have a bounded influence function so that large residuals (spurious data) do not corrupt the parameter estimates $\boldsymbol{\uptheta}$.

If we differentiate the total loss $\mathcal{L}_\mathrm{huber}(\boldsymbol{\uptheta})$ w.r.t.\,$q_{t}$ then 
\begin{linenomath*}
\begin{align}
\frac{\mathrm{d} \mathcal{L}_\mathrm{huber}(\boldsymbol{\uptheta})}{\mathrm{d}q_{t}} = \frac{\partial \mathcal{L}_{c}(\underline{e}_{t})}{\partial \underline{e}_{t}} \frac{\partial \underline{e}_{t}}{\partial q_{t}} = \psi_{c}(\underline{e}_{t})\cdot \biggl(-\frac{1}{S}\biggr) = - \frac{1}{S} \psi_{c}(\underline{e}_{t}). \nonumber
\end{align}
\end{linenomath*}
We can collect the individual entries in a $n \times 1$ vector
\begin{linenomath*}
\begin{align}
\boldsymbol{\updelta}_{n,\mathrm{huber}}(\boldsymbol{\uptheta}) = \frac{\mathrm{d} \mathcal{L}_\mathrm{huber}(\boldsymbol{\uptheta})}{\mathrm{d}\mathbf{q}_{n}} = \biggl( \frac{\mathrm{d} \mathcal{L}_\mathrm{huber}(\boldsymbol{\uptheta})}{\mathrm{d}q_{1}}, \ldots, \frac{\mathrm{d} \mathcal{L}_\mathrm{huber}(\boldsymbol{\uptheta})}{\mathrm{d}q_{n}}\biggr)^{\top},
\label{eqApp:delta_huber}
\end{align}
\end{linenomath*}
and refer to $\boldsymbol{\updelta}_{n,\mathrm{huber}}(\boldsymbol{\uptheta})$ as the Huber loss sensitivity vector, whose entries are the influence-weighted residuals $-\psi_{c}(\underline{e}_{t})/S$.

Algorithm~\ref{algApp:delta_huber} provides a step-by-step recipe for computing the Huber loss sensitivity vector $\boldsymbol{\updelta}_{n,\mathrm{huber}}(\boldsymbol{\uptheta})$
\begin{algorithm}[H]
\caption{Huber loss sensitivity (pseudo-residual) vector $\boldsymbol{\updelta}_{n}(\boldsymbol{\uptheta})$}
\begin{algorithmic}
\State \textbf{Input:} $n$-vectors of observed $\mathbf{y}_{n} = (y_{1},\ldots,y_{n})^{\top}$ and simulated $\mathbf{q}_{n} = (q_{1},\ldots,q_{n})^{\top}$ values/
\State \phantom{\textbf{Input:}} Huber score function, $\psi(\underline{e})$ of Equation \ref{eqApp:psi_huber} with threshold $c = 1.345$.
\State \phantom{\textbf{Input:}} Scalar $\xi = 1/\Phi^{-1}(0.75)$.
\State \textbf{Output:} Huber loss sensitivity vector $\boldsymbol{\updelta}_{n}(\boldsymbol{\uptheta}) \in \mathbb{R}^{n}$. \vspace{1mm}
\Statex \hspace{1em} Compute $n \times 1$ vector of residuals $\mathbf{e}_{n} = \mathbf{y}_{n} - \mathbf{q}_{n}$.
\Statex \hspace{1em} Compute the robust MAD-based scale estimator, $S_{y} = \xi \cdot \mathrm{MAD}(\mathbf{y}_{n})$.
\Statex \hspace{1em} Standardize the residuals, $\underline{\mathbf{e}}_{n} = \mathbf{e}_{n}/S_{y}$.
\Statex \hspace{1em} Compute the gradient contribution, $\delta_{t}(\boldsymbol{\uptheta}) = - \psi(\underline{e}_{t})/S_{y}$ for all $t = 1,\ldots,n$. \vspace{1mm}
\State \textbf{Return:} $\boldsymbol{\updelta}_{n} = \bigl(\delta_{1},\ldots,\delta_{n})^{\top}$.
\end{algorithmic}
\label{algApp:delta_huber}
\end{algorithm}
\noindent and the inset below presents \textsc{Matlab}-style pseudocode.
\begin{tcolorbox}[colback=black!5!white,colframe=black!40!white]
\begin{lstlisting}[style=matlab]
function delta = delta_huber(y_n,q_n)
% DELTA_HUBER Loss sensitivity vector Huber robust loss
  e_n = y_n - q_n;                  % nx1 vector of residuals
  S_y = mad(y_n,1)/norminv(0.75);   % robust scale estimator
  e_ = e_n/S_y;                     % nx1 vector of standardized residuals
  c = 1.345;                        % Huber threshold
  psi = @(x,c) (abs(x)<=c).*(x) ... % Huber score
    + (abs(x)>c).*(c*sign(x));    
  delta = -1/S*psi(e_,c);           % Huber loss sensitivity 
end
\end{lstlisting}
\end{tcolorbox}
\noindent This leaves us with the gradient of $\mathcal{L}_\mathrm{huber}$ with respect to the parameters
\begin{linenomath*}
\begin{align}
\mathbf{g}_{n,\mathrm{huber}}(\boldsymbol{\uptheta}) & = \frac{\partial \mathbf{q}_{n}}{\partial \boldsymbol{\uptheta}^{\top}} \frac{\partial \mathcal{L}_\mathrm{huber}(\boldsymbol{\uptheta})}{\partial \mathbf{q}_{n}} = \mathbf{J}^{\top}_{q}(\boldsymbol{\uptheta})\boldsymbol{\updelta}_{n,\mathrm{huber}}(\boldsymbol{\uptheta}).
\label{eqApp:g_nhuber(uptheta)}
\end{align}
\end{linenomath*}
M-estimators such as the Huber loss are less sensitive to spurious data, heavy tails, and modest departures from distributional assumptions than common maximum likelihood and least squares estimators.

\subsection{Flow-duration curve}\label{subsecApp:FDC}
The flow‐duration curve or FDC relates the exceedance probability of streamflow, $\mathbb{P}(Y > y)$, to its magnitude, $y$. This is also known as the \emph{survival function} in statistics, and the \emph{reliability function} $R_Y(y)$ in engineering
\begin{linenomath*}
\begin{align}
R_{Y}(y) & = \mathbb{P}(Y > y) = \int_{y}^{\infty}f_{Y}(t)\,\text{d}t = 1 - \int_{-\infty}^{y}f_{Y}(t)\,\text{d}t = 1 - F_{Y}(y), \nonumber
\end{align}
\end{linenomath*}
and is simply equal to the complement of the streamflow cumulative distribution function (cdf) $F_Y(y)$ \citep{vogel1994}. The FDC plays a central role in flood frequency analysis, hydrologic model evaluation, water quality management, and the design of hydroelectric power systems \citep{sadegh2016}. Existing approaches for comparing measured and simulated FDCs typically examine only \emph{partial} characteristics such as slope
\citep{yadav2007,sawicz2011,mcmillan2017}, selected percentile flows
\citep{vogel1994}, concavity indices \citep{zhang2016}, or high-to-low flow ratios \citep{olden2003,sadegh2015}. These methods extract only fragments of information and lack a unifying theoretical foundation.

\citet{vrugt2024b} recently introduced a framework for converting the FDC into a \emph{hydrograph functional}, a watershed signature embodied as a strictly proper scoring rule. Let $Q_{\boldsymbol{\uptheta}} = (q_{1},\ldots,q_{n})$ denote the cdf of the simulated streamflow series under parameter vector $\boldsymbol{\uptheta} = (\theta_1,\ldots,\theta_d)^{\top}$. Using the identity $F_{Y}(y) = 1 - R_{Y}(y)$, the continuous ranked exceedance probability score can be expressed as the divergence
\begin{linenomath*}
\begin{align}
d_\text{FDC}(Q,F) & = \int_{\,0}^{\, \infty}\bigl(R_{F}(z) - R_{Q}(z)\bigr)^{2}\,\text{d}z, \nonumber
\end{align}
\end{linenomath*}
which is nonnegative and equal to zero only when $R_{Q} = R_{F}$. Following \cite{thorarinsdottir2013} this divergence admits a decomposition into an energy-distance term measuring between-distribution variability and two terms capturing within-distribution variability
\begin{linenomath*}
\begin{align}
d_\text{FDC}(Q,F) & = \mathbb{E}_{Q,F}\bigl[\vert q - y \vert \bigr] - \ltfrac{1}{2}\bigl(\mathbb{E}_{Q}\bigl[\vert q - q^{\ast} \vert \bigr] + \mathbb{E}_{F}\bigl[\vert y - y^{\ast} \vert \bigr] \bigr),
\label{eqApp:d_FDC(Q,F)_def}
\end{align}
\end{linenomath*}
where $(q, q^{\ast})$ and $(y, y^{\ast})$ are independent copies of the simulated and observed discharge, respectively. 

For finite time series $q_{1},\ldots,q_{n}$ and $y_{1},\ldots,y_{n}$, we can approximate the FDC divergence in Equation 
\ref{eqApp:d_FDC(Q,F)_def} with the following Monte Carlo estimator \citep{vrugt2024b}
\begin{linenomath*}
\begin{align}
d_\text{FDC}(Q,F) & = \frac{1}{n^{2}} \sum_{i=1}^{n}\sum_{j=1}^{n} \vert q_{i} - y_{j} \vert - \frac{1}{2}\frac{1}{n^{2}} \sum_{i=1}^{n}\sum_{j=1}^{n} \bigl\{ \vert q_{i} - q_{j} \vert + \vert y_{i} - y_{j} \vert \bigr\}.
\label{eqApp:d_FDC(Q,F)_monte_carlo}
\end{align}
\end{linenomath*}
This divergence defines a proper loss function  $\mathcal{L}_\mathrm{fdc}(\boldsymbol{\uptheta}) = d_\mathrm{FDC}(Q,F)$, which captures the full distributional information encoded in the flow duration curve \citep{vrugt2024b}. Specifically, it is (i) mathematically well understood and strictly nonnegative; (ii) expressed in physical units of discharge (e.g., m\textsuperscript{3}\,s\textsuperscript{-1} or mm\,d\textsuperscript{-1}); (iii) sensitive to the \emph{entire} distribution of streamflow magnitudes, rather than only its mean or variance \citep{ferson2008,vrugt2024b}. Then, $\mathcal{L}_\mathrm{fdc}(\boldsymbol{\uptheta})$ reduces to the absolute error when comparing two point masses.

To compute the derivative of $\mathcal{L}_\mathrm{fdc}$ with respect to simulated streamflows $\mathbf{q}_{n} = (q_{1}, \ldots, q_{n})^{\top}$, we examine Equation \ref{eqApp:d_FDC(Q,F)_monte_carlo} in more detail. The term $\lvert y_{i} - y_{j} \rvert$ in the second double summation operator does not depend on $\mathbf{q}_{n}$ and therefore does not contribute to the derivative. Differentiating the remaining terms with respect to a particular $q_{t}$ yields 
\begin{linenomath*}
\begin{align}
\frac{\partial \mathcal{L}_\mathrm{FDC}}{\partial q_{t}} & = \frac{1}{n^{2}}\sum_{j=1}^{n} \sign(q_{t} - y_{j}) - \frac{1}{2}\frac{1}{n^{2}}\sum_{j=1}^{n}\bigl( \sign(q_{t} - q_{j}) - \sign(q_{j} - q_{t})\bigr).
\nonumber
\end{align}
\end{linenomath*}
We can write $\sign(q_{t} - q_{j}) = - \sign(q_{j} - q_{t})$ and the second summation term simplifies to
\begin{linenomath*}
\begin{align}
- \frac{1}{2n^{2}} \sum_{j=1}^{n}\bigl\{ \sign(q_{t} - q_{j}) - \bigl(-\sign(q_{t} - q_{j})\bigr)\bigr\} = - \frac{1}{n^{2}} \sum_{j=1}^{n} \sign(q_{t} - q_{j}). \nonumber
\end{align}
\end{linenomath*}
Thus, we yield
\begin{linenomath*}
\begin{align}
\frac{\partial \mathcal{L}_\mathrm{fdc}}{\partial q_{t}} & =
\frac{1}{n^{2}}\sum_{j=1}^{n}\sign(q_{t} - y_{j}) - \frac{1}{n^{2}} \sum_{j=1}^{n}\sign(q_{t} - q_{j}), \qquad t = 1,\ldots,n. \nonumber
\end{align}
\end{linenomath*}
Next, we organize the partial derivatives $\partial \mathcal{L}_\mathrm{fdc}/\partial q_{1},\ldots,\partial \mathcal{L}_\mathrm{fdc}/\partial q_{n}$ in a single vector
\begin{linenomath*}
\begin{align}
\boldsymbol{\updelta}_{n,\mathrm{fdc}}(\boldsymbol{\uptheta}) & = \frac{1}{n^{2}}
\colvec{1}{c}{ \; \sum\limits_{j=1}^{n} \sign(q_{1} - y_{j}) - \sum\limits_{j=1}^{n} \sign(q_{1} - q_{j}) \; \\[1mm]
\; \vdots \; \\[1mm]
\; \sum\limits_{j=1}^{n} \sign(q_{n} - y_{j}) - \sum\limits_{j=1}^{n} \sign(q_{n} - q_{j}) \; }.
\label{eqApp:delta_fdc}
\end{align}
\end{linenomath*}
The entries are easy to compute, piecewise smooth, and fully consistent with the definition of the FDC divergence in~\ref{eqApp:d_FDC(Q,F)_def}.

Algorithm~\ref{algApp:delta_fdc} provides a step-by-step recipe for computing the FDC loss sensitivity vector $\boldsymbol{\updelta}_{n,\mathrm{fdc}}(\boldsymbol{\uptheta})$
\begin{algorithm}[H]
\caption{Flow duration curve loss sensitivity vector $\boldsymbol{\updelta}_{n}(\boldsymbol{\uptheta})$}
\begin{algorithmic}
\State \textbf{Input:} $n$-vectors of observed $\mathbf{y}_{n} = (y_{1},\ldots,y_{n})^{\top}$ and simulated $\mathbf{q}_{n} = (q_{1},\ldots,q_{n})^{\top}$ values.
\State \textbf{Output:} FDC loss sensitivity vector $\boldsymbol{\updelta}_{n}(\boldsymbol{\uptheta}) \in \mathbb{R}^{n}$. \vspace{1mm}
\For{$i = 1$ \textbf{to} $n$}
\Statex \hspace{1em} Compute cross-differences with observed discharge, $a_{i} = n^{-2} \sum\nolimits_{j=1}^{n} \sign(q_{i} - y_{j})$.
\Statex \hspace{1em} Compute differences within simulated flows $b_{i} = n^{-2}\sum\nolimits_{j=1}^{n} \sign(q_{i} - q_{j})$.
\Statex \hspace{1em} Combine the FDC score, $\delta_{i} = a_{i} - b_{i}$.
\EndFor \vspace{1mm}
\State \textbf{Return:} $\boldsymbol{\updelta}_{n} = \bigl(\delta_{1},\ldots,\delta_{n})^{\top}$.
\end{algorithmic}
\label{algApp:delta_fdc}
\end{algorithm}
\noindent and the inset below presents \textsc{Matlab}-style pseudocode.
\begin{tcolorbox}[colback=black!5!white,colframe=black!40!white]
\begin{lstlisting}[style=matlab]
function delta = delta_fdc(y_n,q_n)
% DELTA_FDC Loss sensitivity vector flow duration curve
  n = numel(q_n);               % # elements
  delta = zeros(n,1);           % initialize FDC score
  [q_ns,p] = sort(q_n);         % sorted flows, permutation vector 
  y_ns = y_n(p);                % permute discharge data
  delta_s = sign(q_ns-y_ns)/n;  % sorted space
  delta(P) = delta_s;           % FDC loss sensitivity
end
\end{lstlisting}
\end{tcolorbox}

\noindent This leaves us with the gradient of the FDC loss $\mathcal{L}_\mathrm{fdc}$ with respect to the parameters
\begin{linenomath*}
\begin{align}
\mathbf{g}_{n,\mathrm{fdc}}(\boldsymbol{\uptheta}) & = \frac{\partial \mathbf{q}_{n}}{\partial \boldsymbol{\uptheta}^{\top}} \frac{\partial \mathcal{L}_\mathrm{fdc}(\boldsymbol{\uptheta})}{\partial \mathbf{q}_{n}} = \mathbf{J}^{\top}_{q}(\boldsymbol{\uptheta})\boldsymbol{\updelta}_{n,\mathrm{fdc}}(\boldsymbol{\uptheta}).
\label{eqApp:g_nfdc(uptheta)}
\end{align}
\end{linenomath*}
This completes our derivation of analytic expressions for the score vectors associated with the various loss functions considered in this work.

\newpage

\section[\appendixname~\thesection]{Derivatives of summary statistics of KGE loss}\label{sec:AppendixD}
\renewcommand{\theequation}{\thesection.\arabic{equation}}
\setcounter{equation}{0}
In this Appendix we derive analytic expressions for $\partial m_{q}/\partial q_{t}$, $\partial s_{q}/\partial q_{t}$ and $\partial r_{qy}/\partial q_{t}$ of the KGE loss function. 

Let $\mathbf{q}_{n} = (q_{1},\ldots,q_{n})^{\top}$ denote the $n \times 1$ vector of modeled streamflows. We compute the sample mean and sample variance of the simulated data
\begin{linenomath*}
\begin{align}
m_{q} & = \frac{1}{n} \sum_{t=1}^{n} q_{t}, \qquad \text{and} \qquad s^{2}_{q} = \frac{1}{n-1} \sum_{t=1}^{n}(q_{t} - m_{q})^{2}. \nonumber
\end{align}
\end{linenomath*}
Similarly, for the corresponding discharge observations $\mathbf{y}_{n} =(y_{1},\ldots,y_{n})^{\top}$ we write
\begin{linenomath*}
\begin{align}
m_{y} & = \frac{1}{n} \sum_{t=1}^{n} y_{t}, \qquad \text{and} \qquad s_{y}^{2} = \frac{1}{n-1} \sum_{t=1}^{n}(y_{t} - m_{y})^{2}. \nonumber
\end{align}
\end{linenomath*}
The sample covariance $C_{qy}$ and Pearson correlation coefficient $r_{qy}$ are equal to
\begin{linenomath*}
\begin{align}
C_{qy} = \frac{1}{n-1} \sum_{t=1}^{n} (q_t - m_{q})(y_t - m_{y}), \qquad \text{and} \qquad r_{qy} & = \frac{C_{qy}}{s_{q} s_{y}}.
\label{eqApp:C_{qy}}
\end{align}
\end{linenomath*}

\subsection{Analytic derivative of \texorpdfstring{$\partial m_{q}/\partial q_{t}$}{∂mq/∂qt}}
We first differentiate the sample mean of  simulated discharge $m_{q}$ with respect to simulated discharge
\begin{linenomath*}
\begin{align}
\frac{\partial m_{q}}{\partial q_{t}} & = \frac{\partial}{\partial q_{t}} \biggl(\frac{1}{n} \sum_{t=1}^{n} q_{t}\biggr) = \frac{1}{n}.
\label{eqApp:dm_{q}/dq_t}
\end{align}
\end{linenomath*}

\subsection{Analytic derivative of \texorpdfstring{$\partial s_{q}/\partial q_{t}$}{∂sq/∂qt}}
Next, we differentiate the sample variance of simulated discharge $s^{2}_{q}$ with respect to $q_{t}$
\begin{linenomath*}
\begin{align}
\frac{\partial s^{2}_{q}}{\partial q_{t}} & = \frac{\partial}{\partial q_{t}}\biggl(\frac{1}{n-1} \sum_{t=1}^{n}(q_{t} - m_{q})^{2} \biggr) \nonumber \\
& = \frac{1}{n-1} \sum_{i=1}^{n} 2(q_{i} - m_{q})
\biggl(\frac{\partial q_{i}}{\partial q_{t}} - \frac{\partial m_{q}}{\partial q_{t}} \biggr) \nonumber \\
& = \frac{2}{n-1} \biggl[ (q_{t} - m_{q})\biggl(1 - \frac{1}{n}\biggr) - \sum_{i\neq t} (q_{i} - m_{q})\biggl(\frac{1}{n}\biggr) \biggr]. \nonumber
\intertext{Since $\sum_{i=1}^{n} (q_{i} - m_{q}) = 0$, the second term cancels and we yield}
\frac{\partial s^{2}_{q}}{\partial q_{t}} & = \frac{2}{n-1} (q_{t} - m_{q}). \nonumber
\intertext{Using the chain rule we arrive at an expression for $\partial s_{q}/\partial q_{t}$ as follows}
\frac{\partial s_{q}}{\partial q_{t}} & = \frac{1}{2 s_{q}} \frac{\partial s^{2}_{q}}{\partial q_{t}} = \frac{q_{t} - m_{q}}{(n-1)s_{q}}.
\label{eqApp:ds_{q}/dq_t}
\end{align}
\end{linenomath*}

\subsection{Analytic derivative of \texorpdfstring{$\partial r_{qy}/\partial q_{t}$}{drqy/dqt}}
As last step, we must find an expression for the derivative of the sample correlation coefficient $r_{qy}$ with respect to simulated streamflows. We revisit the sample covariance $C_{qy}$ of Equation \ref{eqApp:C_{qy}} and differentiate this expression with respect to $q_{t}$  
\begin{linenomath*}
\begin{align}
\frac{\partial C_{qy}}{\partial q_{t}} & = \frac{\partial}{\partial q_{t}}\biggl(\frac{1}{n-1} \sum_{t=1}^{n} (q_t - m_{q})(y_t - m_{y})\biggr) \nonumber \\
& = \frac{1}{n-1}
\biggl[ (y_t - m_{y}) - \frac{1}{n}\sum_{i=1}^{n}(y_{i} - m_{y}) \biggr]. \nonumber
\intertext{As $\sum_{i=1}^{n}(y_{i} - m_{y}) = 0$ the summation term will cancel}
\frac{\partial C_{qy}}{\partial q_{t}} & = \frac{y_{t} - m_{y}}{n-1}.
\label{eqApp:dC_{q,y}/dq_t}
\intertext{Next, we enter Equation \ref{eqApp:C_{qy}} and apply the quotient rule}
\frac{\partial r_{qy}}{\partial q_{t}} & = \frac{1}{s_{q} s_{y}} \frac{\partial C_{qy}}{\partial q_{t}} - \frac{C_{qy}}{s^{2}_{q} s_{y}} \frac{\partial s_{q}}{\partial q_{t}}. \nonumber
\intertext{If we substitute Equations \ref{eqApp:ds_{q}/dq_t} and \ref{eqApp:dC_{q,y}/dq_t} into the above expression we yield}
\frac{\partial r_{qy}}{\partial q_{t}} & = \frac{1}{s_{q} s_{y}} \frac{y_{t} - m_{y}}{n-1} - \frac{C_{qy}}{s^{2}_{q} s_{y}} \frac{q_{t} - m_{q}}{(n-1)s_{q}}, \nonumber
\intertext{and this expression simplifies to}
\frac{\partial r_{qy}}{\partial q_{t}} & = \frac{y_{t} - m_{y}}{(n-1)s_{q} s_{y}} - r_{qy}\frac{(q_{t} - m_{q})}{(n-1)s^{2}_{q}}. 
\label{eqApp:drdqt}
\end{align}
\end{linenomath*}
This concludes our derivation of the identities used for $\partial m_{q}/\partial q_{t}$, $\partial s_{q}/\partial q_{t}$ and $\partial r/\partial q_{t}$ in the main text.

\clearpage

\bibliographystyle{abbrvnat}
\bibliography{refs_JAV}

@article{shen2023,
  title = {Differentiable modelling to unify machine learning and physical models for geosciences},
  author = {Shen, Chaopeng and Appling, Alison P and Gentine, Pierre and Bandai, Toshiyuki and Gupta, Hoshin and Tartakovsky, Alexandre and Baity-Jesi, Marco and Fenicia, Fabrizio and Kifer, Daniel and Li, Li and others},
  journal = {Nature Reviews Earth \& Environment},
  volume = {4},
  number = {8},
  pages = {552--567},
  year = {2023},
  publisher = {Nature Publishing Group UK London}
}

@article{feng2023,
  author = {Feng, Dapeng and Beck, Hylke and Lawson, Kathryn and Shen, Chaopeng},
  title = {The suitability of differentiable, physics-informed machine learning hydrologic models for ungauged regions and climate change impact assessment},
  journal = {Hydrology and Earth System Sciences},
  volume = {27},
  pages = {2357--2373},
  year = {2023},
  doi = {10.5194/hess-27-2357-2023}
}

@article{feng2022,
  author = {Feng, Dapeng and Liu, Jiangtao and Lawson, Kathryn and Shen, Chaopeng},
  title = {Differentiable, Learnable, Regionalized Process-Based Models With Multiphysical Outputs Can Approach State-Of-The-Art Hydrologic Prediction Accuracy},
  journal = {Water Resources Research},
  year = {2022},
  volume = {58},
  number = {10},
  pages = {e2022WR032404},
  doi = {10.1029/2022WR032404}
}

@article{sorooshian1983b,
  author  = {Sorooshian, Soroosh and Gupta, Vijay K.},
  title   = {Automatic Calibration of Conceptual Rainfall--Runoff Models: The Question of Parameter Observability and Uniqueness},
  journal = {Water Resources Research},
  volume  = {19},
  number  = {1},
  pages   = {260--268},
  year    = {1983},
  doi     = {10.1029/WR019i001p00260}
}

@article{sorooshian1993,
  author  = {Sorooshian, S. and Duan, Q. and Gupta, V. K.},
  title   = {Calibration of Rainfall--Runoff Models: Application of Global Optimization to the Sacramento Soil Moisture Accounting Model},
  journal = {Water Resources Research},
  volume  = {29},
  number  = {4},
  pages   = {1185--1194},
  year    = {1993},
  doi     = {10.1029/92WR02617}
}

@inproceedings{sorooshian1980b,
  author    = {Sorooshian, S.},
  title     = {Comparison of Two Direct Search Algorithms Used in Calibration of Rainfall--Runoff Models},
  booktitle = {IFAC Proceedings Volumes},
  volume    = {13(3)},
  pages     = {477--485},
  year      = {1980},
  publisher = {Elsevier},
  doi       = {10.1016/S1474-6670(17)65355-9}
}

@article{guptavk1985,
  author = {Gupta, Vijai Kumar and Sorooshian, Soroosh},
  title = {Uniqueness and observability of conceptual rainfall-runoff model parameters: The percolation process examined},
  journal = {Water Resources Research},
  volume = {19},
  number = {1},
  pages = {269-276},
  doi = {https://doi.org/10.1029/WR019i001p00269},
  url = {https://agupubs.onlinelibrary.wiley.com/doi/abs/10.1029/WR019i001p00269},
  eprint = {https://agupubs.onlinelibrary.wiley.com/doi/pdf/10.1029/WR019i001p00269},
  year = {1983}
}

@article{duan1994,
  author  = {Duan, Qingyun and Sorooshian, Soroosh and Gupta, Hoshin V.},
  title   = {Optimal use of the SCE-UA global optimization method for calibrating watershed models},
  journal = {Journal of Hydrology},
  year    = {1994},
  volume  = {158},
  number  = {3--4},
  pages   = {265--284},
  doi     = {10.1016/0022-1694(94)90057-4}
}

@article{hsu1995,
  author = {Hsu, Kuang-Lin and Gupta, Hoshin V. and Sorooshian, Soroosh},
  title = {Artificial neural network modeling of the rainfall--runoff process},
  journal = {Water Resources Research},
  year = {1995},
  volume = {31},
  number = {10},
  pages = {2517--2530},
  doi = {10.1029/95WR01955}
}

@article{zhou2021,
  title = {A review of the Xinanjiang model developments and applications},
  author = {Zhou, Jing and Bao, Hongjun and Xu, Chong-Yu},
  journal = {Hydrological Sciences Journal},
  volume = {66},
  number = {10},
  pages = {1629--1646},
  year = {2021}
}

@book{beven2012,
  title = {Rainfall-Runoff Modelling: The Primer},
  author = {Beven, Keith},
  year = {2012},
  publisher = {Wiley-Blackwell}
}

@article{moore2007,
  title = {The PDM rainfall-runoff model},
  author = {Moore, R. J.},
  journal = {Hydrology and Earth System Sciences},
  volume = {11},
  number = {1},
  pages = {483--499},
  year = {2007}
}

@article{jakeman1993,
  title = {Predicting daily flows in ungauged catchments: A comparison of methods},
  author = {Jakeman, A. J. and Hornberger, G. M.},
  journal = {Water Resources Research},
  volume = {29},
  number = {12},
  pages = {405--419},
  year = {1993}
}

@article{zhang2014,
  author = {Hao Zhang and Adrian Sandu},
  title = {{FATODE}: A Library for Forward, Adjoint, and Tangent Linear Integration of {ODE}s},
  journal = {SIAM Journal on Scientific Computing},
  volume = {36},
  number = {5},
  pages = {C504--C523},
  year = {2014},
  month = {jan},
  doi = {10.1137/130912335},
  url = {http://epubs.siam.org/doi/10.1137/130912335}
}

@inproceedings{kingma2015,
  title = {Adam: A Method for Stochastic Optimization},
  author = {Kingma, Diederik P. and Ba, Jimmy},
  booktitle = {Proceedings of the 3rd International Conference on Learning Representations (ICLR)},
  year = {2015},
  url = {https://arxiv.org/abs/1412.6980},
}

@book{griewank2008,
  title = {Evaluating Derivatives: Principles and Techniques of Algorithmic Differentiation},
  author = {Griewank, Andreas and Walther, Andrea},
  year = {2008},
  publisher = {SIAM}
}

@article{baydin2018,
  title={Automatic differentiation in machine learning: a survey},
  author={Baydin, A. G. and Pearlmutter, B. A. and Radul, A. A. and Siskind, J. M.},
  journal = {Journal of Machine Learning Research},
  year = {2018},
  volume = {18},
  number = {153},
  pages = {1--43}
}

@book{naumann2011,
  title = {The Art of Differentiating Computer Programs: An Introduction to Algorithmic Differentiation},
  author = {Naumann, U.},
  year = {2011},
  publisher = {SIAM}
}

@misc{jax2018,
  title = {JAX: Autograd and XLA},
  author = {Google},
  year = {2018},
  howpublished = {\url{https://github.com/google/jax}}
}

@article{kramer1981,
  author  = {Kramer, M. A. and Calo, J. M. and Rabitz, H.},
  title   = {An improved computational method for sensitivity analysis: {G}reen's function method with {AIM}},
  journal = {Applied Mathematical Modelling},
  year    = {1981},
  volume  = {5},
  pages   = {432--441},
  doi     = {10.1016/0307-904X(81)90069-4}
}

@article{turanyi1990,
  author  = {Tur\'anyi, T.},
  title   = {Sensitivity analysis of complex kinetic systems: Tools and applications},
  journal = {Journal of Mathematical Chemistry},
  year    = {1990},
  volume  = {5},
  number  = {3},
  pages   = {203--248},
  doi     = {10.1007/BF01166355}
}

@article{fisher2015,
  author  = {Fisher, J. and Henzinger, T. A.},
  title   = {Executable cell biology},
  journal = {Nature Biotechnology},
  year    = {2007},
  volume  = {25},
  number  = {11},
  pages   = {1239--1249},
  doi     = {10.1038/nbt1356}
}

@article{gutenkunst2007,
  author  = {Gutenkunst, R. N. and Waterfall, J. J. and Casey, F. P. and Brown, K. S. and Myers, C. R. and Sethna, J. P.},
  title   = {Universally sloppy parameter sensitivities in systems biology models},
  journal = {PLoS Computational Biology},
  year    = {2007},
  volume  = {3},
  number  = {10},
  pages   = {e189},
  doi     = {10.1371/journal.pcbi.0030189}
}

@article{rabitz1989,
  author  = {Rabitz, H. and Kramer, M. and Dacol, D.},
  title   = {Sensitivity analysis in chemical kinetics},
  journal = {Annual Review of Physical Chemistry},
  year    = {1989},
  volume  = {40},
  pages   = {419--461},
  doi     = {10.1146/annurev.pc.40.100189.002223}
}

@book{walter1997,
  author    = {Walter, E. and Pronzato, L.},
  title     = {Identification of Parametric Models},
  publisher = {Springer},
  year      = {1997},
  doi       = {10.1007/978-1-4612-2020-6}
}

@article{cacuci1981,
  author  = {Cacuci, D. G.},
  title   = {Sensitivity theory for nonlinear systems},
  journal = {Journal of Mathematical Physics},
  year    = {1981},
  volume  = {22},
  number  = {12},
  pages   = {2794--2802},
  doi     = {10.1063/1.524385}
}

@article{hindmarsh2005,
  author  = {Hindmarsh, A. C. and Brown, P. N. and Grant, K. E. and Lee, S. L. and Serban, R. and Shumaker, D. E. and Woodward, C. S.},
  title   = {SUNDIALS: Suite of nonlinear and differential/algebraic equation solvers},
  journal = {ACM Transactions on Mathematical Software},
  year    = {2005},
  volume  = {31},
  number  = {3},
  pages   = {363--396},
  doi     = {10.1145/1089014.1089020}
}

@article{depauw2006,
  author = {Dirk J.W. De Pauw and Peter A. Vanrolleghem},
  title = {Practical aspects of sensitivity function approximation for dynamic models},
  journal = {Mathematical and Computer Modelling of Dynamical Systems},
  volume = {12},
  number = {5},
  pages = {395--414},
  year = {2006},
  publisher = {Taylor \& Francis},
  doi = {10.1080/13873950600723301},
  URL = {https://doi.org/10.1080/13873950600723301},
  eprint = {https://doi.org/10.1080/13873950600723301}
}

@article{perumal2011,
  author  = {Perumal, Adamna and Gunawan, Raoul},
  title   = {Understanding dynamics using sensitivity analysis: caveat and solution},
  journal = {BMC Systems Biology},
  volume  = {5},
  number  = {41},
  pages   = {1--19},
  year    = {2011},
  doi     = {10.1186/1752-0509-5-41}
}

@article{ma2018,
  author  = {Ma, Yingbo and Dixit, Vaibhav and Innes, Michael and Guo, Xingjian and Rackauckas, Christopher},
  title   = {A comparison of automatic differentiation and continuous sensitivity analysis for derivatives of differential equation solutions},
  journal = {arXiv preprint arXiv:1812.01892},
  year    = {2018},
  url     = {https://arxiv.org/abs/1812.01892}
}

@article{borggaard2000,
  author  = {Borggaard, Jeff and Pattison, Ken},
  title   = {On the efficient solution to the continuous sensitivity equation for large ODE systems},
  journal = {SIAM Journal on Scientific Computing},
  volume  = {22},
  number  = {4},
  pages   = {1204--1223},
  year    = {2000},
  doi     = {10.1137/S1064827599352136}
}

@article{bradbury2018,
  title = {{JAX}: composable transformations of {P}ython+{N}um{P}y programs},
  author = {Bradbury, James and Frostig, Roy and Hawkins, Peter and Johnson, Matthew and Leary, Chris and Maclaurin, Dougal and Necula, Radu and Paszke, Adam and VanderPlas, Jake and Wanderman-Milne, Scott and Zhang, Qiao},
  journal = {Version 0.1},
  year = {2018},
  note = {URL: \url{https://github.com/google/jax}}
}

@inproceedings{abadi2016,
  title = {{TensorFlow}: A system for large-scale machine learning},
  author = {Abadi, Mart{\'\i}n and Barham, Paul and Chen, Jianmin and Chen, Zhifeng and Davis, Andy and Dean, Jeffrey and others},
  booktitle = {12th USENIX Symposium on Operating Systems Design and Implementation (OSDI)},
  pages = {265--283},
  year = {2016}
}

@inproceedings{paszke2017,
  author = {Paszke, Adam and Gross, Sam and Chintala, Soumith and Chanan, Gregory and Yang, Edward and DeVito, Zachary and Lin, Zeming and Desmaison, Alban and Antiga, Luca and Lerer, Adam},
  booktitle = {NIPS 2017 Workshop on Autodiff},
  location = {Long Beach, California, USA},
  title = {Automatic Differentiation in PyTorch},
  year = {2017}
}

@inproceedings{paszke2019,
  title = {PyTorch: An imperative style, high-performance deep learning library},
  author = {Paszke, A. and others},
  booktitle = {Advances in Neural Information Processing Systems},
  year = {2019}
}

@article{condon2021,
  author  = {Condon, Laura E. and Atchley, Adam L. and Flores, Ariana and Maxwell, Reed M.},
  title   = {Data for continental-{US} hydrologic modeling: The {CONUS}404 data products},
  journal = {Earth System Science Data},
  volume  = {13},
  pages   = {1547--1575},
  year    = {2021},
  doi     = {10.5194/essd-13-1547-2021}
}

@article{maxwell2013,
  author = {Maxwell, Reed M.},
  title = {A terrain-following grid transform and preconditioner for parallel, large-scale, integrated hydrologic modeling},
  journal = {Advances in Water Resources},
  volume = {53},
  pages = {109--117},
  year = {2013},
  doi = {10.1016/j.advwatres.2012.10.001}
}

@article{kuffour2020,
  author = {Kuffour, Benjamin N. O. and Engdahl, Nicholas B. and Woodward, Carol S. and Condon, Laura E. and Kollet, Stefan and Maxwell, Reed M.},
  title = {Simulating coupled surface-subsurface flows with ParFlow v3.5.0: capabilities, applications, and ongoing development of an open-source, massively parallel, integrated hydrologic model},
  journal = {Geoscientific Model Development},
  volume = {13},
  number = {3},
  pages = {1373--1397},
  year = {2020},
  doi = {10.5194/gmd-13-1373-2020}
}

@article{maxwell2015,
  author = {Maxwell, Reed M. and Condon, Laura E. and Kollet, Stefan J.},
  title = {A high-resolution simulation of groundwater and surface water over most of the continental US with the integrated hydrologic model ParFlow v3},
  journal = {Geoscientific Model Development},
  volume = {8},
  number = {3},
  pages = {923--937},
  year = {2015},
  doi = {10.5194/gmd-8-923-2015}
}

@article{feng2024,
  author  = {Feng, Dapeng and Beck, Hylke and de Bruijn, Jens and Sahu, Reetik Kumar and Satoh, Yusuke and Wada, Yoshihide and Liu, Jiangtao and Pan, Ming and Lawson, Kathryn and Shen, Chaopeng},
  title   = {Deep dive into hydrologic simulations at global scale: harnessing the power of deep learning and physics-informed differentiable models ($\delta$HBV-globe1.0-hydroDL)},
  journal = {Geoscientific Model Development},
  volume  = {17},
  number  = {18},
  pages   = {7181--7198},
  year    = {2024},
  doi     = {10.5194/gmd-17-7181-2024}
}

@article{kling2012,
  author  = {Kling, Harald and Fuchs, Markus and Paulin, Maria},
  title   = {Runoff conditions in the upper {Danube} basin under an ensemble of climate change scenarios},
  journal = {Journal of Hydrology},
  volume  = {424--425},
  pages   = {264--277},
  year    = {2012},
  doi     = {10.1016/j.jhydrol.2012.01.011}
}

@Article{vrugt2025b,
  author = {Vrugt, Jasper A. and Diks, Cees G. H.},
  title = {The Learning Rate Is Not a Constant: Sandwich-Adjusted {M}arkov chain {M}onte {C}arlo Simulation},
  journal = {Entropy},
  volume = {27},
  year = {2025},
  number = {10},
  article-number = {999},
  url = {https://www.mdpi.com/1099-4300/27/10/999},
  PubMedID = {41148957},
  ISSN = {1099-4300},
  doi = {10.3390/e27100999}
}

@article{zhao1963,
  author  = {Zhao, Renjun and Zhuang, Y.},
  title   = {Regional Patterns of Rainfall-Runoff Relationship},
  journal = {Journal of Hohai University (Natural Sciences)},
  year    = {1963},
  volume  = {S2},
  pages   = {53-68},
  note    = {In Chinese},
}

@article{jayawardena2000,
  author    = {Jayawardena, A. and Zhou, M.},
  title     = {A modified spatial soil moisture storage capacity distribution curve for the {X}inanjiang model},
  journal   = {Journal of Hydrology},
  volume    = {227},
  number    = {1-4},
  pages     = {93-113},
  year      = {2000},
  doi       = {10.1016/S0022-1694(99)00173-0},
  url       = {https://doi.org/10.1016/S0022-1694(99)00173-0}
}

@inproceedings{zhao1980,
  author    = {Zhao, Renjun and Zhuang, Y. and Fang, L. and Liu, X. and Zhang, Q.},
  title     = {The {X}inanjiang Model},
  booktitle = {Proceedings of the Oxford Symposium on Hydrological Forecasting},
  year      = {1980},
  address   = {Oxford, England},
  note      = {UNESCO-WMO Symposium, November 1980},
}

@article{zhao1992,
  author    = {Zhao, Ren Jun},
  title     = {The {X}inanjiang model applied in {C}hina},
  journal   = {Journal of Hydrology},
  year      = {1992},
  volume    = {135},
  number    = {1-4},
  pages     = {371-381},
  doi       = {10.1016/0022-1694(92)90096-E},
  publisher = {Elsevier},
  url       = {https://doi.org/10.1016/0022-1694(92)90096-E}
}

@techreport{Frazier2023a, 
  title = {Reliable {B}ayesian Inference in Misspecified Models},
  url = {http://arxiv.org/abs/2302.06031}, 
  DOI = {10.48550/arXiv.2302.06031}, 
  abstractNote = {We provide a general solution to a fundamental open problem in Bayesian inference, namely poor uncertainty quantification, from a frequency standpoint, of Bayesian methods in misspecified models. While existing solutions are based on explicit Gaussian approximations of the posterior, or computationally onerous post-processing procedures, we demonstrate that correct uncertainty quantification can be achieved by replacing the usual posterior with an intuitive approximate posterior. Critically, our solution is applicable to likelihood-based, and generalized, posteriors as well as cases where the likelihood is intractable and must be estimated. We formally demonstrate the reliable uncertainty quantification of our proposed approach, and show that valid uncertainty quantification is not an asymptotic result but occurs even in small samples. We illustrate this approach through a range of examples, including linear, and generalized, mixed effects models.},  
  number = {arXiv:2302.06031}, 
  publisher = {arXiv}, 
  author = {D.T. Frazier and R. Kohn and C. Drovandi and D. Gunawan}, 
  year = {2023}, 
  institution = {Monash University}
}

@book{hampel1986,
  title = {Robust Statistics: The Approach Based on Influence Functions},
  author = {Hampel, Frank R.},
  isbn = {9780471632382},
  lccn = {85009428},
  series = {Probability and Statistics Series},
  url = {https://books.google.com/books?id=KXWMNAAACAAJ},
  year = {1986},
  publisher = {Wiley}
}

@article{vrugt2025a,
  author = {Vrugt, Jasper A. and Cees G. H. Diks and Ramon de Punder and Peter Gr\"{u}nwald},
  journal = {ARC Geophysical Research},
  publisher = {Academic Research Community (ARC) Alliance for the geosciences},
  title = {A Sandwich with Water: Bayesian \& Frequentist Uncertainty Quantification under Model Misspecification},
  year = {2025},
  volume = {16},
  url = {https://janeway.uncpress.org/ARC-GR/article/id/1824/},
  number = {1},
  doi  = {10.5149/ARC-GR.1824}
}

@article{vrugt2024b,
  author = {Vrugt, Jasper A.},
  title = {Distribution-Based Model Evaluation and Diagnostics: Elicitability, Propriety, and Scoring Rules for Hydrograph Functionals},
  journal = {Water Resources Research},
  volume = {60},
  number = {6},
  pages = {e2023WR036710},
  doi = {10.1029/2023WR036710},
  url = {https://agupubs.onlinelibrary.wiley.com/doi/abs/10.1029/2023WR036710},
  eprint = {https://agupubs.onlinelibrary.wiley.com/doi/pdf/10.1029/2023WR036710},
  note = {e2023WR036710 2023WR036710},
  year = {2024}
}

@misc{derrico2024,
  author = {D'Errico, John},
  title = {Adaptive Robust Numerical Differentiation},
  note = {{MATLAB} Central File Exchange. Retrieved March 29, 2024},
  year = {2024},
  url = {https://www.mathworks.com/matlabcentral/fileexchange/13490-adaptive-robust-numerical-differentiation}
}

@phdthesis{hampel1968,
  title = {Contribution to the theory of robust estimation},
  school = {University of California, Berkeley},
  author = {Hampel, F. R.},
  year = {1968},
  publisher = {University of California},
  location = {Berkeley, California, USA},
  url = {},
  doi = {}
}

@article{hampel1971,
  author = {Frank R. Hampel},
  title = {{A General Qualitative Definition of Robustness}},
  volume = {42},
  journal = {The Annals of Mathematical Statistics},
  number = {6},
  publisher = {Institute of Mathematical Statistics},
  pages = {1887-1896},
  year = {1971},
  doi = {10.1214/aoms/1177693054},
  url = {https://doi.org/10.1214/aoms/1177693054}
}

@article{hampel1974,
  author = {Hampel, Frank R.},
  doi = {10.1080/01621459.1974.10482962},
  journal = {Journal of the American Statistical Association},
  number = {346},
  pages = {383-393},
  title = {The Influence Curve and its Role in Robust Estimation},
  url = {https://app.dimensions.ai/details/publication/pub.1058301356},
  volume = {69},
  year = {1974}
}

@inproceedings{tukey1960a,
  title = {A survey of sampling from contaminated distributions},
  booktitle = {In Contributions to Probability and Statistics: Essays in Honor of Harold Hotelling},
  editors = {Olkin, I. and Ghurye, S. G. and Hoeffding, W. and Madow, W. G. and Mann, H. B.},
  author = {John W. Tukey},
  page = {448-485},
  publisher = {Stanford University Press},
  year = {1960}
}

@article{huber1964,
  author = {Peter J. Huber},
  title = {{Robust Estimation of a Location Parameter}},
  volume = {35},
  journal = {The Annals of Mathematical Statistics},
  number = {1},
  publisher = {Institute of Mathematical Statistics},
  pages = {73-101},
  year = {1964},
  doi = {10.1214/aoms/1177703732},
  url = {https://doi.org/10.1214/aoms/1177703732}
}

@article{liang1986,
  author = {Liang, Kung-Yee and Zeger, Scott L.},
  title = {Longitudinal data analysis using generalized linear models},
  journal = {Biometrika},
  volume = {73},
  number = {1},
  pages = {13-22},
  year = {1986},
  month = {04},
  issn = {0006-3444},
  doi = {10.1093/biomet/73.1.13},
  url = {https://doi.org/10.1093/biomet/73.1.13},
  eprint = {https://academic.oup.com/biomet/article-pdf/73/1/13/679793/73-1-13.pdf}
}

@article{huber1973,
  author = {Peter J. Huber},
  title = {{Robust Regression: Asymptotics, Conjectures and Monte Carlo}},
  volume = {1},
  journal = {The Annals of Statistics},
  number = {5},
  publisher = {Institute of Mathematical Statistics},
  pages = {799-821},
  year = {1973},
  doi = {10.1214/aos/1176342503},
  url = {https://doi.org/10.1214/aos/1176342503}
}

@book{godambe1991,
  title = {Estimating Functions},
  author = {Godambe, V.P.},
  isbn = {9780198522287},
  lccn = {lc91007676},
  series = {Oxford science publications},
  url = {https://books.google.nl/books?id=V7P2wAEACAAJ},
  year = {1991},
  publisher = {Clarendon Press}
}

@article{levenberg1944,
  author = {Levenberg, K.},
  title = {A method for the solution of certain non-Linear problems in least squares},
  journal = {Quarterly of Applied Mathematics},
  volume = {2},
  number = {2},
  pages = {164-168},
  publisher = {American Mathematical Society},
  year = {1944},
}

@incollection{huber1967,
  address = {{Berkeley, CA}},
  title = {The Behavior of Maximum Likelihood Estimates under Nonstandard Conditions},
  volume = {1: Statistics},
  booktitle = {Proceedings of the {{Fifth Berkeley Symposium}} on {{Mathematical Statistics}} and {{Probability}}},
  publisher = {{University of California Press}},
  author = {Huber, Peter J.},
  year = {1967},
  pages = {221-233}
}

@book{huber1981, 
  place = {Cambridge}, 
  series = {Wiley Series in Probability and Statistics}, 
  title = {Robust Statistics}, 
  doi = {10.1002/0471725250}, publisher = {John Wiley \& Sons}, 
  author = {Huber, Peter J.}, 
  year = {1981}
}

@article{mameli2015,
  title = {Higher-order asymptotics for scoring rules},
  journal = {Journal of Statistical Planning and Inference},
  volume = {165},
  pages = {13-26},
  year = {2015},
  issn = {0378-3758},
  doi = {10.1016/j.jspi.2015.03.005},
  url = {https://www.sciencedirect.com/science/article/pii/S0378375815000567},
  author = {Valentina Mameli and Laura Ventura}
}

@article{giummole2018,
  doi = {10.1007/s11749-018-0597-z},
  year = {2018},
  month = {jul},
  publisher = {Springer Science and Business Media {LLC}},
  volume = {28},
  number = {3},
  pages = {728-755},
  author = {F. Giummol{\`{e}} and V. Mameli and E. Ruli and L. Ventura},
  title = {Objective {B}ayesian inference with proper scoring rules},
  journal = {{TEST}}
}

@inproceedings{dawid2005,
  title = {The Geometry of Decision Theory},
  author = {Philip Dawid and Steffen L. Lauritzen},
  booktitle = {Proceedings of the Second International Symposium on Information Geometry and its Applications},
  editors = {Brody, D. C. \&  Hughston, L. P.},
  year = {2006},
  publisher = {},
  address = {Tokyo, Japan},
  pages = {22-28}
}

@article{godambe1960,
  author = {Godambe, V. P.},
  title = {{An Optimum Property of Regular Maximum Likelihood Estimation}},
  volume = {31},
  journal = {The Annals of Mathematical Statistics},
  number = {4},
  publisher = {Institute of Mathematical Statistics},
  pages = {1208-1211},
  year = {1960},
  doi = {10.1214/aoms/1177705693},
  url = {https://doi.org/10.1214/aoms/1177705693}
}

@article{ferson2008,
  title = {Model validation and predictive capability for the thermal challenge problem},
  journal = {Computer Methods in Applied Mechanics and Engineering},
  volume = {197},
  number = {29},
  pages = {2408-2430},
  year = {2008},
  note = {Validation Challenge Workshop},
  issn = {0045-7825},
  doi = {10.1016/j.cma.2007.07.030},
  url = {https://www.sciencedirect.com/science/article/pii/S0045782507005105},
  author = {Ferson, Scott and Oberkampf, William L. and  Ginzburg, Lev}
}

@article{thorarinsdottir2013,
  author = {Thorarinsdottir, Thordis L. and Gneiting, Tilmann and Gissibl, Nadine},
  title = {Using Proper Divergence Functions to Evaluate Climate Models},
  journal = {SIAM/ASA Journal on Uncertainty Quantification},
  volume = {1},
  number = {1},
  pages = {522-534},
  year = {2013},
  doi = {10.1137/130907550},
  url = {https://doi.org/10.1137/130907550},
  eprint = {https://doi.org/10.1137/130907550}
}

@TechReport{burnash1973,
author = {Burnash, Robert J. C. and Ferral, R. Larry and McGuire, Richard A.},
editor = {},
institution = {Joint Federal-State River Forecast Center: US Department of Commerce, National Weather Service and CA Department of Water Resources},
address = {Sacramento, CA},
title = {A generalized streamflow simulation system: Conceptual modeling for digital computers},
series = {},
year = {1973},
edition = {},
issn = {},
doi = {},
url = {},
language = {ENGLISH}
}

@article{vrugt2022b,
  author = {Vrugt, Jasper A. and Yumi de Oliveira, D. and Schoups, Gerrit and Diks, Cees G. H.},
  title = {On the use of distribution-adaptive likelihood functions: {G}eneralized and universal likelihood functions, scoring rules and multi-criteria ranking},
  journal = {Journal of Hydrology},
  volume = {615},
  pages = {128542},
  year = {2022},
  issn = {0022-1694},
  doi = {10.1016/j.jhydrol.2022.128542},
  url = {https://www.sciencedirect.com/science/article/pii/S002216942201112X}
}

@book{boyle2001,
title = {Multicriteria calibration of hydrological models (PhD thesis)},
author = {Boyle, D. P.},
publisher = {Department of Hydrology and Water Resources, University of Arizona, Tucson, AZ},
series = {},
volume = {},
year = {2001},
isbn = {},
}

@article{clark2008,
author = {Clark, Martyn P. and Slater, Andrew G. and Rupp, David E. and Woods, Ross A. and Vrugt, Jasper A. and Gupta, Hoshin V. and Wagener, Thorsten and Hay, Lauren E.},
title = {Framework for Understanding Structural Errors ({FUSE}): A modular framework to diagnose differences between hydrological models},
journal = {Water Resources Research},
volume = {44},
number = {12},
pages = {},
doi = {10.1029/2007WR006735},
url = {https://agupubs.onlinelibrary.wiley.com/doi/abs/10.1029/2007WR006735},
eprint = {https://agupubs.onlinelibrary.wiley.com/doi/pdf/10.1029/2007WR006735},
year = {2008}
}

@article{duan1992,
author = {Duan, Qingyun and Sorooshian, Soroosh and Gupta, Vijai},
title = {Effective and efficient global optimization for conceptual rainfall-runoff models},
journal = {Water Resources Research},
volume = {28},
number = {4},
pages = {1015-1031},
doi = {10.1029/91WR02985},
url = {https://agupubs.onlinelibrary.wiley.com/doi/abs/10.1029/91WR02985},
eprint = {https://agupubs.onlinelibrary.wiley.com/doi/pdf/10.1029/91WR02985},
year = {1992}
}

@article{gupta1985,
title = {The relationship between data and the precision of parameter estimates of hydrologic models},
journal = {Journal of Hydrology},
volume = {81},
number = {1},
pages = {57-77},
year = {1985},
issn = {0022-1694},
doi = {10.1016/0022-1694(85)90167-2},
url = {http://www.sciencedirect.com/science/article/pii/0022169485901672},
author = {Vijai K. Gupta and Soroosh Sorooshian}
}

@article{gupta1998,
author = {Gupta, Hoshin Vijai and Sorooshian, Soroosh and Yapo, Patrice Ogou},
title = {Toward improved calibration of hydrologic models: Multiple and noncommensurable measures of information},
journal = {Water Resources Research},
volume = {34},
number = {4},
pages = {751-763},
doi = {10.1029/97WR03495},
url = {https://agupubs.onlinelibrary.wiley.com/doi/abs/10.1029/97WR03495},
eprint = {https://agupubs.onlinelibrary.wiley.com/doi/pdf/10.1029/97WR03495},
year = {1998}
}

@article{gupta2009,
title = {Decomposition of the mean squared error and {NSE} performance criteria: Implications for improving hydrological modelling},
journal = {Journal of Hydrology},
volume = {377},
number = {1},
pages = {80-91},
year = {2009},
issn = {0022-1694},
doi = {10.1016/j.jhydrol.2009.08.003},
url = {http://www.sciencedirect.com/science/article/pii/S0022169409004843},
author = {Hoshin V. Gupta and Harald Kling and Koray K. Yilmaz and Guillermo F. Martinez}
}

@article{kirchner2006,
author = {Kirchner, James W.},
title = {Getting the right answers for the right reasons: Linking measurements, analyses, and models to advance the science of hydrology},
journal = {Water Resources Research},
volume = {42},
number = {3},
pages = {},
doi = {10.1029/2005WR004362},
url = {https://agupubs.onlinelibrary.wiley.com/doi/abs/10.1029/2005WR004362},
eprint = {https://agupubs.onlinelibrary.wiley.com/doi/pdf/10.1029/2005WR004362},
year = {2006}
}

@article{knoben2018,
author = {Knoben, Wouter J. M. and Woods, Ross A. and Freer, Jim E.},
title = {A Quantitative Hydrological Climate Classification Evaluated With Independent Streamflow Data},
journal = {Water Resources Research},
volume = {54},
number = {7},
pages = {5088-5109},
doi = {10.1029/2018WR022913},
url = {https://agupubs.onlinelibrary.wiley.com/doi/abs/10.1029/2018WR022913},
eprint = {https://agupubs.onlinelibrary.wiley.com/doi/pdf/10.1029/2018WR022913},
year = {2018}
}

@article{marquardt1963,
 issn = {03684245},
 url = {http://www.jstor.org/stable/2098941},
 author = {Donald W. Marquardt},
 journal = {Journal of the Society for Industrial and Applied Mathematics},
 number = {2},
 pages = {431-441},
 publisher = {Society for Industrial and Applied Mathematics},
 title = {An Algorithm for Least-Squares Estimation of Nonlinear Parameters},
 volume = {11},
 year = {1963}
}

@article{mcmillan2017,
author = {McMillan, Hilary and Westerberg, Ida and Branger, Flora},
title = {Five guidelines for selecting hydrological signatures},
journal = {Hydrological Processes},
volume = {31},
number = {26},
pages = {4757-4761},
doi = {10.1002/hyp.11300},
url = {https://onlinelibrary.wiley.com/doi/abs/10.1002/hyp.11300},
eprint = {https://onlinelibrary.wiley.com/doi/pdf/10.1002/hyp.11300},
year = {2017}
}

@article{nash1970,
title = {River flow forecasting through conceptual models part {I} - {A} discussion of principles},
journal = {Journal of Hydrology},
volume = {10},
number = {3},
pages = {282-290},
year = {1970},
issn = {0022-1694},
doi = {10.1016/0022-1694(70)90255-6},
url = {https://www.sciencedirect.com/science/article/pii/0022169470902556},
author = {J.E. Nash and J.V. Sutcliffe}
}

@article{olden2003,
author = {Olden, Julian D. and Poff, N. L.},
title = {Redundancy and the choice of hydrologic indices for characterizing streamflow regimes},
journal = {River Research and Applications},
volume = {19},
number = {2},
pages = {101-121},
doi = {10.1002/rra.700},
url = {https://onlinelibrary.wiley.com/doi/abs/10.1002/rra.700},
eprint = {https://onlinelibrary.wiley.com/doi/pdf/10.1002/rra.700},
year = {2003}
}

@article{sadegh2015,
author = {Sadegh, Mojtaba and Vrugt, Jasper A. and Xu, Chonggang and Volpi, Elena},
title = {The stationarity paradigm revisited: Hypothesis testing using diagnostics, summary metrics, and {DREAM}$_\text{(ABC)}$},
journal = {Water Resources Research},
volume = {51},
number = {11},
pages = {9207-9231},
doi = {10.1002/2014WR016805},
url = {https://agupubs.onlinelibrary.wiley.com/doi/abs/10.1002/2014WR016805},
eprint = {https://agupubs.onlinelibrary.wiley.com/doi/pdf/10.1002/2014WR016805},
year = {2015}
}

@article{sadegh2016,
title = {The soil water characteristic as new class of closed-form parametric expressions for the flow duration curve},
journal = {Journal of Hydrology},
volume = {535},
pages = {438-456},
year = {2016},
issn = {0022-1694},
doi = {10.1016/j.jhydrol.2016.01.027},
url = {http://www.sciencedirect.com/science/article/pii/S0022169416000457},
author = {M. Sadegh and J.A. Vrugt and H.V. Gupta and C. Xu}
}

@Article{sawicz2011,
author = {Sawicz, K. and Wagener, T. and Sivapalan, M. and Troch, P. A. and Carrillo, G.},
title = {Catchment classification: empirical analysis of hydrologic similarity based on catchment function in the eastern USA},
journal = {Hydrology and Earth System Sciences},
volume = {15},
year = {2011},
number = {9},
pages = {2895-2911},
url = {https://www.hydrol-earth-syst-sci.net/15/2895/2011/},
doi = {10.5194/hess-15-2895-2011}
}

@article{schoups2010a,
author = {Schoups, G. and Vrugt, J. A. and Fenicia, F. and van de Giesen, N. C.},
title = {Corruption of accuracy and efficiency of {M}arkov chain {M}onte {C}arlo simulation by inaccurate numerical implementation of conceptual hydrologic models},
journal = {Water Resources Research},
volume = {46},
number = {10},
pages = {},
doi = {10.1029/2009WR008648},
url = {https://agupubs.onlinelibrary.wiley.com/doi/abs/10.1029/2009WR008648},
eprint = {https://agupubs.onlinelibrary.wiley.com/doi/pdf/10.1029/2009WR008648},
year = {2010}
}

@article{schoups2010b,
author = {Schoups, Gerrit and Vrugt, Jasper A.},
title = {A formal likelihood function for parameter and predictive inference of hydrologic models with correlated, heteroscedastic, and non-{G}aussian errors},
journal = {Water Resources Research},
volume = {46},
number = {10},
pages = {},
doi = {10.1029/2009WR008933},
url = {https://agupubs.onlinelibrary.wiley.com/doi/abs/10.1029/2009WR008933},
eprint = {https://agupubs.onlinelibrary.wiley.com/doi/pdf/10.1029/2009WR008933},
year = {2010}
}

@article{sorooshian1980,
author = {Sorooshian, Soroosh and Dracup, John A.},
title = {Stochastic parameter estimation procedures for hydrologic rainfall-runoff models: Correlated and heteroscedastic error cases},
journal = {Water Resources Research},
volume = {16},
number = {2},
pages = {430-442},
doi = {10.1029/WR016i002p00430},
url = {https://agupubs.onlinelibrary.wiley.com/doi/abs/10.1029/WR016i002p00430},
eprint = {https://agupubs.onlinelibrary.wiley.com/doi/pdf/10.1029/WR016i002p00430},
year = {1980}
}

@article{sorooshian1983,
author = {Sorooshian, Soroosh and Gupta, Vijai Kumar and Fulton, James Lloyd},
title = {Evaluation of Maximum Likelihood Parameter estimation techniques for conceptual rainfall-runoff models: Influence of calibration data variability and length on model credibility},
journal = {Water Resources Research},
volume = {19},
number = {1},
pages = {251-259},
doi = {10.1029/WR019i001p00251},
url = {https://agupubs.onlinelibrary.wiley.com/doi/abs/10.1029/WR019i001p00251},
eprint = {https://agupubs.onlinelibrary.wiley.com/doi/pdf/10.1029/WR019i001p00251},
year = {1983}
}

@article{vogel1994,
author = {Richard M. Vogel  and Neil M. Fennessey },
title = {Flow-duration Curves. I: New Interpretation and Confidence Intervals},
journal = {Journal of Water Resources Planning and Management},
volume = {120},
number = {4},
pages = {485-504},
year = {1994},
doi = {10.1061/(ASCE)0733-9496(1994)120:4(485)},
url = {},
eprint = {}
}

@article{yadav2007,
title = {Regionalization of constraints on expected watershed response behavior for improved predictions in ungauged basins},
journal = {Advances in Water Resources},
volume = {30},
number = {8},
pages = {1756-1774},
year = {2007},
issn = {0309-1708},
doi = {10.1016/j.advwatres.2007.01.005},
url = {http://www.sciencedirect.com/science/article/pii/S0309170807000140},
author = {Maitreya Yadav and Thorsten Wagener and Hoshin Gupta}
}

@article{yapo1996,
title = {Automatic calibration of conceptual rainfall-runoff models: sensitivity to calibration data},
journal = {Journal of Hydrology},
volume = {181},
number = {1},
pages = {23-48},
year = {1996},
issn = {0022-1694},
doi = {10.1016/0022-1694(95)02918-4},
url = {http://www.sciencedirect.com/science/article/pii/0022169495029184},
author = {Patrice O. Yapo and Hoshin Vijai Gupta and Soroosh Sorooshian}
}

@article{zhang2016,
title = {Multi-metric calibration of hydrological model to capture overall flow regimes},
journal = {Journal of Hydrology},
volume = {539},
pages = {525-538},
year = {2016},
issn = {0022-1694},
doi = {10.1016/j.jhydrol.2016.05.053},
url = {http://www.sciencedirect.com/science/article/pii/S0022169416303286},
author = {Yongyong Zhang and Quanxi Shao and Shifeng Zhang and Xiaoyan Zhai and Dunxian She}
}

\end{document}